# Modeling Transformative AI Risks (MTAIR) Project

Sam Clarke

Ben Cottier

Aryeh Englander

Daniel Eth

David Manheim

Samuel Dylan Martin

Issa Rice

# Table of Contents

# Table of Figures

# List of Tables

# Acknowledgments

The MTAIR project (formerly titled, "AI Forecasting: What Could Possibly Go Wrong?") was originally funded through the Johns Hopkins University Applied Physics Laboratory (APL), with team members outside of APL working as volunteers. While APL funding was only for one year, the non-APL members of the team have continued work on the project, with additional support from the EA Long-Term Future Fund (except for Daniel Eth, whose funding came from FHI). Aryeh Englander has also continued working with the project under a grant from the Johns Hopkins Institute for Assured Autonomy (IAA).

The project is led by Daniel Eth (FHI), David Manheim, and Aryeh Englander (APL). The original APL team included Aryeh Englander, Randy Saunders, Joe Bernstein, Lauren Ice, Sam Barham, Julie Marble, and Seth Weiner. Non-APL team members include Daniel Eth (FHI), David Manheim, Ben Cottier, Sammy Martin, Jérémy Perret, Issa Rice, Ross Gruetzemacher (Wichita State University), Alexis Carlier (FHI), and Jaime Sevilla.

We would like to thank a number of people who have graciously provided feedback and discussion on the project. These include (apologies to anybody who may have accidentally been left off this list): Ashley Llorens (formerly APL, currently at Microsoft), I-Jeng Wang (APL), Jim Scouras (APL), Helen Toner, Rohin Shah, Ben Garfinkel, Daniel Kokotajlo, and Danny Hernandez, as well as several others who prefer not to be mentioned. We are also indebted to several other people who have provided feedback and discussion on some or all of these chapters, including Neel Nanda, Adam Shimi, Edo Arad, Lukas Finnveden, Jennifer Lin, Ben Snodin, Evan Hubinger, Chris van Merwijk, Ozzie Gooen, Richard Ngo, and Paul Christiano. Last, but certainly not least, Wes Cowley, for his excellent editing of the final version of the report, without whom the full report would not exist.

The chapter on takeover scenarios (§8) was partially inspired by similar work by Kaj Sotala [1]. All errors are our own.

Note: while this report is very long, and the chapters do reference each other, they are also largely self-contained. Chapters can be read in whatever order you prefer, and you don't need to read the whole thing to make sense of parts that you're interested in reading.

# 1 Introduction

## David Manheim, Aryeh Englander

Numerous books, articles, and blog posts have laid out reasons to think that AI might pose catastrophic or existential risks for the future of humanity. However, these reasons often differ both in their details and in their main conceptual arguments, and other researchers have questioned or disputed many of the key assumptions and arguments.

The disputes and associated discussions can often become quite long and complex, and they can involve many different arguments, counterarguments, sub-arguments, implicit assumptions, and references to other discussions or debated positions. Many of the relevant debates and hypotheses are also subtly related to each other.

In 2019, Ben Cottier and Rohin Shah created the following hypothesis map [2], which provided a useful starting point for untangling and clarifying some of these interrelated hypotheses and disputes:

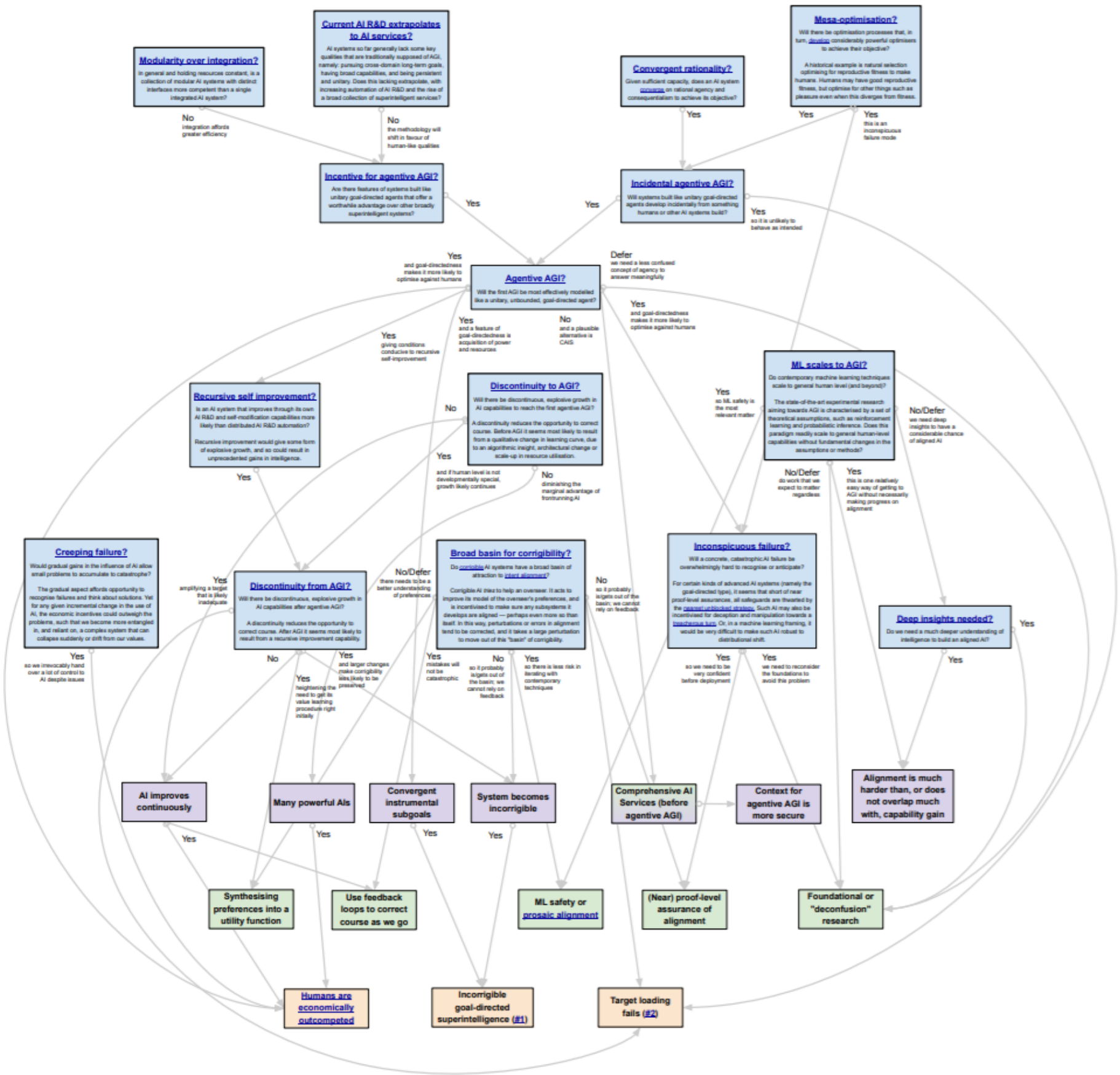

*Figure 1: Cottier and Shah's hypothesis map*

The Modelling Transformative AI Risks (MTAIR) project is an attempt to build on this earlier work by including additional hypotheses, debates, and uncertainties, and by including more recent research. Eventually, we are also hoping to convert the work we have done building on Cottier and Shah's informal diagram into a quantitative model that can incorporate explicit probability estimates, measures of uncertainty, relevant data, and other quantitative factors or analyses in a way that might be useful for planning or decision-making.

In this document, we present preliminary outputs from this project along with some of our future plans. Although the project is still a work in progress, we believe we are now at a stage where we can productively engage the community, both to contribute to the relevant discourse and to solicit feedback, critiques, and suggestions.

The remainder of this chapter gives a brief conceptual overview of our approach and a high-level walkthrough of the hypothesis map that we have developed. Subsequent chapters go into more detail on different parts of this model. At this stage, we are interested in presenting our tentative conclusions and in feedback on the portions of the model that we are presenting in detail.

## 1.1 Conceptual Approach

There are two primary parts to the MTAIR project. The first, which is now concluded, involved creating a qualitative map ("model") of key hypotheses, cruxes, and relationships. The second part, which is now partly underway, is to convert our qualitative map into a quantitative model, eventually to include estimates elicited from experts, in a way that can be useful for decision-making.

**Mapping key hypotheses:** As mentioned above, this part of the project involves an ongoing effort to map out the key hypotheses and debate cruxes relevant to risks from transformative AI and high-level machine intelligence (HLMI), in a manner comparable to and building upon the earlier diagram [2] by Ben Cottier and Rohin Shah. As shown in the conceptual diagram below (Figure 2), the idea is to create a qualitative map showing how the various disagreements and hypotheses (blue nodes) are related to each other, how different proposed technical or governance agendas (green nodes) relate to different disagreements and hypotheses, and how all of those factors feed into the likelihood that different catastrophe scenarios (red nodes) might materialize.

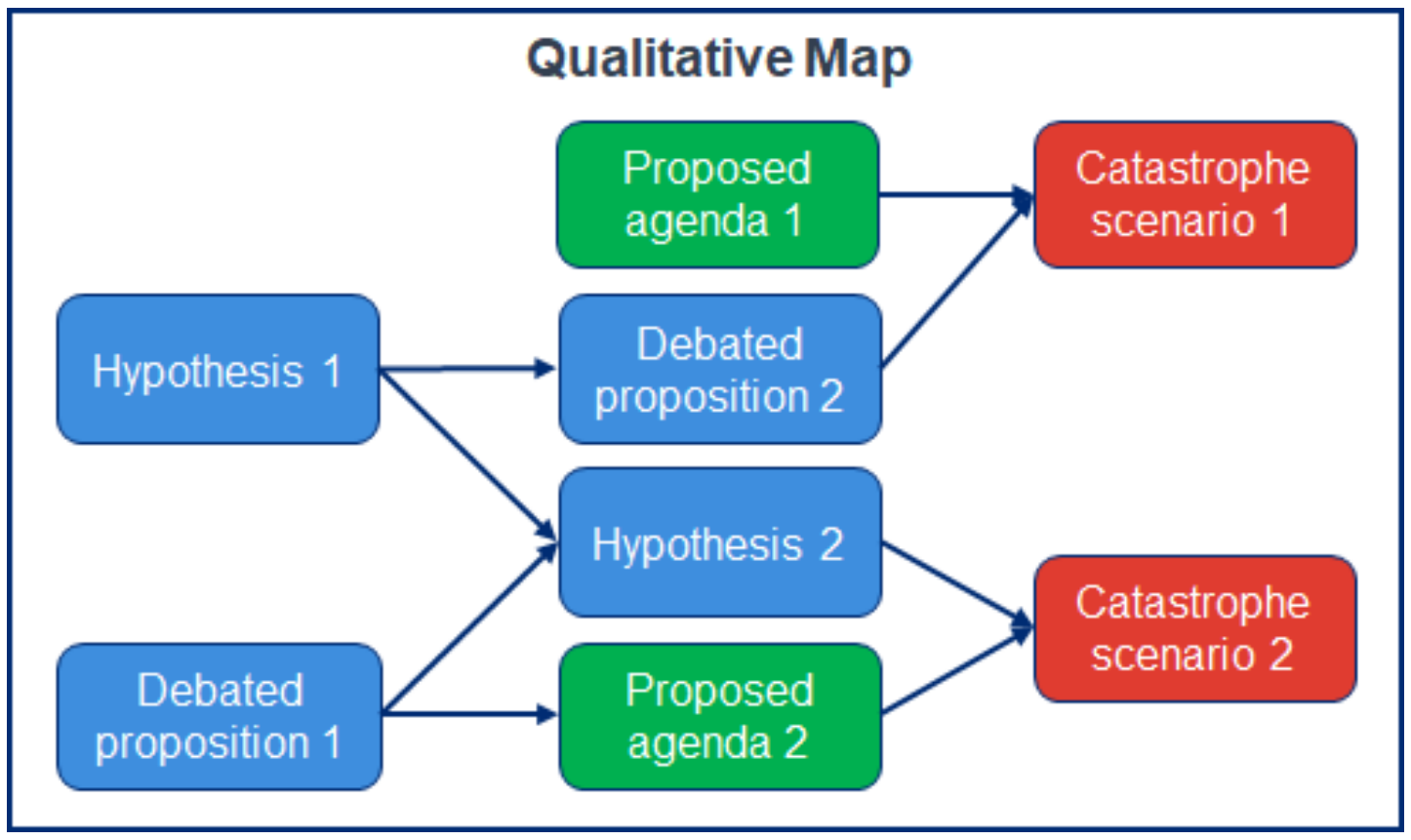

*Figure 2: Structure of the qualitative map*

**Quantification and decision analysis:** Our longer-term plan is to convert our hypothesis map into a quantitative model (Figure 3) that can be used to calculate decision-relevant probability estimates. For example, a completed model could output a roughly estimated probability of transformative AI arriving by a given date, a given catastrophe scenario materializing, or a given approach successfully preventing a catastrophe.

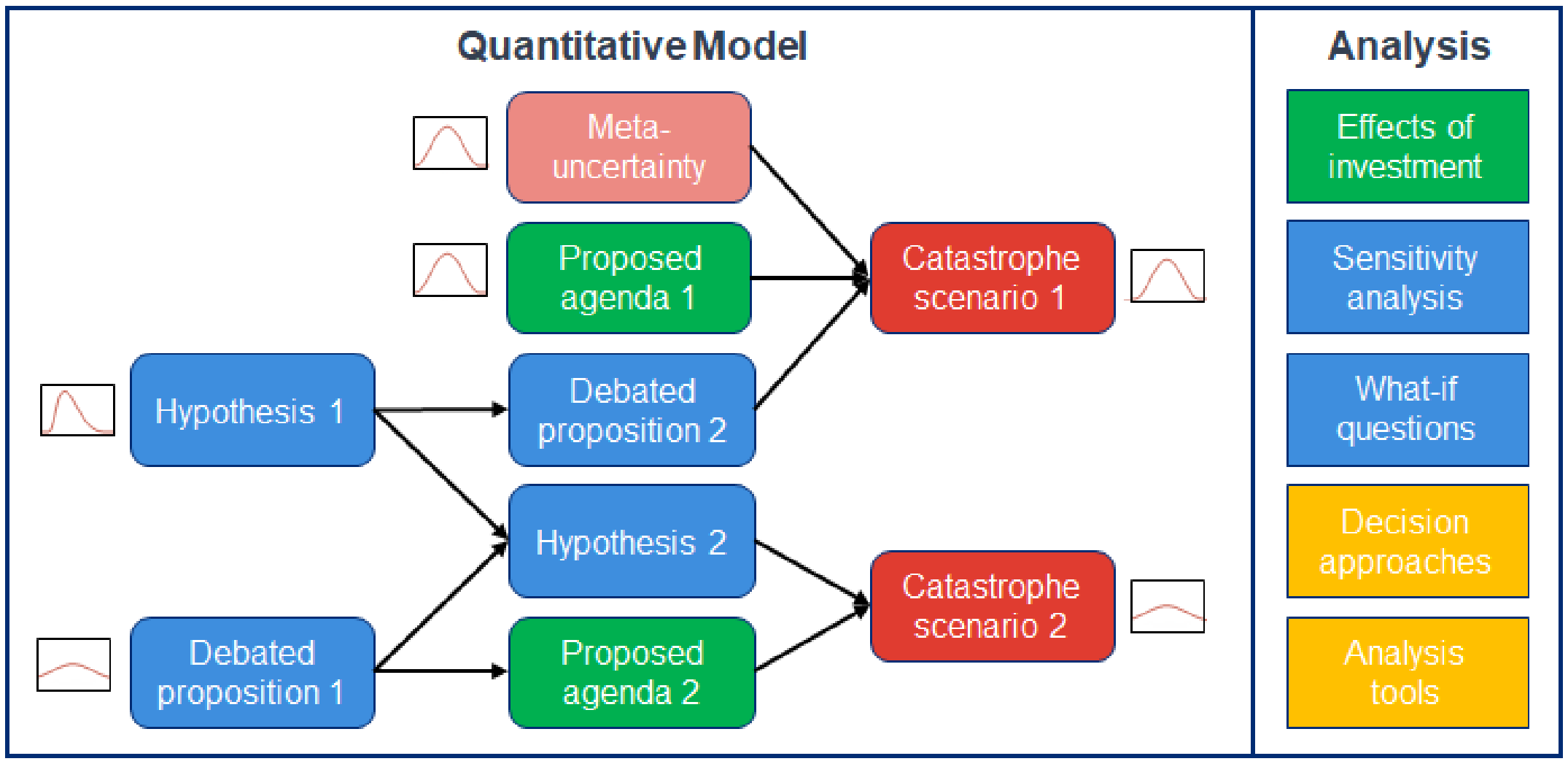

*Figure 3: Structure of the quantitative map*

The basic idea is to take into account any available data along with probability estimates or structural beliefs elicited from relevant experts (which users can modify or replace with their own estimates as desired). Once this model is fully implemented, we can then calculate Monte Carlo–based probability estimates for downstream nodes of interest based either on a subset or a weighted average of expert opinions, or on specific claims about the structure or quantities of interest, or on a combination of the above. Finally, even if the outputs are not accepted, we can use the indicative values as inputs for a variety of analysis tools or formal decision-making techniques. For example, we might consider the choice to pursue a given alignment strategy and use the model as an aid to think about how the investment payoff changes if we believe hardware progress will accelerate or if we presume that there is relatively more existential risk from nearer-term failures.

## 1.2 Model Overview

The next several chapters dive into the details of our current qualitative model. Each has been written by team members involved in crafting that particular part of the model, as different team members or groups of team members worked on different parts of the model.

The structure of each part of the model is primarily based on a literature review and the understanding of the team members, along with considerable feedback and input from researchers outside the team. As noted above, this work will hopefully continue to gather input from the community and lead to further discussions. At the same time, the various parts of the model are interrelated. Daniel Eth led the work of integrating the individual parts of the model.

## 1.2.1 Note on Implementation and Software:

At present, we are using Analytica, a "visual software environment for building, exploring, and sharing quantitative decision models that generate prescriptive results." The models displayed in the rest of this sequence were created using this software program. Note: If you have Windows, you can download the free version of Analytica. We hope to make the model files available, if not publicly, at least on request. Editing the full model unfortunately needs the expensive licensed version of Analytica. Please contact David Manheim, Sammy Martin, or Aryeh Englander for more information or requests for access.

## 1.2.2 How to Read Analytica Models

Before presenting an overview of the actual model, and as a reference for later chapters, we present a brief explanation of how the models work and how they can be read. Analytica models are composed of different types of nodes with the relationships between nodes represented as directed edges (arrows) (Figure 4). The two primary types of nodes in our model are variable nodes and modules. Variable nodes are usually oval or rounded rectangles[1] **without** bolded outlines; these correspond to key hypotheses, cruxes of disagreement, or other parameters of interest. Modules, represented by rounded rectangles **with** bolded outlines, are "sub-models" that contain their own sets of nodes and relationships. In our model we also sometimes use small square nodes to visually represent AND, OR, or NOT relationships. In the software, a far wider set of ways to combine outputs from nodes are available and will be used in our model—but these are difficult to represent visually.

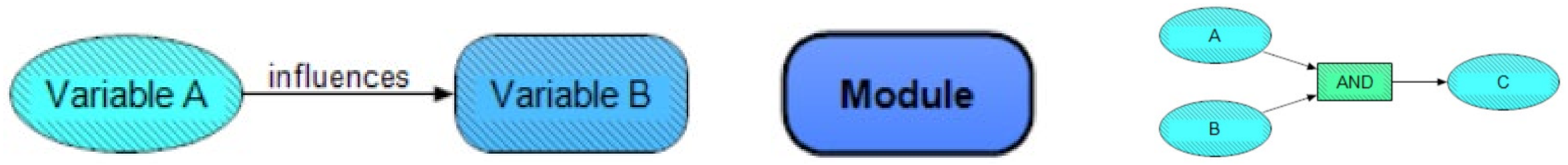

*Figure 4: Module types in the model*

Arrows represent directions of probabilistic influence, in the sense that information about the first node influences the probability estimate for the second. For example, an arrow from Variable A to Variable B indicates that the probability of B depends at least in part on the probability of A. It is important to note that the model is not a causal model *per se*. An edge from one node to another does not necessarily imply that the first *causes* the second, but rather that there is some relationship between them such that information about the first *informs* the probability estimate for the second. Some edges do represent causal relationships but only insofar as those relationships are important for informing probability estimates.

Different parts of the model use various color schemes to group nodes that share certain characteristics, but color does not have any formal meaning in Analytica and is not necessary to make sense of the model. The color schemes for individual parts of the model will be explained as needed, but they can be safely ignored if they become confusing.

[1] There are technical differences between ovals and rounded rectangle types of variable nodes, but for the purposes of the qualitative mapping part of the project—the focus of this work—the two are interchangeable.

Other things to note:

- In some of the diagrams there are small arrowheads leading into or out of certain nodes but which do not point to any other node in the diagram (Figure 5). These indicate that there are nodes elsewhere in the model that depend on this node or that this node depends on.

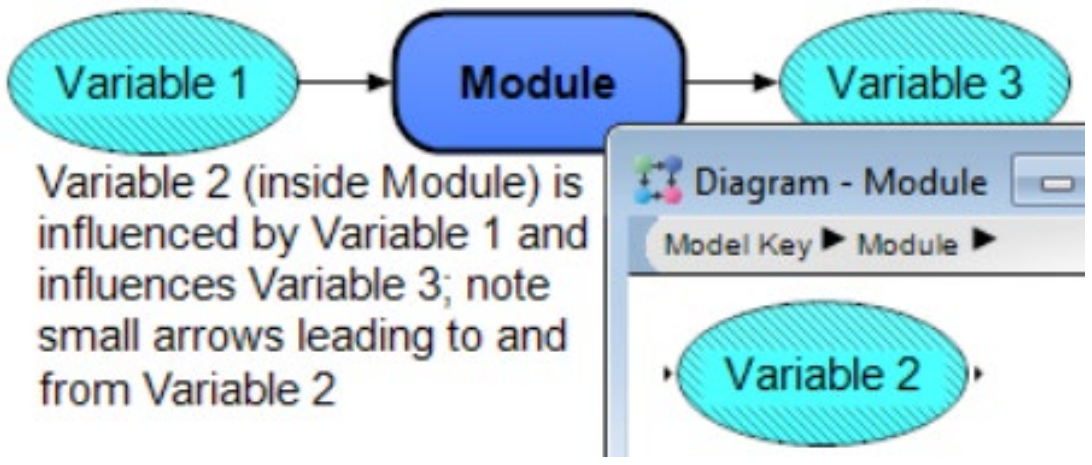

*Figure 5: Representing dependencies in the model*

- Alias nodes are copies of nodes that link back to the original "real" node and are mainly useful for display or readability purposes (Figure 6). We use alias nodes in many parts of our diagrams, especially when a node from one module influences or is influenced by some important node(s) elsewhere in the model. Analytica indicates that a node is an alias by displaying the node name in italics.

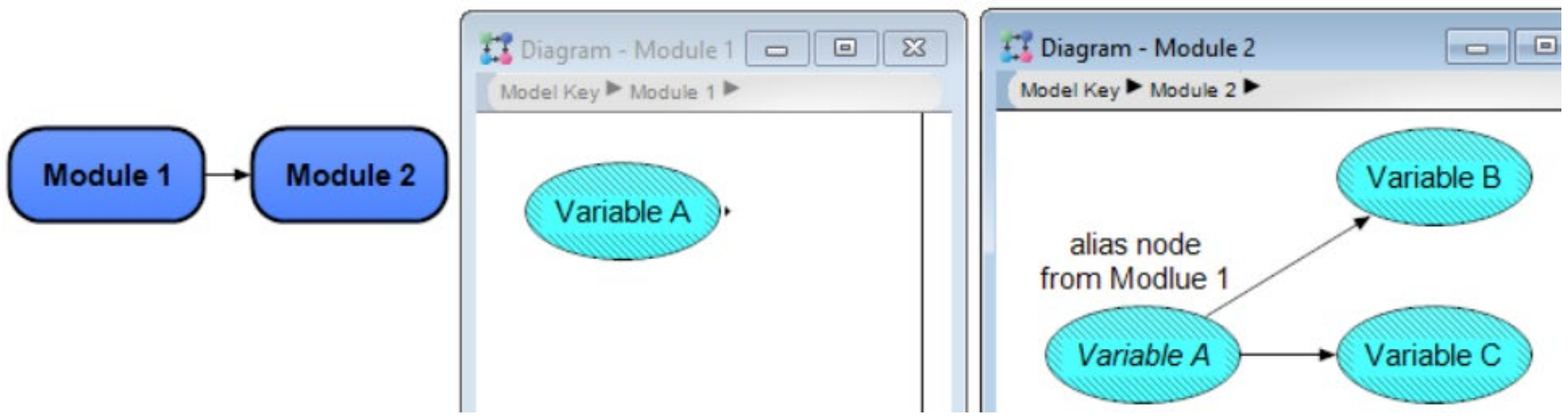

*Figure 6: Alias nodes in the model*

- Our model is technically a directed acyclic graph. However, there are a few places in the model diagrams where Analytica confusingly displays bidirectional arrows between modules even though the direction of influence only goes in one direction (Figure 7). This is because Analytica uses arrows to indicate not only direction of influence but also that one module contains an alias node from a different model. For example, the direction of influence in the image below is from Variable A in Module 1 to Variable B in Module 2, but Analytica displays a bidirectional arrow between the modules because Module 1 also contains an alias node from Module 2.

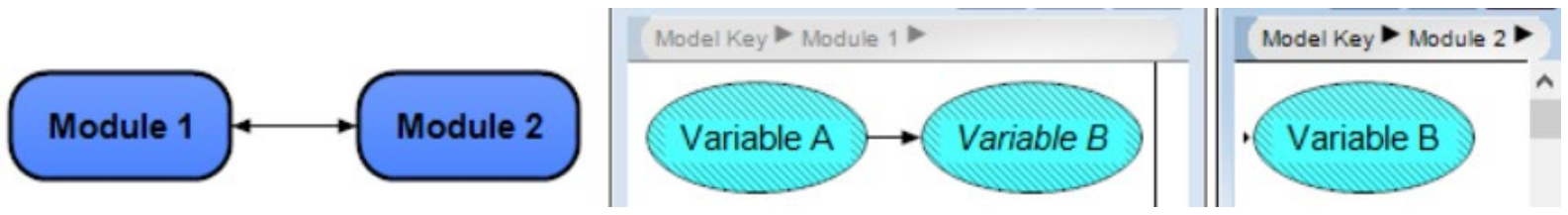

*Figure 7: Bidirectional arrows in the model*

## 1.2.3 Top-Level Model Walkthrough

The image below (Figure 8) represents the top-level diagram of our current model. Most of the nodes in this diagram are their own separate modules, each with their own set of nodes and relationships. Most of these modules will be discussed in much more detail in later chapters.

Below, we highlight key potential nodes and the related questions, and discuss how they are interrelated at a high level. This overview, which in part explains the diagram below, hopes to provide a basic outline of what later chapters discuss in much more detail.

**Important note for interpreting the model:** The arrows represent the direction of inference in the model, rather than underlying causal relationships. Similarly, the relationships between the modules reflect dependencies between individual nodes in the modules, rather than indicating notional suggestions about relationships between the concepts represented by the modules themselves.

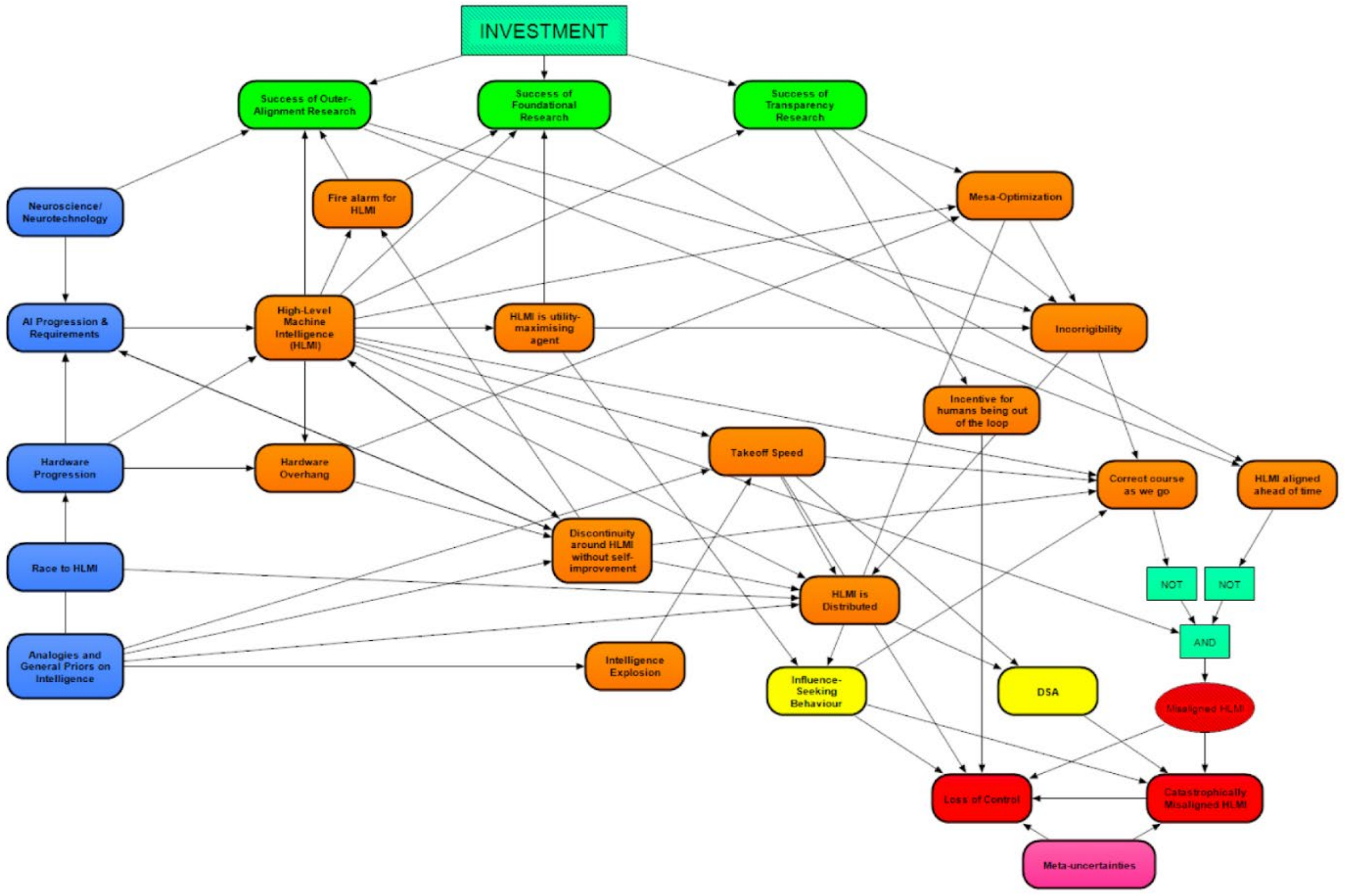

*Figure 8: Top-level view of the model*

The blue nodes on the left represent technical or other developments or future progress areas that are potentially relevant inputs to the rest of the model. They are Neuroscience / Neurotechnology, AI Progression & Requirements (§3.2), Hardware Progression (§3.1), and Race to HLMI. Finally, Analogies and General Priors on Intelligence (§2), which address many assumptions and arguments by analogy from domains like human evolution, are used to ground debates about AI takeoff or timelines. These are the key inputs for understanding progress towards high-level machine intelligence (HLMI)[2].

The main internal portions of the model (largely in orange), represent the relationships between different hypotheses and potential medium-term outcomes. Several key parts of this, which are discussed in later chapters, include paths to HLMI (§3) and the inputs to it, in the blue modules), takeoff/discontinuities (§4), and mesa-optimization (§5). Impacting these are different safety agendas (§6) (along the top in green).

Finally, the nodes on the bottom right represent conditions leading to failure (yellow) and failure modes (red). For instance, the possibility of misaligned HLMI (bottom right in red) motivates the critical question of how the misalignment can be prevented. Two possibilities are modeled (orange nodes, right): The first is that HLMI is aligned ahead of time (using outer alignment, inner alignment [3] and, if necessary, foundational research [4], as discussed in other research on alignment). The second possibility is that we can "correct course as we go," for instance, by using an alignment method that ensures the HLMI is corrigible [5].

While our model has intermediate outputs (which when complete will include estimates of HLMI timelines and takeoff speed), its principal outputs are the predictions for the modules marked in red. Catastrophically Misaligned HLMI (§7.5) covers scenarios involving a single HLMI or a coalition achieving a decisive strategic advantage (DSA) over the rest of the world and causing an existential catastrophe. Loss of Control (§7.6) covers creeping-failure scenarios, including those that don't require a coalition or individual AI to seize a DSA.

### 1.2.4 The Model is (Already) Wrong

We expect that readers will disagree with us, and with one another, about various points. As mentioned, the above is only a high-level overview, and many items in it are contentious or unclear—which is exactly why we are trying to map it more clearly.

Throughout this work, we attempt to model disagreements and how they relate to each other, as shown in the earlier notional outline for mapping key hypotheses. As a concrete example, whether HLMI will be agentive, itself a debate, influences whether it is plausible that the HLMI will attempt to self-modify or design successors. The feasibility of either modification or successor design is another debate, and this partly determines the potential for very fast takeoff, influencing the probability of a catastrophic outcome. As the example illustrates, the values and the connections between the nodes are all therefore subject to potential disagreement, which must be represented in order to model the risk. Further and more detailed examples are provided in the following chapters.

[2] Note that HLMI is viewed as a precursor to and a likely cause of transformative AI. For this reason, in the model, we discuss HLMI, which is defined more precisely in later chapters.

## 1.3 Further Chapters and Feedback

The upcoming chapters will cover the internals of these modules, which are only outlined at a very high level here, starting with the chapter on Analogies and General Priors on Intelligence (§2) followed by Paths to HLMI (§3).

If you think any of this is useful, or if you already disagree with some of our claims, we are very interested in feedback and disagreements and hope that the report provides an introduction to productive discussions. We are especially interested in places where the model does not capture any expert views or fails to include an uncertainty which could be an important crux. Similarly, if any explanations seem confused or confusing, flagging this is useful—both to help us clarify and to ensure it doesn't reflect an actual disagreement. It may also be useful to consider things which are modeled but are *not* cruxes or are obvious, since others may disagree.

Also, if this seems interesting or related to any other work you are doing to map or predict the risks, please be in touch—we are still open to additional collaborators or outside feedback for the next parts of the project.

# 2 Analogies and General Priors on Intelligence

Issa Rice, Samuel Dylan Martin

This chapter explains our effort to incorporate basic assumptions and arguments by analogy about intelligence, which are used to ground debates about AI takeoff and paths to high-level machine intelligence (HLMI[3]). In the overall model, this module, *Analogies and General Priors on Intelligence*, is one of the main starting points, and influences modules (covered in subsequent chapters) addressing the possibilities of a *Discontinuity around HLMI* (§4.1.1) or an *Intelligence Explosion* (§4.1.2), as well as *AI Progression* (§3.2) and HLMI *Takeoff Speed* (§4.3.1).

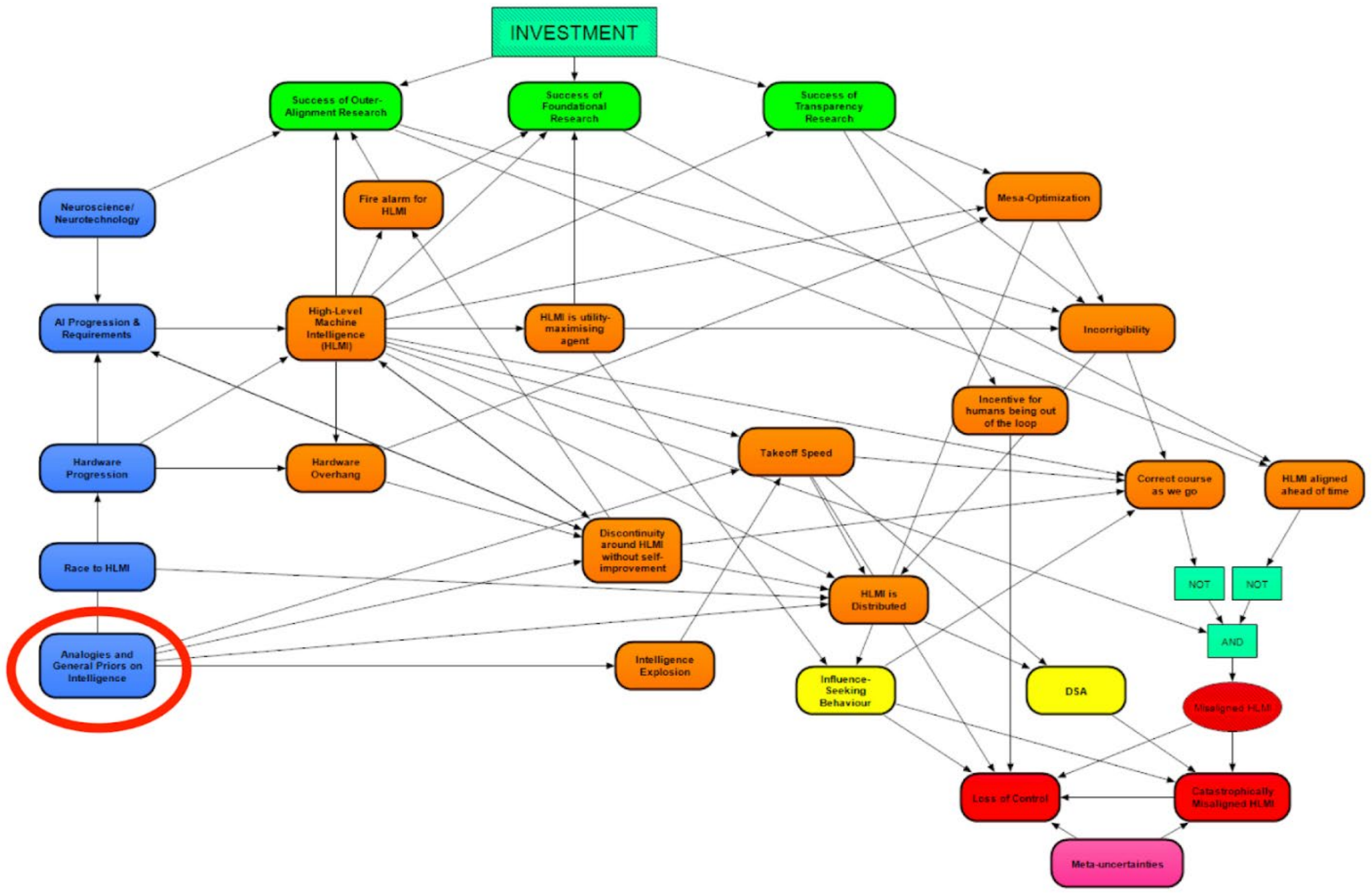

*Figure 9: Model focus for the Analogies and Priors module*

---

[3] We define HLMI as machines that are capable of performing (either individually or collectively) almost all economically relevant information-processing tasks that are performed by humans, or quickly (relative to humans) learning to perform such tasks. We are using the term "high-level machine intelligence" here instead of the related terms "human-level machine intelligence," "artificial general intelligence," or "transformative AI," since these other terms are often seen as baking in assumptions about either the nature of intelligence or advanced AI that are not universally accepted.

The *Analogies and General Priors on Intelligence* module addresses various claims about AI and the nature of intelligence:

- The difficulty of marginal intelligence improvements at the approximate "human level" (i.e., around HLMI)
- Whether marginal intelligence improvements become increasingly difficult beyond HLMI at a rapidly growing rate or not
    - "Rapidly growing rate" is operationalized as becoming difficult exponentially or faster than exponentially
- Whether there is a fundamental upper limit to intelligence that is not significantly above the human level
- Whether, in general, further improvements in intelligence tend to be bottlenecked by previous improvements in intelligence rather than some external factor (such as the rate of physics-limited processes)

These final outputs are represented by the four terminal nodes at the bottom of the module.

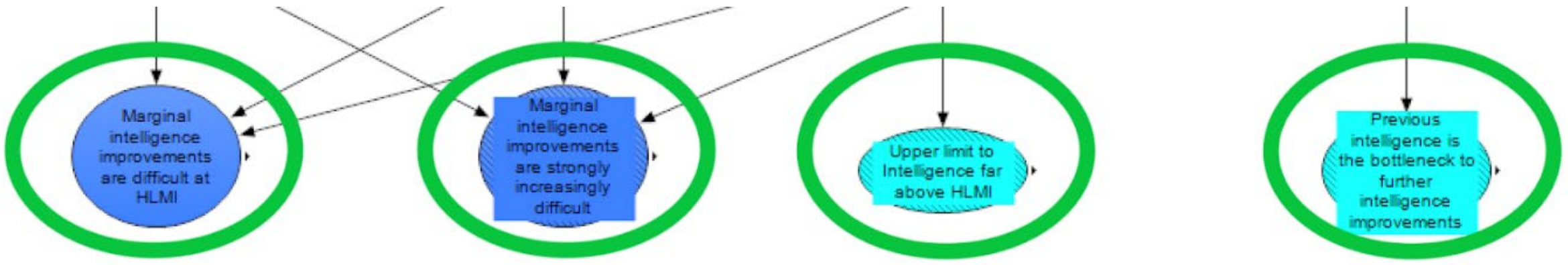

*Figure 10: Outputs of the Analogies and Priors module*

These four claims depend on arguments that analogize the development of HLMI to other domains. Each of these arguments is given a specific submodule in our model, and each argument area is drawn from our review of the existing literature. In the "outputs" section (§2.3) at the end of this chapter, we explain why we chose these four as cruxes for the overall debate around HLMI development.

Each argument area is shown in Figure 11 below, a zoomed-out view of the entire *Analogies and General Priors on Intelligence* module. The argument areas are human biological evolution, machine learning, human cultural evolution, the overall history of AI, and the human brain. Each of these areas is an example of either the development of intelligence or of intelligence itself, and each might be a useful analogy for HLMI.

We also incorporate broad philosophical claims about the nature of intelligence as informative in this module. These claims act as priors before the argument areas are investigated and cover broad issues like whether the concept of general intelligence is coherent and whether the capabilities of agents are strongly constrained by things other than intelligence.

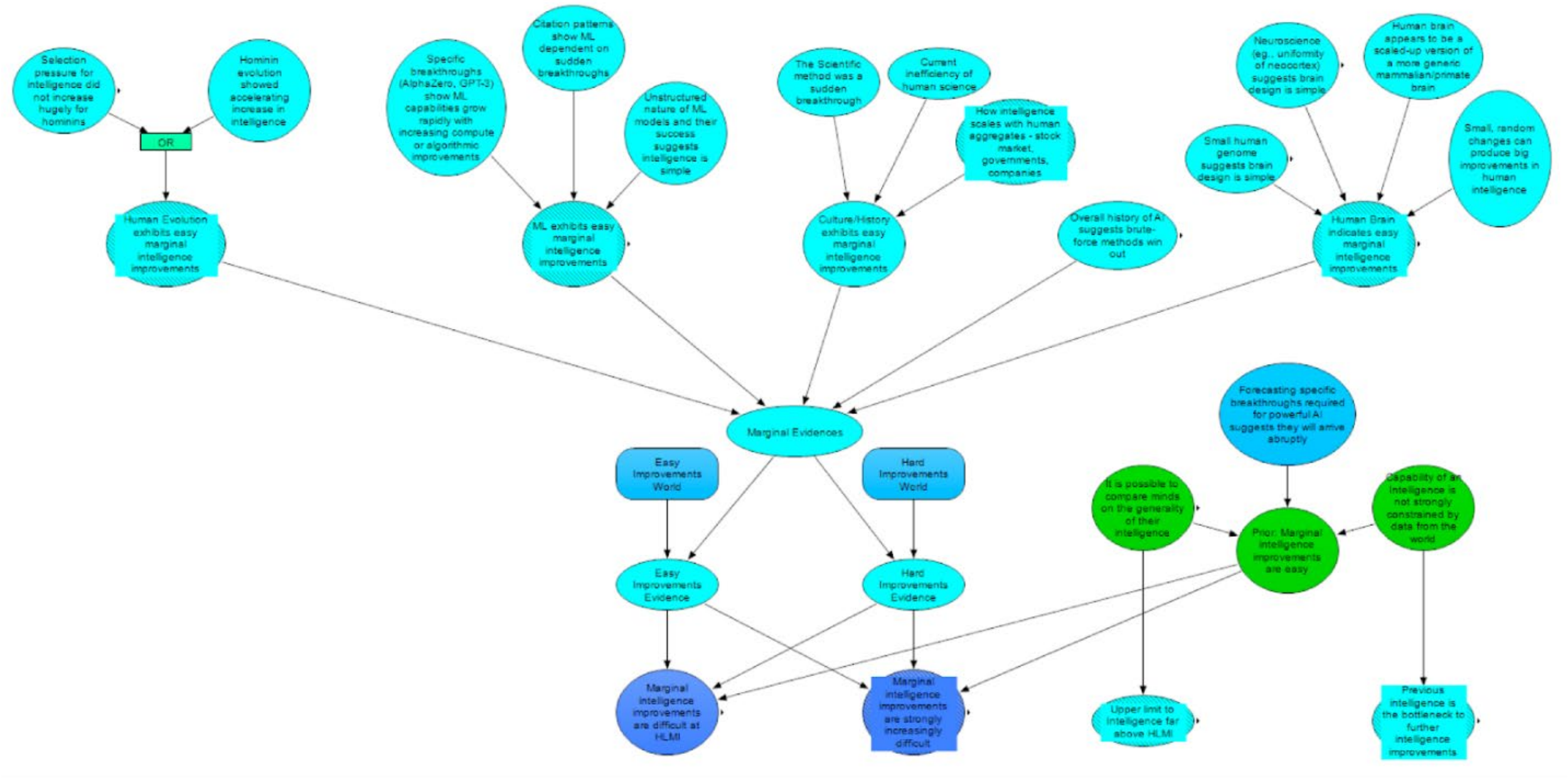

*Figure 11: Full view of the Analogies and Priors module*

## 2.1 Module Overview

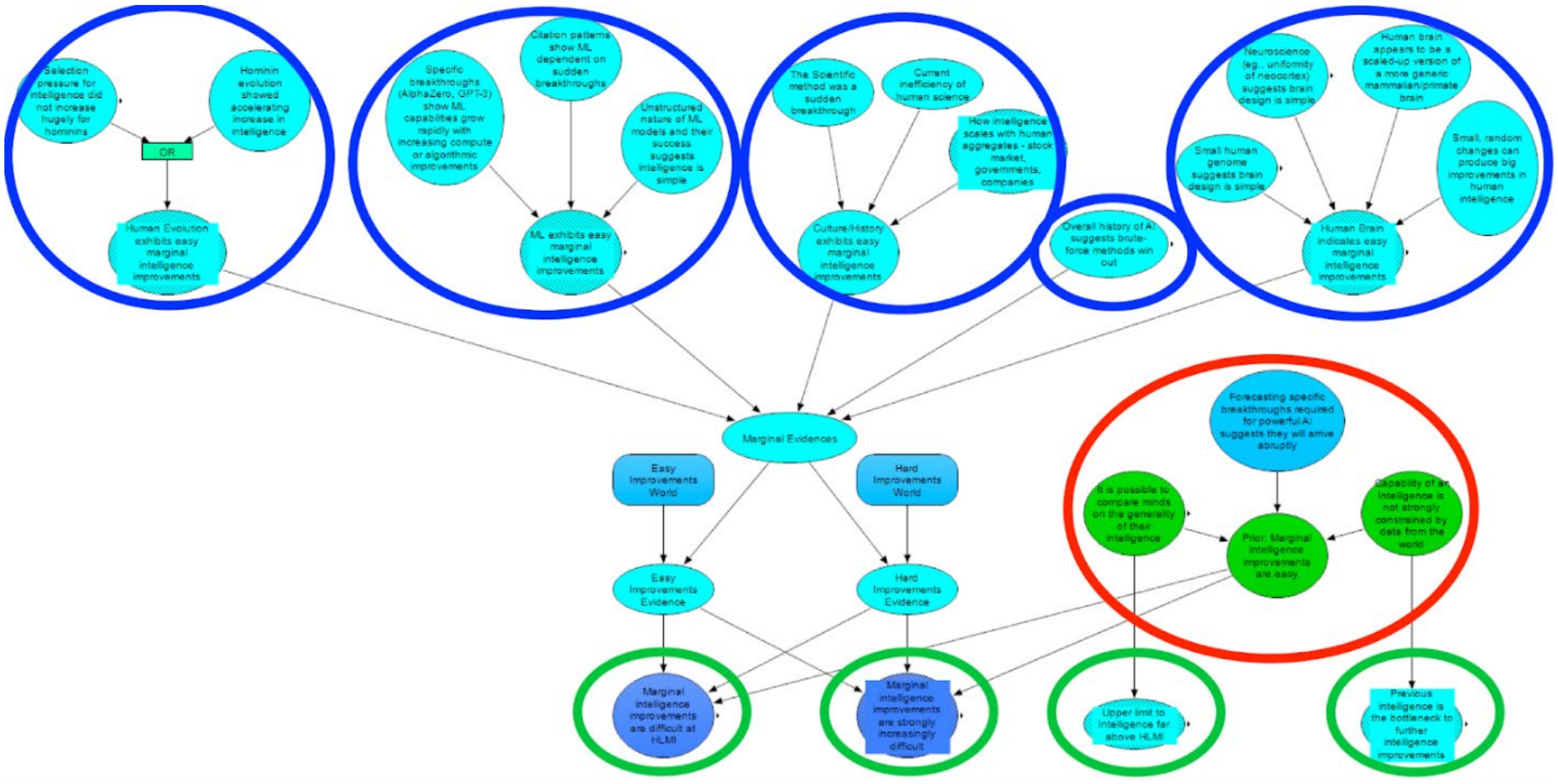

*Figure 12: Cruxes, general priors, and outputs of the Analogies and Priors module*

In this diagram of the overall module, the groups of nodes circled in blue are (left to right), Human Evolution (§2.2.1), Machine Learning (§2.2.2), Human Culture and History (§2.2.3), History of AI (§2.2.4), and the Human Brain (§2.2.5), which feed forward to a classifier that estimates the ease of marginal intelligence improvements. The red circle shows General Priors (§2.2.6), and the green-circled nodes are the final model outputs (§2.3) as discussed above. We discuss each of these sections in this order in the rest of this chapter.

## 2.2 The Cruxes

Each of the argument area submodules, as well as the General Priors submodule, contains cruxes (represented by nodes) that ultimately influence this module's terminal nodes.

### 2.2.1 Human Evolution

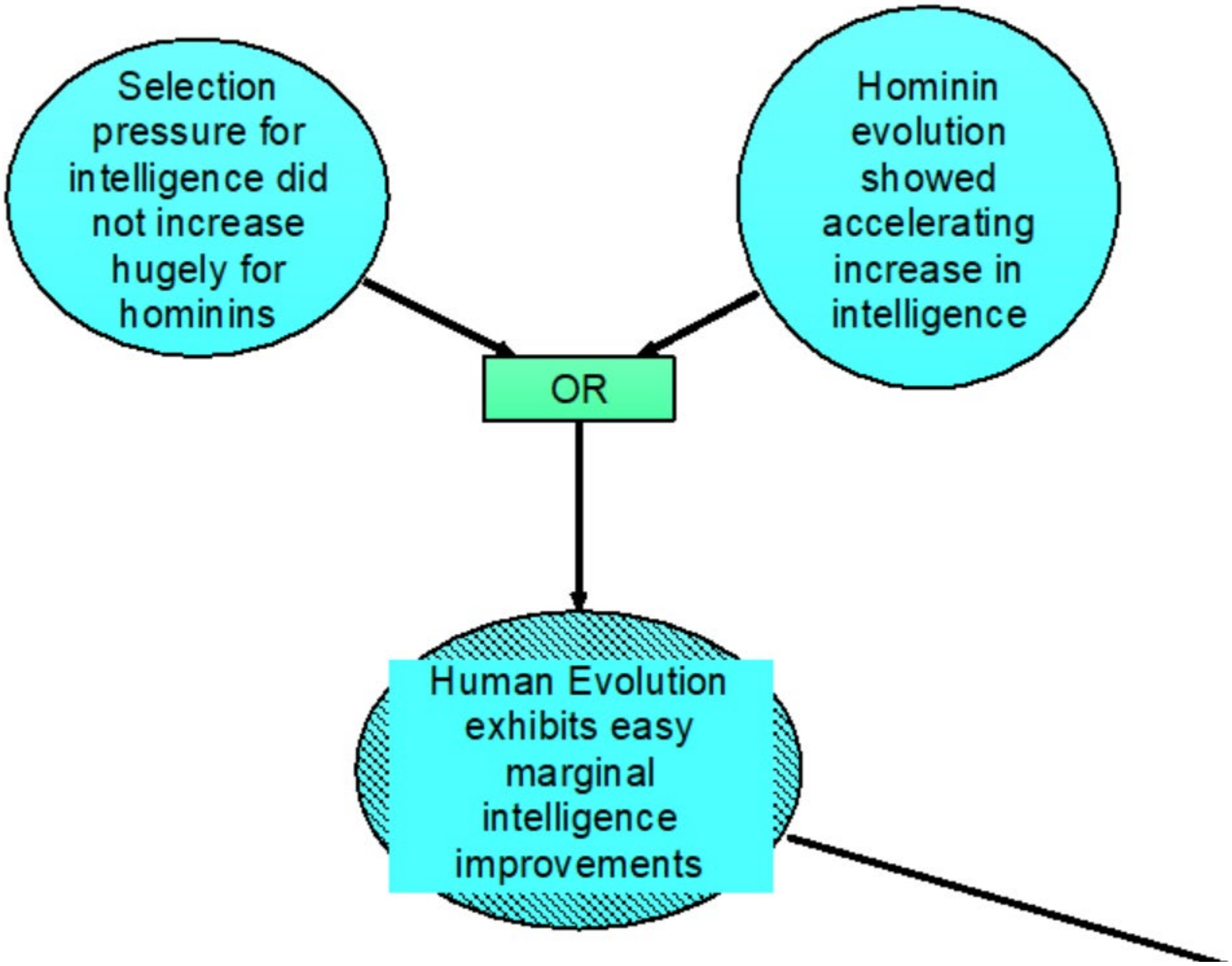

*Figure 13: Human evolution cruxes*

Human evolution is one of the areas most commonly analogized to HLMI development. Eliezer Yudkowsky argued in [Intelligence Explosion Microeconomics] [6] that human evolution demonstrated accelerating returns on cognitive investment, which suggests marginal intelligence improvements were (at least) not rapidly increasingly difficult during this process.

There are two sources of potential evidence which could support these claims about evolution. The first (left-hand node in Figure 13) is if hominin evolution did not involve hugely more selection pressure for intelligence than did the evolution of other, ancestral primates. It is generally assumed that hominins (all species more closely related to *Homo sapiens* than to chimpanzees, e.g., all species from *Australopithecus* and *Homo*) saw much faster rises in intelligence than what had happened prior to primates. If this unprecedented rise in intelligence was not due to drastically changing selection pressures for intelligence, then that would provide evidence that this fast rise in intelligence did not involve evolution "solving" increasingly more difficult problems than what came before. On the other hand, if selection pressures for intelligence had been marginal (or less) up until this point and were suddenly turned way up, then the fast rise in intelligence could be squared with evolution solving more difficult problems (as the increase in intelligence could then be less than proportional to the increase in selection pressures for intelligence).

Even if selection pressures for intelligence were marginal before hominins, we could still obtain evidence that human evolution exhibited easy marginal intelligence improvements—if we observe a rapid acceleration of intelligence *during* hominin evolution up to *Homo sapiens* (right-hand node in Figure 13). Such an acceleration of intelligence would be less likely if intelligence improvements became rapidly more difficult as the human level was approached.

We must note, however, that the relevance of these evolutionary considerations for artificial intelligence progress is still a matter of debate. Language, for instance, evolved very late along the human lineage, while AI systems have been trained to deal with language from much earlier in their relative development. It is unknown how differences such as this would affect the difficulty-landscape of developing intelligence. The amount to update based on analogies to evolution, however, is not handled by this specific submodule, but instead by the classifier (mentioned above and described below in the section on outputs [§2.3]).

Sources:
Likelihood of discontinuous progress around the development of AGI [7]
Takeoff speeds—The sideways view [8]
Hanson-Yudkowsky AI Foom Debate [9]
Thoughts on Takeoff Speeds [10]

## 2.2.2 Machine Learning

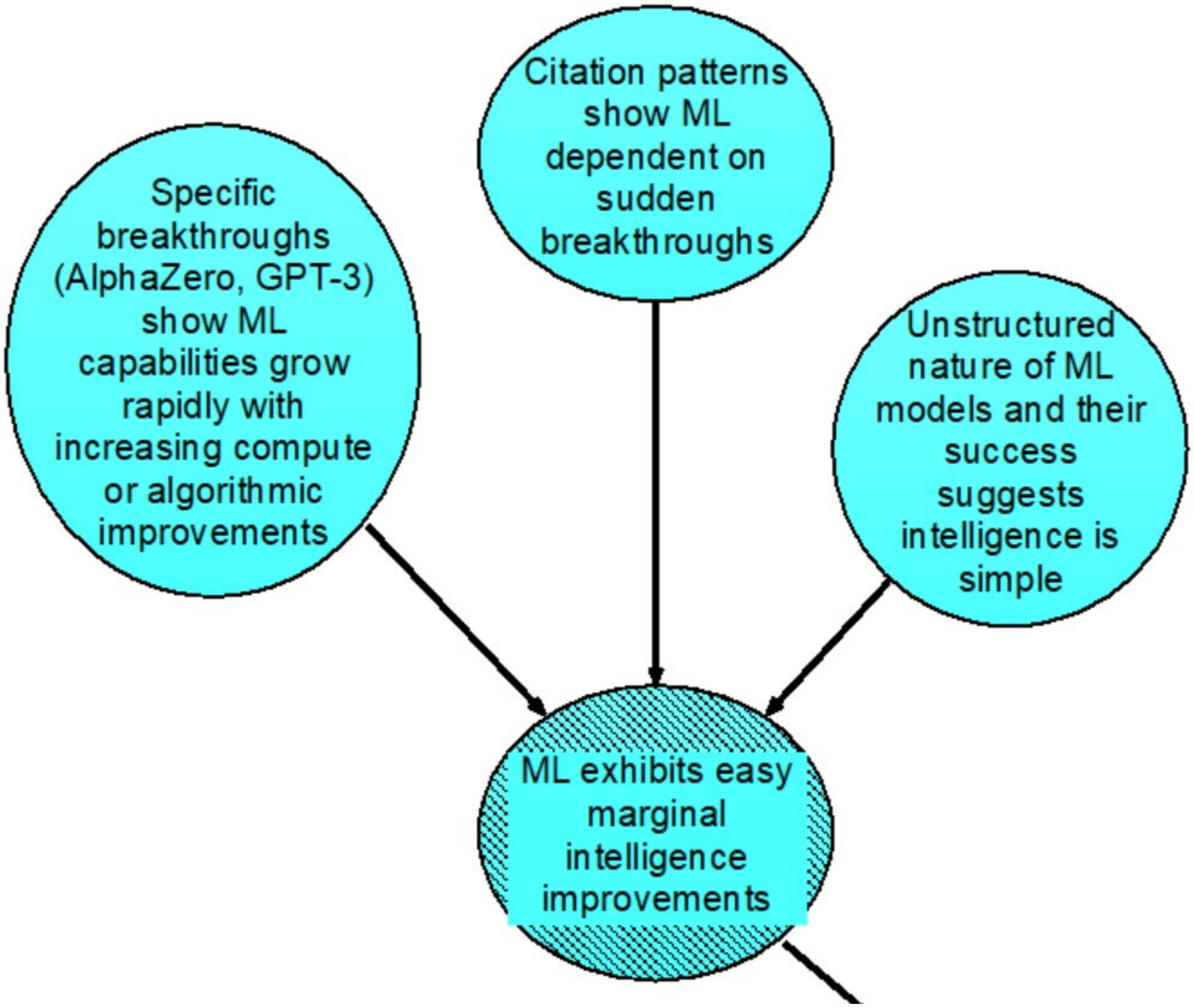

*Figure 14: Machine learning cruxes*

Much of the debate around HLMI has focused on what, if any, are the implications of current progress in machine learning (ML) for how easy improvements will be at the level of HLMI. Some see developments like AlphaGo or GPT-3 as examples of (and evidence for) the claim that marginal intelligence improvements are not increasingly difficult, though others disagree with those conclusions.

For the first of the ways that we could conclude that current ML exhibits easy marginal intelligence improvements, see this argument [11] (Figure 15):

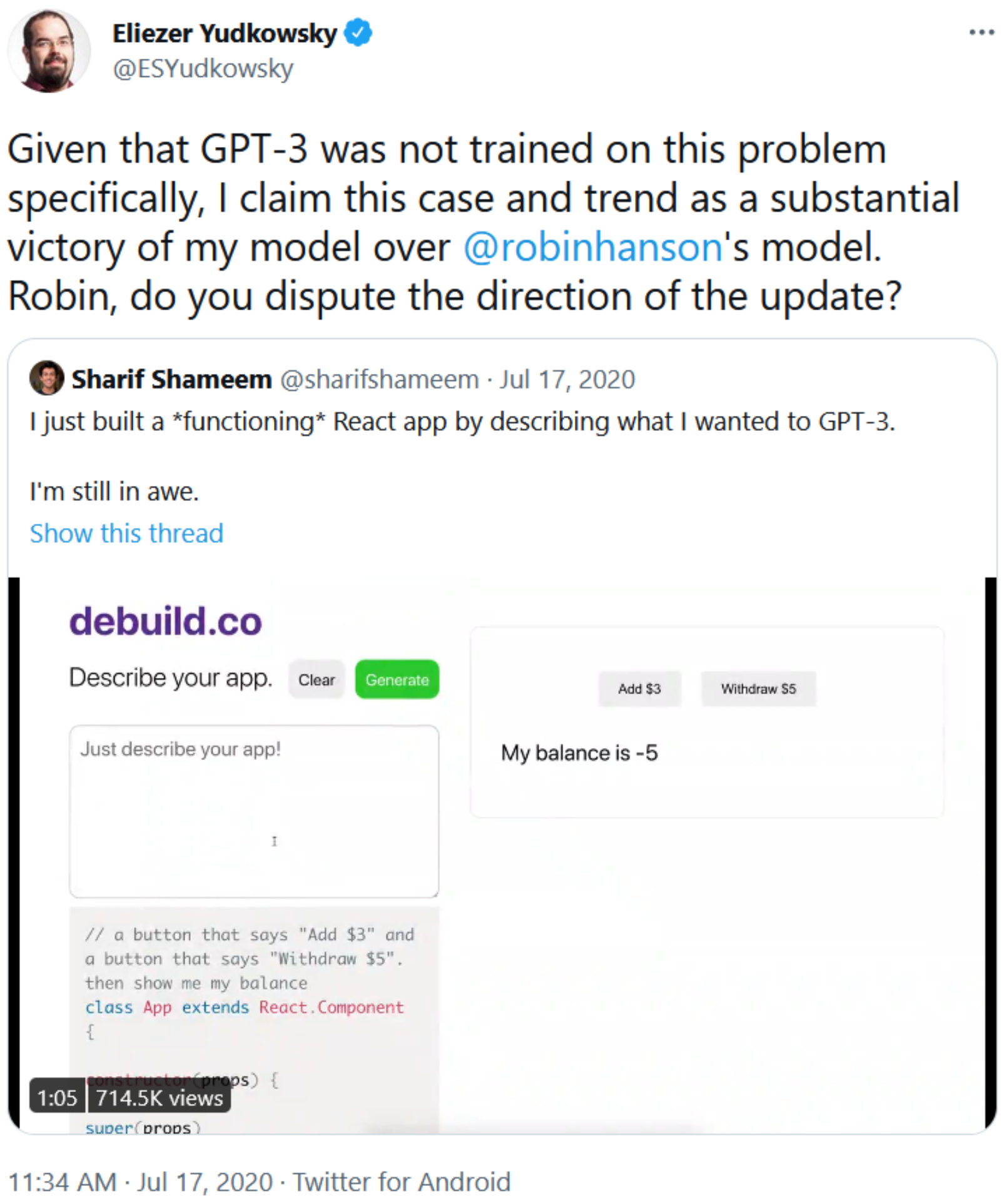

*Figure 15: Eliezer Yudkowsky tweet on GPT-3 progress*

Which we understand to mean,

> GPT-3 is general enough that it can write a functioning app given a short prompt, despite the fact that it is a relatively unstructured transformer model with no explicitly coded representations for app-writing. The fact that GPT-3 is this capable suggests that ML models scale in capability and generality very rapidly with increases in computing power or minor algorithm improvements...

In order to understand whether claims like Yudkowsky's are true, we must understand the specific nature of the breakthroughs made by cutting-edge ML systems like GPT-3 or AlphaZero and the limits of what these systems can do (left node in Figure 14).

Claims about current ML systems are related to a broader, more qualitative claim that the general success of ML models indicates that the fundamental algorithms needed for general intelligence are less complex than we might think (right node in Figure 14). Specific examples of humanlike thinking or reasoning in neural networks, for example OpenAI's discovery of multimodal neurons [12], lend some support to this claim.

Alternatively, Robin Hanson claims [13] that if machine learning was developing in sudden leaps, we would expect to see a pattern of citations in ML research where a few breakthrough papers received a very disproportionately large amount of citations. If Hanson is right about this, and if in reality citations aren't distributed in an unusually concentrated pattern in ML compared to other fields, then we have reason to expect marginal intelligence improvements from ML are hard (middle node in Figure 14).

Sources:
Hanson-Yudkowsky AI Foom Debate [9]
Searching for Bayes-Structure [14]
Will AI undergo discontinuous progress? [15]
Conceptual issues in AI safety: the paradigmatic gap [16]
The Scaling Hypothesis [17]
Eliezer Yudkowsky on AlphaGo's Wins [18]
GPT-2 As Step Toward General Intelligence [19]
GPT-3, Bloviator: OpenAI's language generator has no idea what it's talking about [20]
GPT-3: a disappointing paper [21]

### 2.2.3 Human Culture and History

Another source of evidence is human history and cultural evolution. During the Hanson-Yudkowsky debate [9], Eliezer Yudkowsky argued that the scientific method is an organizational and methodological insight that suddenly allowed humans to have much greater control over nature, and that this is evidence that marginal intelligence improvements are easy and that AI systems will similarly have breakthroughs where their capabilities suddenly increase (left-hand node in Figure 16).

In Intelligence Explosion Microeconomics [6], Yudkowsky also identified specific limitations of human science that wouldn't limit AI systems (middle node in Figure 16). For example, human researchers need to communicate using slow and imprecise human language. Wei Dai has similarly argued [22] that AI systems have greater economies of scale than human organizations because they do not hit a Coasean ceiling [23]. We might expect that a lot of human intelligence is "wasted," as organizations containing humans are not proportionately more intelligent than their members, due to communication limits that won't exist in HLMI (right-hand node). If these claims are right, simple organizational insights radically improved humanity's practical abilities, but we still face many organizational limitations an AI would not have to deal with. This suggests marginal improvements in practical abilities could be easy for an AI. On the other hand, even early, relatively unintelligent AIs don't face human limitations such as the inefficiency of interpersonal communication. This means AI might have already baked in whatever gains can be achieved from methodological improvements before reaching HLMI.

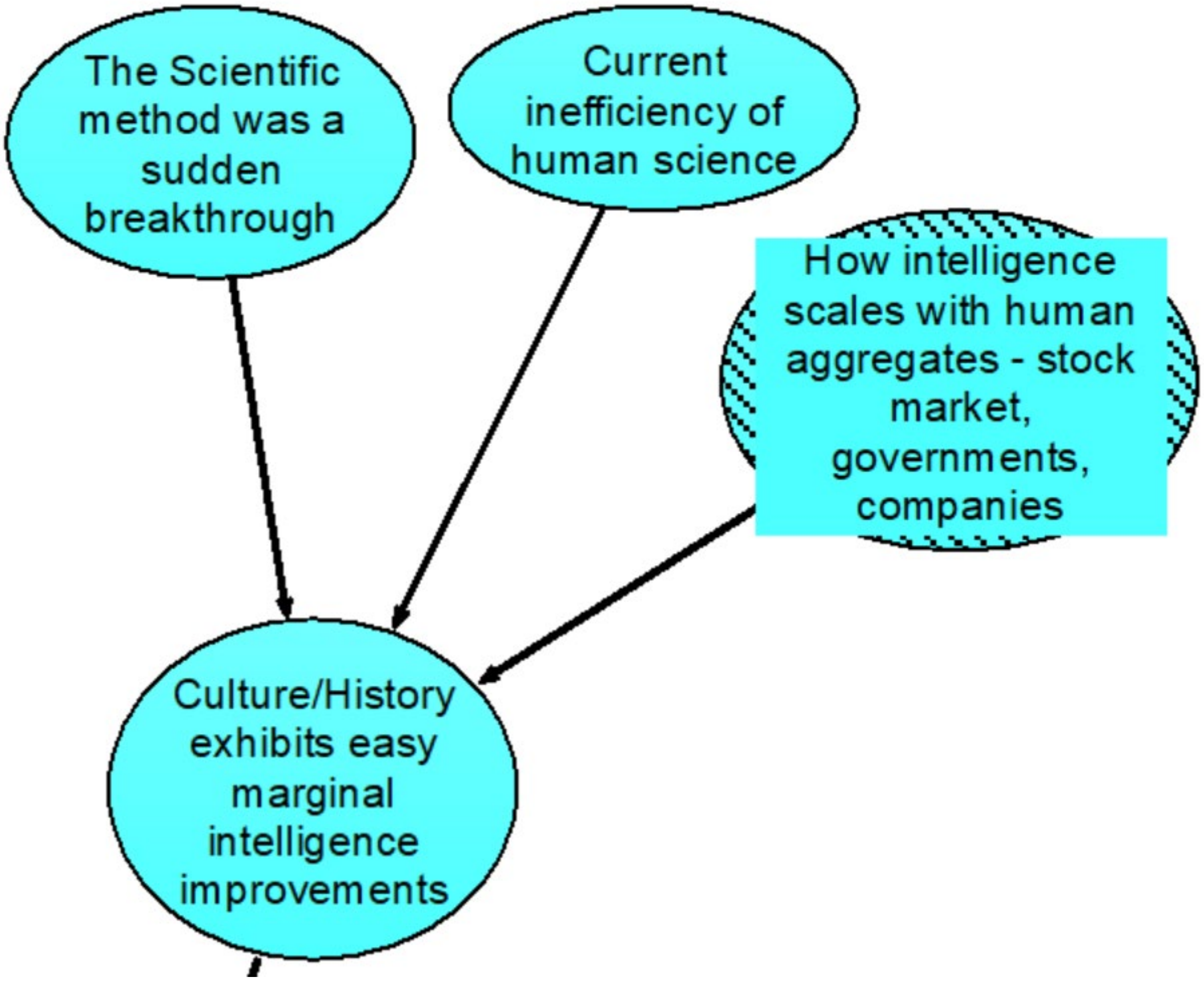

*Figure 16: Human culture and history cruxes*

Sources:
Hanson-Yudkowsky AI Foom Debate (search “you look at human civilization and there’s this core trick called science”) [9]
Debating Yudkowsky (point 5 responds to Eliezer) [24]
Intelligence Explosion Microeconomics (3.5) [6]
AGI and Economies of Scale [22]
Continuing the takeoffs debate [25]

## 2.2.4 History of AI

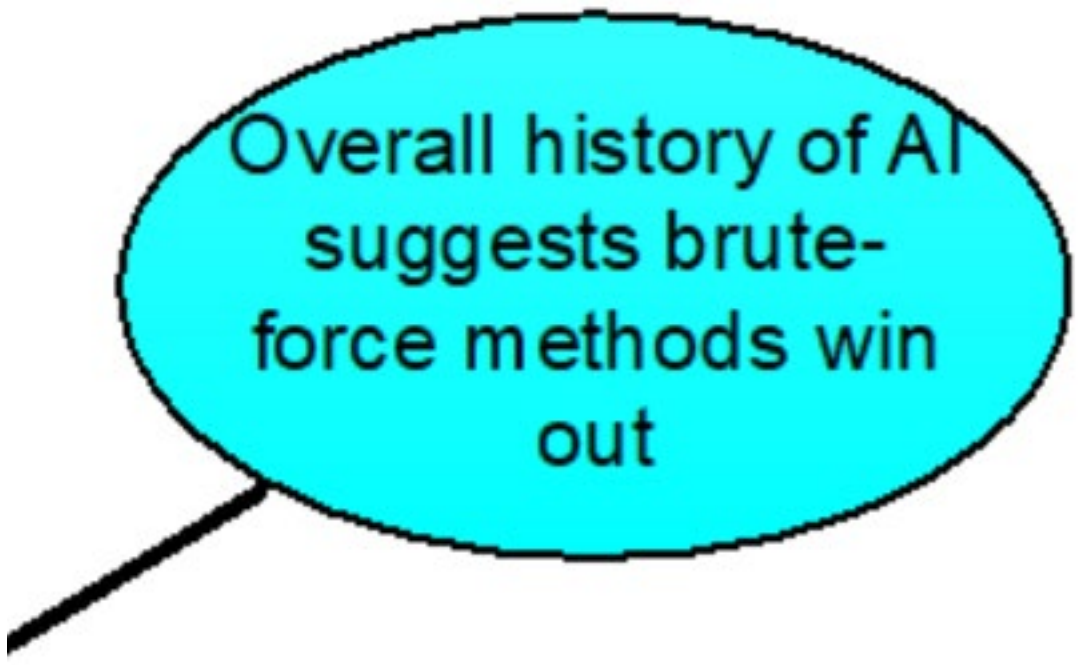

*Figure 17: History of AI crux*

This section covers reference-class forecasting based on the history of AI, going back before the current machine learning paradigm. Principally, what Richard Sutton referred to as the bitter lesson [26]:

> The biggest lesson that can be read from 70 years of AI research is that general methods that leverage computation are ultimately the most effective, and by a large margin... Seeking an improvement that makes a difference in the shorter term, researchers seek to leverage their human knowledge of the domain, but the only thing that matters in the long run is the leveraging of computation.

The bitter lesson is much broader than machine learning—it also includes, for example, the success of deep (relatively brute-force) search methods in computer chess. If the bitter lesson is true in general, it would suggest that we can get significant capability gains from scaling up models. That in turn should lead us to update towards marginal intelligence improvements being easier. The conjecture is that we can continually scale up compute and data to get smooth improvements in search and learning. If this is true, then plausibly scaling up compute and data will also produce smooth increases in general intelligence.

Sources:
Bitter lesson [26]

## 2.2.5 The Human Brain

Biological details of the human brain provide another source of intuitions about the difficulty of marginal improvements in intelligence. Eliezer Yudkowsky has argued that [9] (search for "750 megabytes of DNA" on that page) the small size of the human genome suggests that brain design is simple (this is similar to the Genome Anchor [27] hypothesis in Ajeya Cotra's AI timelines report [28], where the number of parameters in the machine learning model is anchored to the number of bytes in the human genome) (left-hand node in Figure 18).

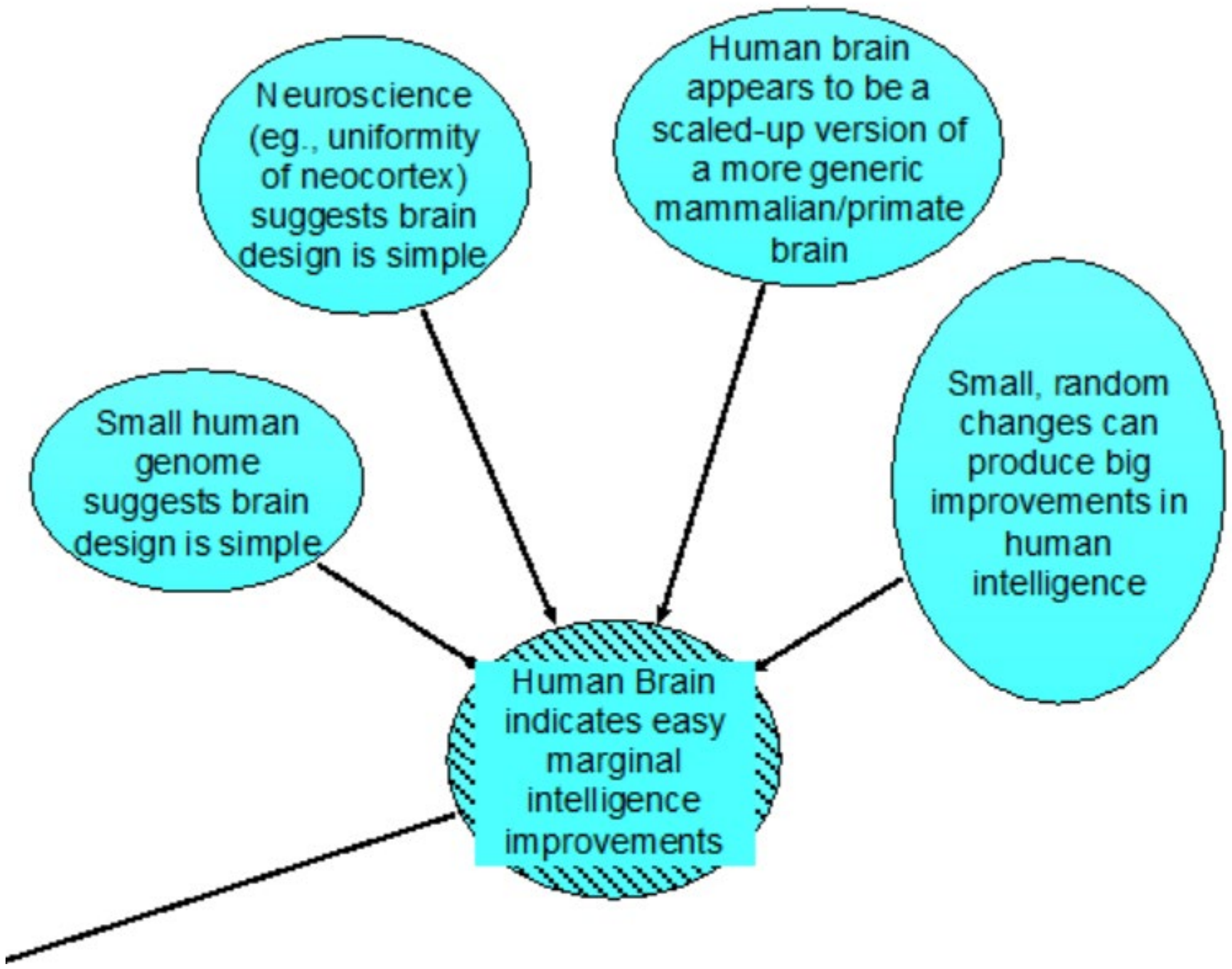

*Figure 18: Human brain cruxes*

If the human neocortex is uniform in architecture [29], or if cortical neuron count strongly predicts general intelligence, this also suggests there is a simple "basic algorithm" for intelligence (possibly analogous to a common algorithm [30] already used in ML). The fact that the neocortex can be divided into different brain regions with different functions and that the locations of these different regions are conserved across individuals is evidence against such a simple, uniform algorithm, but on the other hand, the ability of neurons in certain regions to be recruited by other regions (e.g., in ferrets that have had retinal projections redirected to the auditory thalamus [31], or in blind humans that can learn to echolocate via mouth clicks and apparently using brain regions typically devoted to vision [32]) is an argument in favor. If the human brain employs a simple algorithm, then we should think it more likely that there are other algorithms that can be rapidly scaled to produce highly intelligent behavior. These all fall under the evidence from neuroscience node (middle left node in Figure 18).

Additionally, it has been argued that the human brain is qualitatively very similar to the brains of other mammals or primates, and that the only major difference is that the human brain has been scaled-up [33] from a more generic mammal or primate brain (where scale is largely a function of cortical neuron count, but can also include other factors relevant for information-processing capacity [34], such as neuron packing density and axonal conduction velocity). If this is true, then it is a reason to suspect that intelligence improvements do not get particularly difficult around the human level, as evolving human intelligence did not depend on evolving new, intricate brain parts, and instead was achieved primarily through evolution turning a knob on brain scale. On the other hand, some have argued that current neuroscience techniques are inadequate to determine the hierarchy of information processing in brains [35], and thus the superficial physical resemblance of human brains to nonhuman primate/mammal brains does not tell us much about the algorithmic similarity between them. These arguments relate to the middle-right node in Figure 18.

Variation in scientific and other intellectual ability among humans (compare von Neumann to an average human) who share the same basic brain design also suggests improvements are easy at the HLMI level. Similarly, the fact that mood or certain drugs (stimulants [36] and psychedelics [37]) can sometimes improve human cognitive performance implies that humans aren't at a relative maximum; if we were, simple blunt changes to our cognition should almost always degrade performance (and rare reports of people gaining cognitive abilities after brain damage [38] provide a potentially even more extreme version of this argument). All of these provide evidence for the claim that small, random changes can produce big improvements in human intelligence (right-hand node in Figure 18).

Sources:
Human genome [39]
Hanson-Yudkowsky Debate [9]
Source code size vs learned model size in ML and in humans? [40]
Investigation into the relationship between neuron count and intelligence across differing cortical architectures [41]
Neurons And Intelligence: A Birdbrained Perspective [42]
Jeff Hawkins on neuromorphic AGI within 20 years [43]

## 2.2.6 General Priors

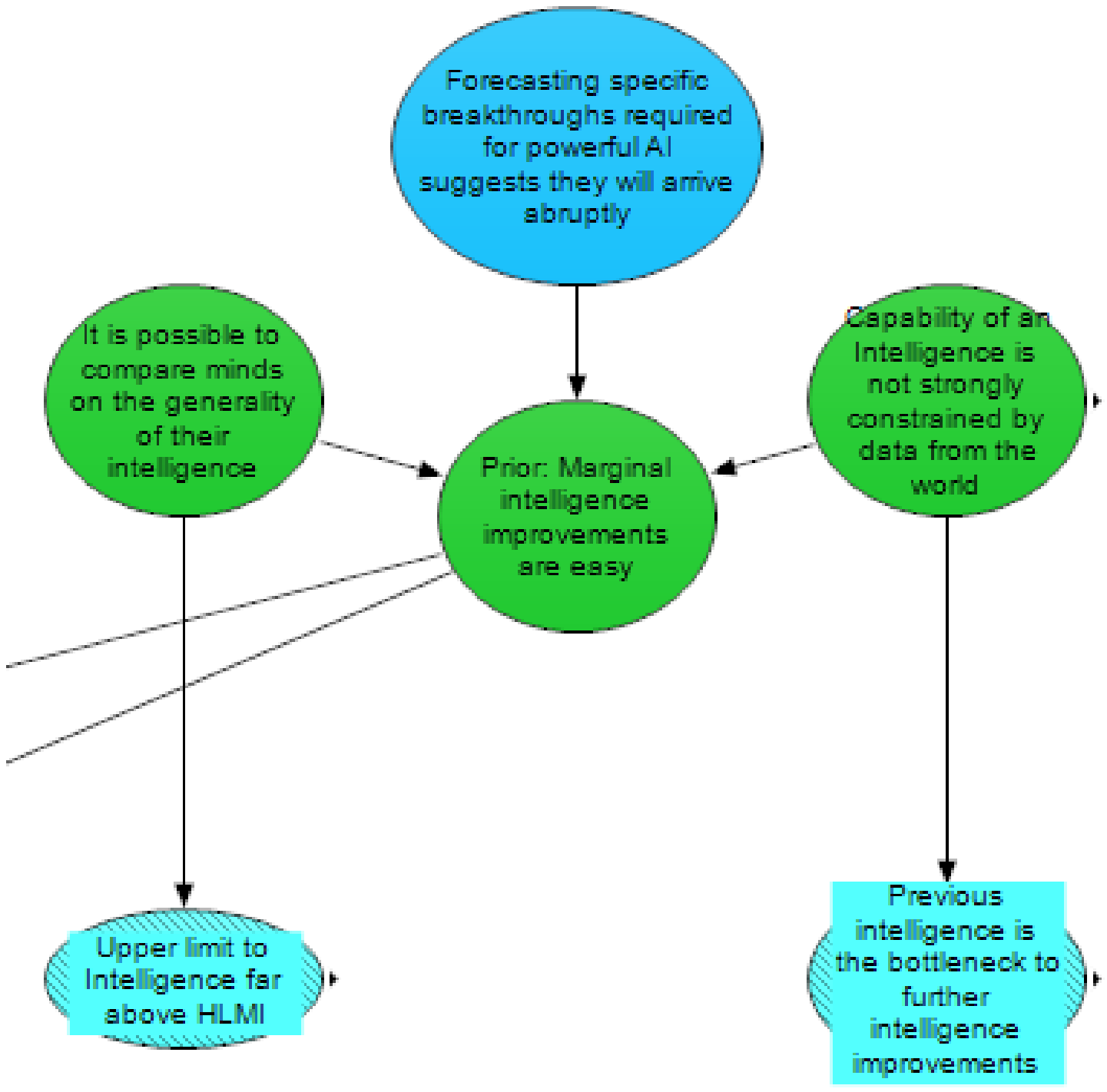

*Figure 19: General priors*

One's beliefs about the possibility of an intelligence explosion are likely influenced to a large degree by general priors about the nature of intelligence: whether general intelligence [44] is a coherent concept, whether nonhuman animal species can be compared by level of general intelligence, and so on. Claims like "intelligence isn't one thing that can just be scaled up or down—it consists of a bunch of different modules that are put together in the right way," imply that it is not useful to compare minds on the basis of general intelligence. For instance, François Chollet [45] denies the possibility of an intelligence explosion in part based on these considerations. As well as affecting the difficulty of marginal intelligence improvements, general priors (including the possibilities that *previous intelligence is a bottleneck to future improvements* and that there exists an *upper limit to intelligence*) also matter because they are potential defeaters of a fast progress/intelligence explosion scenario.

Sources:
The implausibility of intelligence explosion [45]
A reply to Francois Chollet on intelligence explosion [46]
General intelligence [44]
General Priors

## 2.3 The Outputs

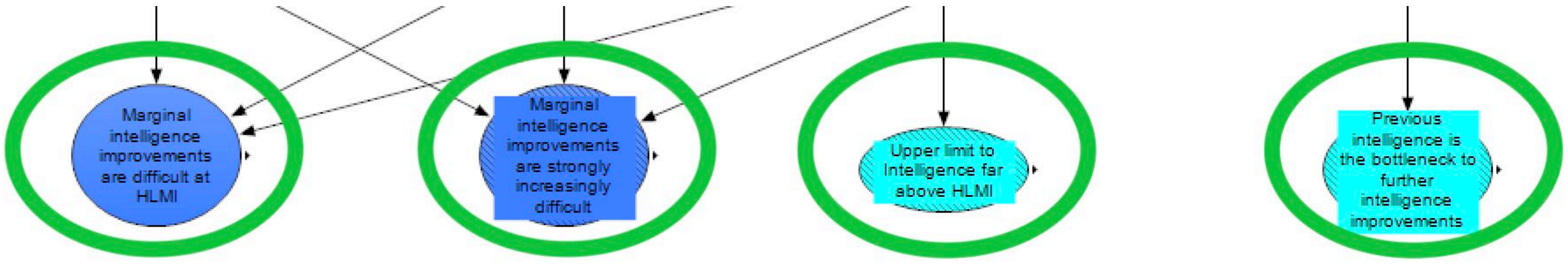

*Figure 20: Outputs of the Analogies and Priors module*

The empirical cruxes mentioned above influence the outputs of this module, which further influence downstream nodes in other modules. The cruxes this module outputs are

- marginal intelligence improvements are difficult at HLMI,
- marginal intelligence improvements are strongly increasingly difficult,
- the upper limit to intelligence is far above HLMI, and
- previous intelligence is a bottleneck to future intelligence improvements.

These cruxes influence probabilities the model places on different paths to HLMI and takeoff scenarios.

For the two nodes related to the difficulty of marginal intelligence improvements, we implement a naïve Bayes classifier in the Analytica model (i.e., a probabilistic classifier that applies Bayes' theorem with strong independence assumptions between the features). Each claim about one of the domains analogous to HLMI (e.g., in the domain of the human brain, the claim that the neocortex is uniform) is more likely to be true in a world where marginal intelligence improvements *in general* are either easy or hard. Therefore, when taken together, the analogy areas provide evidence about the difficulty of marginal intelligence improvements for HLMI.

This use of a naïve Bayes classifier enables us to separate out the prior likelihoods of the original propositions, such as that the human neocortex is uniform, and these propositions' relevance to HLMI (the likelihood of being true in a world with easy vs hard marginal intelligence improvements).

This extra step of using Bayes classification is useful because the claims about the analogy domains themselves are often much more certain than their degree of relevance to claims about the ease of improvements to HLMI. Whether there was a rapid acceleration in intelligence during hominin evolution is something that can be assessed by domain experts or a review of the relevant literature, but the relevance of this claim to HLMI is a separate question we are much less certain about. Using the naïve Bayes classifier allows us to separate these two factors out.

### 2.3.1 Difficulty of Marginal Intelligence Improvements

How difficult marginal intelligence improvements are at the level of HLMI and how rapidly they become more difficult at higher levels of intelligence are some of the most significant cruxes of disagreement among those trying to predict the future of AI. If marginal intelligence improvements are difficult around HLMI, then HLMI is unlikely to arrive in a sudden burst. If marginal intelligence improvements rapidly become increasingly difficult beyond HLMI, it is unlikely there will be an intelligence explosion post-HLMI.

For an example [6] from Yudkowsky of discussion relating to whether marginal intelligence improvements are "strongly increasingly difficult":

> The Open Problem posed here is the quantitative issue: whether it's possible to get sustained returns on reinvesting cognitive improvements into further improving cognition.

("Strongly increasingly difficult" has often been operationalized as "exponentially increasingly difficult," [47] as in that way it serves as a defeater for the "sustained returns" we might expect from powerful AI accelerating the development of AI.)

Paul Christiano has also claimed [8] models of progress which include "key insights [that]… fall into place" are implausible, which we model as a claim that marginal intelligence improvements at HLMI are *difficult* (since if they are difficult, a few key insights will not be enough). For a direct example of a claim that the difficulty of marginal intelligence improvements is a key crux, see this quote by Robin Hanson [48]:

> So I suspect this all comes down to, how powerful is architecture in AI, and how many architectural insights can be found how quickly? If there were say a series of twenty deep powerful insights, each of which made a system twice as effective, just enough extra oomph to let the project and system find the next insight, it would add up to a factor of a million. Which would still be nowhere near enough, so imagine a lot more of them, or lots more powerful.

In our research, we have found that what is most important in many AI takeoff models is their stance on whether marginal intelligence improvements are difficult at HLMI and if they become much more difficult beyond HLMI. Some models, for instance, claim there is a secret sauce for intelligence such that once it is discovered, progress becomes easy. In the Takeoff Speeds and Discontinuities section (§4) of the model, we more closely examine the relationships between claims about the ease of marginal intelligence improvements and progress around HLMI.

Why did we attempt to identify an underlying key crux in this way? It is clear that beliefs about AI takeoff relate to beliefs from arguments by analogy (in this chapter) in fundamental ways, for example:

Paul Christiano et al. [8]: Human evolution is not an example of massive capability gain given constant optimization for intelligence + *(other factors)* ➔ [Implicit Belief A about nature of AI Progress] ➔ Continuous Change model of Intelligence Explosion, likely not highly localized

Eliezer Yudkowsky et al. [6]: Human evolution is an example of massive capability gain given constant optimization for intelligence + *(other factors)* ➔ [Implicit Belief B about nature of AI Progress] ➔ Discontinuous Change model of Intelligence Explosion, likely highly localized

We have treated implicit beliefs A and B as being about the difficulty of marginal intelligence improvements at HLMI and beyond.

We have identified “previous intelligence is a bottleneck” and “there is an upper limit to intelligence” as two other important cruxes. They are commonly cited as defeaters for scenarios that involve any kind of explosive growth due to an acceleration in AI progress. Both of these appear to be cruxes between skeptics and non-skeptics of HLMI and AI takeoff.

### 2.3.2 Previous Intelligence is a bottleneck

The third major output of this module is whether, in general, further improvements in intelligence tend to be bottlenecked by the current intelligence of our systems rather than some external factor (such as the need to run experiments and wait for real-world data).

This output is later used in the intelligence explosion module (§4.1.2) : if such an external bottleneck exists, we are unlikely to see rapid acceleration of technological progress through powerful AI improving our ability to build yet more powerful AI. There will instead be drag factors preventing successor AIs from being produced without reference to the outside world. This view is summarized by François Chollet [45]:

> If intelligence is fundamentally linked to specific sensorimotor modalities, a specific environment, a specific upbringing, and a specific problem to solve, *then you cannot hope to arbitrarily increase the intelligence of an agent merely by tuning its brain—no more than you can increase the throughput of a factory line by speeding up the conveyor belt*. Intelligence expansion can only come from a co-evolution of the mind, its sensorimotor modalities, and its environment.

In short, this position claims that positive feedback loops between successive generations of HLMI could not simply be closed, but instead would require feedback from environmental interaction that can’t be arbitrarily sped up. While Chollet appears to be assuming such bottlenecks will occur due to necessary interactions with the physical world, it is possible in principle that such bottlenecks could occur in the digital world—for instance, brain emulations that were sped up by 1M times would find all computers (and thus communications, access to information, calculators, simulations, etc.) slowed down by 1M times (from their perspective), potentially creating a drag on progress.

A somewhat more tentative version of this claim is that improvements in intelligence (construed here as "ability to do applied science") require a very diverse range of discrete skills and access to real-world resources. Ben Garfinkel [49] makes this point:

> ...there's really a lot of different tasks that feed into making progress in engineering or areas of applied science. There's not really just this one task of "Do science." Let's take, for example, the production of computing hardware… I assume that there's a huge amount of different works in terms of how you're designing these factories or building them, how you're making decisions about the design of the chips. Lots of things about where you're getting your resources from. Actually physically building the factories.

### 2.3.3 Upper Limit to Intelligence

Finally, this module has an output for whether there is a practical upper limit to intelligence that is not significantly above the human level. This could be true if there are physical barriers to the development of greater intelligence. Alternatively, it seems more likely to be effectively true if we cannot even compare the abilities of different minds along a general metric of "intelligence" (hence a "no" answer to "it is possible to compare minds on the generality of their intelligence" leads to a "yes" to this crux). For the "it is not possible to compare minds on the generality of their intelligence" claim, from François Chollet [45]:

> The first issue I see with the intelligence explosion theory is a failure to recognize that intelligence is necessarily part of a broader system—a vision of intelligence as a "brain in jar" that can be made arbitrarily intelligent independently of its situation.

Chollet argues that human (and all animal) intelligence is "hyper-specialized" and situational to a degree that makes comparisons of general intelligence much less useful than they first appear.

> If intelligence is a problem-solving algorithm, then it can only be understood with respect to a *specific* problem. In a more concrete way, we can observe this empirically in that all intelligent systems we know are highly specialized. ... The brain has hardcoded conceptions of having a body with hands that can grab, a mouth that can suck, eyes mounted on a moving head that can be used to visually follow objects (the vestibulo-ocular reflex [50]), and these preconceptions are required for human intelligence to start taking control of the human body. It has even been convincingly argued, for instance by Chomsky [51], that very high-level human cognitive features, such as our ability to develop language, are innate.

A strong version of the modularity of mind hypothesis, "massive modularity," also implies that intelligence is hyper-specialized at extremely specific tasks [52]. If true, massive modularity would lend support to the claim that "intelligence… can only be understood with respect to a specific problem," which in turn suggests that intelligence cannot be increased independent of its situation.

## 2.4 Conclusion

This chapter has explained the structure and reasoning behind one of the starting points of the MTAIR model, *Analogies and General Priors*. This module connects conclusions about the nature of HLMI to very basic assumptions about the nature of intelligence and analogies to domains other than HLMI about which we have greater experience.

We have made the simplifying assumption to group the conclusions drawn from the general priors and the different analogy domains into four outputs, which we think characterize the important variables needed to predict HLMI development and post-HLMI takeoff. In later chapters, we will explain how those outputs are used to make predictions about HLMI takeoff and development.

The next chapter, Paths to High-Level Machine Intelligence, attempts to forecast when HLMI will be developed and by what route.

# 3 Paths To High-Level Machine Intelligence

Daniel Eth

This chapter explains parts of our model most relevant to paths to high-level machine intelligence (HLMI). We define HLMI as machines that are capable, either individually or collectively, of performing almost all economically relevant information-processing tasks that are performed by humans, or quickly (relative to humans) learning to perform such tasks. Since many corresponding jobs (such as managers, scientists, and startup founders) require navigating the complex and unpredictable worlds of physical and social interactions, the term HLMI implies very broad cognitive capabilities, including an ability to learn and apply domain-specific knowledge and social abilities.

We are using the term “high-level machine intelligence” here instead of the related terms “human-level machine intelligence,” “artificial general intelligence,” or “transformative AI” because these other terms are often seen as baking in assumptions about either the nature of intelligence or advanced AI that are not universally accepted.

In relation to the model as a whole, this chapter focuses on the modules circled in red in Figure 21:

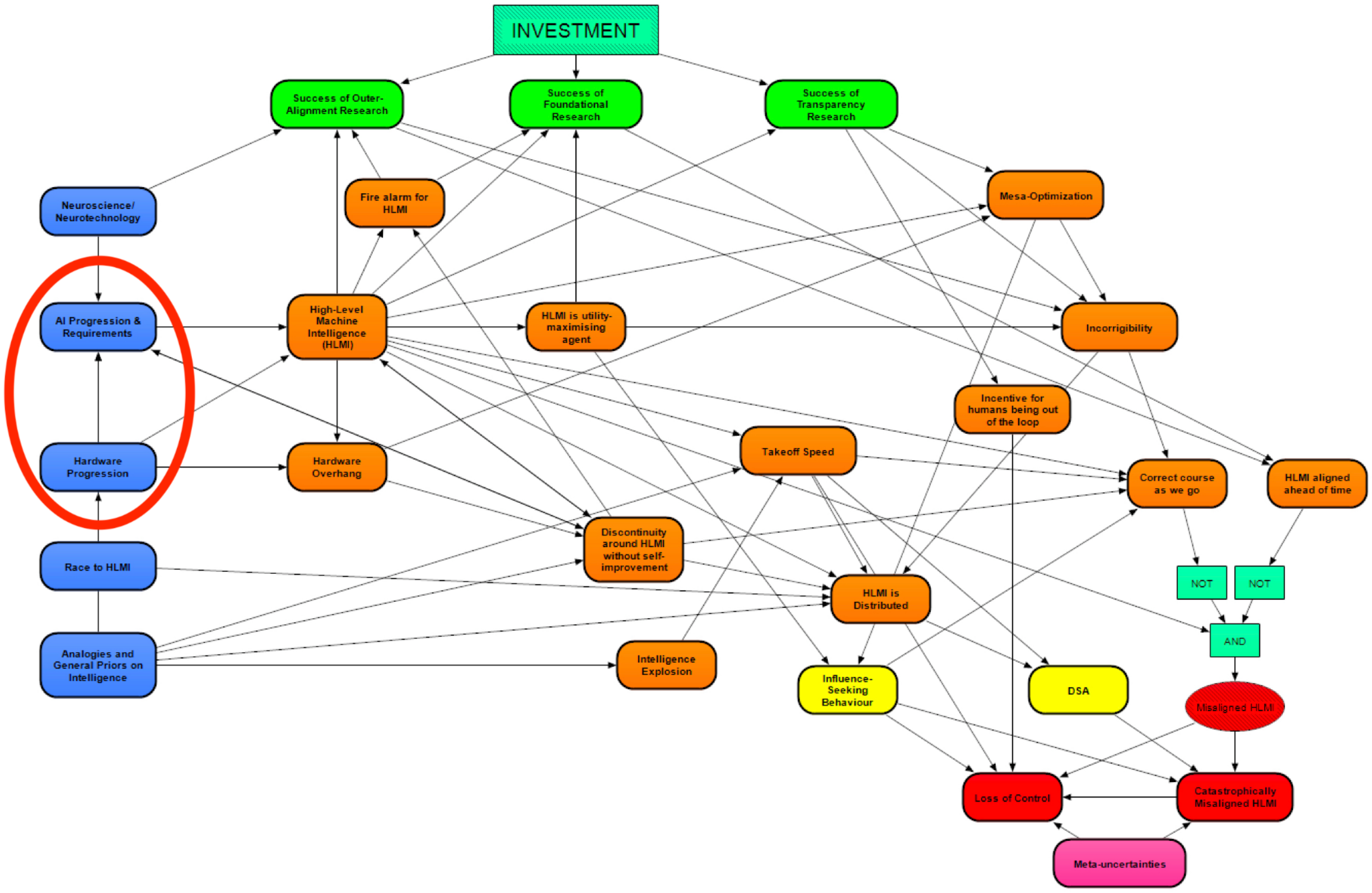

*Figure 21: Model focus for the Paths to HLMI modules*

The module *AI Progression & Requirements* (§3.2) investigates when we should expect HLMI to be developed, as well as what kind of HLMI we should expect (e.g., whole brain emulation [53] (§3.2.1.5), HLMI from current deep learning (DL) methods (§3.2.1.2), etc.). These questions about the timing and kind of HLMI are the main outputs from the sections in this chapter and influence downstream parts of the model. The timing question, for instance, determines how much time there is for safety agendas to be solved. The question regarding the kind of HLMI, meanwhile, affects many further cruxes, including which safety agendas are likely necessary to solve in order to avoid failure modes as well as the likelihood of HLMI being distributed versus concentrated.

The module *Hardware Progression* (§3.1) investigates how much hardware will likely be available for a potential project towards HLMI (as a function of time). This module provides input for the *AI Progression & Requirements* module (§3.2), which also receives significant input from the *Analogies and General Priors on Intelligence* module described in the previous chapter.

We will start our examination here with the *Hardware Progression* module, and then discuss the module for *AI Progression & Requirements*.

# 3.1 Hardware Progression

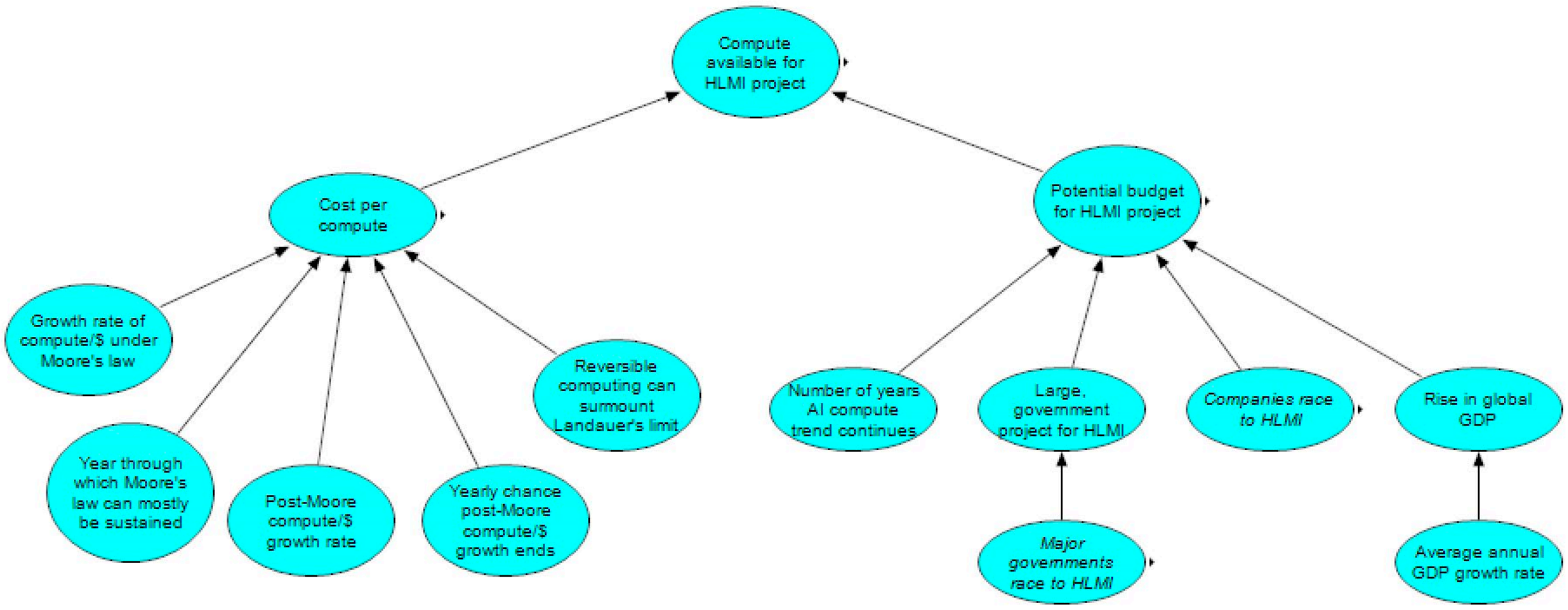

*Figure 22: Hardware progression*

The output from this section is *compute available for an HLMI project*, which varies by year and is determined by dividing the *potential budget for an HLMI project* by the *cost per compute* (both as a function of the year).

## 3.1.1 Cost per Compute

Zooming in on the cost portion, we see the following subgraph (Figure 23):

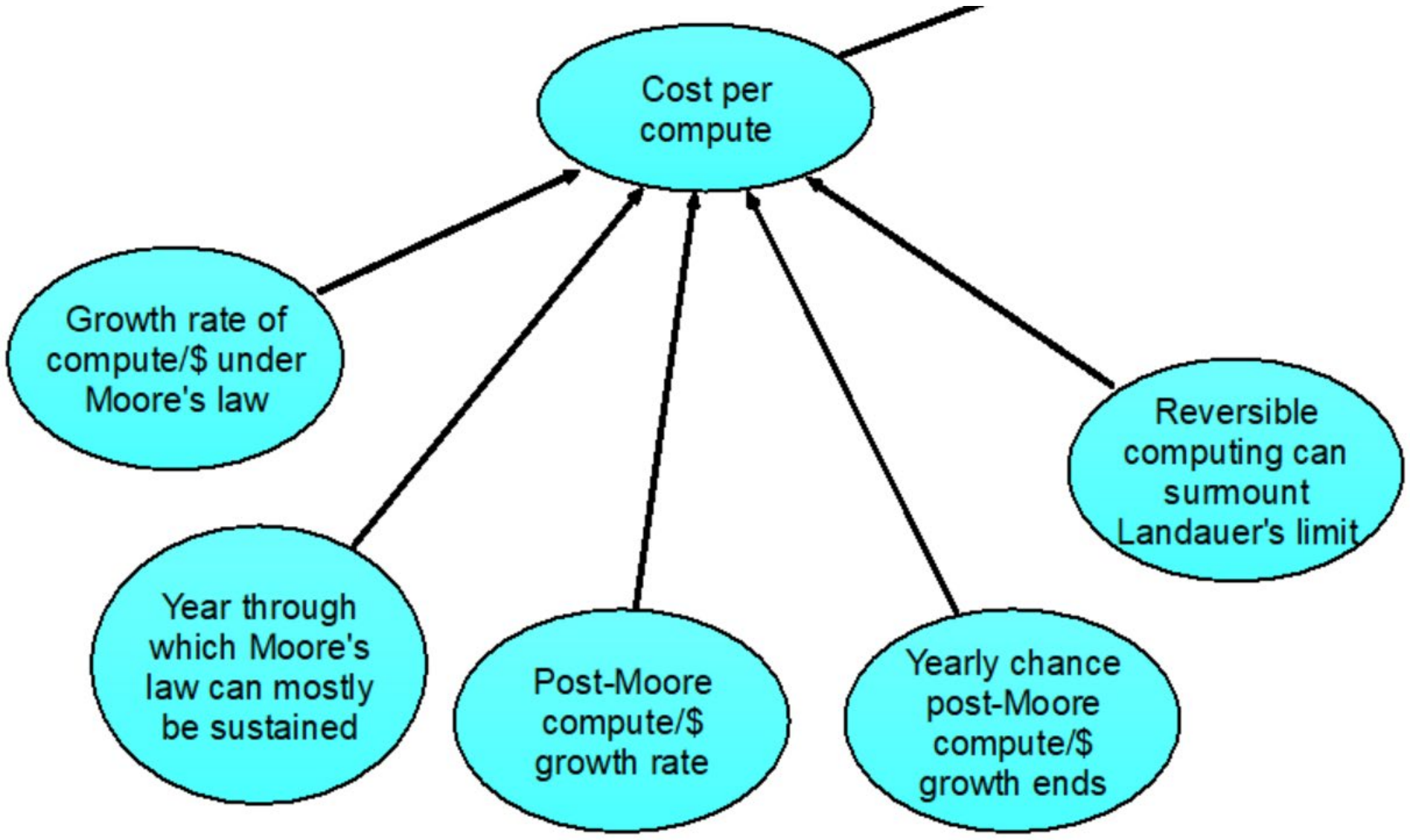

*Figure 23: Cost per compute*

The *cost per compute* (over time) is the output of this subgraph and is determined by the other listed nodes. Starting from the current cost of compute, the compute/$ is expected to continue rising until the trend of increasing transistor density on 2D Si chips (i.e., Moore's law) runs out of steam. Note that we're using "Moore's law" in the colloquial sense to refer to approximately exponential growth within the 2D Si chip paradigm, not the more specific claim of a 1–2-year doubling time for transistor count. Also note that, at this stage, we do not differentiate between CPU, GPU, or ASIC compute, as similar trends apply to all of them. Both the *growth rate of compute/$ under the future of Moore's law* and the *year through which Moore's law can mostly be sustained* are uncertainties within our model [54].

After this paradigm ends, compute/$ may increase over time at a new (uncertain) rate (*post-Moore compute/$ growth rate*). Such a trend could perhaps be sustained by a variety of mechanisms: new hardware paradigm(s) (such as 3D Si chips, optical computing, or spintronics), specialized hardware (leading to more effective compute for the algorithms of interest), a physical-manufacturing revolution (such as via atomically precise manufacturing), or pre-HLMI AI-led hardware improvements. Among technology forecasters, there is large disagreement about the prospects for post-Moore computational growth, with some forecasters [55] predicting Moore-like or faster growth in compute/$ to continue post-Moore, while others [56] expect compute/$ to slow considerably or plateau.

Even if post-Moore growth is initially substantial, however, we would expect compute/$ to eventually run into some limit and plateau (or slow to a crawl), due to either hard technological limits or economic limits related to increasing R&D and fab costs [57]. The possibility of an eventual leveling off of compute/$ is handled by the model in two different ways. First, there is assumed to be an uncertain, *yearly chance post-Moore compute/$ growth ends*. Second, Landauer's limit [58] (the thermodynamic

limit relating bit erasure and energy use) may present an upper bound for compute/$, which the model assumes will happen unless *reversible computing can surmount Landauer's limit*.

It should be noted that the model does not consider that specialized hardware may present differential effects for different paths to HLMI, nor does it consider how quantum computing might affect progression towards HLMI, nor the possibility of different paradigms post-Moore seeing different compute/$ growth rates. Such effects may be important but appear complex and difficult to model.

### 3.1.2 Potential Budget for HLMI Project

The budget section, meanwhile, has this subgraph (Figure 24):

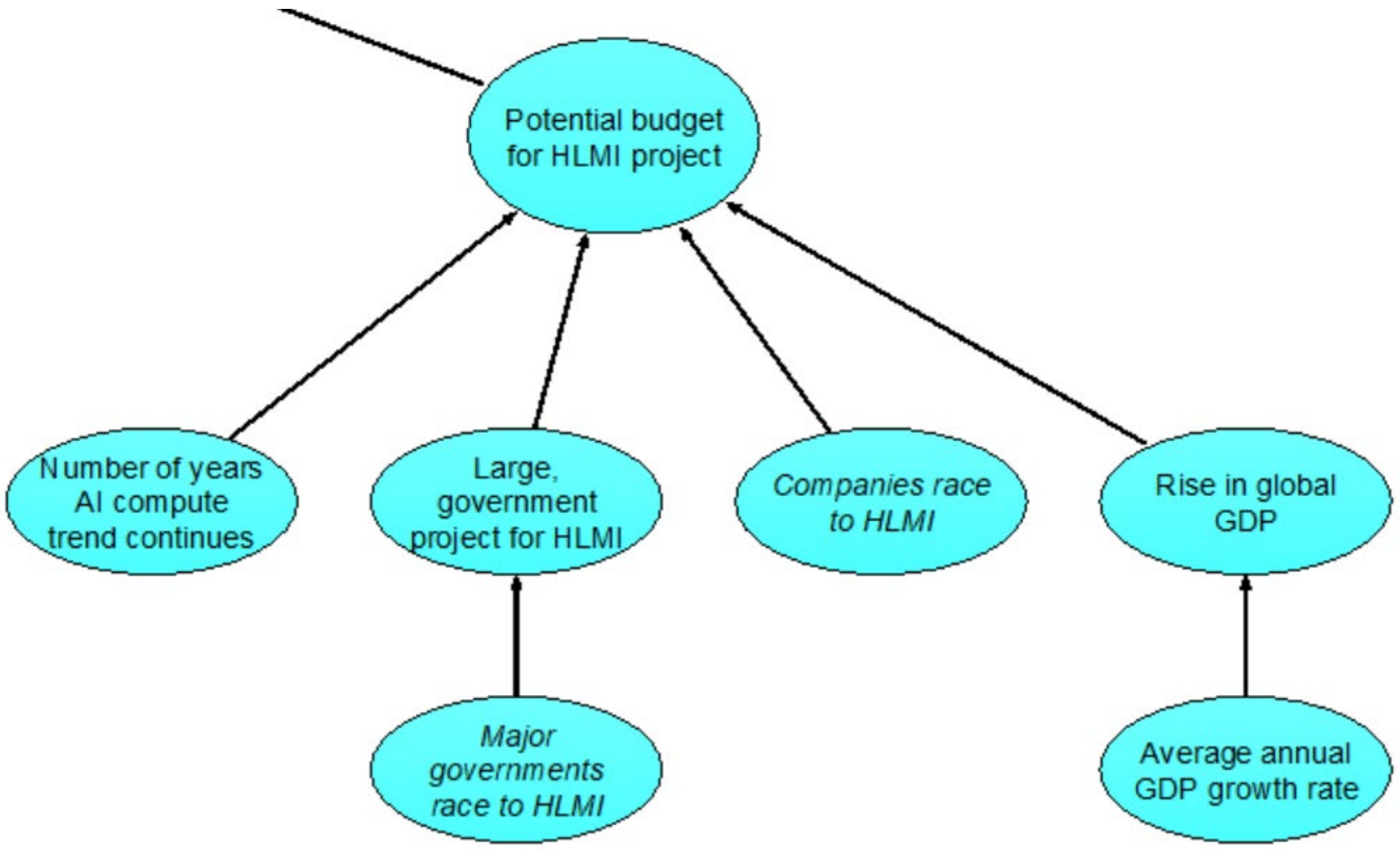

*Figure 24: Potential budget for HLMI*

The *potential budget for an HLMI project*, which is the primary output of this subgraph (used in combination with the output from the previous section on *cost per compute* to derive the compute available for an HLMI project) is modeled as follows. In 2021, the budget is set to that of the most expensive AI project to date. The recent, quickly rising *AI compute trend* [59] *may or may not* [60] *continue for some number of years*, with the modeled budget rising accordingly. After that, the budget for HLMI is currently modeled as generally rising with the *rise in global GDP* (determined by the *average annual GDP growth rate*), except if *companies race to HLMI* (in which case we assume the budget will grow to be similar, in proportion of global GDP, to tech firms' R&D budgets), or if there's a *large government project for HLMI* (in which case we assume budgets will grow to ~1% of the richest country's GDP, in line with [61] the Apollo program and Manhattan Project). We think such a large government project is particularly likely if *major governments race to HLMI*.

We should note that our model does not consider the possibility of budgets being cut between now and the realization of HLMI; while we think such a situation is plausible (especially if there are further AI winters), we expect such budgets would recover before reaching HLMI, and thus aren't clearly relevant for the potential budget leading up to HLMI. The possibility of future AI winters implying longer timelines through routes other than hardware (e.g., if current methods "don't scale" to HLMI and further paradigm shifts don't occur for a long time) is, at least implicitly, covered in the module on *AI Progression & Requirements* (§3.2).

As our model is acyclic, it doesn't handle feedback loops well. We think it is worth considering, among other possible feedback loops, how pre-HLMI AI may affect GDP or how hardware spending and costs may affect each other (such as through economies of scale & learning effects, as well as simple supply and demand), though these effects aren't currently captured in our model.

## 3.2 AI Progression & Requirements

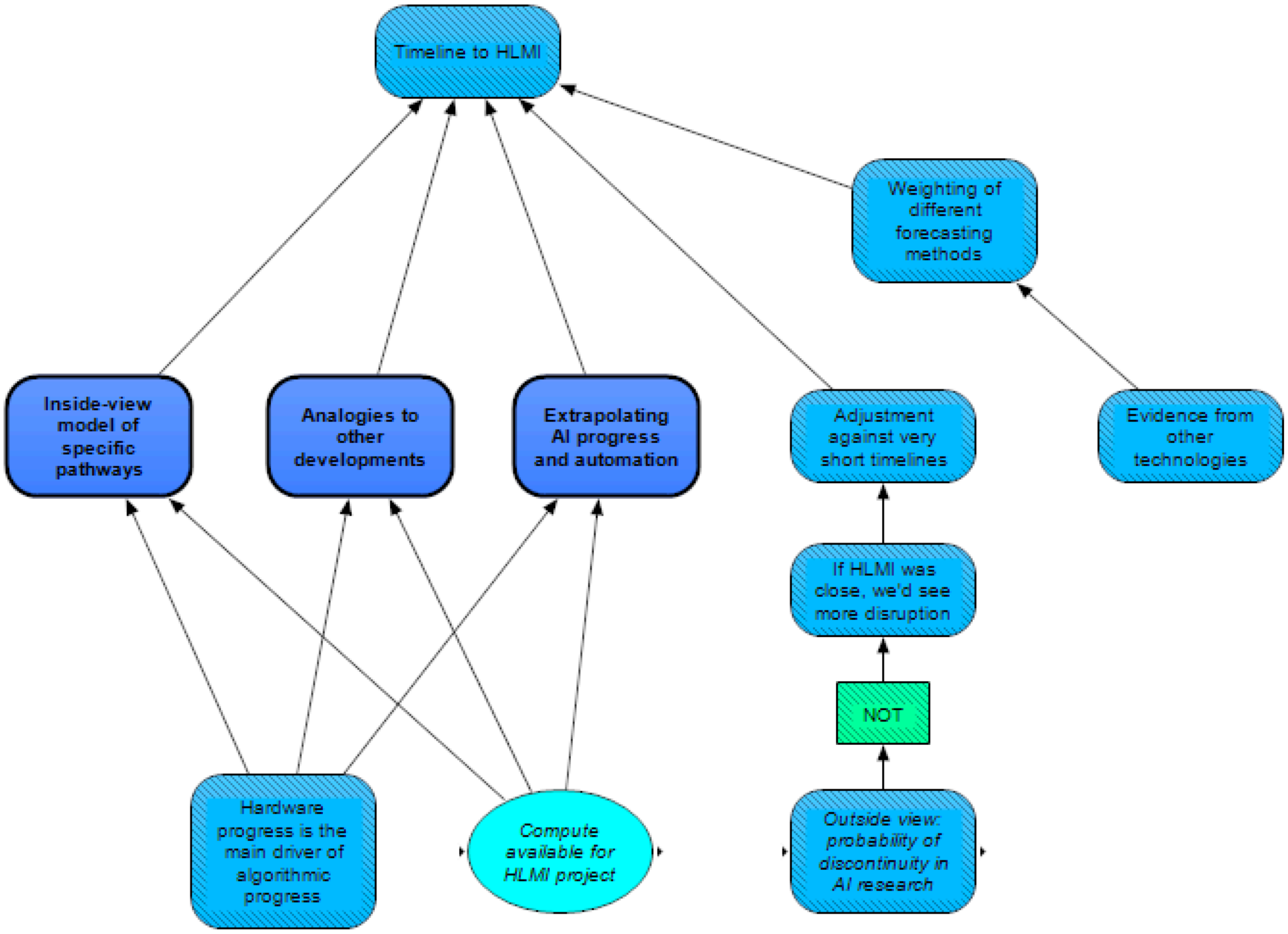

*Figure 25: AI Progression and Requirements and Path to HLMI*

The module on *AI Progression & Requirements* uses a few different methods for estimating the *timeline to HLMI*: a gears-level, *inside-view model of specific pathways* (§3.2.1) (which considers how various possible pathways to HLMI might succeed) as well as outside-view methods (*analogies to other developments* (§3.2.2.1) and *extrapolating AI progress and automation* (§3.2.2.2)). All of these methods are influenced by the *compute available for an HLMI project* (described in the section above), as well as whether or not [62] *hardware progress is the main driver of algorithmic progress*.

Estimates from these approaches are then combined, currently as a linear combination (*weighting of different methods of forecasting timelines*), with the weighting based on *evidence from other technologies* regarding which technological forecasting methods are most appropriate. A (possible) *adjustment against very short timelines* is also made, depending on whether one believes that *if HLMI was close, we'd see more disruption* from AI, which in turn is assumed to be less likely if there's a larger *probability of discontinuity in AI research* (since such a discontinuity could allow HLMI to "come out of nowhere").

### 3.2.1 Inside-View Model of Specific Pathways

This module represents the most intricate of our methods for estimating the arrival of HLMI. Here, several possible approaches to HLMI are modeled separately, and the date for the arrival of HLMI is taken as the earliest date that one of these approaches succeeds. Note that such an approach may introduce an optimizer's curse [63] (from the OR gate); if the different pathways have independent errors, then the earliest modeled pathway may be expected to have unusually large error in favor of being early, underestimating the actual timeline. On the other hand, the estimations of the different pathways themselves each contain the combination of different requirements (AND gates), such as requirements for adequate software and hardware, and this dynamic introduces the opposite bias—the last requirement to fall into place may be expected to have an error in favor of being late. Instead of correcting for these two biases, we simply note that they exist in opposite directions and we are uncertain about which is the larger bias.

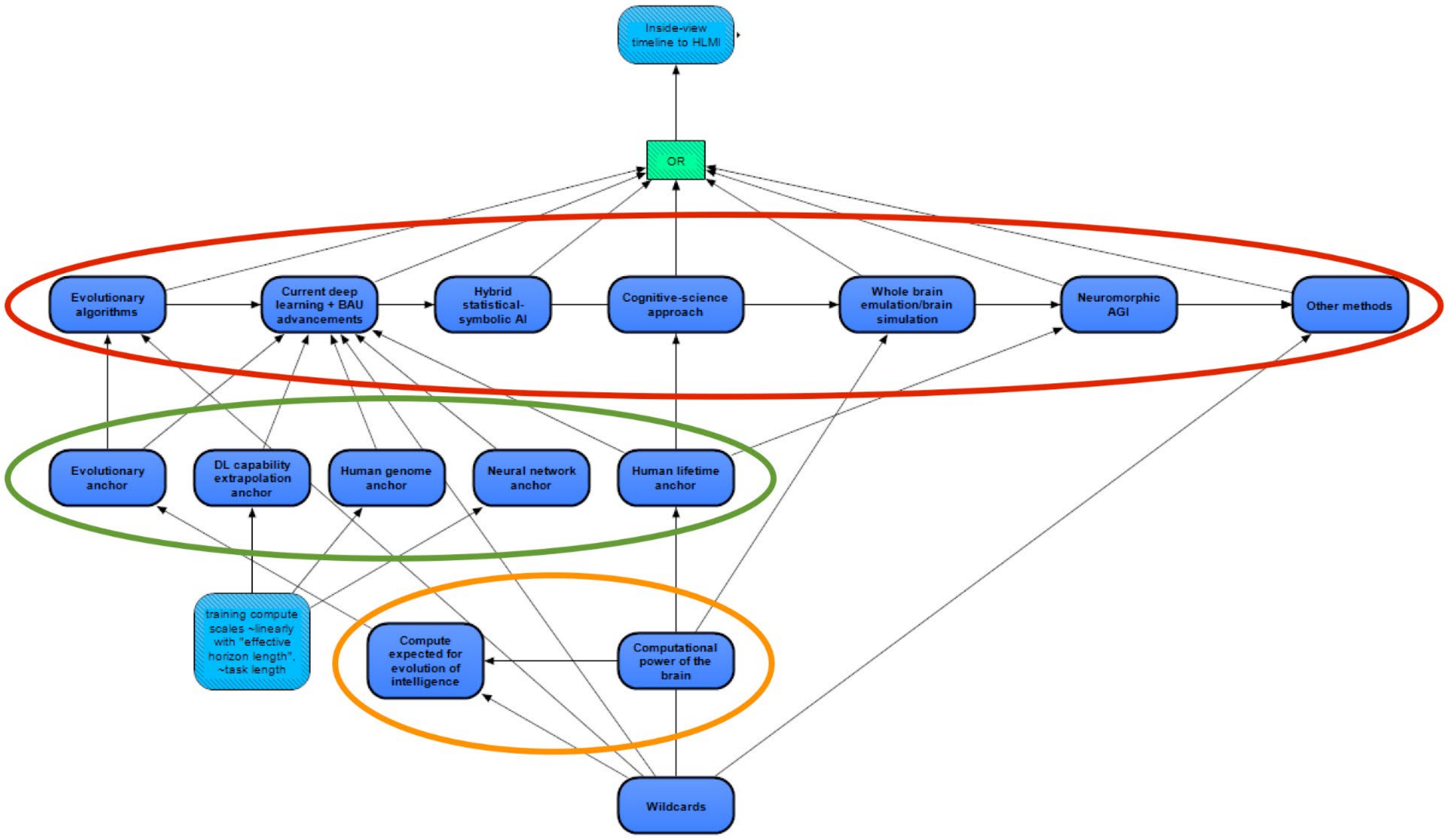

*Figure 26: Inside-view HLMI timeline*

The different methods to HLMI in our model (circled in red in Figure 26 and expanded upon in their individual subsections below) are:

- *Evolutionary algorithms* (§3.2.1.1)—either similar to evolutionary algorithms today, a direct simulation of virtual evolution, or some middle ground. While current evolutionary algorithms are very different from virtual evolution, we have grouped these together instead of grouping current evolutionary algorithms with current deep learning methods, largely because in our model, both of the evolutionary methods rely on similar hardware estimation methods.

- *Current deep learning plus business-as-usual advancements* (§3.2.1.2)—HLMI achieved through methods similar to those in deep learning today; the key features of such methods are large, deep neural nets, trained from a largely "blank-slate" initial state and optimized via local search techniques such as SGD (though crucially, to avoid double counting, not including evolutionary algorithms).

- *Hybrid statistical-symbolic AI* (§3.2.1.3)—approaches marrying statistical methods and more intentionally designed symbolic methods, such as in GOFAI.

- *Cognitive-science approach* (§3.2.1.4*)*—approaches to HLMI heavily relying on cognitive science or developmental psychology (likely in combination with deep learning).

- *Whole brain emulation / brain simulation* (§3.2.1.5)—emulating a particular person's brain *in silico* (for WBE) or simulating a generic human brain (for brain simulation).

- *Neuromorphic AGI* (§3.2.1.6)—HLMI created using many of the "low-level" processes or architectures of the brain but without being put together into a virtual brain with particularly humanlike intelligence.

- *Other methods* (§3.2.1.7)—a catch-all for unanticipated or other potential paths to HLMI.

Each method is generally assumed to be achieved once both the hardware and software requirements are met. The software requirements of each of the methods are largely dependent on different considerations, while the hardware requirements are more dependent on different weighting or modifications of the same set of "anchors" (circled in green in Figure 26).

These anchors are plausibly reasonable estimates for the computational difficulty of training an HLMI, based on different analogies or other reasoning that allow for anchoring our estimates. Four of the five anchors come from Ajeya Cotra's Draft report on AI timelines [28] (with minor modifications), and we add one more. These five anchors (elaborated upon below) are:

- *Evolutionary anchor*—an estimate based on the "compute-equivalent" that was "used" in evolution from the first animals with nervous systems to humans (considering both the "brain-compute" used and the compute necessary to simulate the environment sufficiently). This estimate includes both upward and downward adjustments for possible anthropic effects and for human engineers potentially outcompeting evolution.

- *Deep learning capability extrapolation anchor*—an estimate of how much compute would be needed to achieve broad, human-level capabilities given an extrapolation of how capabilities in current deep learning systems scale with compute.

- *Human genome anchor*—an estimate of the compute needed to train a current ML system sufficiently, based on setting the parameter count to the size of the human genome (arguably the "code" of human intelligence), used in combination with empirically derived scaling laws and other relevant biological considerations.

- *Neural network anchor*—an estimate of the compute needed to train a current ML system sufficiently, similar to the human genome anchor, but instead with the parameter count set by considerations related to the compute-equivalent used in the human brain.

- *Human lifetime anchor*—an estimate of how much compute-equivalent occurs in "training" a human brain from birth to adulthood, increased by a factor for human infants being "pretrained" by evolution.

Also similar to Cotra, we plan to use lognormal distributions (instead of point estimates) around our anchors for modeling compute requirements due to uncertainty, spanning orders of magnitude, about these anchors' "true" values as well as uncertainty around the anchors' applicability without modifications. A couple of these anchors are influenced by upstream nodes related to the evolution of intelligence and the computational power of the brain (circled in orange in Figure 26; these "upstream" nodes are lower down on the figure since the information in the figure flows bottom to top).

Additionally, one final module includes "wildcards" that are particularly speculative but, if true, would be game changers. The nodes within this wildcards module have been copied over to other relevant modules (to be discussed below) as alias nodes.

We discuss the modules for the various methods to HLMI below, elaborating on the relevant anchors (and more upstream nodes) where relevant.

#### 3.2.1.1 Evolutionary Algorithms

The development of evolved HLMI would require both *enough hardware* and a *suitable environment or other dataset*, given the evolutionary algorithm. Whether there will be enough hardware by a specific date depends on both the required hardware (approximated here by the *evolutionary anchor*, discussed below) and how much hardware is available at that date (calculated in the *Hardware Progression* module [§3.1]). The ability to create a suitable environment/dataset by a certain date depends, in large part, on the *hard paths hypothesis* [64]—that it's rare for environments to straightforwardly select for general intelligence.

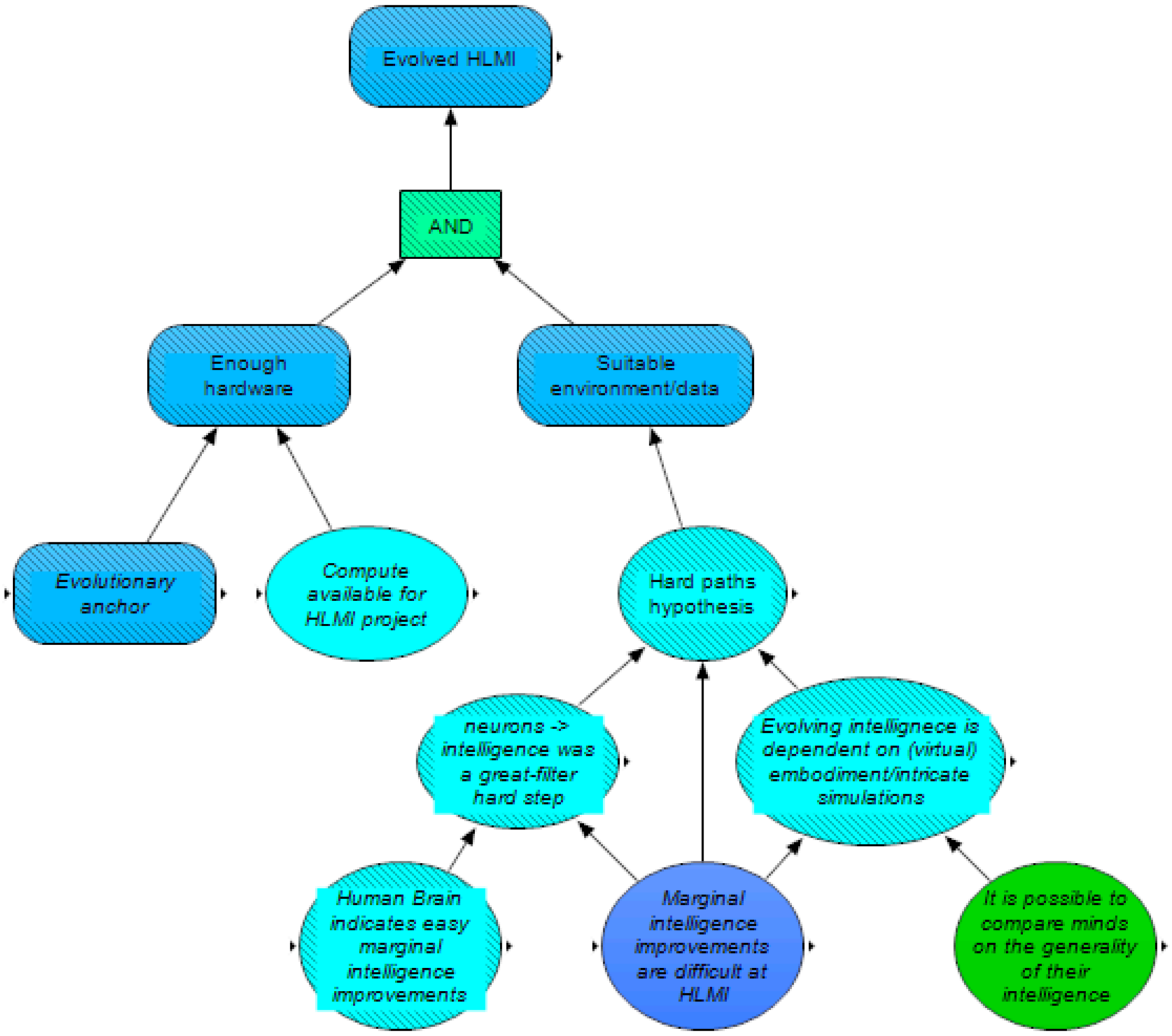

*Figure 27: HLMI from evolutionary algorithms*

This hypothesis is further influenced by other cruxes. If somewhere along the progression from the first neurons to human intelligence, there was a *"hard step"* [65] *(a la the great filter)*, then such a hypothesis is more likely true. Such a hard step would imply that the vast majority of planets that evolve animal-like life with neuron-like parts never go on to evolve a technologically intelligent lifeform. This scenario could be the case if Earth (or some portion of Earth's history) was particularly unusual in some environmental factors that selected for intelligence. Due to anthropic effects, we cannot rule out such a hard step simply based on the existence of (human) technological intelligence on Earth. Such a hard step is less likely to be the case if *marginal intelligence improvements are easy around the human level*, and we consider *evidence from the human brain* (such as whether the human brain appears to be a "scaled up" version of a more generic mammalian or primate brain) to be particularly informative here (both of these cruxes are copied over from our module on *Analogies and General Priors on Intelligence* [§2]).

Additionally, the hard paths hypothesis is influenced by whether *evolving intelligence would depend on embodiment* (e.g., in a virtual body and environment). Proponents of this idea have argued that gaining sufficient understanding of the physical world (and/or developing cognitive modules for such understanding), including adequate symbol grounding [66] of important aspects of the world and (possibly necessary) intuitive physical reasoning (e.g., that humans use for certain mathematical and engineering insights), would require an agent to be situated in the world (or in a sufficiently realistic simulation), such that the agent can interact with the world (or simulation). Opponents of such claims often argue that multimodal learning will lead to such capabilities, without the need for embodiment. Some proponents further claim that even if embodiment is not necessary in principle for evolving intelligence, it may still be necessary in practice if other environments tend to lack the complexity or other attributes that sufficiently select for intelligence.

If evolving intelligence depends on embodiment, this would significantly decrease the potential paths to evolving intelligence, as HLMI would therefore only be able to be evolved in environments in which it was embodied and potentially only in select simulated environments that possessed key features. The necessity of embodiment appears to be less likely if *marginal intelligence improvements are easy*. In particular, the necessity of embodiment appears less likely if *minds can be compared on the generality of their intelligence* (also copied over from our module on *Analogies and General Priors on Intelligence* [§2]), as the alternative implies "general intelligence" isn't a coherent concept, and intelligence is then likely best thought of as existing only with respect to specific tasks or goals. In such a situation, AI systems interacting in or reasoning about the physical world would likely need to have the development of their cognitive faculties strongly guided by interactions in such a world (or a close replica).

Finally, even if there is no evolutionary hard step between neurons and intelligence, and embodiment is not necessary for evolving intelligence, the hard paths hypothesis may still be true; some have argued [67], for example, that curriculum learning [68] may be necessary for achieving HLMI. It therefore seems reasonable to assume that the harder marginal intelligence improvements are around the HLMI level (even holding constant the other mentioned cruxes), the lower the chance we should give to an individual environment or dataset leading to the evolution of intelligence, and thus the higher the credence we should give the hard paths hypothesis.

As mentioned above, on the compute side, the requirements for evolved HLMI are dependent on the *evolutionary anchor*. This anchor is determined by an upstream submodule (Figure 28). Our estimate for this anchor is similar to its analog in Cotra's report [28], but with a couple of key differences:

Here, this anchor is determined by taking the amount of *"compute-equivalent" expected for the evolution of intelligent life* such as humans (considering both neural activity and simulating an environment) and dividing it by a *speedup factor* due to human engineers potentially being able to outcompete biological evolution towards this goal. Cotra's report did not include such a speedup factor, as she was using her evolutionary anchor as somewhat of an upper bound. Here, on the other hand, we are instead concerned with an all-things-considered estimate of how difficult it would be to evolve HLMI.

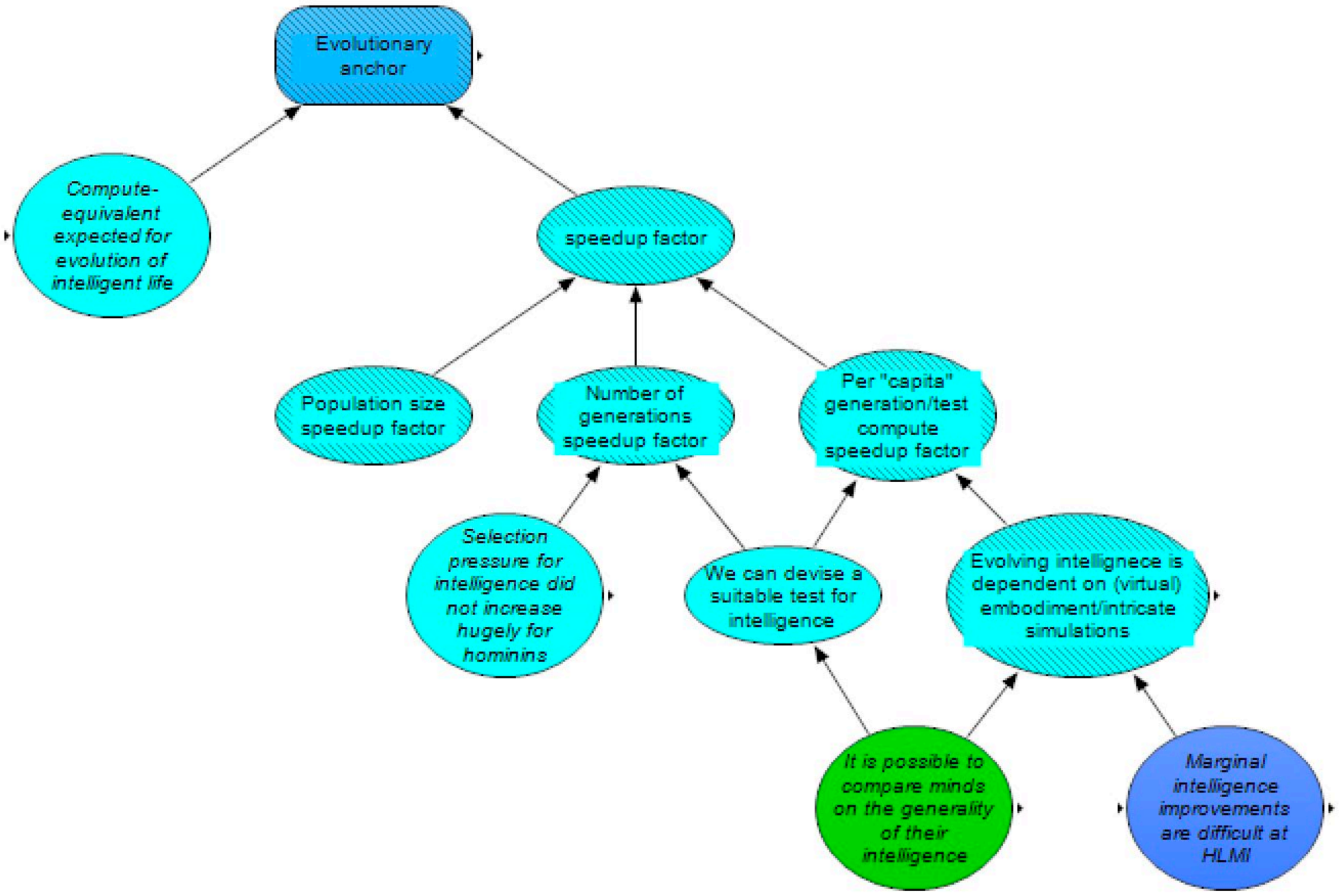

*Figure 28: Evolutionary anchor*

Our *speedup factor* considers three sources of increased speed:

- *Population size speedup factor*—the population of Earth's animals is not set by minimizing the amount of compute-equivalent necessary to evolve an intelligent species, but instead by Earth's carrying capacity; human engineers, meanwhile, could set population sizes with the goal of minimizing necessary compute to evolve intelligence.

- *Number of generations speedup factor*—much of Earth's macroevolutionary history was presumably not optimizing for an intelligent species in as few generations as possible, and there are many tricks that human engineers could employ to select for and possibly achieve this goal quicker. In particular, if evolution was *hardly selecting for intelligence before hominins*, then a more intentional effort could likely be much faster. Additionally, if we can design *tests for intelligence*, then we could potentially create stronger selection pressure for intelligence, reducing the simulated time needed.

- *Per "capita" speedup factor*—for biological evolution, organisms are "run" from birth until they die, despite much of that time not necessarily presenting much selection effect. Human engineers might be more efficient here—especially if *we can devise a test for intelligence* (in which case the generation time might be whatever is required to complete a short test) and/or if *evolving intelligence does not depend on embodiment* (in which case much of the information processes in animals' brains are likely, in principle, unimportant for evolving intelligence, and thus human engineers could reduce compute requirements by crafting much simpler datasets that don't lead to costly adaptations for embodiment).

Note that there is somewhat of a tradeoff between these three factors—a smaller population, for instance, would presumably imply more generations needed. Each of these factors should therefore be set to minimize the product of the three variables instead of only the corresponding variable itself.

The expected amount of compute-equivalent required to evolve intelligence on Earth (the left node in Figure 28, not considering the *speedup factor*) is estimated from a further upstream submodule, *compute expected for evolution of intelligence*, which is represented in Figure 29.

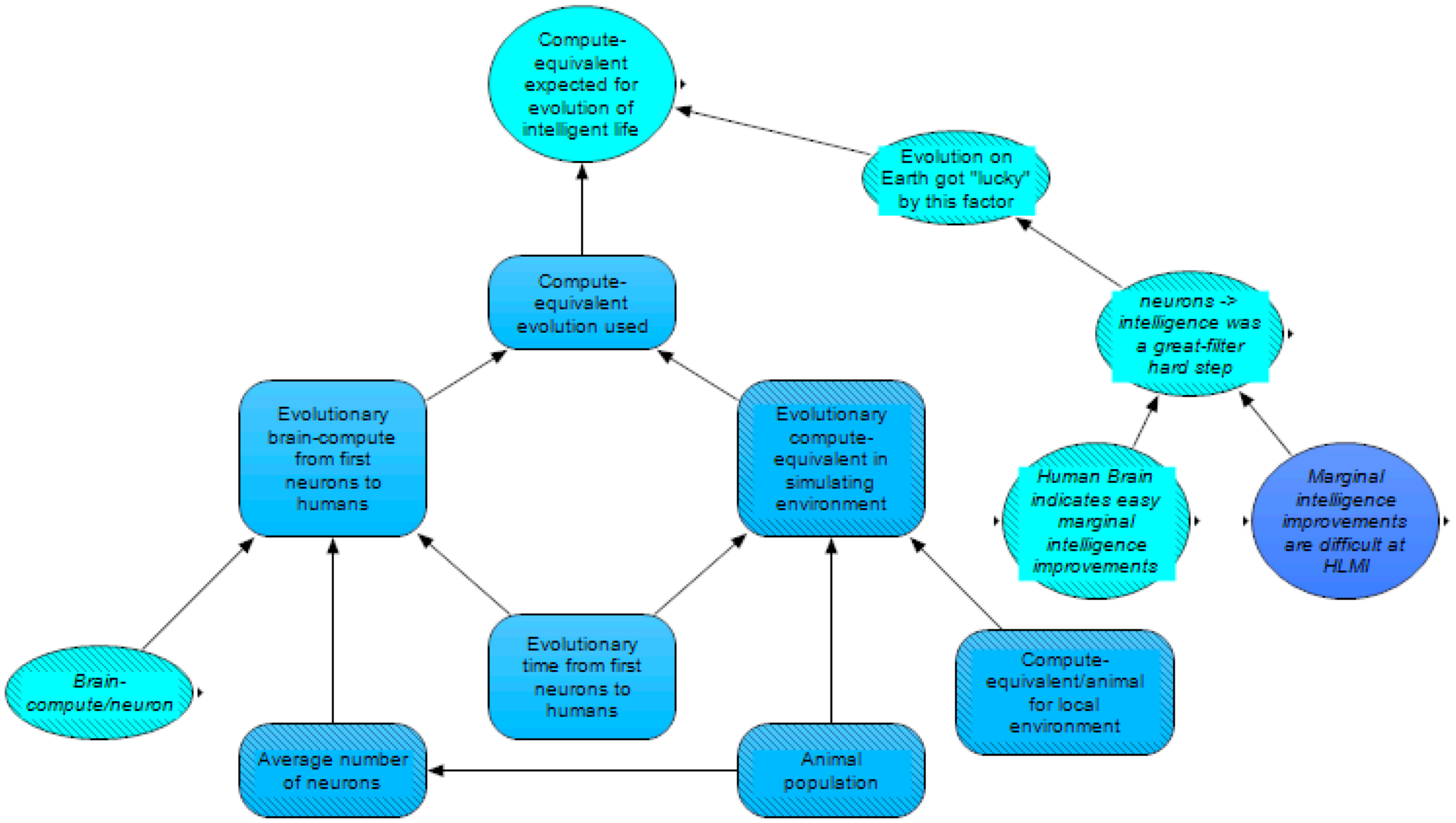

*Figure 29: Compute expected for evolution of intelligence*

This estimate is made up of two main factors which are multiplied together: the *compute-equivalent evolution used* on Earth, and a factor related to the possibility that *evolution on Earth got "lucky."* The latter of these factors corresponds to the aforementioned possibility that there was a hard step between the first neurons and humans—this factor is set to one if either there was no hard step or if there was a hard step relating strictly to Earth's environment being unusual; otherwise, the factor is set to a number higher than one, corresponding to how much faster evolution of intelligence (from first neurons) took compared to, naïvely, how long such a process would be expected to take given an Earth-like environment and infinite time. That is, the factor is more or less set to an estimate to the question, "for every Earth-like planet that evolves a lifeform with technological intelligence, how many Earth-like planets evolve a lifeform with something like Precambrian jellyfish intelligence?"

The factor for *compute-equivalent evolution used*, meanwhile, is broken up into the sum of two factors: the *brain-compute from first neurons to humans* (i.e., based on an estimate of how much compute each nervous system "performs," though note the definition here is somewhat inexact) and the *compute-equivalent used in simulating the environment* (i.e., an estimate of the minimum compute necessary to simulate an environment for the evolution of animals on Earth). The environmental compute-equivalent factor has traditionally been neglected in analyzing the compute-equivalent for simulating evolution, as previous research has assumed that much more necessary compute occurs in brains than in the surrounding environment (for instance, because the human brain uses much more brain-compute than the compute in even high-resolution video games, and humans can function in such environments, including interacting with each other). While this assumption may be reasonable for the functioning of large animals such as humans, it is not immediately obvious that it also applies to much smaller animals such as *C. elegans*. Furthermore, environmental complexity, dynamism, and a large action-space are typically thought to be important for the evolution of intelligence, and a more complex and dynamic environment, especially in which actors are free to take a very broad set of actions, may be more computationally expensive to simulate.

Both the brain-compute estimate and the environmental compute-equivalent estimate are determined by multiplying the *evolutionary time from first neurons to humans* by their respective populations, and then by the compute-equivalent per time of each member within these populations. For the brain-compute estimate, the population of interest is the neurons, so the relevant factors are *average number of neurons* (in the world) and the *brain-compute per neuron* (as has been estimated [69] by Joseph Carlsmith). For the environmental compute-equivalent estimate, the relevant factors are the *average population of animals* (of course, related to the average population of neurons), and the *compute required to simulate the local environment* for the average animal (under the idea that the local environment surrounding animals may have to be simulated in substantially higher resolution than parts of the global environment which no animal occupies).

To recap, adding together the brain-compute estimate and the environmental compute-equivalent estimate yields the total estimate for compute-equivalent evolution used on Earth, and this estimate is then multiplied by the "luckiness" factor for the compute-equivalent expected to be necessary for evolving intelligence. The evolutionary anchor is then determined by multiplying this factor by a speedup factor. The date by which enough hardware is available for evolved HLMI is set to be the first date the available compute for an HLMI project exceeds the evolutionary anchor, and evolved HLMI is assumed to be created on the first date that there is enough hardware and a suitable environment or dataset (which is strongly influenced by the hard paths hypothesis).

#### 3.2.1.2 Current Deep Learning plus Business-As-Usual Advancements

This module, which represents approaches towards HLMI similar to most current approaches used in deep learning, is more intricate than the others, though it also relies on some similar cruxes to the above.

Similar to evolutionary algorithms, we model achieving this milestone at the first date that there is both *enough hardware* (given the available algorithms) and *sufficient data* (quantity, quality, and kind, again given the available algorithms) for training the HLMI.

The ease of achieving the necessary data is strongly influenced by the *hard paths hypothesis* [64] (circled in red in Figure 30), as before.

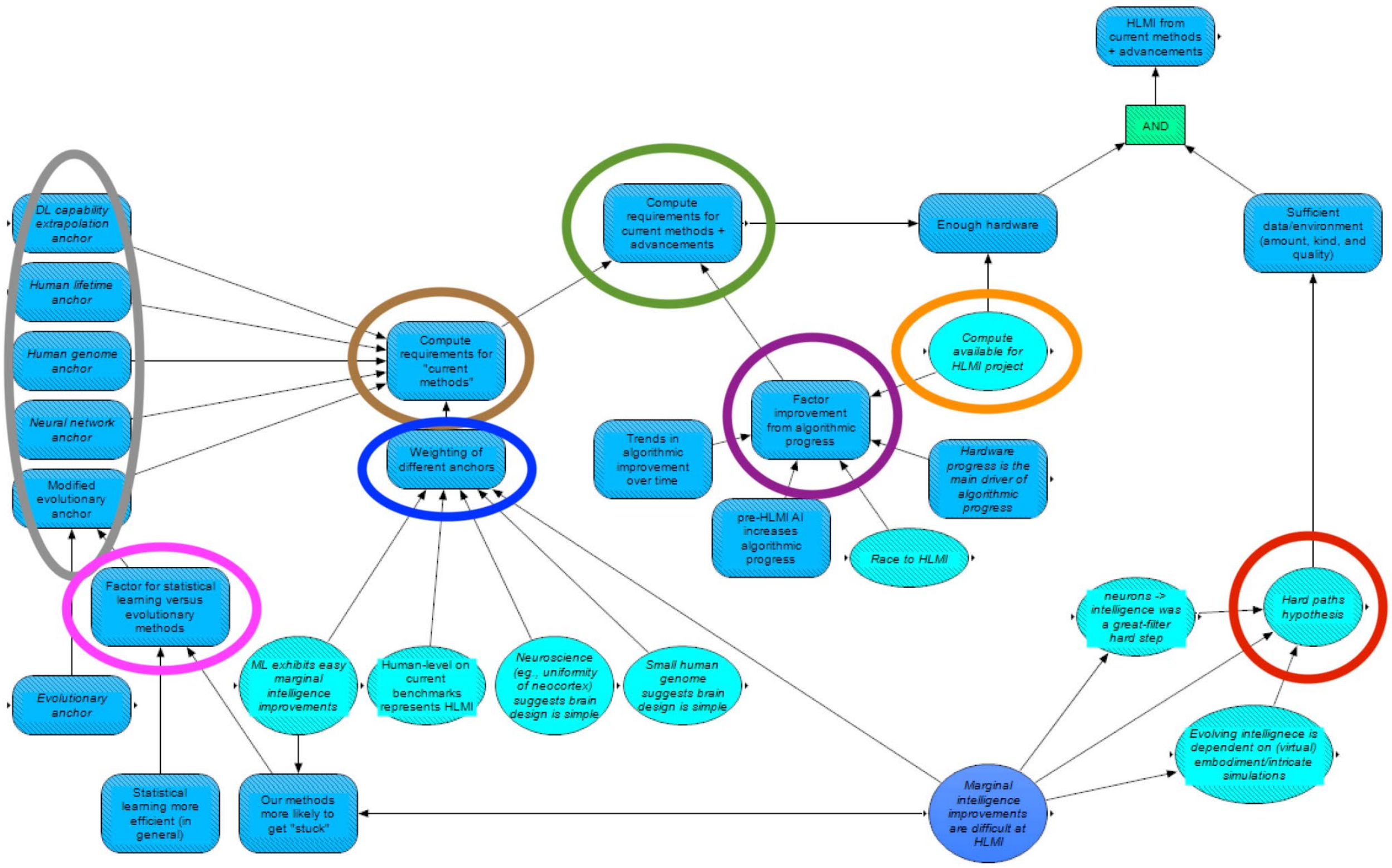

*Figure 30: HLMI from current methods and business-as-usual advancements*

Whether enough hardware will exist for HLMI via these methods (by a certain date) is determined by both the *amount of hardware available* (circled in orange) and the *hardware requirements* (circled in green). As before, the hardware available is taken from the *Hardware Progression* (§3.1) module. The hardware requirements, in turn, are determined by the *compute requirements for "current methods"* in deep learning to reach HLMI (circled in brown), modified by a *factor for algorithmic progress* (circled in purple). Expected algorithmic improvements are modeled as follows.

First, baseline algorithmic improvements are extrapolated from *trends in such improvements*. This extrapolation is performed, by default, in terms of time. However, if the crux *hardware progress is the main driver of algorithmic progress* resolves positively, then the extrapolation is instead based on hardware progress, proxied with *compute available for an HLMI project*. Next, the baseline algorithmic improvements are increased if there is a *race to HLMI* (in which case it is assumed investment into algorithmic advancements will increase), or if it is assumed that *pre-HLMI AI will enable* faster AI algorithmic progress [70].

Compute requirements for "current methods" are estimated as a linear combination of the estimates provided by the anchors (circled in grey). The *evolutionary anchor* is *modified by a factor* (circled in pink) for whether our statistical learning algorithms would be better or worse than the evolutionary anchor in terms of finding intelligence. The main argument in favor of our statistical methods is that *statistical learning is generally more efficient than evolution*, and the main argument against is that our *statistical methods may be more likely to get* "stuck" [67] either via strongly separated local minima or due to goal hacking and Goodhart's Law becoming dead-ends for sufficiently general intelligence (which may be less likely if *marginal improvements in intelligence* near HLMI are easier, and particularly less likely if *ML shows evidence of easy marginal intelligence gains*).

Our estimates for the human lifetime anchor, human genome anchor, and neural network anchor are all calculated using the similar logic as in Cotra's report [28]. The human lifetime anchor involves estimating the brain-compute that a human brain performs in "training" from birth to adulthood—that is, (86 billion neurons in a human brain)*(brain-compute per neuron per year)*(18 years)—and multiplying this number by a factor for newborns being "pretrained" by evolution. This pretraining factor could be based on the extent to which ML systems today are "slower-learners" [61] compared to humans, or the extent to which human-engineered artifacts tend to be less efficient [71] than natural analogs.

The human genome anchor and neural network anchor, meanwhile, are calculated in somewhat similar manners to each other. In both, the modeled amount of compute to train an HLMI can be broken down into the amount of data to train the system, times the compute used in training per amount of data. The estimated number of data points needed can be calculated from the number of parameters the AI uses, via empirically derived scaling laws between parameters and data points for training, with the number of parameters calculated differently for the different anchors: for the human genome anchor, it's set to the bytes in the human genome, and for the neural network anchor, it's set based on expectations from scaling up current neural networks to use similar compute as the brain (with a modifying factor). The compute used in training per data point for both of these anchors, meanwhile, is calculated as the brain-compute in the human brain, times a modifying factor for the relative efficiency of human-engineered artifacts versus their natural analogs, times the amount of "subjective time" generally necessary to determine if a model perturbation improves or worsens performance (where "subjective time" is set from the amount of compute the AI uses, such that the compute the AI uses equals the brain-compute the human brain uses, times the modifying factor, times the subjective time). For this last factor regarding the subjective time for determining if a model perturbation is beneficial or harmful (called the "effective horizon length" [72]), the appropriate value for the human genome anchor is arguably related to animal generation times, while for the neural network anchor, the appropriate value is arguably more related to the amount of time it takes humans to receive environmental feedback for their actions (i.e., perhaps seconds to years, though for meta-learning or other abilities evolution selected for, effective horizon lengths on par with human generation times may be more appropriate). For a more in-depth explanation of the calculations related to these anchors, see Cotra's report [28] or Rohin Shah's summary [73] of the report.

Finally, our model includes an anchor for *extrapolating the cutting edge of current deep learning algorithms* (e.g., GPT-3 [74]) along various benchmarks in terms of compute—effectively, this anchor assumes that deep learning will (at least with the right data) "spontaneously" [17] overcome its current hurdles [75] (such as building causal world models, understanding compositionality, and performing abstract symbolic reasoning) in such a way that progress along these benchmarks will continue their current trends in capabilities vs compute as these hurdles are overcome.

The human genome anchor, neural network anchor, and DL capability extrapolation anchor all rely on the concept of the effective horizon length, which potentially provides a relationship between task lengths and training compute. Resultantly, disagreements related to these relationships are cruxy. One view is that, all else equal, effective horizon length is generally similar to task length, and training compute scales linearly with effective horizon length. If this is not the case, however, and either effective horizon length grows sublinearly with task length, or training scales sublinearly with effective horizon length, then all three of these anchors could be substantially shorter. While our model currently has a node for whether *training time scales approximately linearly with effective horizon length and task length*, we are currently uncertain how to model these relationships if this crux resolves negatively. An intuition pump in favor of sublinearity between training time and task length is that if one has developed the qualitative capabilities necessary to write a 10-page book, one also likely has the qualitative capabilities necessary to write a 1,000-page book; writing a 1,000-page book might itself take 100 times as long as writing a 10-page book, but training to write a 1,000-page book presumably does not require 100 times longer than training to write a 10-page book.

The *weighting* of these five anchors (circled in blue in Figure 30) is also a subject of dispute. We model greater weighting towards lower anchors if *marginal intelligence improvements around HLMI are easy*, and we additionally weight several of the specific anchors higher or lower depending on other cruxes. Most importantly, the weighting of the *DL capability extrapolation anchor* is strongly dependent on whether *human level on current benchmarks represents HLMI* and if *ML exhibits easy marginal intelligence improvements* (implying such extrapolations are more likely to hold until human level as compute is ramped up). The *human lifetime anchor* is weighted higher if *neuroscience suggests the brain design is simple* (since then, presumably, the human mind is not heavily fine-tuned due to pretraining). Additionally, if the *small human genome suggests that the brain design is simple*, then the *human genome anchor* is weighted higher (otherwise, we may assume that the complexity of the brain is higher than we might assume from the size of the genome, in which case, this anchor doesn't make a lot of sense).

Let's consider how a couple of archetypal views about current methods in deep learning can be represented in this model. If one believes that DL will reach HLMI soon after matching the compute in the human brain (for instance, due to the view [76] that we could make up for a lack of fine-tuning from evolution with somewhat larger scale), then this would correspond to high weight on the *human lifetime anchor*, in combination with a "no" resolution for the *hard paths hypothesis*. On the other end of the spectrum, if one were to believe that "current approaches in DL can't scale to HLMI," then this would presumably correspond to either a sufficiently strong "yes" to the *hard paths hypothesis* (that is, "current methods could scale if we had the right environment or data, but such data will not be created/aggregated, and the wrong kind of data won't lead to HLMI"), or a high weight to the *modified evolutionary anchor*, with, presumably, either a very large factor for *our methods getting "stuck"* ("evolution was only able to find intelligence due to evolution having advantages over our algorithms"), or a large factor for *"luckiness" of evolution* ("evolution got astronomically lucky to reach human-level intelligence, and even with evolutionary amounts of compute, we'd still be astronomically unlikely to reach HLMI").

### 3.2.1.3 Hybrid Statistical-Symbolic AI

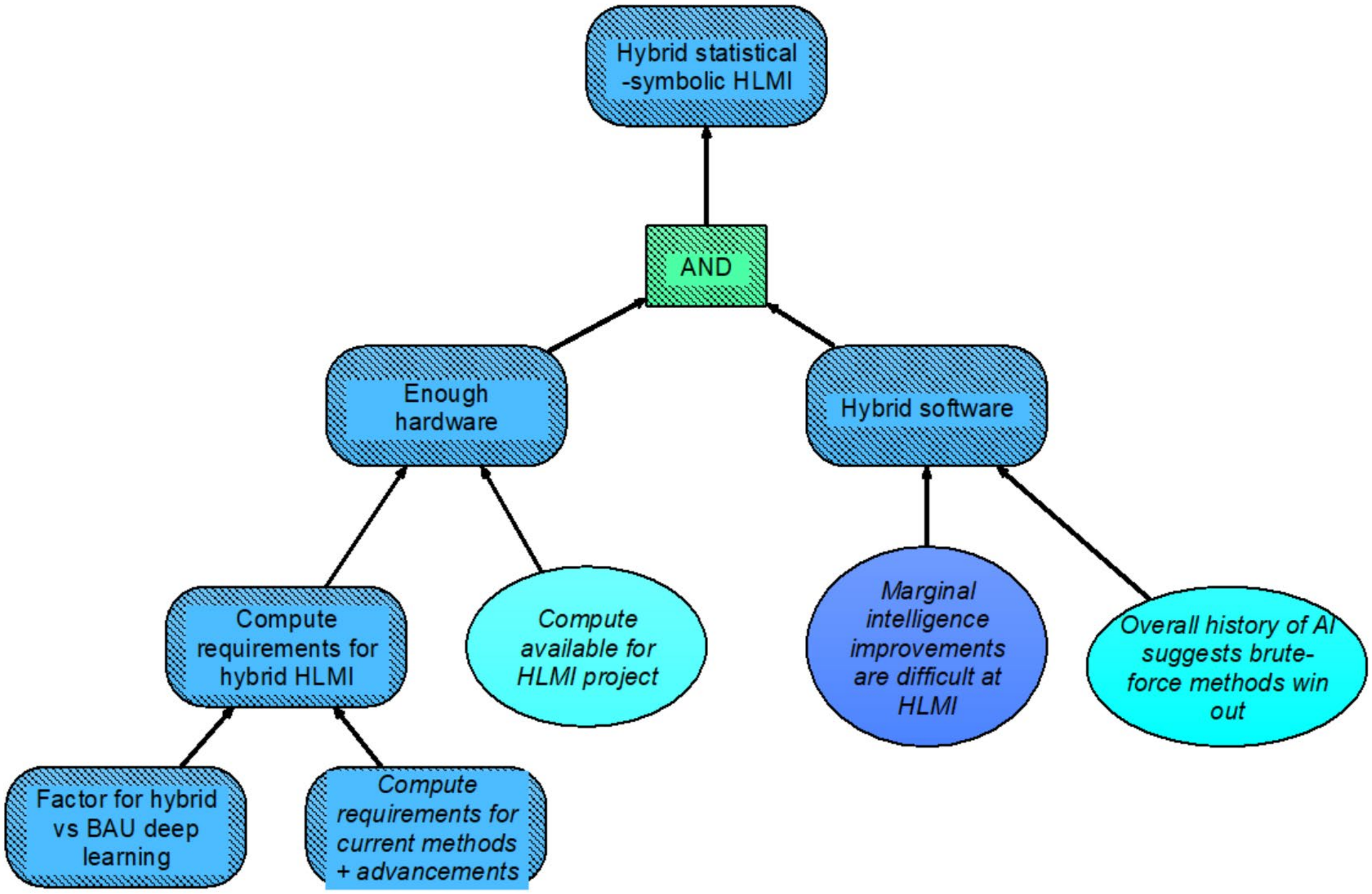

Figure 31: Hybrid statistical-symbolic AI

Many people who are doubtful about the ability of current statistical methods to reach HLMI instead think [77] that a hybrid statistical-symbolic approach (e.g., DL + GOFAI) could be more fruitful. Such an approach would involve achieving the necessary *hardware* (similar to before) and creating the necessary symbolic methods, data, and other aspects of the *software*. Here, we model the required hardware as being related to the *required hardware for current deep learning + BAU (business-as-usual) advancements* (described in the above section), modified with a *factor for hybrid methods requiring more/less compute*. As we are not sure how best to model achieving the necessary software (we are open to relevant suggestions), our model of this is very simple—we assume that such software is probably easier to develop if *marginal intelligence improvements are easier*, and further that the more one buys the bitter lesson [26], that there is a history of more naïve, compute-leveraging, *brute-force methods winning* compared to more manually crafted methods, the less one should suspect such hybrid software is feasible.

### 3.2.1.4 Cognitive-Science Approach

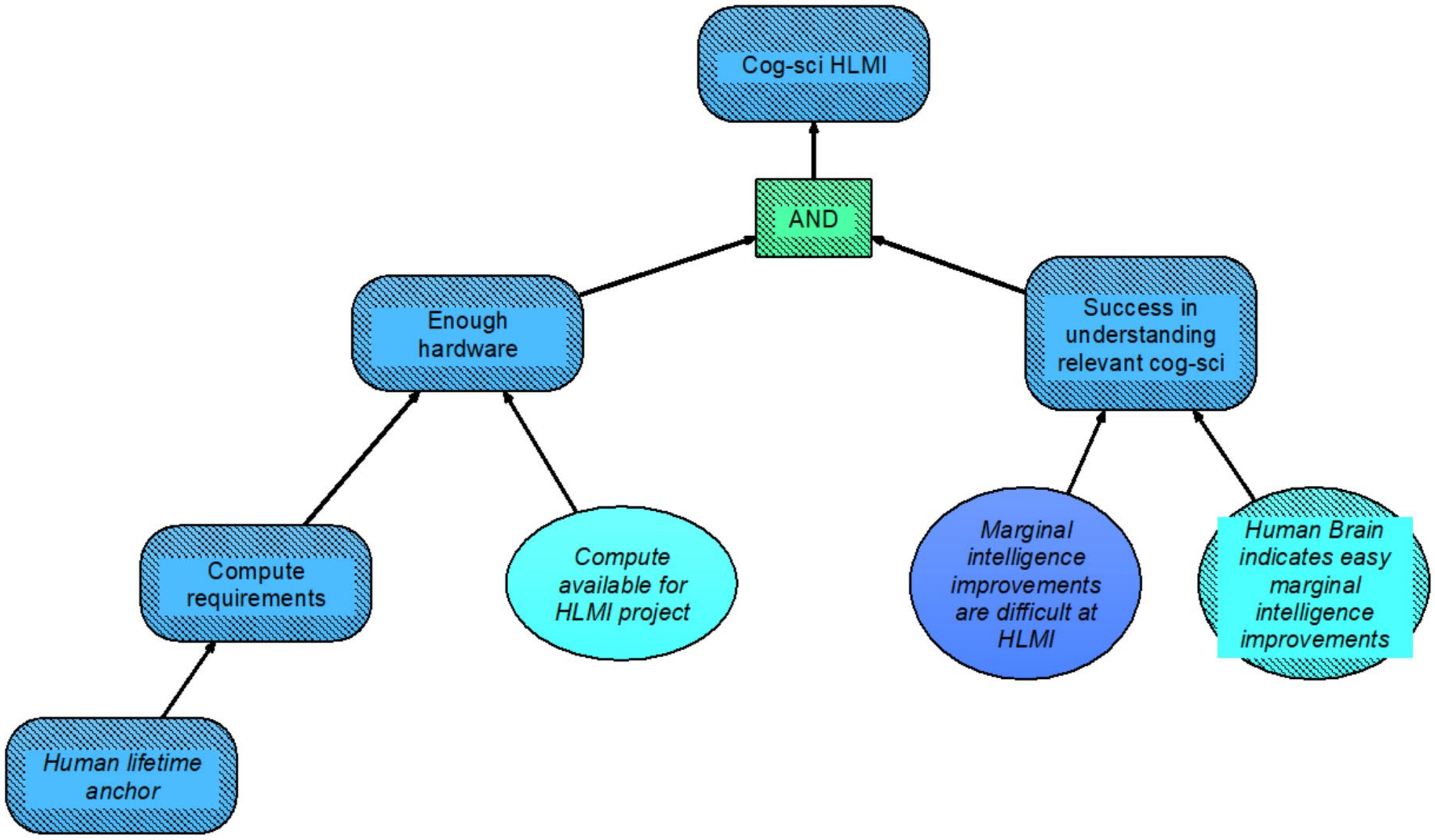

*Figure 32: Cognitive-science approach*

Similar to the above section, some people who are doubtful about the prospects of deep learning have suggested [77] insights from cognitive science and developmental psychology would enable researchers to achieve HLMI. Again, this approach requires the relevant *hardware* and *software* (where the software is now dependent on understanding the relevant cognitive science). The hardware requirement here is estimated via the *human lifetime anchor*, as the cognitive-science approach attempts to imitate the learning processes of the human mind more closely. Here again, we are unsure how to model the likelihood in software success (ideas for approaching this question would be appreciated). For the time being, we assume that, similar to other methods, the *easier marginal intelligence is*, the more likely this method will work, and in particular *evidence from the brain* would be particularly informative here, as a simpler human brain would presumably imply an easier time copying certain key features of the human mind.

### 3.2.1.5 Whole Brain Emulation/Brain Simulation

These methods would involve running a model of a brain (in the case of WBE, of a specific person's brain, and in the case of brain simulation, of a generic human brain) on a computer, by modeling the structure of the brain and the information-processing dynamics of the lower-level parts of the brain, as well as integrating this model with a body inside an environment (either a virtual body and environment, or a robotic body with I/O to the physical world). To be considered successful for our purposes here, the model would have to behave humanlike (fulfilling the criteria for HLMI), but (in the case of WBE) would not need to act indistinguishable from the specific individual whose brain was being emulated, nor would it have to accurately predict the individual's behavior. We can imagine, by analogy to weather forecasting, that accurately forecasting the weather is much harder than simply forecasting weather-like behavior, and similarly the task outlined here is likely far easier than creating a WBE with complete fidelity to an individual's behavior, especially given the potential for chaos.

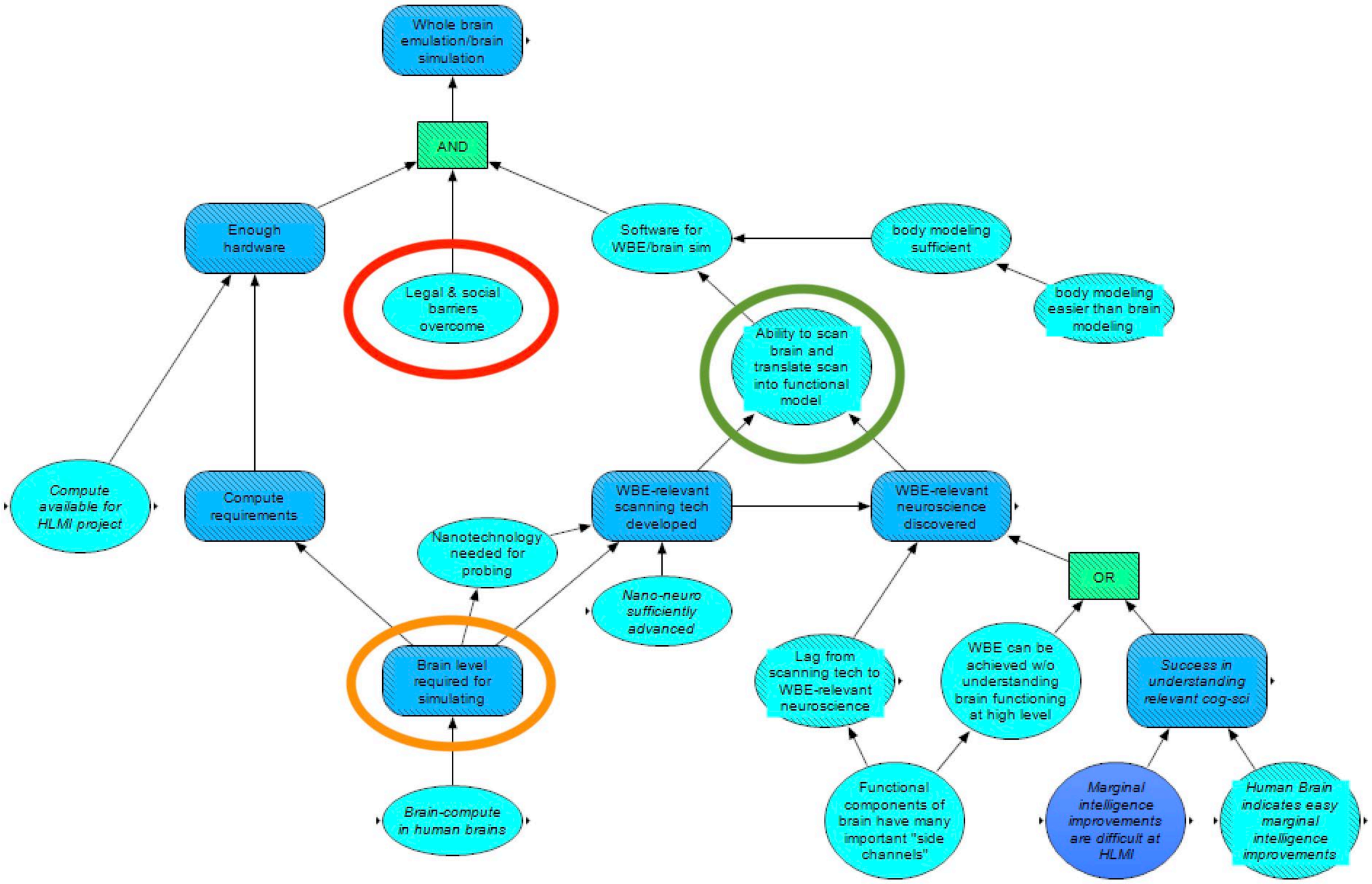

*Figure 33: Whole brain emulation/simulation*

As before, our Analytica model (Figure 33) considers cruxes for fulfilling *hardware* and *software* requirements, but we also include a node here for *overcoming the legal and social barriers* (circled in red). This is because many people find the idea of brain emulation/simulation either ethically fraught or intuitively disturbing and unpalatable, and it is therefore less likely to be funded and pursued even if technologically feasible.

As before, the hardware question is determined by *availability* and *requirements*, with requirements dependent on the *scale* [78] *at which the brain must be simulated* (e.g., spiking neural network, concentrations of neurotransmitters in compartments, and so on [79])—circled in orange. We consider that the more *brain-compute that the human brain uses* (number of neurons times brain-compute per neuron), the lower the scale we should assume must be simulated, as important information processing is then presumably occurring in the brain at lower scales (interactions on lower scales tend to be faster and larger in absolute number); however, the amount of brain-compute the brain uses is modeled here as a lower bound on computational costs, as emulating or simulating might require modeling details less efficiently than how the brain instantiates its compute (in the same way that a naïve simulation of a calculator—simulating the electronic behavior in the various transistors and other electronic parts—would be much more computationally expensive than simply using the calculator itself).

On the software side, success would be dependent on the ability to *scan a brain and translate the scan into a functional model* (circled in green) and to *model a body sufficiently* (with one potential crux for whether *body modeling is easier than brain modeling*). Scanning the brain sufficiently, in turn, is dependent on *WBE-relevant scanning technology* enabling the creation of a structural model of the brain (at whatever scale is necessary to model), while translating the scan into a model would depend on the discovery of *WBE-relevant neuroscience* that allows for describing the information-processing dynamics of the parts of the brain. Note that, depending on the scale necessary to consider for the model, such "neuroscience" might include understanding of significantly lower-level behavior than what's typically studied in neuroscience today.

Discovering the relevant neuroscience, in turn, would depend on a few factors. First, relevant scanning technology would need to be created so that the dynamics of the brain could be probed adequately. Such scanning technology would presumably include some similar scanning technology as mentioned above for creating a structural model, though would also likely go beyond this (our model collapses these different types of scanning technology down to one node representing all the necessary scanning technology being created). Second, after such scanning technology was created, there would plausibly be a *lag from the scanning technology to the neuroscience*, as somewhat slow, physical research would presumably need to be carried out with the scanning technology to discover the relevant neuroscience.

Third, neuroscientific success would depend on brain emulation or simulation being *achievable without understanding the brain functioning at a high level*, or there would need to be success in *understanding sufficient cognitive science* (which we proxy with the same understanding necessary for a cognitive-science HLMI [§3.2.1.4]). Such higher-level understanding may be necessary for validating the proper functioning of components on various scales (e.g., brain regions, large-scale brain networks, etc.) before integrating them.

We plan to model both the length of the lag and whether WBE can be achieved without understanding the higher-level functioning of the brain as dependent on whether there are *many important "side channels"* for the I/O behavior of the functional components of the brain. If there are many such side channels, then we expect finding all of them and understanding them sufficiently would likely take a long time, increasing the lag considerably. Furthermore, validating the proper functioning of parts in this scenario would likely require reference to proper higher-level functioning of the brain, as the relevant neural behavior would resultantly be incredibly messy. On the other hand, if there are not many such side channels, discovering the proper behavior of the functional parts may be relatively quick after the relevant tools for probing are invented, and appropriate higher-level functioning may emerge spontaneously from putting simulated parts together in the right way. Arguments in favor [80] of there being many side channels typically make the point that the brain was biologically evolved, not designed, and evolution will tend to exploit whatever tools it has at its disposal in a messy way, without heed to legibility. Arguments against [81] there being many such side channels typically take the form that the brain is fundamentally an information-processing machine, and therefore its functional parts, as signal processors, should face strong selection pressure for maintaining reliable relationships between inputs and outputs—implying relatively legible and clean information-processing behavior of the parts.

Developing the necessary scanning technology for WBE, meanwhile, also depends on a few factors. The difficulty of developing the technology would depend on the brain level required to model (mentioned above). Additionally, if the scanning technology *requires* [82] *nanotechnology* (likely dependent on the brain level required for simulating), then an important crux here is whether *sufficient nano-neurotechnology is developed*.

### 3.2.1.6 Neuromorphic AGI (NAGI)

To avoid one point of confusion between this section and the two previous sections: the previous section (WBE/brain simulation [§3.2.1.5]) is based on the idea of creating a virtual brain that operates similar to a biological brain, and the section before that (cognitive-science approach [§3.2.1.4]) is based on the idea of creating HLMI that uses many of the higher-level processes of the human mind but that isn’t instantiated on a lower level in a similar manner to a biological brain (e.g., such an approach might have modules that perform similar functions to brain regions but without anything modeling bio-realistic neurons). In contrast, this section (neuromorphic AGI) discusses achieving HLMI via methods that use lower-level processes from the brain but without putting them together in a manner to create something particularly human mind-like (e.g., it may use models of bio-realistic neurons but in ways at least somewhat dissimilar to human brain structures).

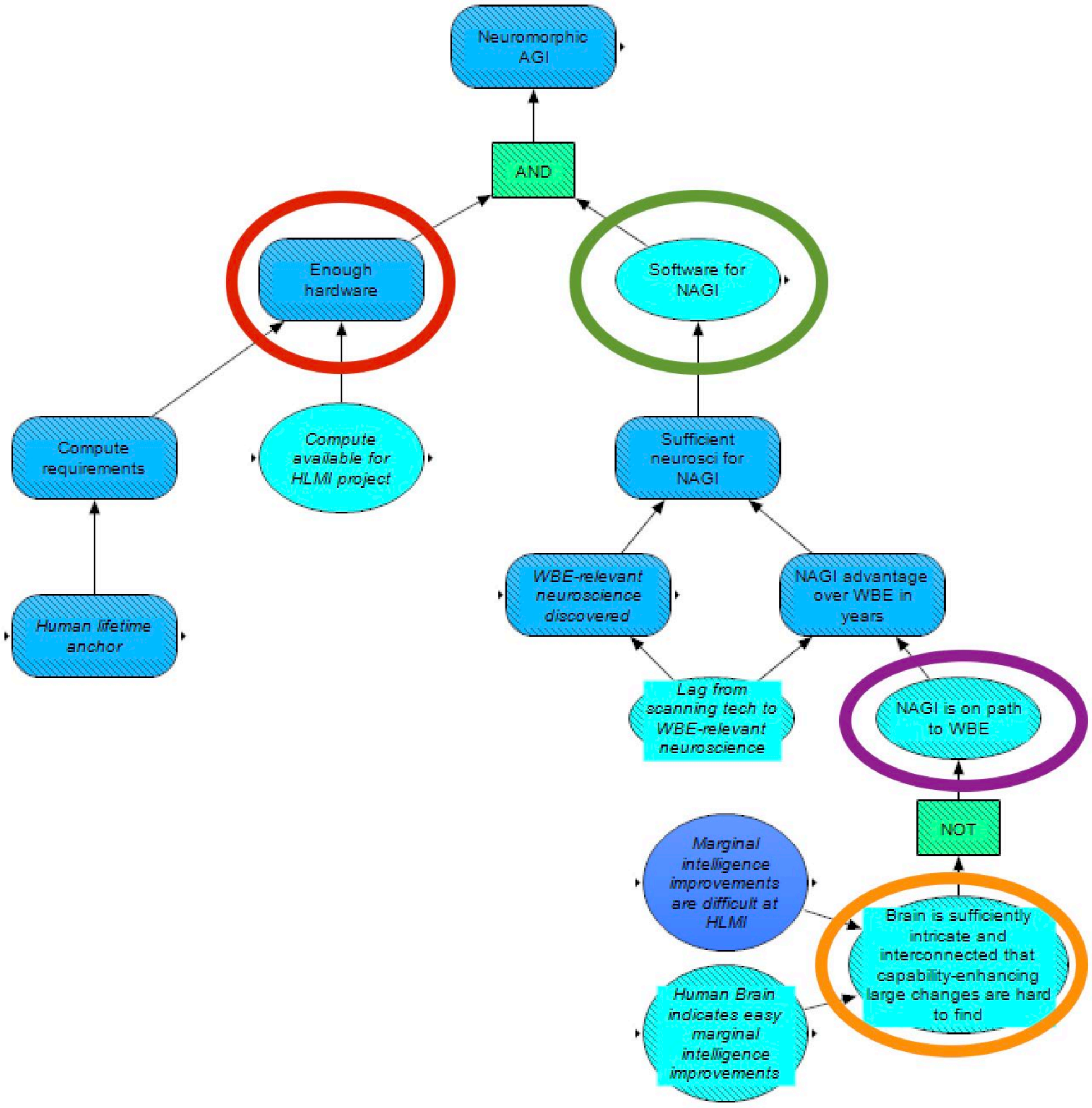

*Figure 34: Neuromorphic AGI*

Similar to other methods, neuromorphic AGI (NAGI) is modeled as being achieved if the relevant *hardware* (circled in red in Figure 34) and *software* (circled in green) are achieved. As with the cognitive-science approach, the *human lifetime anchor* is used for the *compute requirement* (which is compared against the *available compute* for whether there is enough hardware) because this method attempts to imitate more humanlike learning.

On the software side, the main crux is whether *NAGI is on the path to WBE* (circled in purple). If so, then the relevant *neuroscience for NAGI* (sufficient for the software for NAGI) is assumed to be discovered before the *relevant neuroscience for WBE*. The amount of time *advantage that NAGI has over WBE* in terms of relevant neuroscience under this condition is then assumed to be less than or equal to the *lag from scanning technology to WBE-relevant neuroscience*; that is, the neurotechnology needed for NAGI is implicitly assumed to be the same as for WBE, but gathering the relevant neuroscience for NAGI is assumed to be somewhat quicker.

Whether NAGI is on the path to WBE, however, is dependent on whether the *brain is sufficiently intricate and interconnected that large changes are almost guaranteed to worsen capabilities* (circled in orange; if so, then NAGI is assumed to not be on the path to WBE, because potential "modifications" of brain architecture to create NAGI would almost definitely hurt capabilities). This crux is further influenced by the *difficulty of marginal intelligence improvements*, with particular emphasis on *evidence from the human brain*.

### 3.2.1.7 Other Methods

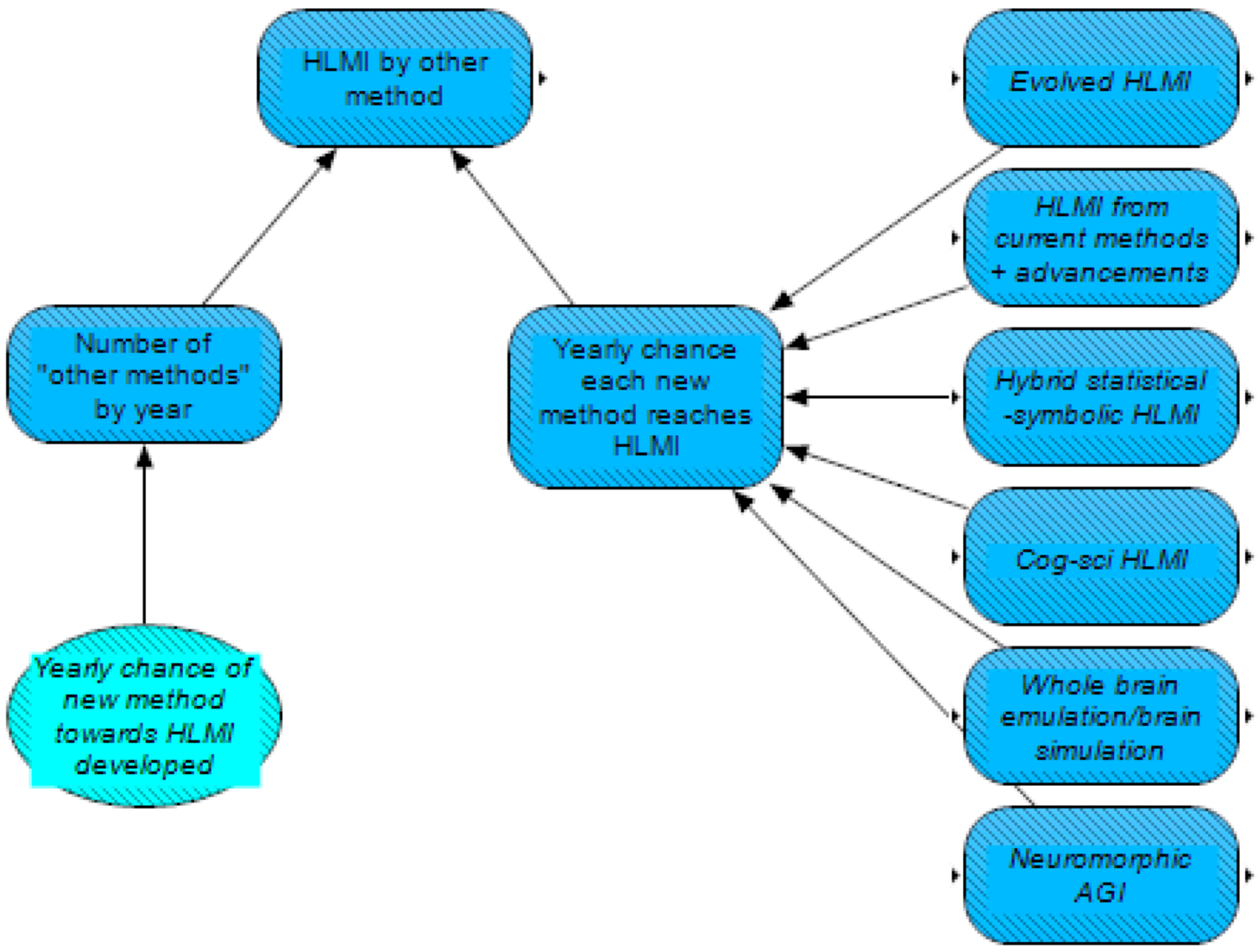

*Figure 35: HLMI by other methods*

As a catch-all for other possible methods, this section depends on the *yearly chance of new methods being developed* and the *yearly chance each of these new methods reaches HLMI*.

As both of these uncertainties are deeply unknown, we plan on taking a naïve, outside-view approach here. (We are open to alternative suggestions or improvements.) The field of AI arguably goes back to 1956, and over these 65 years, we now have around six proposed methods for how to plausibly get to HLMI (i.e., the other six methods listed here). Other characterizations of the different paths may yield a different number, for example by combining the first two methods into one or breaking them up into a few more, but we expect most other characterizations would not tend to drastically differ from ours—presumably within a factor of 2. Taking the assumption of six methods developed within 65 years at face value, this indicates a yearly chance of developing a new method towards HLMI of ~9%. (Though we note this assumption implies the chances of developing a new method is independent each year to the next, which is questionable.) For the second uncertainty (the chance the methods each reach HLMI in a year), the relationship is also unclear. One possible approach is to simply take the average chance of all the other methods (possibly with a delay of ~20 years for the method to "scale up"), but this will likely be informed by further discussion and elicitation.

## 3.2.2 Outside-View Approaches

In addition to the inside-view estimation for HLMI timelines based on specific pathways, we also estimate HLMI arrival via outside-view methods: *analogies to other developments* (§3.2.2.1) and *extrapolating AI progress and automation* (§3.2.2.2).

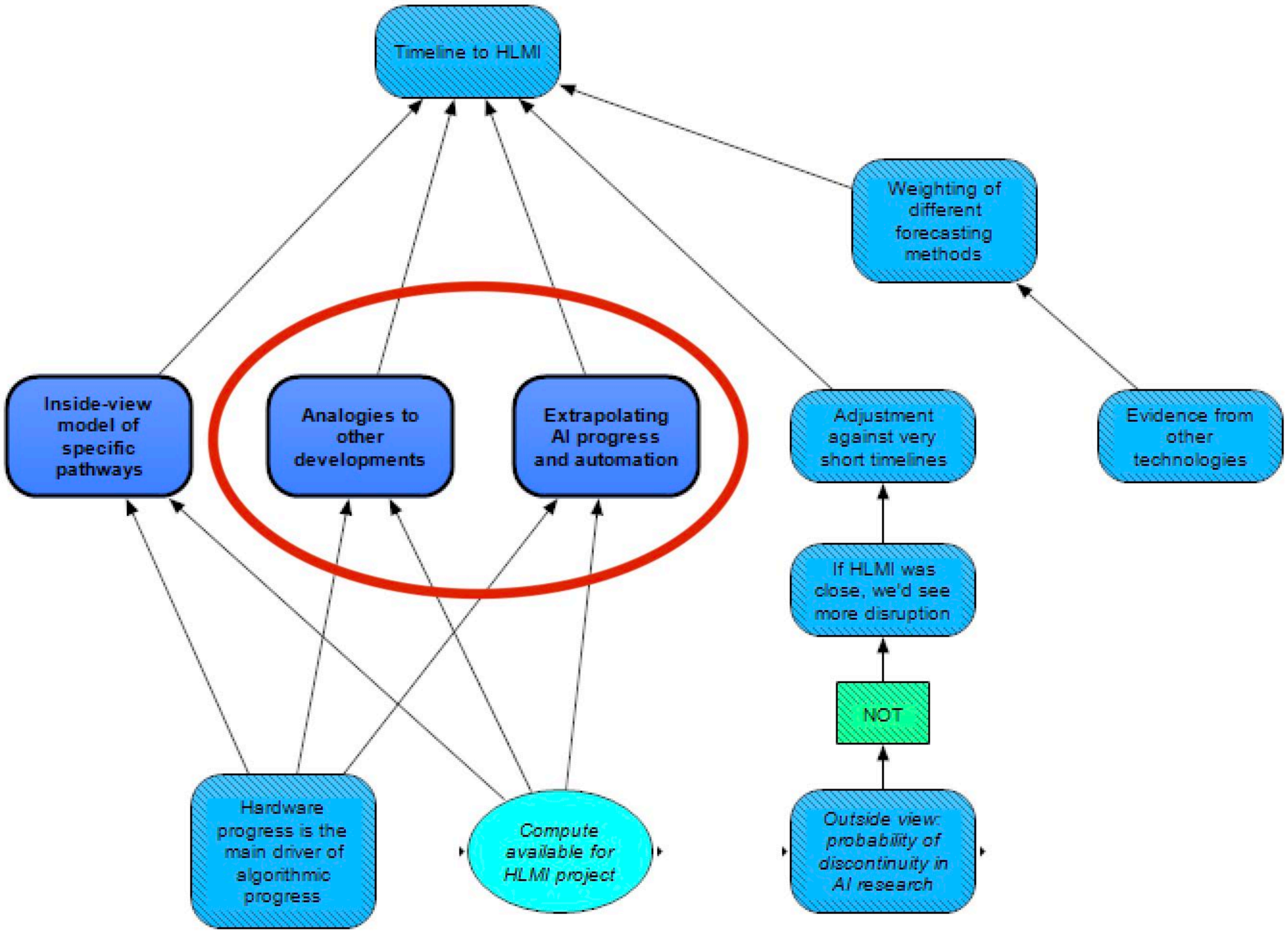

*Figure 36: Outside-view HLMI timeline*

### 3.2.2.1 Analogies to Other Developments

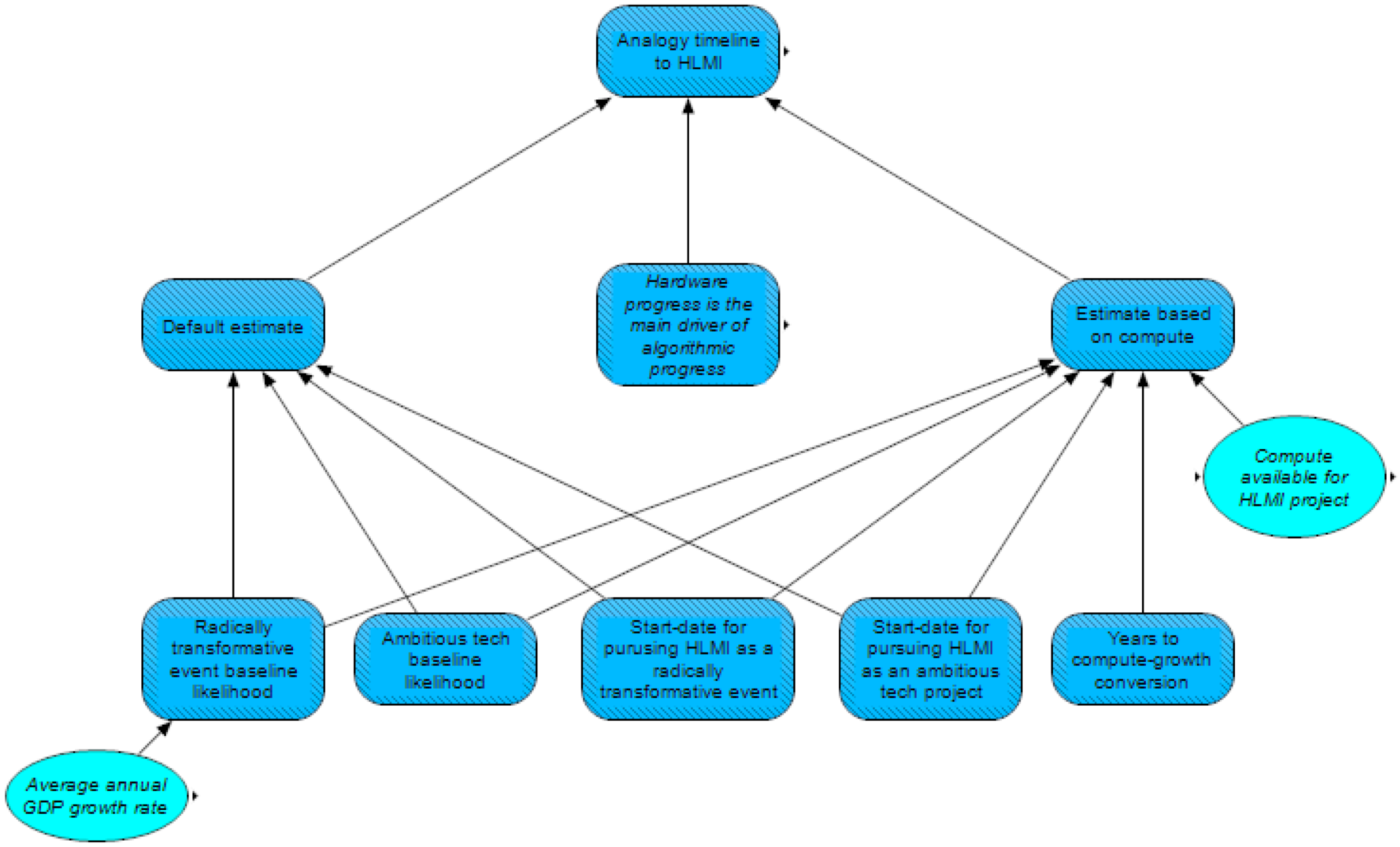

*Figure 37: Analogies to other development*

For this module, we take an approach similar to Tom Davidson's in his Report on Semi-informative Priors [83], though with some simplifications and a few different modeling judgments.

First, the likelihood of HLMI being developed in a given year is estimated by analogizing HLMI to other classes of developments. This initial baseline likelihood is estimated in a very naïve manner—blind to the history of AI, including to the fact that HLMI has not yet been developed, and instead only considering the base rate for "success" in these reference classes. The two classes of events we use (in agreement with Davidson's two preferred classes) are *highly ambitious but presumably physically possible projects seriously pursued by the STEM community* (e.g., harnessing nuclear energy, curing cancer, creating the internet) and *radically transformative events for human work and all of human civilization* (the only examples in this reference class are the Agricultural Revolution and the Industrial Revolution).

For both reference classes, different methods can be used to estimate the baseline yearly likelihood of "success." Here, we sketch out our current plans for making such estimates. For the ambitious-STEM-projects class, the baseline yearly likelihood can be estimated as simply taking the number of successes for such projects (e.g., two for the above three listed ones, as nuclear energy and the internet were successful, while a cure from cancer has not been successful yet) and dividing this number by the sum of the number of years that each project was seriously pursued by the STEM community (e.g., around 5–10 years for pursuing nuclear energy, plus around 50 years for pursuing a cure for cancer, plus perhaps 30 years for the development of the internet, implying, if these three were a representative list, a baseline yearly likelihood of around (2 successes)/(8 + 50 + 30 years) = ~2.3%). Disagreement exists, however, with which projects are fair to include on such a reference-class list as well as when projects may have been "seriously" started and if/when they were successful.

For the radically-transformative-events class, the appropriate calculation is a bit fuzzier. Obviously it wouldn't make sense to simply take a per-year frequentist approach similar to what we did with the ambitious-STEM-projects class—such an estimate would be dominated by the tens of thousands or hundreds of thousands of years it took before the Agricultural Revolution, and it would ignore that technological progress and growth rates have sped up significantly with each transition. Instead, based on the idea [84] that human history can be thought of as divided into a sequence of paradigms with increasing exponential economic growth rates and ~proportionately faster transitions between these paradigms (with the transformative events marking the transition between the different paradigms), we consider that comparisons between paradigms should be performed in terms of economic growth.

That is, if the global economy doubled perhaps ~2-10 times [85] between the first humans and the Agricultural Revolution (depending on whether we count from the beginning of *Homo sapiens* ~300,000 years ago, the beginning of the genus *Homo* ~2 million years ago, or some middle point), and ~8 times [85] between the Agricultural Revolution and the Industrial Revolution, then, immediately after the Industrial Revolution, we might assume that the next transition would similarly occur after around ~5–10 economic doublings. Using similar logic to the ambitious-STEM-projects class, we would perhaps be left with a baseline likelihood of a transformative event per economic doubling of ~2/(6 + 8) = ~14% (taking the average of 2 and 10 for the Agricultural Revolution). If we assume 3% yearly growth in GDP post–Industrial Revolution, the baseline likelihood in terms of economic growth can be translated into a yearly baseline likelihood of ~0.7%. Thus, we have two estimates for baseline yearly likelihoods of the arrival of HLMI, based on two different reference classes.

Second, we consider how the yearly likelihood of developing HLMI may be expected to increase based on how long HLMI has been unsuccessfully pursued. This consideration is based on the expectation that the longer HLMI is pursued without being achieved, the harder it presumably is, and the less likely we may expect it to succeed within the next year. Similar to Davidson, and resembling a modified version of Laplace's rule of succession, we plan to set the probability of HLMI being achieved in a given year as $P = 1/(Y + m)$, where $Y$ is the number of years HLMI has been unsuccessfully pursued, and $m$ is set based on considerations from the baseline yearly likelihood, in a manner described below.

Determining $Y$ is necessarily somewhat arbitrary, as it requires picking a "start date" to the pursuit of HLMI. We consider that the criteria for a start date should be consistent with those for the other developments in the reference class (and need not be the same for the different reference classes). For the ambitious-STEM-project reference class, the most reasonable start date is presumably when the STEM community began seriously pursuing HLMI (to a similar degree that the STEM community started pursuing nuclear energy in the mid-1930s, a cure for cancer in the 1970s, and so on)—arguably this would correspond to 1956 [86]. For the radically-transformative-events reference class, on the other hand, the most reasonable start date is arguably after the Industrial Revolution ended (~1840), as this would be consistent with each of the other transitions being "attempted" from the end of the last transition.

To estimate $m$, we consider that when HLMI is first attempted, we may expect its development to take around as long as implied by the baseline yearly likelihood. Thus, we plan to set $m$ (separately for each of the reference classes) so that the cumulative distribution function of *P* passes the 50% mark in *Y* = 1/(*baseline yearly likelihood*). That is, we assume in the first year that HLMI is pursued, that with 50% odds HLMI will be achieved earlier than implied by the baseline yearly likelihood and with 50% odds it will be achieved later (continuing the examples from above, apparently this would place *m* = 44 for the ambitious-STEM-project reference class, and *m* = 143 for the radically-transformative-events reference class).

Third, we update based on the fact that, as of 2021, HLMI has not yet been developed. For example, if we take the ambitious-STEM-project reference class, and assume that the first date of pursuit is 1956, then for 2022 we set *Y* to (2022 – 1956) = 66 (and 67 for 2023, and so on), and keep *m* at the previous value (implying, continuing from the above dubious example again, a chance of HLMI in 2022 of 1/(66 + 44) = ~0.9%). Similarly for the radically-transformative-events reference class, if HLMI is assumed to initially be pursued in 1840, then for 2022 *Y* is set to 182, and we'd get a chance of HLMI in 2022 of 1/(182 + 143) = ~0.3%.

Finally, to account for the possibility that *hardware progress is the main driver of algorithmic progress*, we duplicate both of these calculations in terms of hardware growth instead of time. In this case, the probability of HLMI being achieved in a given doubling of compute available can be calculated as *P* = 1/(*C* + *n*), where *C* is the number of compute-doublings since HLMI has first been pursued (using a consistent definition for when it initially was pursued as above), and *n* takes the place of *m*, here being valued such that the cumulative distribution function of *P* passes the 50% mark after the number of compute-doublings that we might a priori (when HLMI is first pursued) expect to be needed to achieve HLMI.

For the ambitious-STEM-project class, this latter calculation requires a "conversion" in terms of technological progress between years of pursuit of other ambitious projects and compute-growth for AI. This conversion may be set for the number of years necessary for other projects in the reference class to make similar technological progress as a doubling of compute does for AI.

For the radically-transformative-events reference class, the switch to a hardware-based prediction would simply replace economic growth since the end of the Industrial Revolution with hardware growth since the end of the Industrial Revolution [54] (that is, considering a doubling in compute in the new calculation to take the place of a doubling in global GDP in the old calculation). This comparison may on its face seem absurd, as the first two paradigms are estimated based on GDP, while the third is based on compute (which has recently risen much faster). However, we may consider that the important factor in each paradigm is growth in the main inputs towards the next transition. For the Agricultural Revolution, the most important input was arguably the number of people, and for the Industrial Revolution, it was arguably economic size. Both of these are proxied by GDP, because before the Agricultural Revolution, GDP per capita is assumed to be approximately constant, so growth in GDP would simply track population growth. For the development of HLMI, meanwhile, if it is the case that *hardware progress is the main driver of algorithmic progress*, then the main input towards this transition is compute, so the GDP-based prediction may severely underestimate growth in the relevant metric, and the compute-based estimate may in fact be somewhat justified.

It should be noted that in the compute-based predictions, timelines to HLMI are heavily dependent on what happens to compute going forward, with a potential leveling off of compute implying a high probability of very long timelines.

### 3.2.2.2 Extrapolating AI Progress and Automation

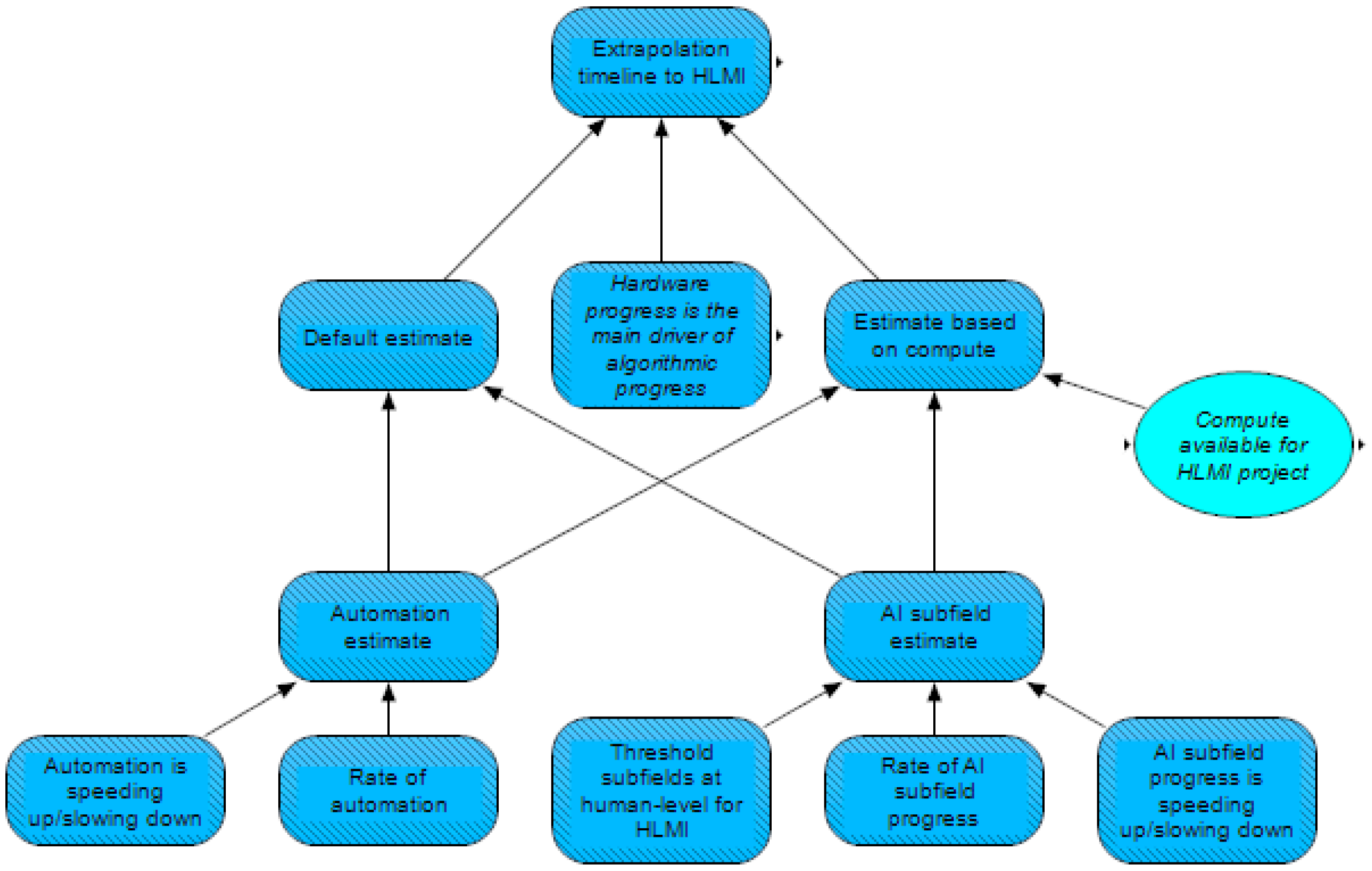

*Figure 38: Extrapolating AI progress and automation*

In this module, HLMI arrival is extrapolated from the rate of automation (i.e., to occur when extrapolated automation hits 100%), and from progress in various AI subfields. Such an extrapolation is perhaps particularly appropriate if one is expecting HLMI to arrive in a piecemeal fashion [87], with different AIs highly specialized for automating specific tasks in a larger economy. Note that questions about the distribution versus concentration of HLMI are handled by a downstream module in our model, covered in a subsequent chapter.

For the extrapolation from automation, this calculation depends on the *rate of automation*, and whether this rate has generally been *speeding up, slowing down, or remaining roughly constant* [88].

For the extrapolation from the progress in various AI subfields [89], [90], a somewhat similar calculation is performed, considering again the *rate of progress* and whether this progress has been *speeding up, slowing down, or remaining roughly constant.* HLMI is assumed to be achieved here when enough subfields are extrapolated to reach human-level performance.

One crux for this calculation, however, is the *threshold for what portion of subfields need to reach human level in the extrapolation* to reach HLMI. One could imagine that all subfields may need to reach human level, as HLMI wouldn't be achieved until AI was as good as humans in all these subfields, but this threshold would introduce a couple problems. First, if certain subfields happen to be dominated by researchers that are more conservative in their estimations, then the most pessimistic subfield could bias the results to be too conservative. Second, it's possible that some subfields are bottlenecked by sufficient success in other subfields and will see very slow progress before these other subfields reach sufficient capabilities. Alternatively, one could imagine that the response from the median subfield may be the least biased and also be the most appropriate for gauging "general" intelligence. On the other hand, if AI achieving human-level competence on half of the domains did not translate into similar competence in other domains, then this would not imply HLMI.

Similar to the above section on analogies to other developments (§3.2.2.1), the extrapolation here is done in two forms: as a default case, it is performed in terms of time, but if *hardware progress is the main driver of algorithmic progress*, then the extrapolation is performed in terms of (the logarithm of [91]) compute. In the latter scenario, a slowdown in compute would lead to a comparable slowdown in AI capability gains, all else equal.

## 3.3 Bringing It All Together

Here, we have examined several methods for estimating the timeline to HLMI: a gears-level model of specific pathways to HLMI, analogizing HLMI to other developments in plausibly similar reference classes, and finally, extrapolating current trends in automation and AI progress. Disagreements exist regarding the proper weight for these different forecasting methods, and further disagreements exist for many factors underlying these estimations, including some factors that appear in several different places (e.g., the progression of hardware (§3.1) and cruxes related to Analogies and General Priors on Intelligence (§2)).

In addition to estimating the timeline to HLMI, our model also allows for estimating the method of HLMI first achieved—though such an estimate only occurs in one of our three forecasting methods (we welcome suggestions for how to make this estimate in model runs where other methods are "chosen" by the Monte Carlo method dice roll, as this information is important for downstream nodes in other parts of our model).

In the next chapter, we will discuss AI takeoff speeds and discontinuities around and after HLMI.

# 4 Takeoff Speeds and Discontinuities

Samuel Dylan Martin, Daniel Eth

The goal of this part of the model is to describe the different potential characteristics of a transition from (pre-)HLMI to superintelligent AI (i.e., "AI takeoff"). We also aim to clarify the relationships between these characteristics and explain what assumptions they are sensitive to.

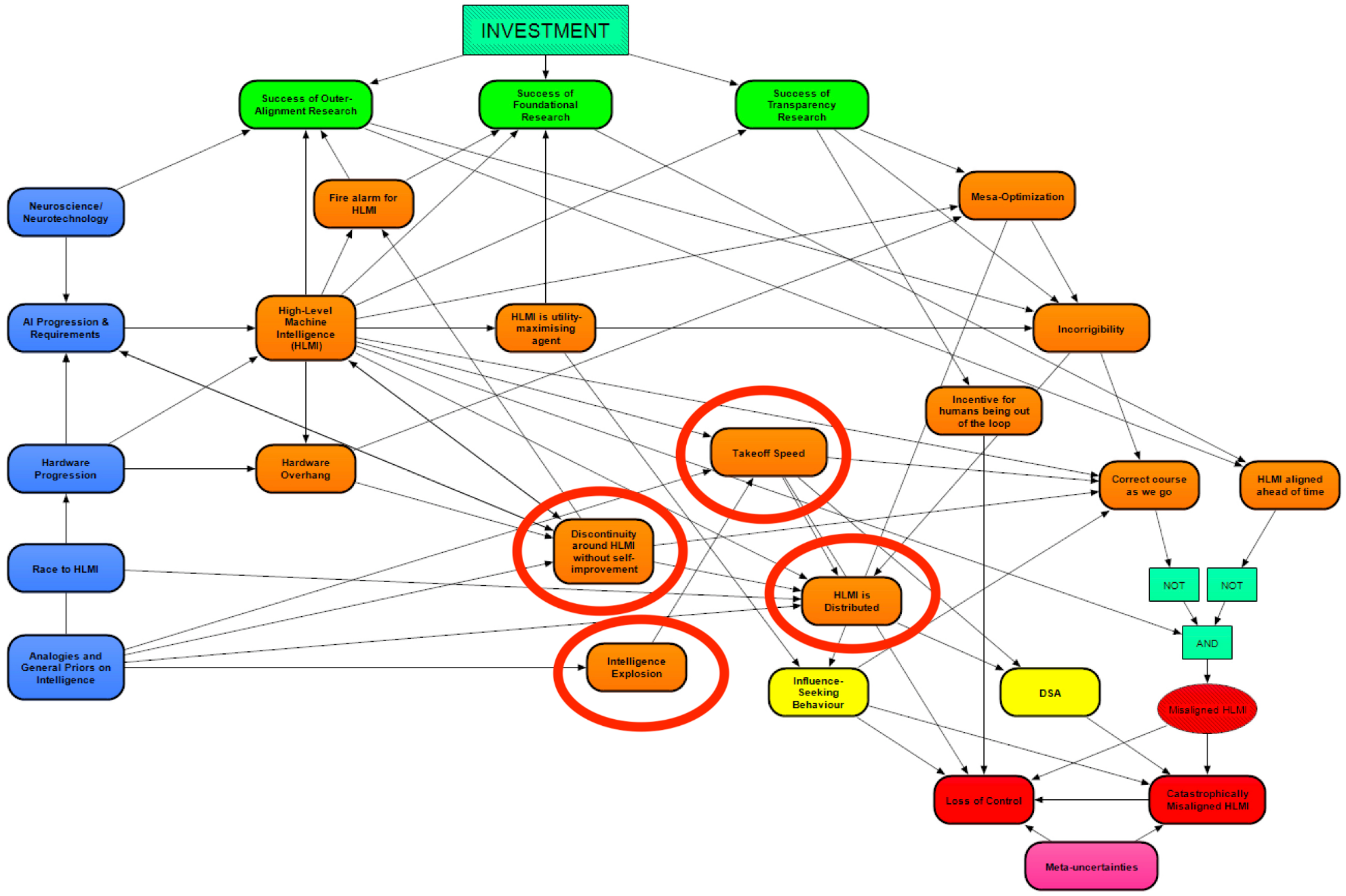

*Figure 39: Module focus for Takeoff Speeds and Discontinuities modules*

As shown in Figure 39, the relevant sections of the model (circled in red) take inputs primarily from these modules:

- *Analogies and General Priors on Intelligence* (§2): this module concerns both arguments that compare HLMI development with other previous developments (for example, the evolution of human intelligence, current progress in machine learning, and past historical or cultural transitions) and broad philosophical claims about the nature of intelligence.
- *High-Level Machine Intelligence (HLMI)* (§3): this module concerns whether HLMI will be developed at all, and if so, which type of HLMI and when.
- *Hardware Overhang*: this module concerns whether, at the time HLMI is developed, the amount of hardware required to run HLMI and the availability of such hardware will lead to a situation of "hardware overhang," where overwhelming hardware resources allow for the first HLMI(s) to be overpowered due to being cheaply duplicated, easily run at a high clock speed, or otherwise rapidly scaled up.

The circled modules in Figure 39 correspond to this chapter's outputs of concern:

- *Intelligence Explosion* (§4.1.2): will a positive feedback loop involving AI capabilities lead these capabilities to grow roughly hyperbolically [92] across a sufficient range, such that the capabilities eventually grow incredibly quickly to an incredibly high level (presumably before plateauing as they approach some fundamental limit)?
- *Discontinuity around HLMI without self-improvement* (§4.1.1): will there be a rapid jump in AI capabilities from pre-HLMI AI to HLMI (for instance, if pre-HLMI AI acts like a machine with missing gears) and/or from HLMI to much higher intelligence (for instance, if a hardware overhang allows for rapid capability gain soon after HLMI)?
- *Takeoff Speed* (§4.3.1): how fast will the global economy (or the next closest thing, if this concept doesn't transfer to a post-HLMI world) grow, once HLMI has matured as a technology?
- HLMI is Distributed (§4.3.2): will AI capabilities in a post-HLMI world be dispersed among many comparably powerful systems? A negative answer to this node indicates that HLMI capabilities will be concentrated in a few powerful systems (or a single such system).

These outputs provide a rough way of characterizing AI takeoff scenarios. While they are non-exhaustive, we believe they are a simple way of characterizing the range of outcomes which those who have seriously considered AI takeoff tend to find plausible. For instance, we can summarize (our understanding of) the publicly espoused views of Eliezer Yudkowsky, Paul Christiano, and Robin Hanson (along with a skeptic position) as shown in Table 1.

*Table 1: Four views on takeoff scenarios*

| | **Eliezer Yudkowsky** | **Paul Christiano** | **Robin Hanson** | **Skeptic** |
|---|---|---|---|---|
| **Intelligence Explosion** | Yes | Yes | No | No |
| **Discontinuity Around HLMI without Self-Improvement** | Yes | No | No | No |
| **Takeoff Speed** | ~hyperbolically increasing<br>(No significant intermediate doublings) | ~hyperbolically increasing<br>(with complete intermediate doublings on the order of ~1 year) | Doubling time of ~weeks to months | Doubling time on the order of ~years or longer |
| **HLMI is Distributed** | No | Yes | Yes | Yes |

These outputs—in addition to outputs from other sections of our model, such as those covering misalignment—impact further downstream sections of our model relevant to failure modes.

In a previous chapter, we explored the *Analogies and General Priors on Intelligence* module (§2), which includes very important input for the modules in this chapter. That module outputs answers to four key questions (which are used throughout this module):

- The difficulty of marginal intelligence improvements at the approximate human level (i.e., around HLMI)
- Whether marginal intelligence improvements become increasingly difficult beyond HLMI at a rapidly growing rate or not
  - "Rapidly growing rate" is operationalized as becoming difficult exponentially or faster than exponentially
- Whether there is a fundamental upper limit to intelligence that is not significantly above the human level
- Whether, in general, further improvements in intelligence tend to be bottlenecked by previous improvements in intelligence rather than some external factor (such as the rate of physics-limited processes)

In the next section, we discuss the *Intelligence Explosion* (§4.1.2) and *Discontinuity around HLMI without self-improvement* (§4.1.1) modules, which are upstream of and which influence the other modules covered in this chapter: *HLMI is Distributed* (§4.3.2) and *Takeoff Speed* (§4.3.1).

# 4.1 Early Modules (Discontinuity and Intelligence Explosion)

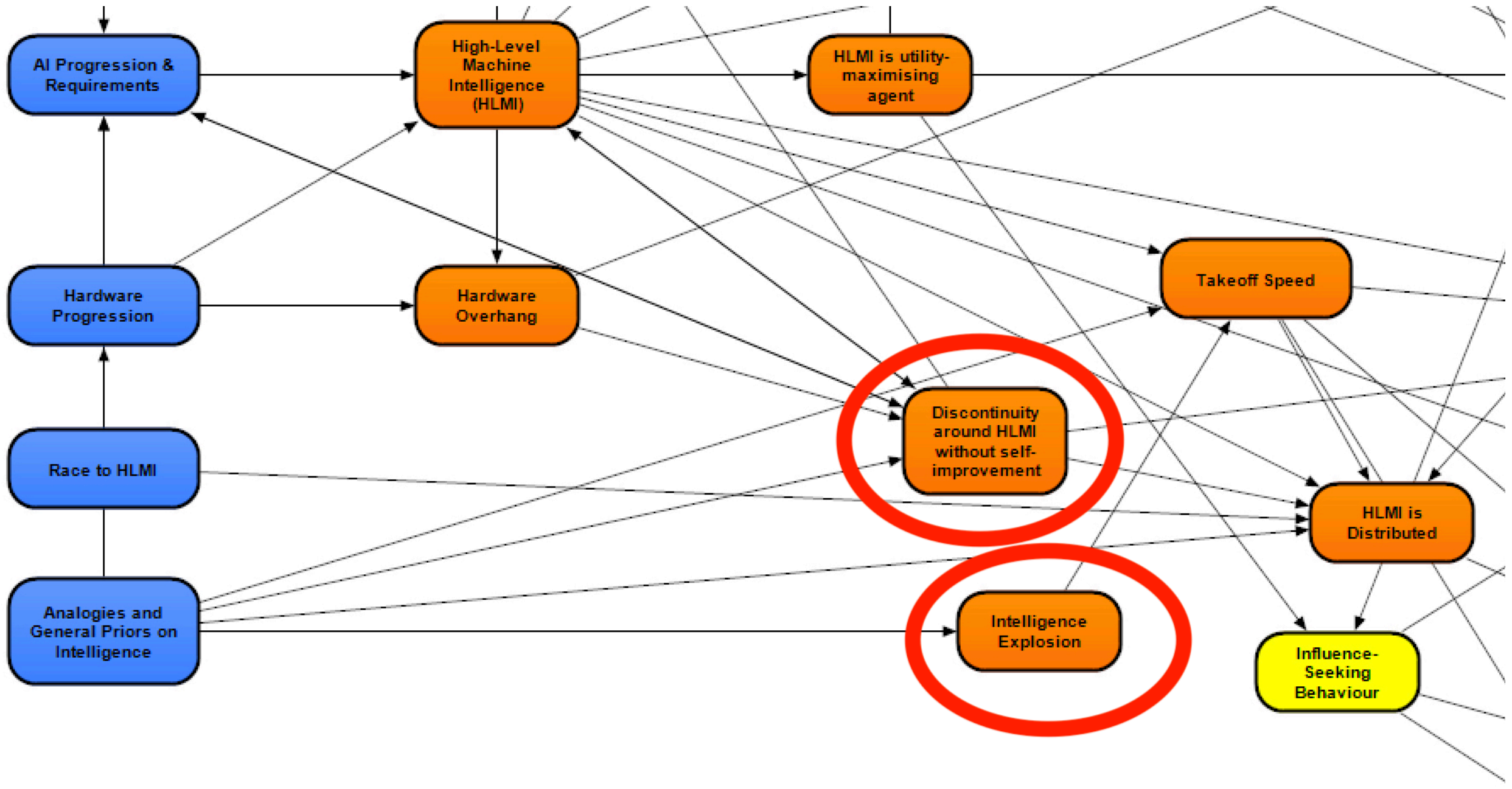

*Figure 40: Discontinuity and Intelligence Explosion modules*

## 4.1.1 Discontinuity around HLMI without Self-Improvement

This module aims to answer the question: will the first HLMI (or a very early HLMI) represent a discontinuity in AI capabilities from what came before? We define a discontinuity as a very large and very sudden jump in AI capabilities, not necessarily a mathematical discontinuity but a phase change caused by a significantly quicker rate of improvement than implied by the previous trend. Note that this module does not consider rapid self-improvement from quick feedback loops (that is considered instead in the module on *Intelligence Explosion* [§4.1.2]) but is instead concerned with large jumps occurring around the time of the HLMI generation of AI systems.

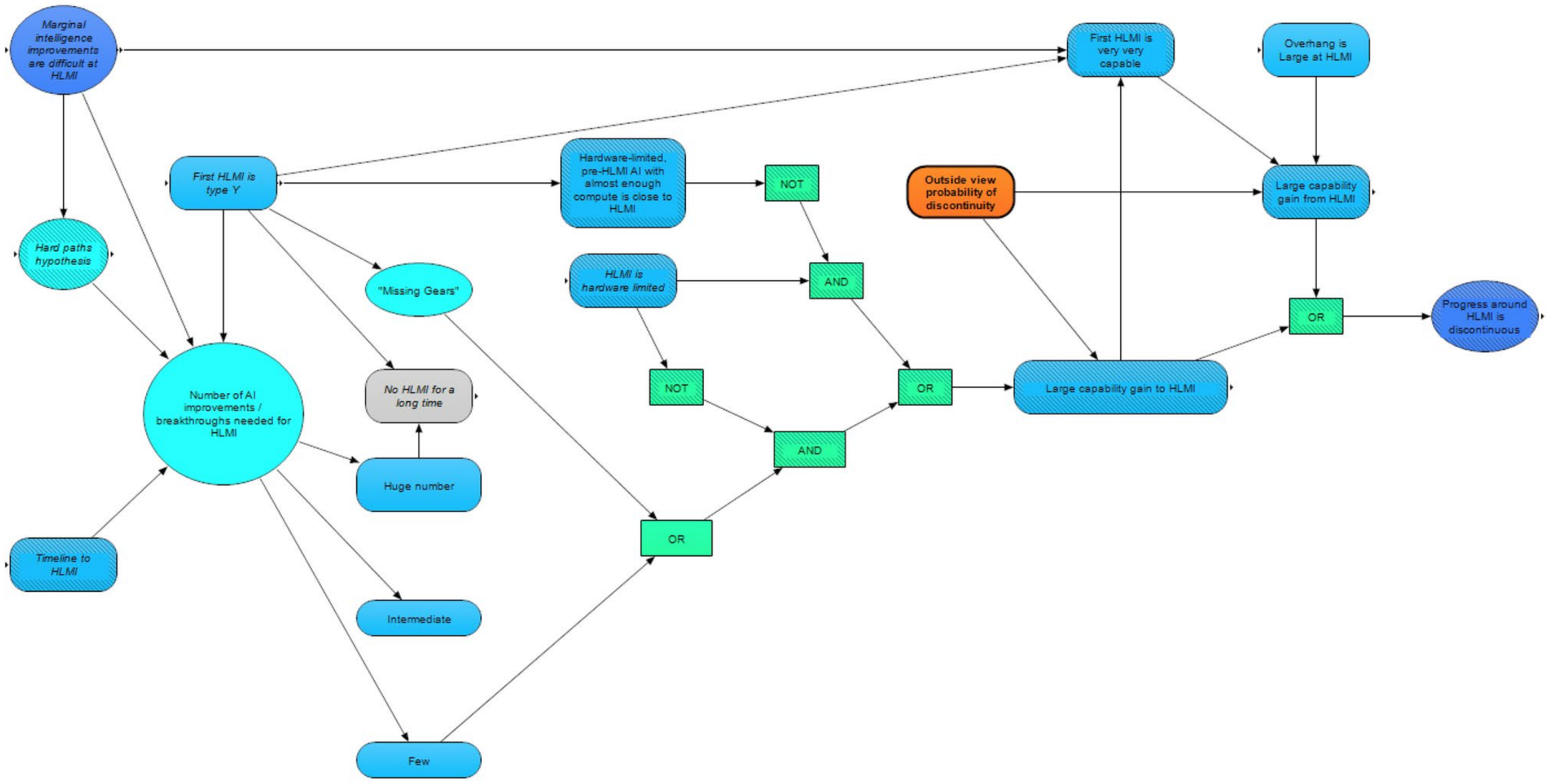

*Figure 41: Discontinuity around HLMI without Self-Improvement*

Such a discontinuity could be from a jump in capabilities to HLMI or a jump from HLMI to significantly higher capabilities (or both). A jump in capabilities from HLMI could occur either if the first HLMI "overshoots" the HLMI level and is very capable (which likely depends on both the type of HLMI and whether marginal intelligence improvements are difficult around HLMI) or if a large hardware overhang allows for the first HLMI to scale (in quality or quantity) in such a way that it is far beyond the HLMI level in capabilities.

Figure 42 shows a zoomed-in view of the two routes (either a large capability gain from HLMI, or a large capability gain to HLMI):

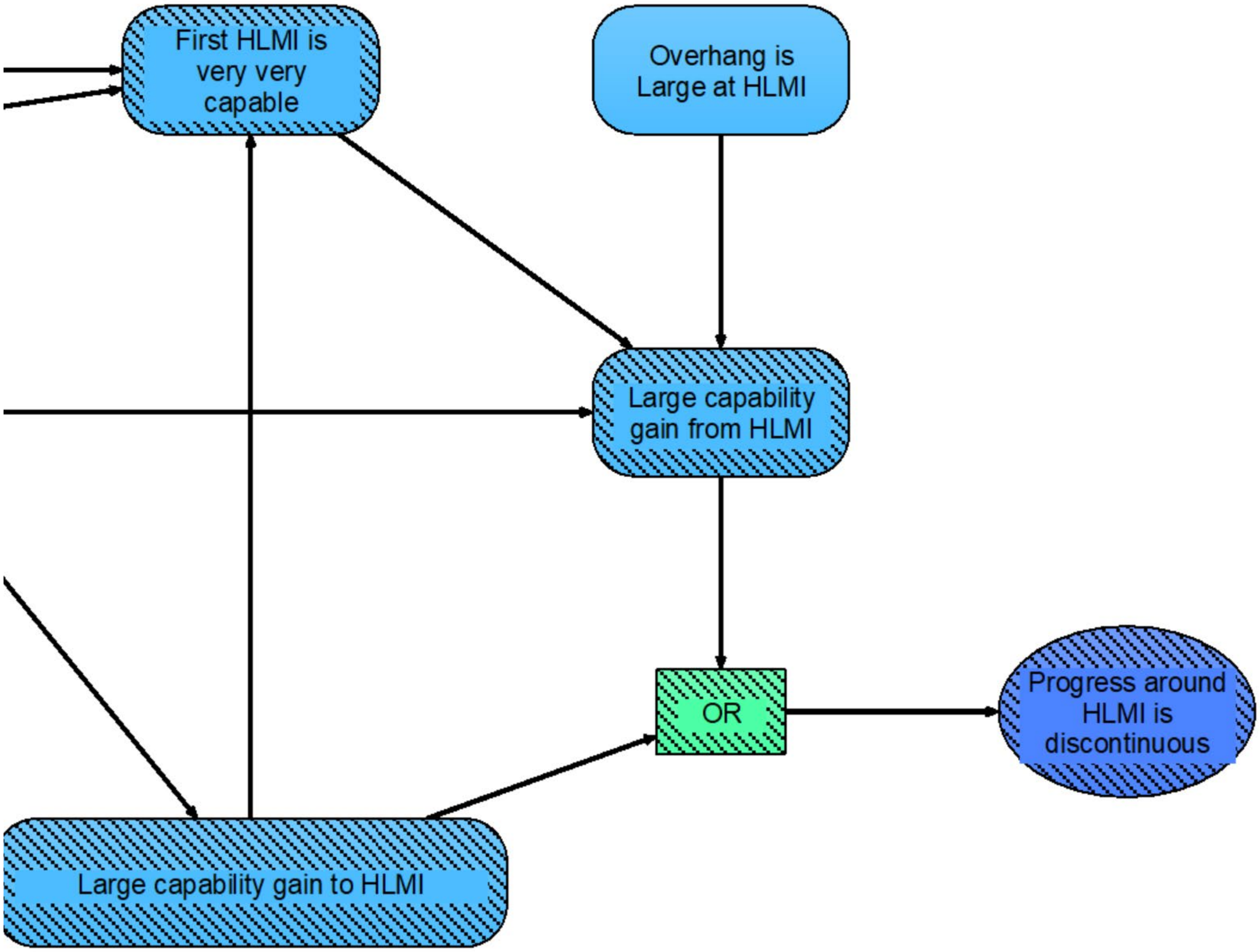

*Figure 42: Discontinuous progress around HLMI*

Regarding capability jumps to HLMI, we see a few pathways, which can be broken down by whether HLMI will ultimately be bottlenecked on hardware or software (i.e., which of hardware or software will be last to fall into place—as determined by considerations in our chapter on *paths to HLMI* (§3)). If HLMI will be bottlenecked on hardware (circled in green in Figure 43), then the question reduces to whether pre-HLMI AI with almost enough compute has almost as strong capabilities as HLMI. To get a discontinuity from hardware-limited HLMI, the relationship between increasing abilities and increasing compute has to diverge from an existing trend to reach HLMI (i.e., hardware-limited, pre-HLMI AI with somewhat less compute is much less capable than HLMI with the required compute). We suspect that whether this crux is true may depend on the type of HLMI in question (e.g., statistical methods might be more likely to gain capabilities if scaled up and run with more compute).

If HLMI is software limited, on the other hand, then instead of hardware, we want to know whether the last software step(s) will result in a large jump in capabilities. This could happen either if there are very few remaining breakthroughs needed for HLMI (circled in magenta in Figure 43) such that the last step(s) correspond to a large portion of the problem, or if the last step(s) act as “missing gears” putting the rest of the system in place (circled in black).

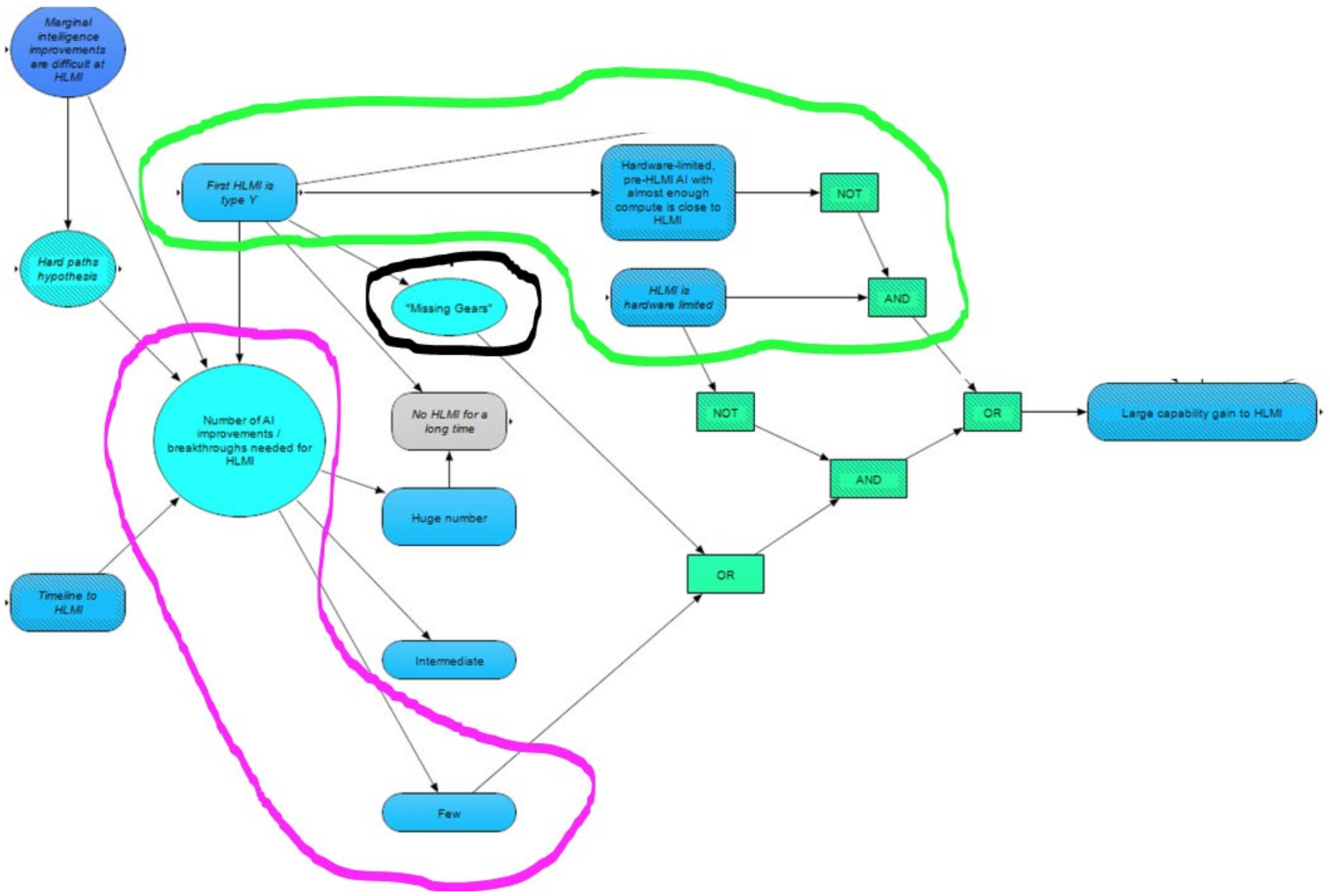

*Figure 43: Large capability gain to HLMI module*

We suspect that whether the last step(s) present a "missing gears" situation is likely to depend on the type of HLMI realized. A (likely) example of "missing gears" would be whole brain emulation (WBE) [53], where 99% of the way towards WBE presumably doesn't get you anything like 99% of the capabilities of WBE. (See here [93] for an extended discussion of the relationship between "fundamental breakthroughs" and "missing gears.") If the "missing gears" crux resolves negatively, however, then determining whether there will be a large capability gain to HLMI is modeled as depending on the number of remaining breakthroughs needed for HLMI.

We make the simplifying assumption that the remaining fundamental breakthroughs are all of roughly comparable size, such that the more breakthroughs needed, the less of a step any individual breakthrough represents. This means that the last breakthrough—the one that gives us HLMI—might either take us from AI that is greatly inferior to HLMI all the way to HLMI (if there are "few" key breakthroughs needed), or just be an incremental improvement on pre-HLMI AI that is already almost as useful as HLMI (if there are an "intermediate" or "huge number" of key breakthroughs needed).

We consider several lines of evidence to estimate whether HLMI requires few or many key breakthroughs.

To start, the type of HLMI being developed influences the number of breakthroughs we expect will be needed. For instance, if HLMI is achieved with current deep learning methods plus business-as-usual advancements (§3.2.1.2), then, *ceteris paribus*, we'd expect fewer breakthroughs needed to reach HLMI than if it is achieved via WBE (§3.2.1.5) [53].

As well as depending on the type of HLMI developed, our model assumes the expected number of breakthroughs needed for HLMI is influenced significantly by the *difficulty of marginal intelligence improvements at HLMI* (in the *Analogies and General Priors on Intelligence* module [§2]). If marginal intelligence improvements are difficult at HLMI, then more separate breakthroughs are probably required for HLMI.

Lastly, the hard paths hypothesis and timeline to HLMI (as determined in the *Paths to HLMI* modules [§3]) each influence our estimate of how many breakthroughs are needed to reach HLMI. The hard paths hypothesis [64] claims that it's rare for environments to straightforwardly select for general intelligence—if this hypothesis is true, then we'd expect more steps to be necessary (e.g., for crafting the key features of such environments). Additionally, short timelines would imply that there are very few breakthroughs remaining, while longer timelines may imply more breakthroughs needing to be found (remember [§1.2.2], it's fine if this logic feels "backwards," as our model is not a causal model *per se*, and instead arrows represent probabilistic influence).

#### 4.1.1.1 How Many Breakthroughs?

We don't place exact values on the number of breakthroughs needed for HLMI in either the "few," "intermediate," or "huge number" cases. This is because we have not yet settled on a way of defining "fundamental breakthroughs," nor of estimating how many would be needed to shift the balance on whether there would be a large capability gain to HLMI.

Our current plan for characterizing the number of breakthroughs is to anchor the "intermediate" answer to "number of remaining breakthroughs" as similar to the number of breakthroughs that have so far occurred in the history of AI [15]. If we identify that there have been three major paradigms in AI so far (knowledge engineering, deep search, and deep learning), and maybe ten times as many decently sized breakthroughs (within deep learning, this means things like convolutional neural networks [CNNs], the transformer architecture, deep Q-networks [DQNs]) to get to our current level of AI capability, then an "intermediate" case would imply similar numbers to come. From this, we have:

- Few breakthroughs: 0–2 major paradigms and 0–9 new breakthroughs
- Intermediate: Similar number of breakthroughs as so far, or somewhat more: 3–9 new paradigms and 10–100 breakthroughs
- Huge number: More than 10 paradigms or 100 breakthroughs

Another way to estimate the number of remaining breakthroughs is to use expert opinion. For example, Stuart Russell identifies 4 remaining fundamental breakthroughs [94] needed for HLMI (although these "breakthroughs" seem more fundamental than those listed above, and might correspond to a series of breakthroughs as defined above):

> "We will need several conceptual breakthroughs, for example in language or commonsense understanding, cumulative learning (the analog of cultural accumulation for humans), discovering hierarchy, and managing mental activity (that is, the metacognition needed to prioritize what to think about next)."

Note that the "Few Breakthroughs" case includes situations where 0 breakthroughs are needed—that is, cases where current deep learning with only minor algorithmic improvements and somewhat increased compute gives us HLMI [95].

### 4.1.2 Intelligence Explosion

This module aims to answer the question of will there eventually be an intelligence explosion? We define an "intelligence explosion" as a process by which HLMI successfully accelerates the rate at which HLMI hardware or software advances, to such a degree that the rate of progress approaches vertical over a large range.

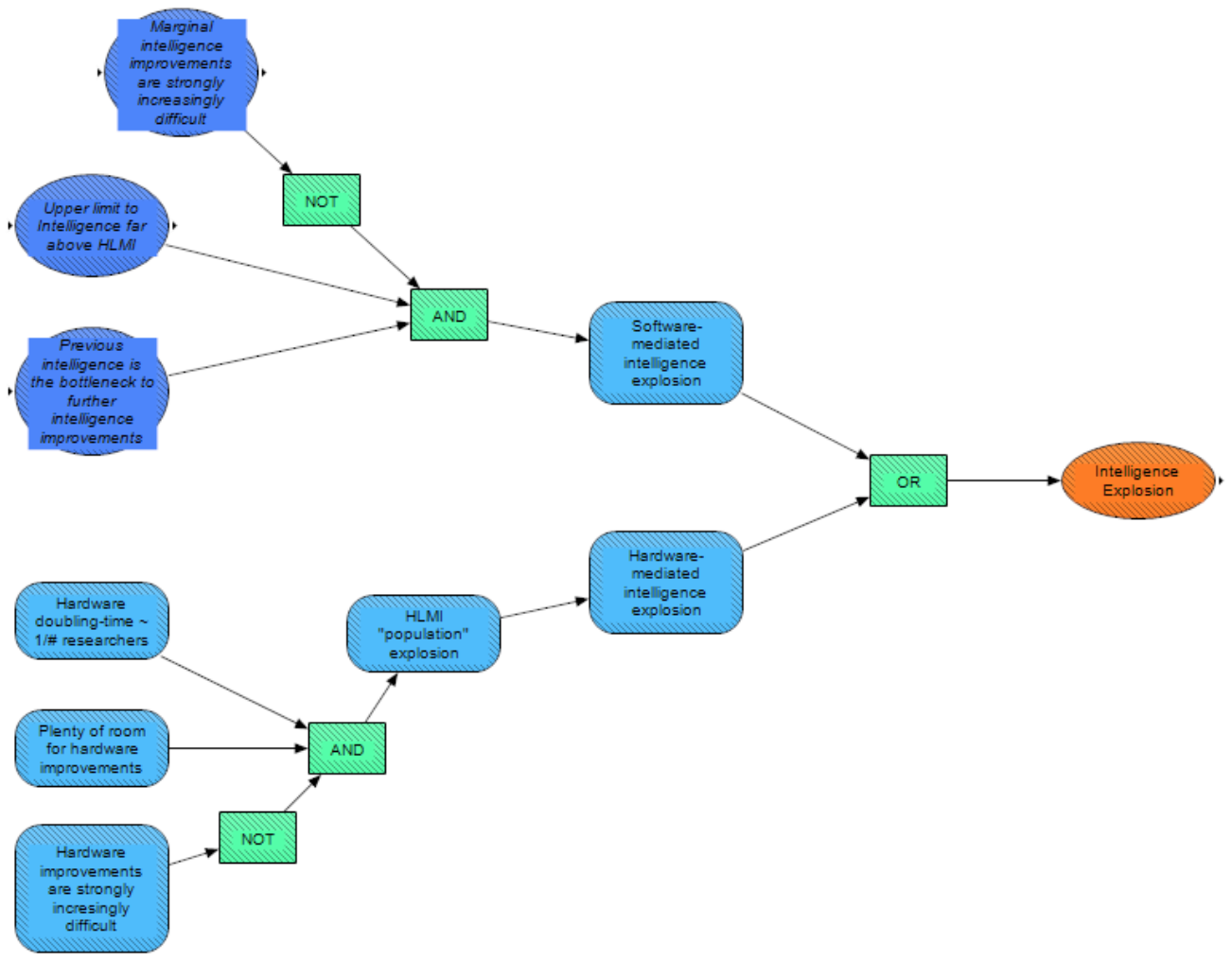

*Figure 44: Intelligence explosion*

In our model, whether an intelligence explosion eventually occurs does not directly depend on what type of HLMI is first developed (as we assume that if one type of HLMI could not achieve an intelligence explosion while another could, even if the first type of HLMI is achieved first, the latter type—if possible to build—will eventually be built and cause an intelligence explosion then). Our model considers two paths to an intelligence explosion—a software-mediated path and a hardware-mediated one.

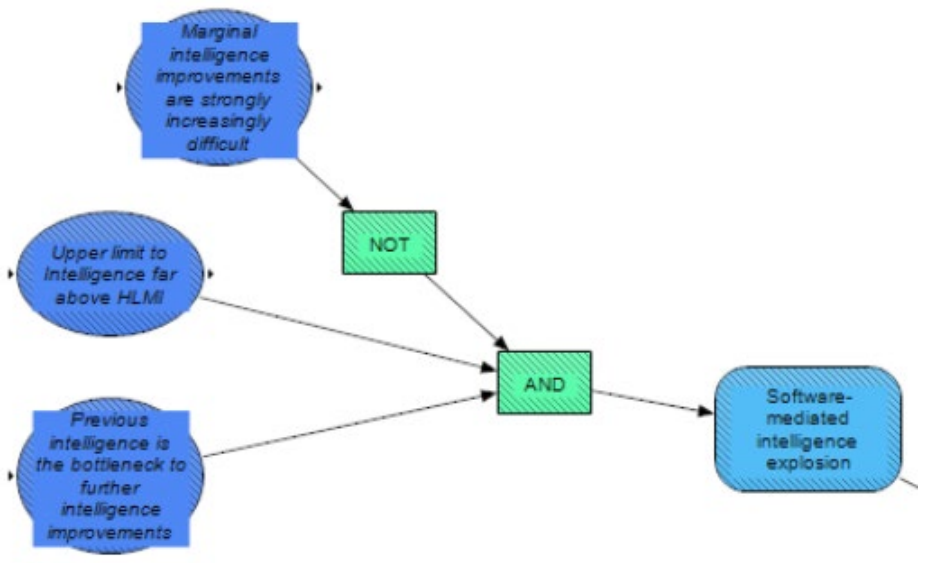

*Figure 45: Software-mediated intelligence explosion*

Under the software-mediated path, our model assumes that, due to AI accelerating the rate of AI progress, there will be explosive growth in intelligence, if HLMI is developed and

- marginal intelligence improvements are not strongly increasingly difficult,
- there are no other theoretical limits to increasing general intelligence or the practical capabilities of a general intelligence (at least none barely above the human level), and
- further intelligence improvements are bottlenecked by previous intelligence (as opposed to, say, physical processes that cannot be sped up).

In such a scenario, the positive feedback loop from "more intelligent AI" to "better AI research (performed by the more intelligent AI)" to "even more intelligent AI still" would be explosive—in effect, HLMI could achieve sustained returns on cognitive reinvestment [6], at least over a sufficiently large range.

In addition to the software-mediated path, an intelligence explosion could occur due to a hardware-mediated path.

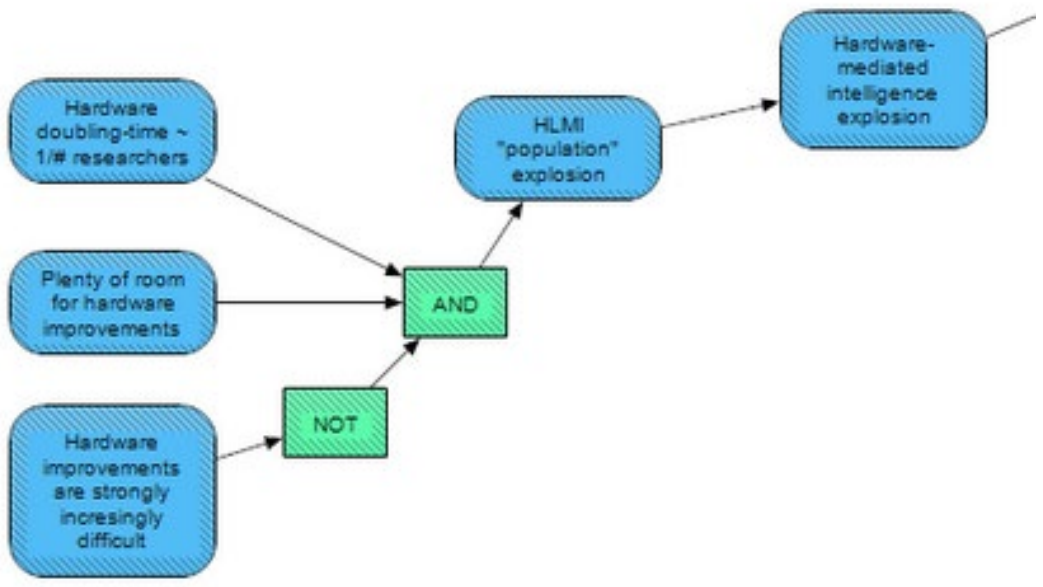

*Figure 46: Hardware-mediated intelligence explosion*

In this scenario, HLMI (doing hardware R&D) would cause an explosion in the amount of hardware, and thus an explosion in the "population" of HLMI [6] (implying more HLMIs to perform hardware research, faster hardware gains, and so on). This phenomenon would require hardware improvements to scale with the number of hardware researchers (with this work being performed by HLMIs), for hardware improvements not to become strongly increasingly difficult, and for there to be plenty of room for more hardware improvements. Such a pathway would allow, at least in principle, the capabilities of AI to explode even if the capability of any given AI (with a fixed amount of hardware) did not explode.

Note that *Intelligence Explosion* as defined in this model does not necessarily refer to an instantaneous switch to an intelligence explosion immediately upon reaching HLMI—an intelligence explosion could occur after a period of slower post-HLMI growth with intermediate doubling times. Questions about immediate jumps in capabilities upon reaching HLMI are handled by the *Discontinuity* module (§4.1.1).

## 4.2 Comparing Discontinuity and Intelligence Explosion

The distinction in our model between a discontinuity and an intelligence explosion can be understood from the graphs in Figure 47, which show rough features of how AI capabilities might advance over time given different resolutions of these cruxes. Note that while these graphs show the main range of views our model can express, they are not exhaustive of these views (e.g., the graphs show the discontinuity going through HLMI, while it's possible that a discontinuity would instead simply go to or from HLMI).

Additionally, the model is a simplification, and we do not mean to imply that progress will be quite as smooth as the graphs imply—we're simply modeling what we expect to be the most important and cruxy features. Take these graphs as qualitative descriptions of possible scenarios as opposed to quantitative predictions—note that the y-axis (AI "intelligence") is an inherently fuzzy concept (which perhaps is better thought of as increasing on a log scale) and that the dotted blue line for "HLMI" might not occupy a specific point as implied here but instead a rough range. Further remember that we're not talking about economic growth, but AI capabilities (which feed into economic growth).

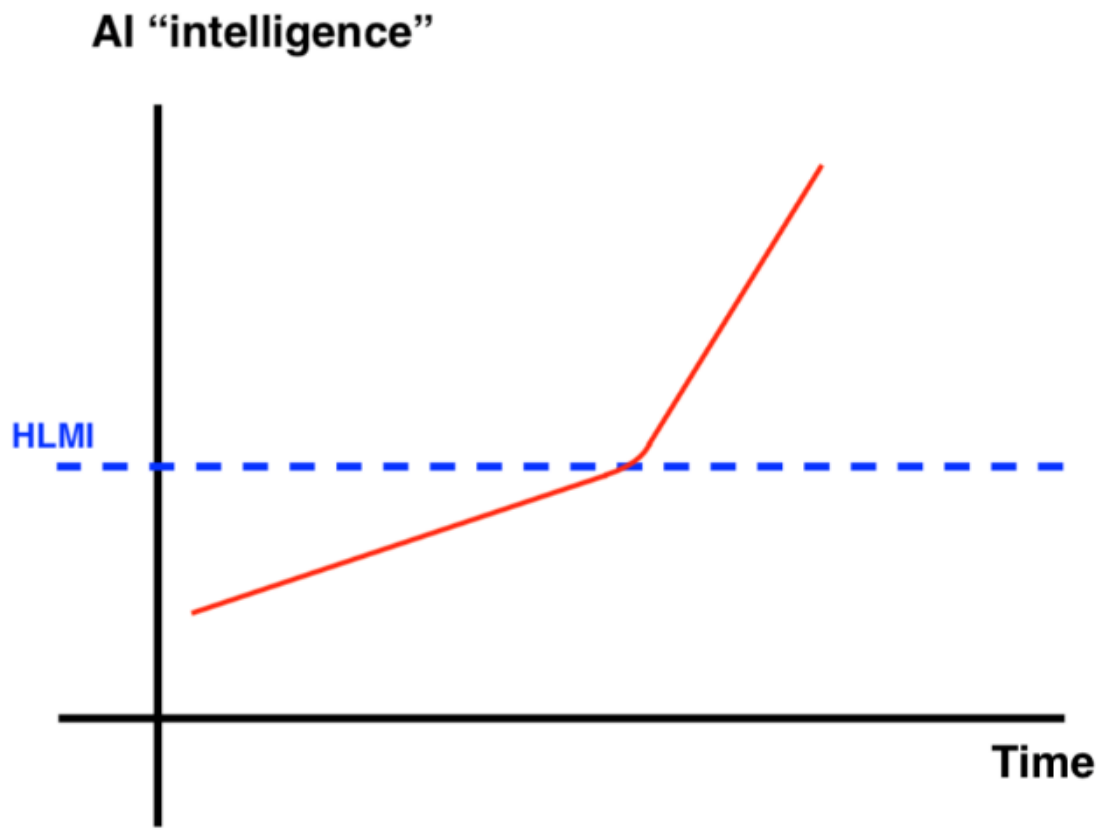

*a) No Intelligence Explosion, No Discontinuity*

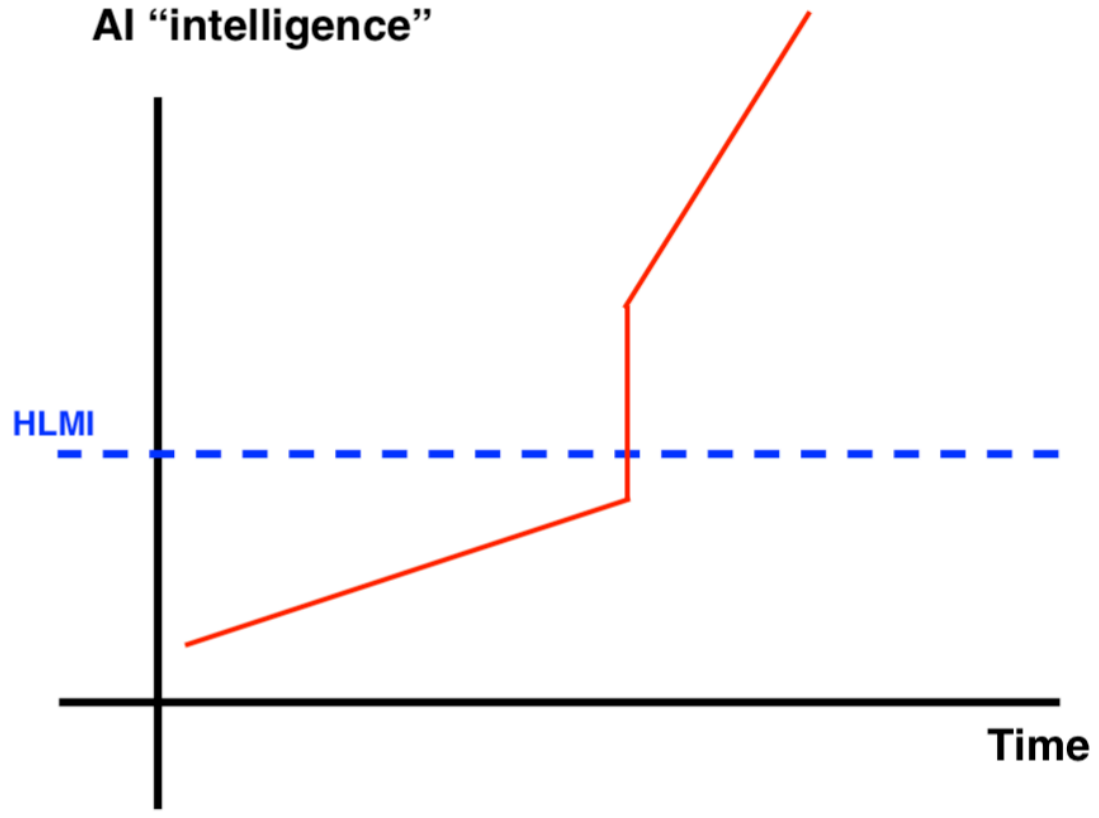

*b) No Intelligence Explosion, Discontinuity*

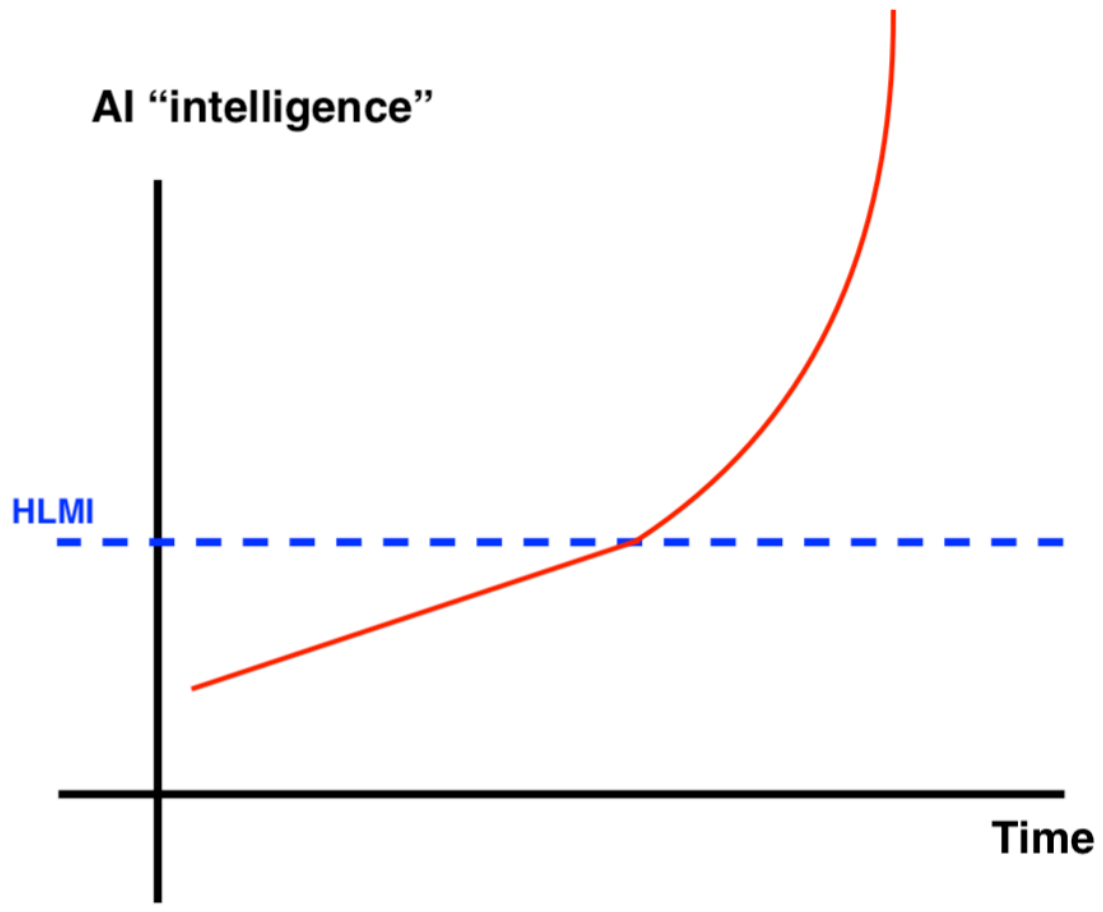

*c) Intelligence Explosion, No Discontinuity*

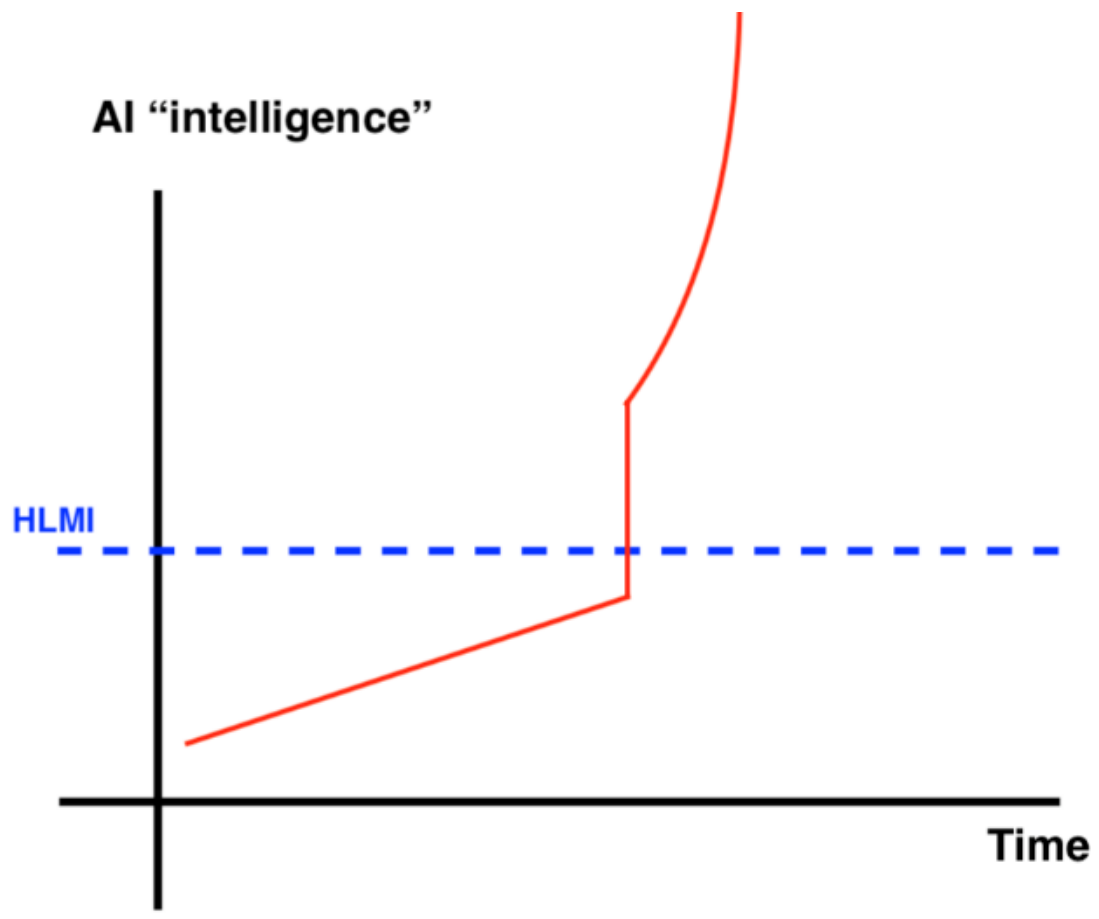

*d) Intelligence Explosion, Discontinuity*

*Figure 47: Comparing discontinuity and intelligence explosion, four scenarios*

# 4.3 Later Modules (Takeoff Speed and HLMI is Distributed)

The modules from the previous section act as inputs to those in this section: *Takeoff Speed* (§4.3.1) and *HLMI is Distributed* (§4.3.2).

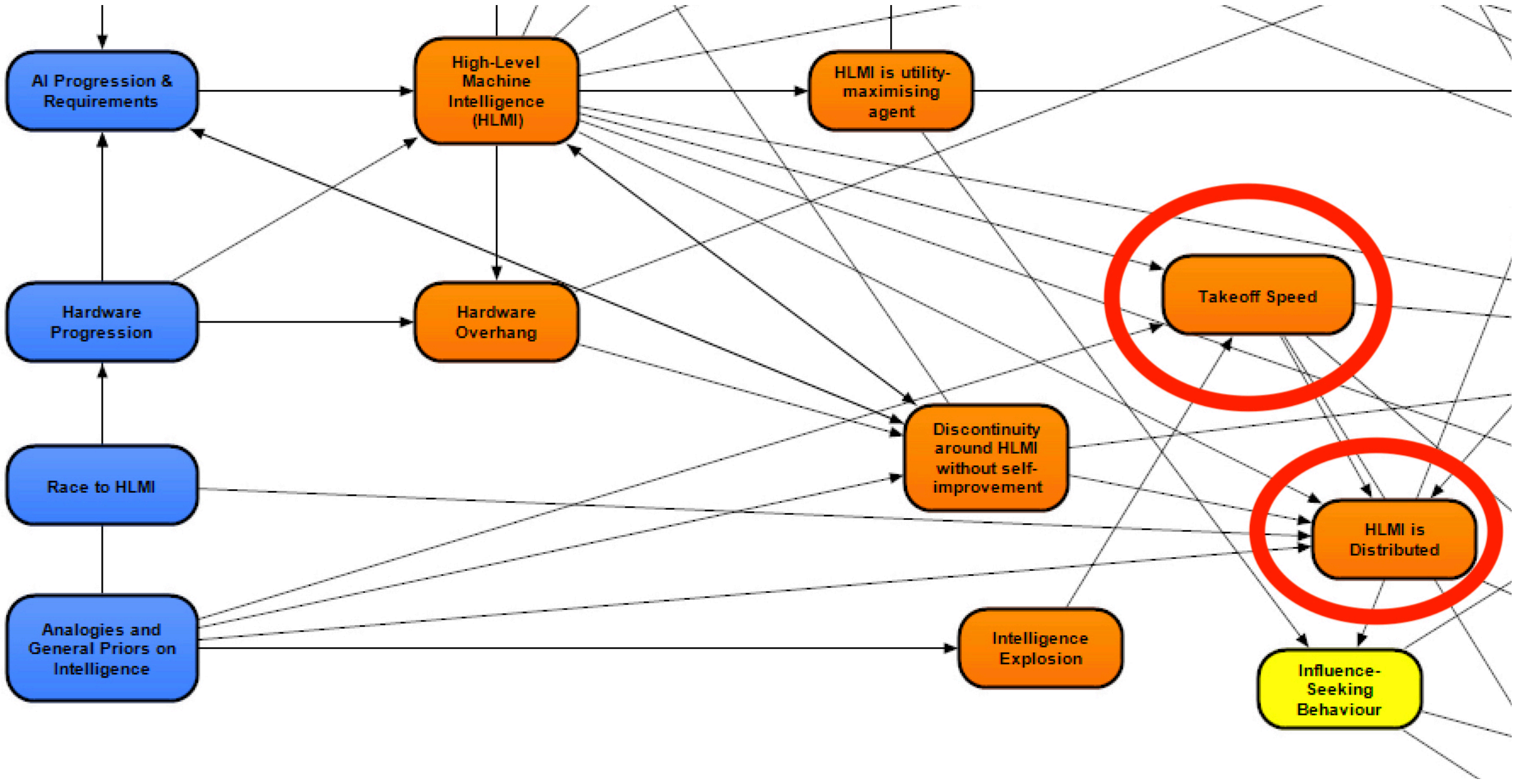

*Figure 48: Takeoff Speed and HLMI is Distributed modules*

## 4.3.1 Takeoff Speed

We have seen how the model estimates the factors which are important for assessing the takeoff speed:

- Will there be an intelligence explosion?
- From *Analogies and General Priors* (§2):
    - How difficult are marginal intelligence improvements at and beyond HLMI?
    - What is the upper limit to intelligence?

This module aims to combine these results to answer the question of what will happen to economic growth post-HLMI? Will there be a new, faster economic doubling time (or equivalent) and, if so, how fast will it be? Alternatively, will growth be roughly hyperbolic (before running into physical limits)? To clarify, while previously we were considering changes in AI capabilities, here we are examining the resultant effects for economic growth (or similar). This discussion is not *per se* premised on considerations of GDP [96] measurement.

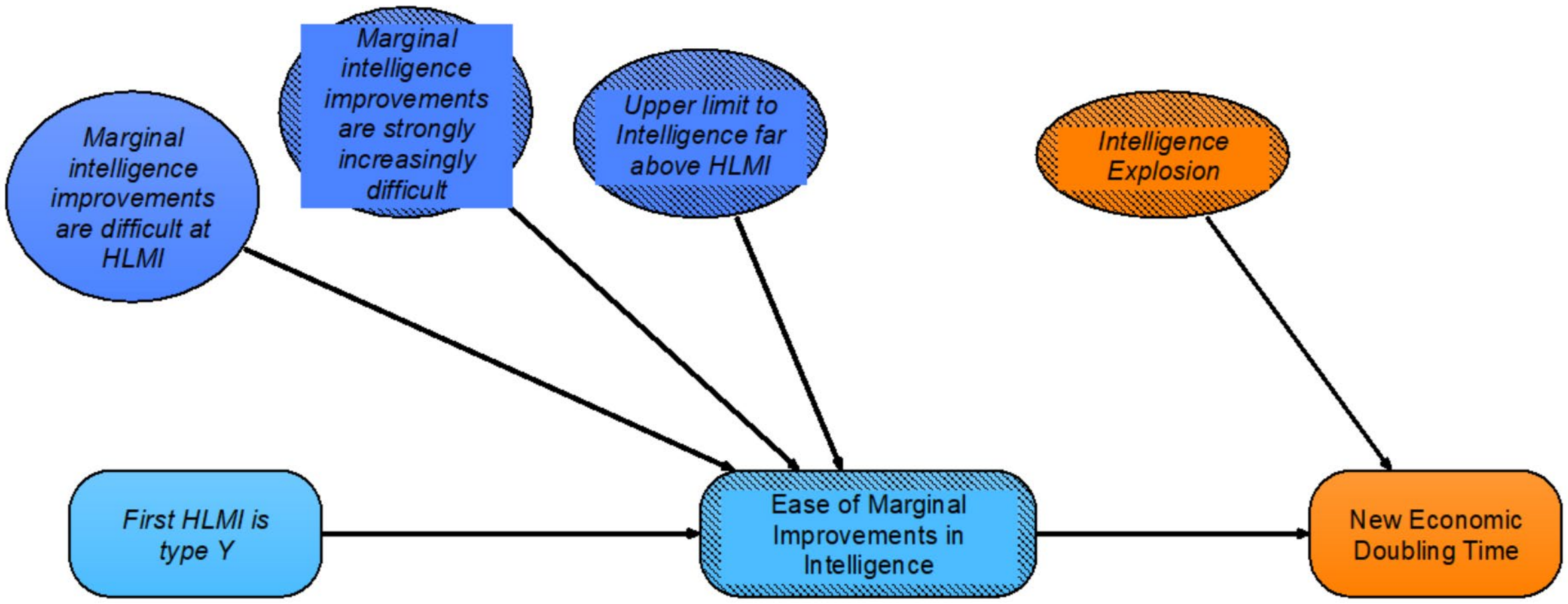

*Figure 49: New economics doubling time*

If the *Intelligence Explosion* module (§4.1.2) indicates an intelligence explosion, then we assume economic growth becomes roughly hyperbolic along with AI capabilities (i.e., increasingly short economic doubling times).

If there is not an intelligence explosion, however, then we assume that there will be a switch to a new mode of exponential economic growth, and we estimate the speedup factor for this new growth rate compared to the current growth rate based largely on outside-view estimates of previous transitions in growth rates (the Agricultural Revolution and the Industrial Revolution—on this outside view alone, we would expect the next transition to bring us an economic doubling time of perhaps days to weeks [84]). This estimate is then updated based on an assessment of the overall ease of marginal improvements post-HLMI.

If

- marginal intelligence improvements are not strongly increasingly difficult,
- there is no significant upper limit to HLMI capability, and
- marginal intelligence improvements are not difficult at HLMI,

then we conclude that the transition to more powerful HLMI looks faster, all else equal, and update our outside-view estimate regarding the economic impact accordingly (and similarly we update towards slower growth if these conditions do not apply). We plan to use these considerations to create a lognormally-distributed estimate of the final growth rate, given that we are uncertain over multiple orders of magnitude regarding the post-HLMI growth rate, even in a world without an intelligence explosion.

The connection between economic doubling time and the overall intelligence/capability of HLMI is not precise. We think our fuzzy assessment is appropriate, however, since we're only looking for a ballpark estimate here (and due to the lognormal uncertainty, the results of our model should be robust to small differences in these parameters).

## 4.3.2 HLMI is Distributed

This module aims to answer the question of is HLMI "distributed by default"? That is, do we expect (ignoring the possibility of a Manhattan Project–style endeavor that concentrates most of the world's initial research effort) to see HLMI capability distributed throughout the world or highly localized into one or a few leading projects?

In later chapters, we will synthesize predictions about the two routes to highly localized HLMI: the route explored in this chapter (i.e., HLMI not being distributed by default) and an alternative route, explored in a later chapter, where most of the world's research effort is concentrated into one project. We expect that if HLMI is distributed by default and research effort is not strongly concentrated into a few projects, then many powerful HLMIs will be around at the same time.

Several considerations in this section are taken from Intelligence Explosion Microeconomics [6] (in section 3.9 "Local versus Distributed Intelligence Explosions").

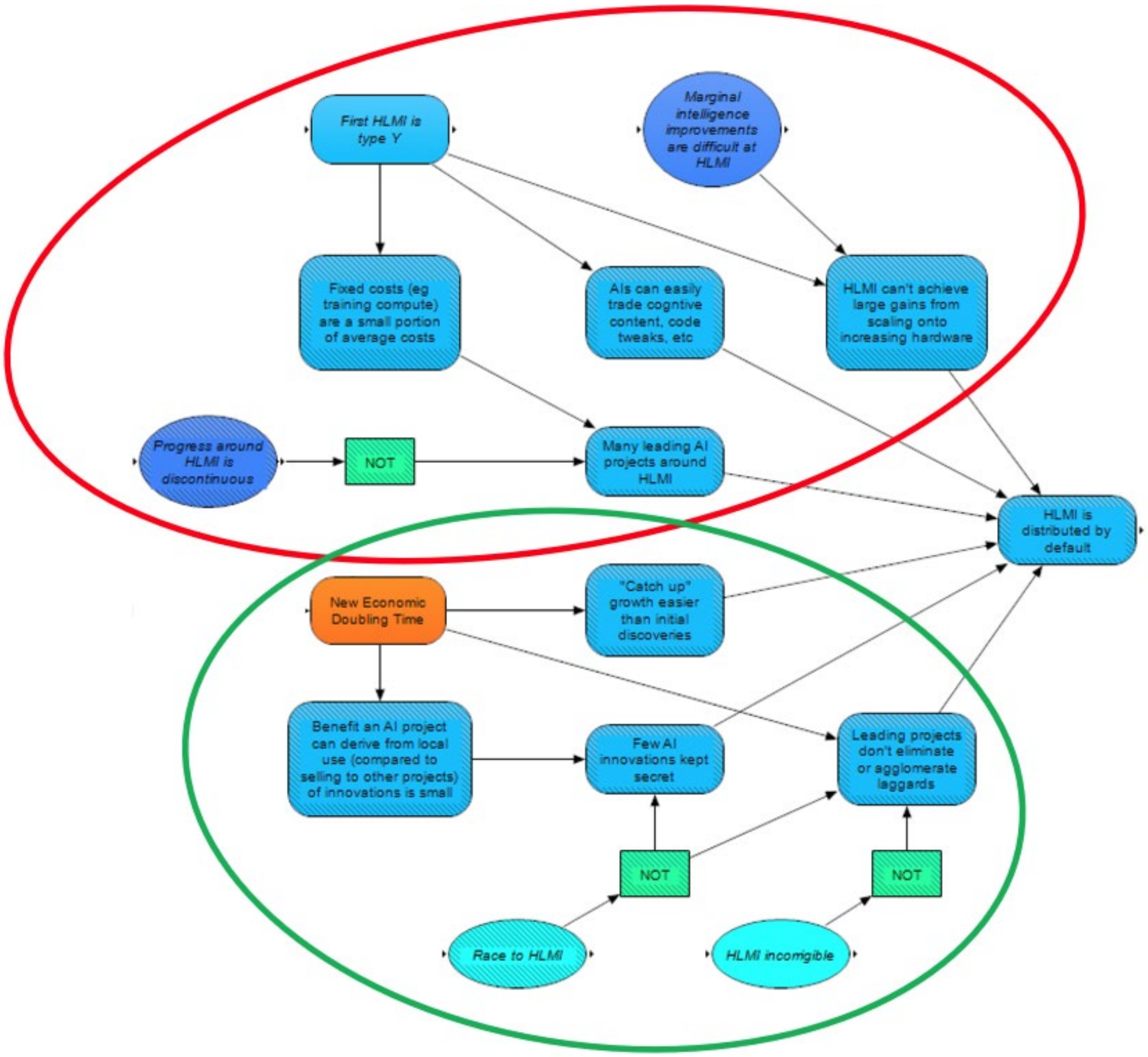

Figure 50: HLMI is distributed by default

Arguments about the degree to which HLMI will be distributed by default can be further broken up into two main categories: those heavily influenced by the economic takeoff speed and the possibility of an intelligence explosion (mostly social factors, circled in green in Figure 50), and those not heavily influenced by the economic takeoff speed (mostly technical factors, circled in red). We should note that while takeoff speed indirectly affects the likelihood of HLMI distribution through intermediate factors, it does not *directly* affect whether HLMI will be distributed; even in the case of an intelligence (and therefore economic) explosion, it's still possible that progress could accelerate uniformly [97] such that no single project has a chance to pull ahead.

Here, we will first examine the factors not tied to takeoff speed, before turning to the ones that are.

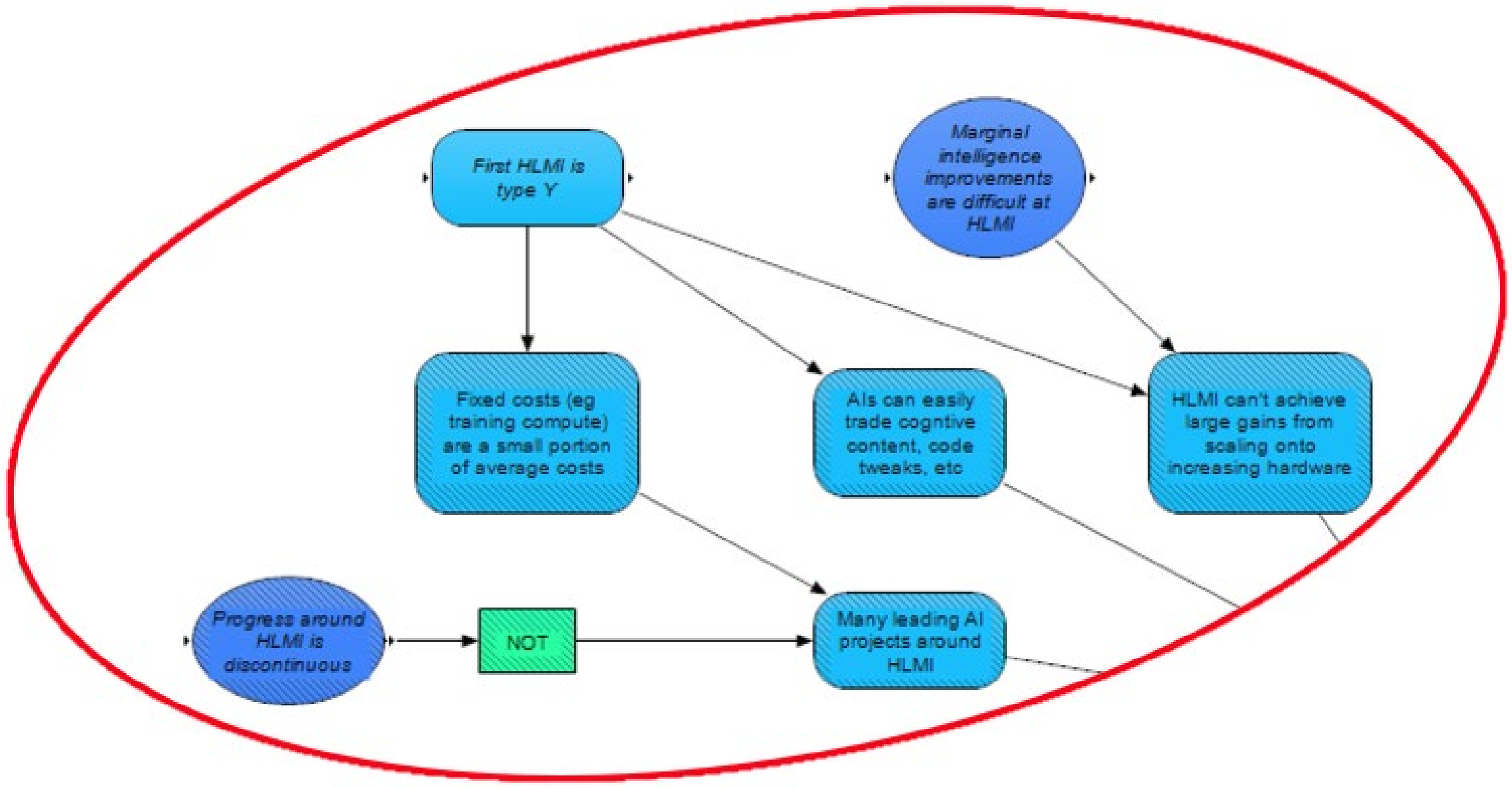

*Figure 51: HLMI can't achieve large gains from hardware scaling*

A significant consideration is whether there will be a discontinuity in AI capabilities around HLMI. If there is a discontinuity, then it is highly likely that HLMI will not initially be distributed by default, because one project will presumably reach the discontinuity first. We model this as there being only one leading AI project to begin with. Even if progress around HLMI is continuous, however, there could still be only a few leading projects going into HLMI, especially if fixed costs are a large portion of the total costs for HLMI (presumably affected by the kind of HLMI), since high fixed costs may present a barrier to there being many competitor projects.

HLMI is also more likely to be distributed if AIs can easily trade cognitive content, code tweaks, and so on (this likelihood is also presumably influenced by the type of HLMI) because if so, the advantages that leading projects hold may be more likely to be distributed to other projects.

Finally, if HLMI can achieve large gains from scaling onto increasing hardware, then we might expect leading projects to increase their leads over competitors, as profits could be reinvested in more hardware (or compute may be seized by other means), and thus HLMI may be expected to be less distributed. We consider that the likelihood of large gains from further hardware is dependent on both the type of HLMI and the difficulty of marginal improvements in intelligence around HLMI (with lower difficulty implying a greater chance of large gains from increasing hardware).

Then, there are the aforementioned factors which are heavily influenced by the takeoff speed (which in turn is influenced by whether there will be an intelligence explosion):

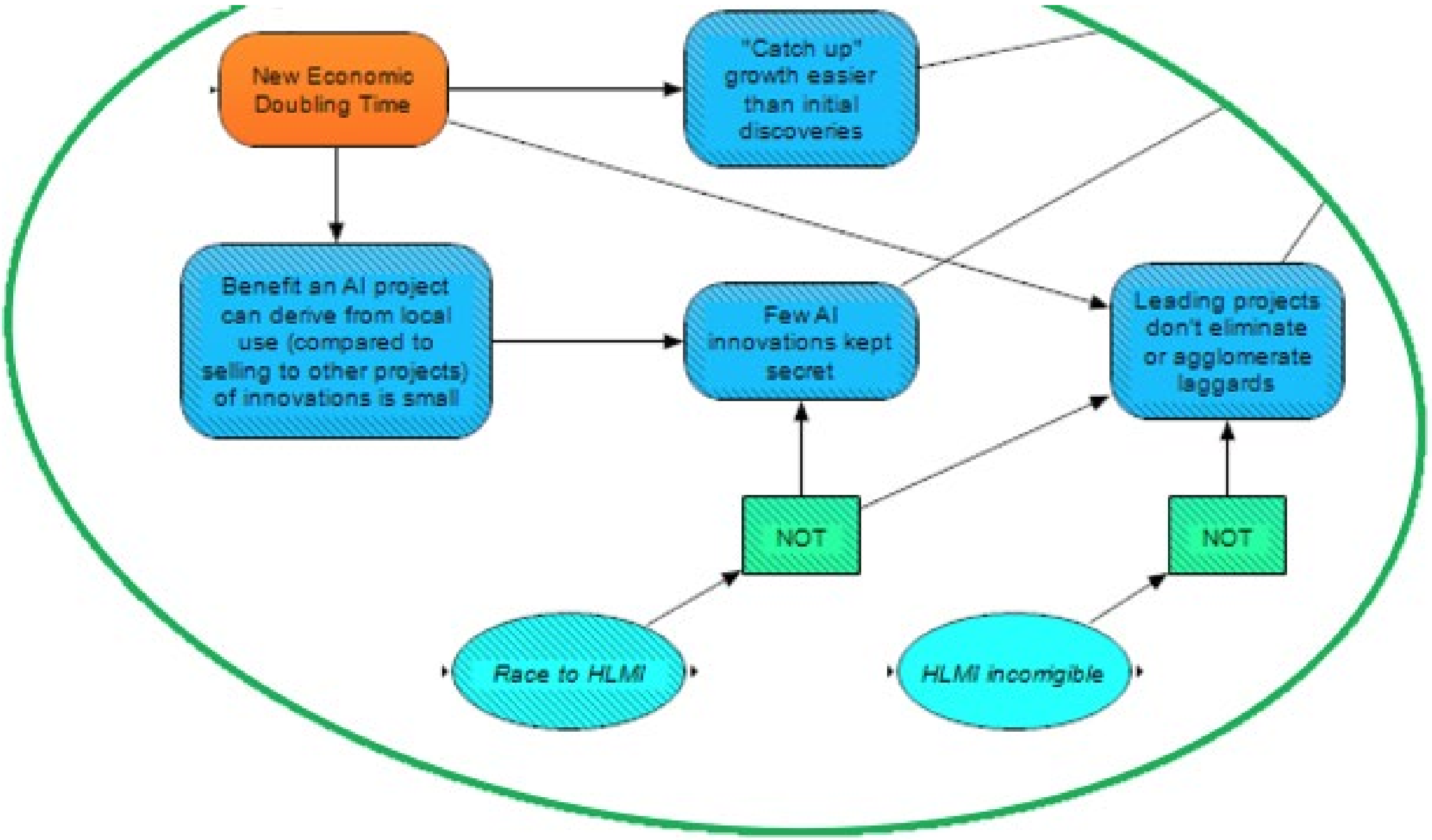

*Figure 52: Factors influenced by takeoff speed*

If catch-up innovation based on imitating successful HLMI projects is easier than discovering methods of improving AI in the first place, then we would expect more distribution of HLMI, as laggards may successfully play catch-up. A faster doubling time—and, in particular, an intelligence explosion—may push against this, as projects that begin to gather a lead may "pull ahead" more easily than others can catch up (we expect a faster takeoff to accelerate cutting-edge AI development by more than it accelerates the rest of the economy).

If major AI innovations tend to be kept secret, then this also pushes against HLMI being distributed. We may consider that a race to HLMI may encourage more secrecy between competitors. Additionally, secrecy may be more likely if AI projects can derive larger benefits from using its innovations locally than from selling its innovations to other projects. Local use may more likely be larger if there are shorter economic doubling times or an intelligence explosion, as such scenarios imply large returns from cognitive reinvestment.

Finally, we consider that distributed HLMI is less likely if leading projects eliminate or agglomerate laggards. Again, a race dynamic probably makes this scenario more likely. Additionally, if HLMI is incorrigible, it might be more likely to "psychopathically" eliminate laggards via actions that projects with corrigible HLMI might opt to avoid.

## 4.4 Conclusion

To summarize, this chapter examines key questions related to AI takeoff: whether there will be a discontinuity in AI capabilities to and/or from HLMI, whether there will be an intelligence explosion due to feedback loops post-HLMI, the growth rate of the global economy post-AI takeoff, and whether HLMI will be distributed by default. These estimates use a mixture of inside-view and outside-view considerations.

In building our model, we have made several assumptions and simplifications, as we’re only attempting to model the main cruxes. Naturally, this does not leave space for every possible iteration on how the future of AI might play out.

In the next chapter, we discuss risks from mesa-optimization.

# 5 Modeling Risks from Learned Optimization

Ben Cottier

This chapter explains how risks from learned optimization are incorporated in our model. The relevant part of the model is mostly based on the Risks from Learned Optimization sequence [98] and paper [99] (henceforth RLO). Although we considered responses and alternate perspectives [100], [101] to RLO in our research, these perspectives are not currently explicitly modeled.

For those not familiar with the topic, a *mesa-optimizer* is a learned algorithm that is itself an optimizer. According to RLO, inner alignment [102] is the problem of aligning the objective of a mesa-optimizer with the *base optimizer's* objective (which may be specified by the programmer). A contrived example [103] supposes we want an algorithm that finds the shortest path through any maze. In the training data, all mazes have doors that are red, including the exit. Inner misalignment arises if we get an algorithm that efficiently searches for the next red door—the capabilities are robust because the search algorithm is general and efficient, but the objective is not robust [104] because it finds red doors rather than the exit.

The relevant part of our model is contained in the *Mesa-Optimization* module (Figure 53).

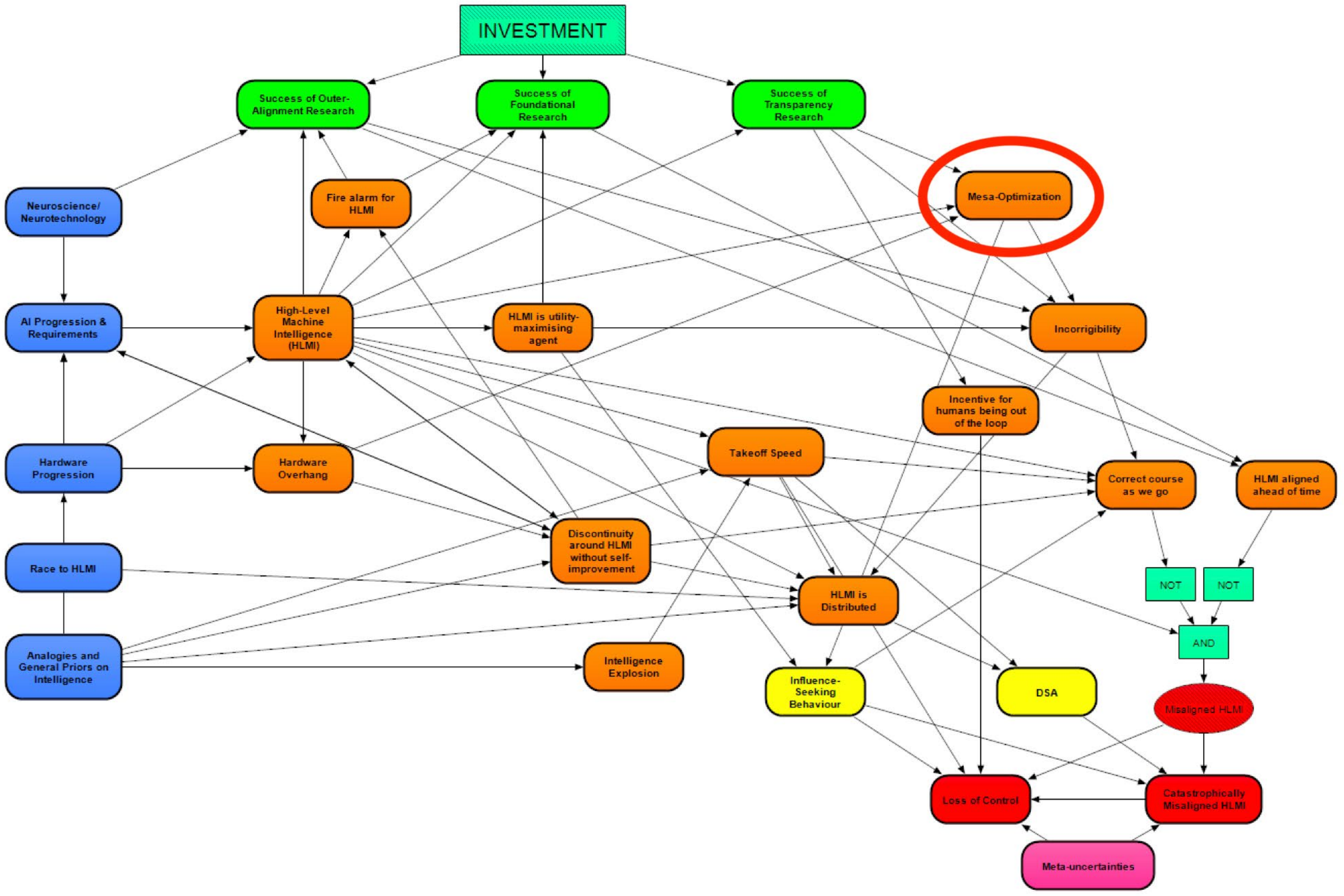

*Figure 53: Model focus for the Mesa-Optimization module*

The output of the *Mesa-Optimization* module is an input to the *Incorrigibility* module (as well as the *Influence-Seeking Behavior* module [§7.4]). The logic is that inner misalignment is one way that HLMI could become incorrigible, which in turn counts strongly against being able to *correct course as we go* in the development of HLMI.

## 5.1 Module Overview

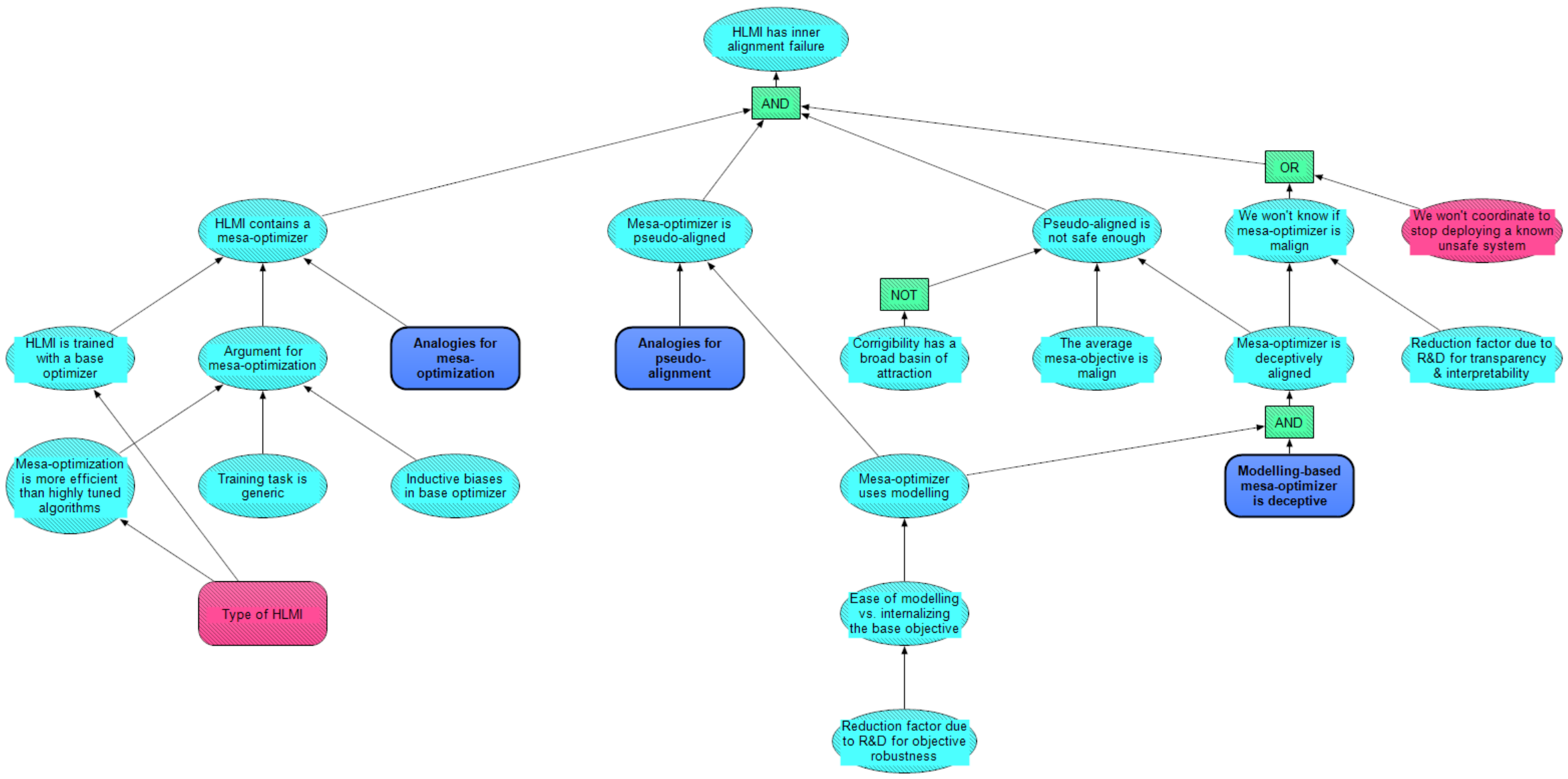

*Figure 54: Mesa-optimization module*

The top-level logic of the *Mesa-Optimization* module is that *HLMI has an inner alignment failure* if

1. The *HLMI contains a mesa-optimizer* (§5.1.1), AND
2. Given (1), the *mesa-optimizer is pseudo-aligned* (i.e., it acts aligned in training) (§5.1.2), AND
3. Given (2), the pseudo-alignment is not sufficient for intent alignment, that is, *it is not safe enough* to make HLMI corrigible (§5.1.3), AND
4. Given (3), *we fail to stop deployment of the unsafe system* (§5.1.4).

In the following sections, we explain how each of these steps are modeled.

### 5.1.1 HLMI Contains a Mesa-optimizer

The output of this section is *HLMI contains a mesa-optimizer*, which depends on three nodes. The left node, *HLMI is trained with a base optimizer*, means that a training algorithm optimized a distinct learned algorithm and that learned algorithm forms all or part of the HLMI system. The crux here is what *type of HLMI you expect*, which comes out of the pathways discussed in the *Paths to HLMI* chapter (§3). For instance, HLMI via current deep learning methods or evolutionary methods will involve a base optimizer, but this is not true of other pathways such as whole brain emulation.

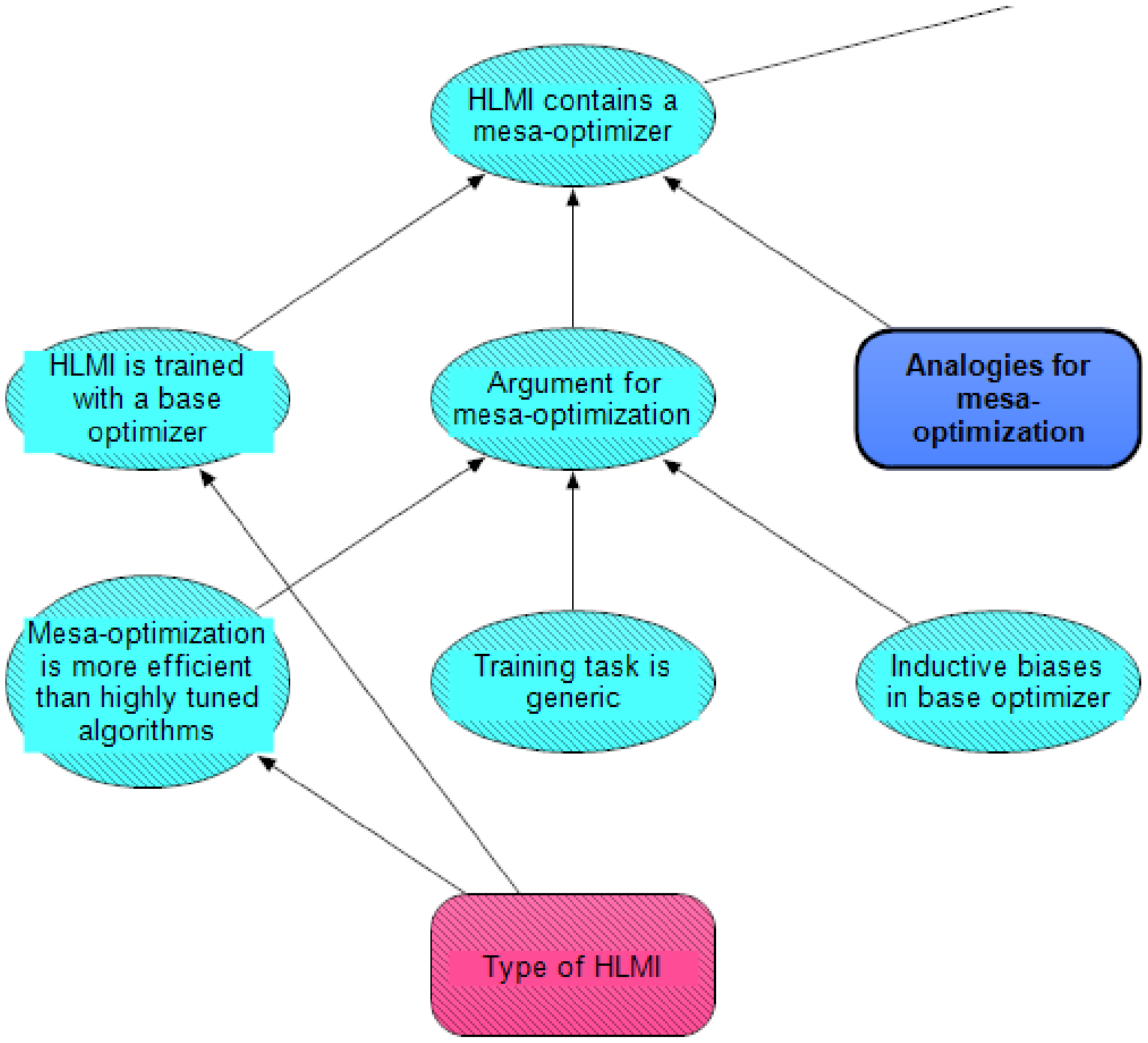

*Figure 55: HLMI contains a mesa-optimizer*

The middle node, *Argument for mesa-optimization*, represents the argument from first principles for why mesa-optimization would occur in HLMI. It is mainly based on the post Conditions for Mesa-Optimization [105]. This is broken down further into three nodes. *Mesa-optimization is more efficient than highly tuned algorithms* distills some claims about the advantages of mesa-optimization compared to systems without mesa-optimization, including that it offers better generalization through search. You could reject that claim on the grounds that sample efficiency will be high enough, such that HLMI consists of a bunch of algorithms that are highly tuned to different domains. You could also argue that some types of HLMI are more prone to be highly tuned than others. For instance, evolutionary algorithms might be more likely to mesa-optimize than machine learning, and machine learning might be more likely to mesa-optimize than hybrid ML-symbolic algorithms.

The middle node, *Training task is generic*, resolves as positive if the learned algorithm is not directly optimized for domain-specific tasks. This scenario is characteristic of pretraining in modern machine learning. For example, the task of predicting the next word in a large text corpus is generic because the text corpus could contain all kinds of content that is relevant to many different domain-specific tasks. In contrast, one could use a more domain-specific dataset (e.g., worded math problems) or objective function (e.g., reward for the quality of a news article summary). This crux is important if mesa-optimizers generalize better than other kinds of algorithms, because then a generic training task would tend to select for mesa-optimizers more. GPT-3 [106] lends credence to this idea, because it demonstrates strong few-shot learning performance (i.e., it is mostly competitive or close to prior state of the art) by simply learning to predict the next word in a text. However, it's uncertain if GPT-3 meets the definition of a mesa-optimizer.

Finally, *Inductive biases in base optimizer* lumps together some factors about inductive bias in the HLMI's architecture and training algorithm which affect the chance of mesa-optimization. For example, the extent that mesa-optimization exists in the space of possible models (i.e., algorithmic range) and the ease by which the base optimizer finds a mesa-optimizer (i.e., reachability). Inductive bias is a big factor in some people's beliefs about inner alignment (see e.g., Inductive biases stick around [107]), but there is disagreement about just how important it is and how it works.

#### 5.1.1.1 Analogies for Mesa-optimization

Finally, evidence for the node *HLMI contains a mesa-optimizer* is also drawn from history and analogies. We modeled this evidence in a submodule, shown below. The submodule is structured as a naïve Bayes classifier: it models the likelihood of the evidence given the hypotheses that an HLMI system does or does not contain a mesa-optimizer. The likelihood updates the following prior: if you only knew the definition of mesa-optimizer and hadn't considered any specific cases, arguments, or evidence for it, what is the probability of an optimizing system containing a mesa-optimizer?

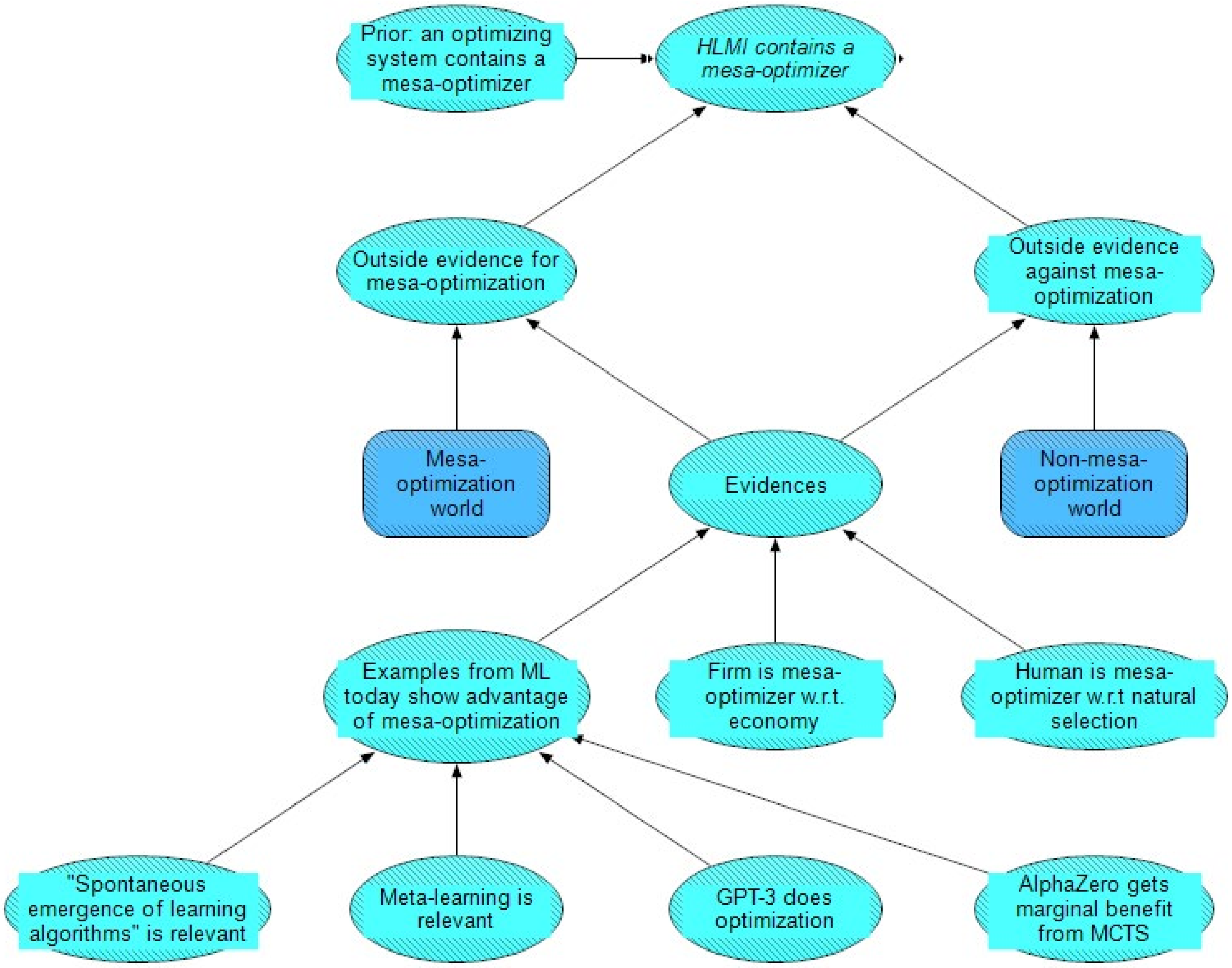

*Figure 56: Analogies impacting whether HLMI contains a mesa-optimizer*

We considered three domains for the evidence (Figure 56): *machine learning today*, *firms with respect to economies*, and *humans with respect to natural selection*. A few more specific nodes are included under *Examples from ML today* because there has been some interesting discussion and disagreement in this area:

- The *"spontaneous emergence of learning algorithms"* out of reinforcement learning algorithms has been cited [108] as evidence for mesa-optimization, but this may not be informative [109].
- *Meta-learning*: learned optimizers have garnered interest in ML research (e.g., this paper [110]). But since this is a deliberate, outer-loop form of learned optimization, it may not affect the likelihood of spontaneous mesa-optimization much.
- GPT-3 does few-shot learning [106] in order to perform novel tasks. John Maxwell's post [111] gives an argument for why it is incentivized for—or may already be—mesa-optimizing.
- It is unclear and debated [112] how much AlphaZero marginally benefits from Monte Carlo tree search, which is a form of mechanistic optimization, compared to just increasing the model size. In turn, it is unclear how much evidence AlphaZero provides for getting better generalization through search, which is argued [105] as an advantage of mesa-optimization.

### 5.1.2 The Mesa-optimizer is Pseudo-aligned

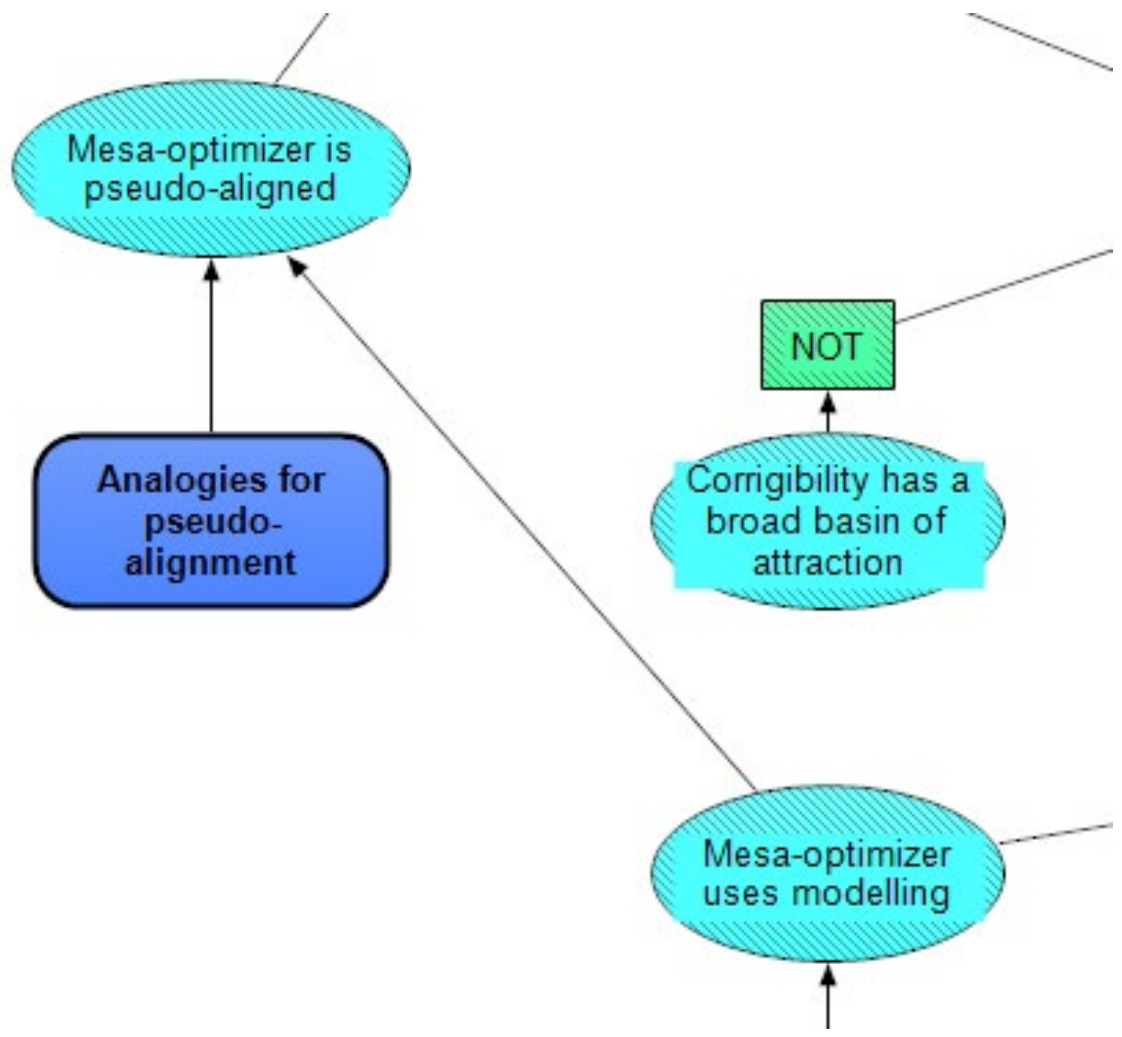

*Figure 57: Mesa-optimizer is pseudo-aligned*

The possibility of pseudo-alignment depends on just two nodes in our model. The structure is simple mainly because pseudo-alignment of any kind seems much more likely than robust alignment, so we haven't found much debate on this point. In general, there are many more ways to perform well on the training distribution which are not robustly aligned with the objective; that is, they would perform significantly worse on that objective under some realistic shift in the distribution. And in practice today, this kind of robustness is a major challenge in ML (Concrete Problems in AI Safety [113] section 7 gives an overview, though it was published back in 2016).

The dependency on the left is a module, *Analogies for pseudo-alignment* (Figure 58), which is structured identically to *Analogies for mesa-optimization* (i.e., with a naïve Bayes classifier, and so on), but the competing hypotheses are *pseudo-alignment* and *robust alignment*, and the analogies are simply *ML systems today*, *Firm*, and *Human*.

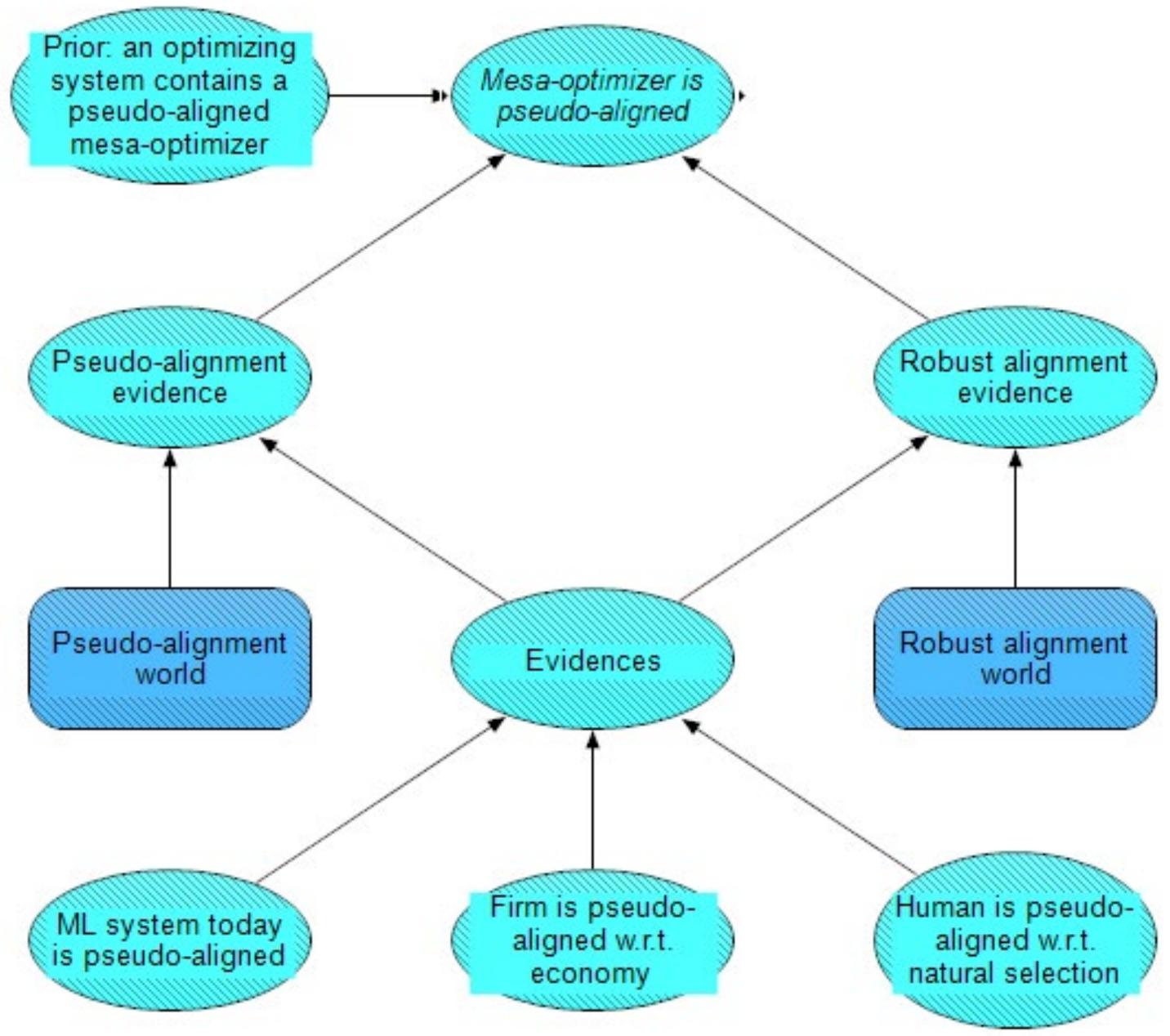

*Figure 58: Analogies for pseudo-alignment of mesa-optimizer*

The second node influencing pseudo-alignment is *Mesa-optimizer uses modeling*. The concept of "modeling" vs. "internalization" introduced in the RLO paper [99] (section 4.4) is relevant to pseudo-alignment. Internalization implies robust alignment, whereas modeling means the mesa-optimizer is pointing to something in its input data and/or its model of the world in order to act aligned. We explain this node and the implications of "modeling" in more detail in the next section.

### 5.1.3 Pseudo-alignment is not Safe Enough

At the top level of this subsection (Figure 59), we include three reasons why pseudo-alignment might not be safe enough and thus count as a failure of inner alignment. First, there is a crux of whether *corrigibility has a broad basin of attraction*. This refers to Paul Christiano's claim [114] that "A sufficiently corrigible agent will tend to become more corrigible and benign over time. Corrigibility marks out a broad basin of attraction towards acceptable outcomes." If Christiano's claim about corrigibility is true, this increases the overall chance that a pseudo-aligned algorithm becomes safe enough before it locks in a path to catastrophe.

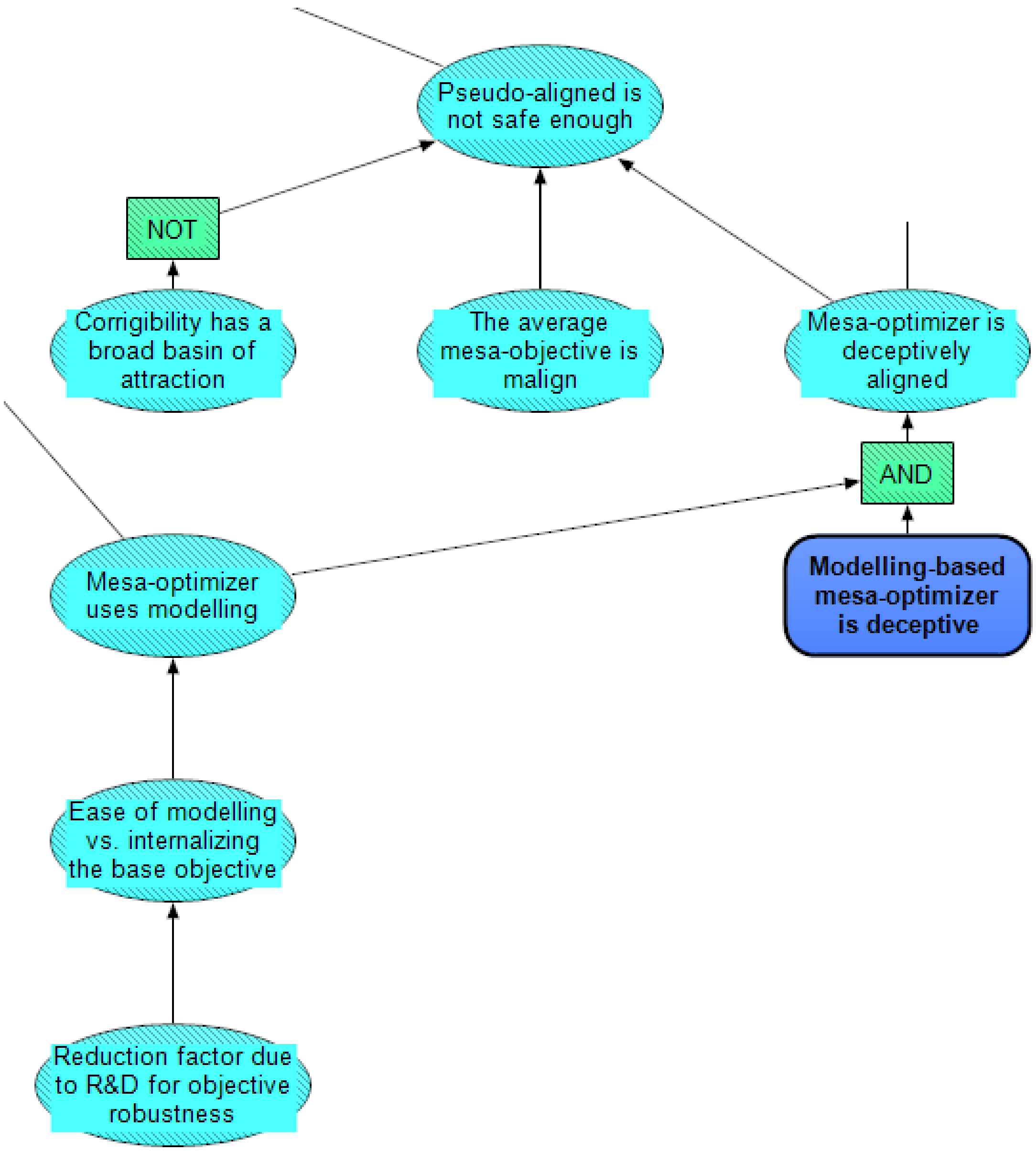

*Figure 59: Pseudo-alignment is not safe enough*

A second crux for the safety of pseudo-alignment is *how malign we expect a mesa-objective to be by default* (by malign, we just mean non-benign, or harmful *in effect*—it doesn't have to be inherently malicious). It's uncertain what mesa-objectives are generally like because they arise internally from the base optimizer, and there is currently scant empirical evidence of mesa-optimization. It's reasonable to expect a proxy objective to be much closer to the base objective than to a random objective, so the danger partly depends on the base objective. Perhaps mesa-objectives will just be weird in benign ways, such as being irrational, or very local [115]. On the other hand, it seems that simple, unbounded, coherent objectives would tend to have lower description length than objectives lacking those attributes, so those objectives may be favored by simplicity bias [116].

Before moving on to the third reason, *deceptive alignment*, let's review the concept of *modeling vs. internalization* introduced in the RLO paper [99] (section 4.4). *Modeling* is where a mesa-optimizer acts aligned with a base objective (in training) by representing the base objective within its world model (using relevant information from its input data) and then deliberately acting aligned with this representation. By understanding the base objective, the mesa-optimizer can behave more aligned, whether or not its own objective is different. (In this case, modeling is similar to a trial employee working hard because their mental model of the company indicates that the company wants them to work hard.) The alternative manner in which a mesa-optimizer can act aligned with a base objective is *internalization*: the base optimizer adjusts the mesa-objective to be more and more aligned until convergence. (In this case, internalization is similar to a trial employee working hard because they intrinsically want to work hard.)

Modeling-based mesa-optimizers are thought to end up *deceptively or corrigibly* [117] aligned as a result of training. Deceptive alignment means the learned algorithm optimizes for the base objective (in training) for instrumental reasons (namely, to deceive the learning algorithm or programmers into thinking it's aligned). Returning to the example of our trial employee, a deceptively aligned trial employee might just want a salary and to slack off, so he will work hard while in trial to be offered the job, and then after being offered the job, only work the minimal amount to not get fired. So, a deceptively aligned algorithm deviates when it becomes favorable (e.g., it figures out that training has ended), removing the threat of being penalized.

The main alternative to a deceptive modeling-based mesa-optimizer that we consider is a *corrigible* modeling-based mesa-optimizer. A corrigible mesa-optimizer optimizes for a *pointer* to the part of its world model that includes the base objective (both in training and deployment). The corrigibility comes from the fact that the base optimizer can modify this pointer to be more accurate. Here, our trial employee cares about being "the perfect employee" both in trial and once offered a job, but they are uncertain how to measure that. So, they might observe employees that have succeeded (the pointer) and learn to imitate that. Note that this is subtly different from the internalizing employee. The internalizing employee doesn't care about the company in itself but values working hard for its own sake (and thus happens to be aligned with the company), whereas the corrigibly modeling employee wants to do whatever they believe the company wants of them, which is to work hard.

Coming back to our model, the third reason that pseudo-alignment might not be safe enough is if the *mesa-optimizer is deceptively aligned* (Figure 60).

We break down deceptive alignment into two main conditions. First, a precondition that the *mesa-optimizer is using modeling*. The key question influencing this precondition is how *easy it is for a mesa-optimizer to model rather than internalize*. We represent this as the likelihood ratio of eventually settling on modeling versus internalization. One argument for modeling being easier is that within the training data for the HLMI, there will likely be a huge amount of rich input data relevant for determining the base objective, and it will therefore be easier to capture this objective function in the HLMI simply by referring to the relevant parts of this data than it will be to build a representation of the objective from scratch. However, this argument depends on just how difficult it is to construct representations that are useful for modeling the objective, which in turn depends both on the nature of the data and the objective. We can also look to analogies here, such as imprinting in animals. For example, rather than have a detailed representation of a mother goose, a gosling imprints on some initial stimulus in its environment, whether that be an actual goose or, say, a human's boots, and then treats the imprinted object as its mother (e.g., it follows the object around).

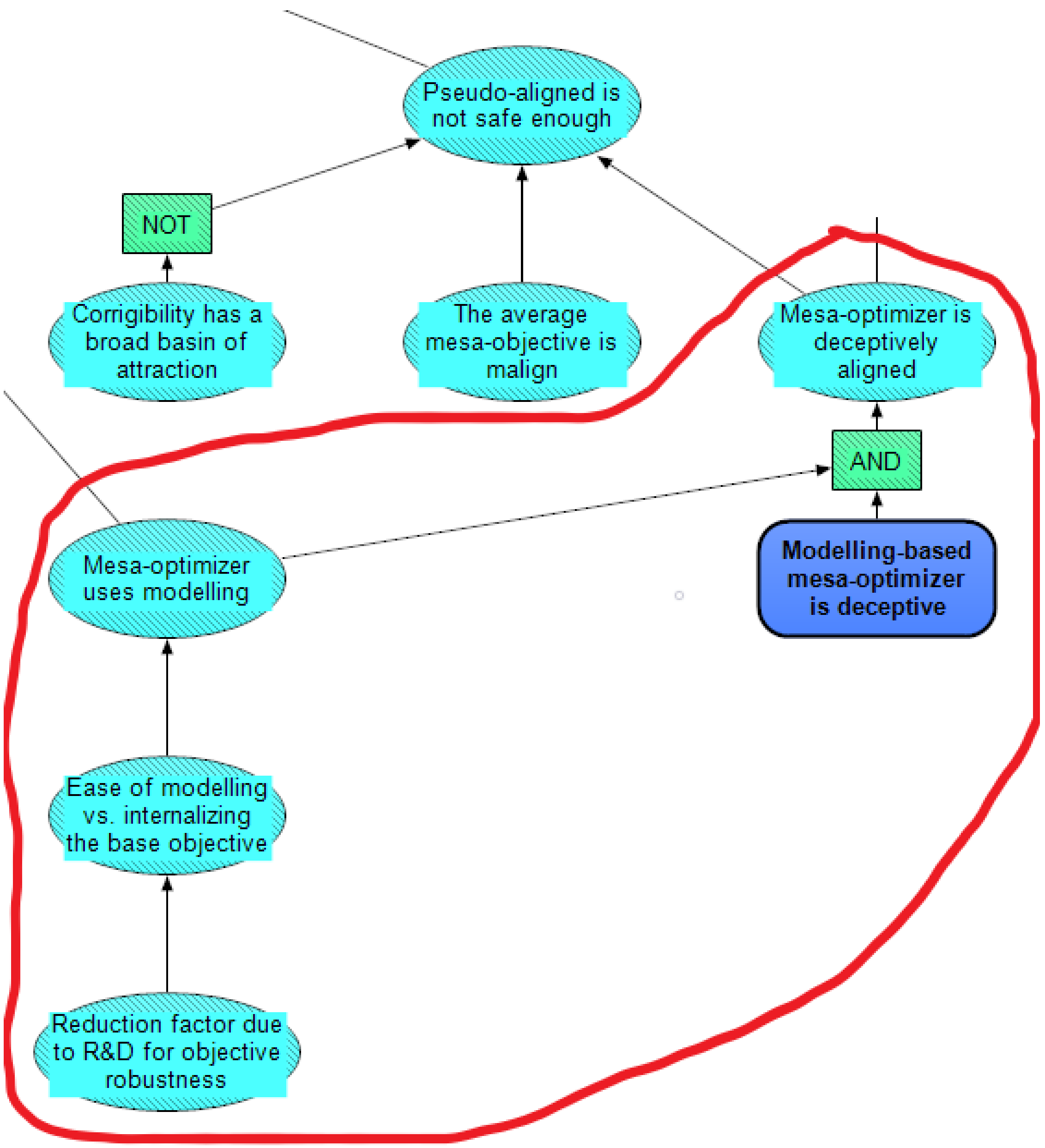

*Figure 60: Mesa-optimizer is deceptively aligned*

Feeding into the ease of modeling vs. internalization node is a *reduction factor due to R&D for objective robustness* (*objective robustness* is terminology from Clarifying inner alignment terminology [3]). This node considers the effect of research for inner alignment on decreasing the difficulty of internalization compared to modeling and outputs a corresponding constant multiplier for the *ease of modeling vs. internalizing the base objective node*.

The second condition for deceptive alignment is simply: *given that a mesa-optimizer uses modeling, it is deceptive*. This possibility is broken down into a submodule of nodes (Figure 61), explained in the next section.

### 5.1.3.1 Deceptive Alignment

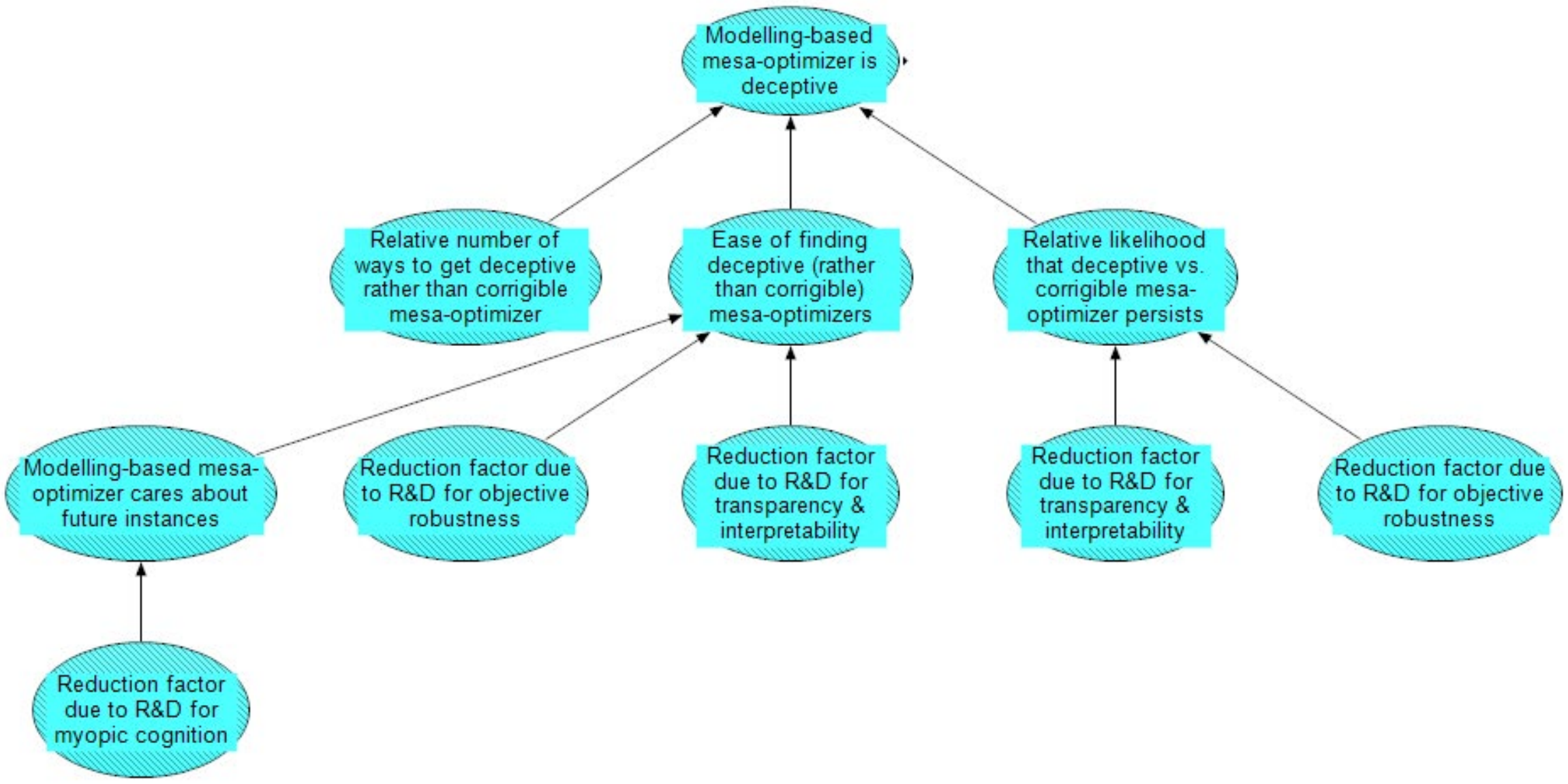

*Figure 61: Modeling-based mesa-optimization is deceptive*

There are three factors feeding into the *modeling-based mesa-optimizer is deceptive* node. First is the *relative number of ways to get a deceptive rather than corrigible mesa-optimizer*. Think of this as a base rate for deceptive (rather than corrigible) mesa-optimization. There is a counting argument [118] for deception here: since deception is instrumentally motivated, it is compatible with almost any proxy goal, whereas for corrigibility, the base optimizer has to find a very specific pointer to relate to the base objective.

The second factor is the *ease of finding deceptive (rather than corrigible) mesa-optimizers*. Similar to reachability [105], this node is represented by the relative likelihood that the base optimizer outputs a deceptive mesa-optimizer at some point in training.

The post Does SGD Produce Deceptive Alignment? [119] details some arguments for and against deceptive algorithms being easier to find than corrigible alignment (specifically for stochastic gradient descent, but much of it is more general). Becoming deceptive requires caring about future instances so that there is an incentive to deceive now for later gain. It also requires reasoning about the training process, so that the mesa-optimizer expects the threat of modification to go away and for deception to pay off. On the other hand, becoming corrigible requires a *robust* pointer to a good model of the base objective—otherwise the mesa-optimizer will just optimize for whatever proxy it has. Furthermore, even if there is information related to the objective in the input data, the mesa-optimizer will need to have uncertainty about precisely what objective it is pointing to. The conditions for deception are claimed to be easier.[4] However, we think this is a large and important uncertainty which might hinge on the nature of the base objective and how the base optimizer works.

Going one level down from the *ease of finding deceptive (rather than corrigible) mesa-optimizers* node, we have *modeling-based mesa-optimizer cares about future instances*. This means the mesa-optimizer is motivated by what happens in future instances of training or in deployment (instead of just its current round of training). We put this here because caring about future instances is a necessary condition for deceptive alignment but not corrigible alignment—if an algorithm does not consider future instances, then it will not see the benefit of defecting later in time, and therefore has no instrumental reason to deceive. The node for the *reduction factor due to R&D for myopic cognition* refers to the idea that methods to induce myopic cognition, such as approval-directed agents [121], could reduce the likelihood of modeling-based mesa-optimizers caring about future instances (myopic cognition is discussed in this post [122] by Richard Ngo).

The third top-level node here is the *relative likelihood that deceptive vs. corrigible mesa-optimizer persists*—that is, the likelihood that once a mesa-optimizer becomes deceptive in one instance, it survives and remains deceptive indefinitely (relative to the likelihood for a corrigible mesa-optimizer). Deceptive algorithms may be favored due to their ability and incentive to adapt to the base objective, because they can reason about the objective and they care about future instances. However, protecting a proxy objective from modification may be difficult. This too seems like an important uncertainty that depends on empirical facts about the base optimizer, such as how stochastic gradient descent works with modern neural nets.

Finally, we include two areas of research that could reduce the likelihood of deceptive alignment. One is *research to improve objective robustness*, such as Relaxed adversarial training for inner alignment [123], where one of the aims is avoiding deception. A key part of making relaxed adversarial training work is transparency tools to guide the training process. Evan Hubinger [124] argues why that is more helpful than inspecting the algorithm *after* deception may have already occurred. *Research into transparency* may help to prevent deceptive alignment in other ways, so this is kept as a separate node.

### 5.1.4 We Fail to Stop Deployment

The last part of the *Mesa-Optimization* module is about whether we will pull back before it's too late, given there is an HLMI with an unsafe mesa-optimizer that hasn't been deployed yet (or trained to sufficiently advanced capabilities that it can "break out"). The way we could pull back in such a situation is if *people are aware that it's unsafe*, AND *they coordinate to stop the threat*. That logic is inverted to an OR in the module (Figure 62) to fit with the top-level output of *HLMI has inner alignment failure*.

[4] Parts of this argument are not explicit in Does SGD Produce Deceptive Alignment? [119]—see also the FLI podcast with Evan Hubinger [120] (search the transcript for "which is deception versus corrigibility").

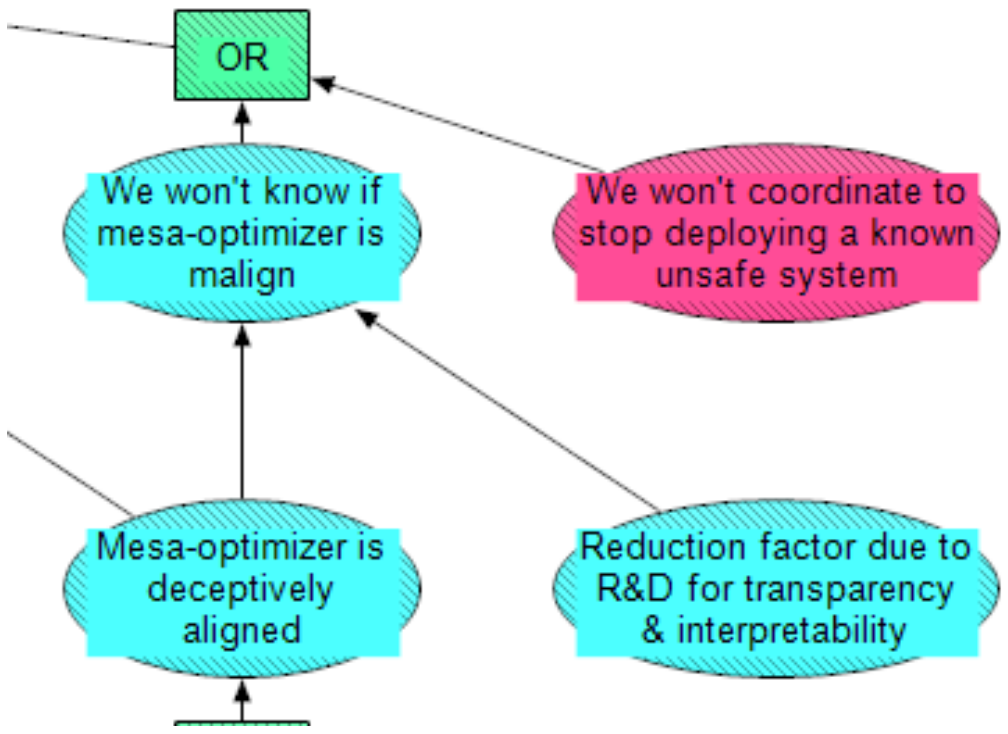

*Figure 62: We fail to stop deployment of HLMI with mesa-optimizers*

*Mesa-optimizer is deceptively aligned* is one of the strongest factors in knowing whether a mesa-optimizer is unsafely pseudo-aligned, because deception works against overseers figuring this out. Meanwhile, *R&D for transparency & interpretability* may help to detect unsafe pseudo-aligned algorithms. On the other hand [125], deceptive algorithms may just exploit weaknesses in the transparency tools, so it is even possible for the reduction factor to be less than 1 (i.e., it *increases* the chance that we fail to see the danger).

## 5.2 Conclusion

In this chapter, we have examined important cruxes and uncertainties about risks from learned optimization, and how they relate to each other. Our model is mostly based on the Risks from Learned Optimization [98] sequence and considers whether mesa-optimization will occur in HLMI at all, whether pseudo-alignment occurs and is dangerous, and whether the mesa-optimizer will be deployed or "break out" of a controlled environment. Some uncertainties we have identified relate to the nature of HLMI, connecting back to *Paths to HLMI* (§3) and to analogies with other domains. Other uncertainties relate to the training task, such as how generic the task is, and whether the input data is large and rich enough to incentivize modeling the objective. Other uncertainties are broadly related to inductive bias, such as whether the base optimizer tends to produce mesa-optimizers with harmful objectives.

The next chapter will look at the effects of AI safety research agendas.

# 6 Modeling the Impact of Safety Agendas

Ben Cottier

We caution that this part of the model is much more of a work in progress than others. At present, it is best described as loosely modeling a few aspects of safety agendas, and the hope is that it can be further developed to be of similar quality as the more complete portions of the model. In many cases, we are unclear about how different research agendas relate to specific types and causes of risk. While this is partly because this portion of the model is a work in progress, it is also because the theory of impact for many safety agendas is still unclear. So, in addition to explaining the model, we will highlight things that are unclear and what would clarify those things.

Modeling of different research areas is contained in the green-colored modules, circled in Figure 63.

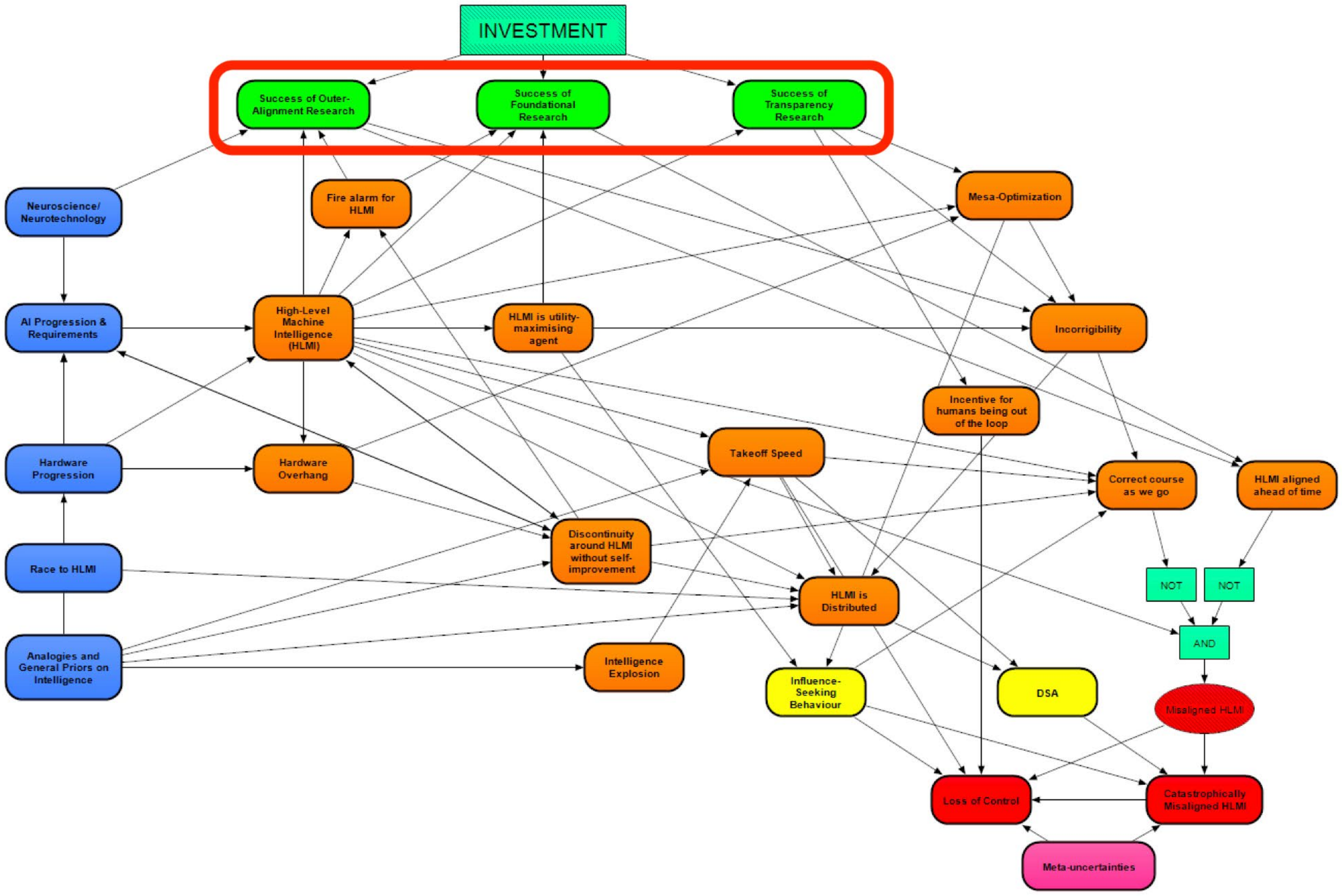

*Figure 63: Model focus for the Impact of Safety Agendas modules*

## 6.1 Key Questions for Safety Agendas

We encountered several uncertainties about safety agendas which have made modeling them relatively difficult. While most of these uncertainties are explained throughout the next section, the following points seem like the most important questions to ask about a safety agenda:

- What is the theory of change? What does success look like, and how does that reduce AI risk? What aspects of alignment is the agenda supposed to address?
- What is assumed about the progression of AI? How much does the agenda rely on a particular AI paradigm?
- What is the expected timeline of the research agenda, and how much does that depend on additional funding or buy-in?
- What are likely effects from work on this agenda even if it doesn't succeed fully? Are there spillover effects for AI capabilities or for other safety agendas? Are there beneficial effects from partial success?

Besides just helping our model, it's worth us highlighting all the benefits of understanding the above for various safety agendas:

- the community can better understand what constitutes success and provide better feedback on the agenda
- the researchers get a better idea of how to steer the research agenda as it progresses
- funders (and researchers) can better evaluate agendas, and in turn prioritize funding or pursuing them
- future research can find gaps in the safety-agenda space more easily

## 6.2 Model Overview

### 6.2.1 Overall Impact of Safety Research

The key outcome to focus on for our model of research impact is *Misaligned HLMI*, circled in purple in the figure below. This node is obviously important to the final risk scenarios that we model, which will be covered in more detail in the next chapter (§7). Looking at the inputs to this node in Figure 64, the model says we can avoid *Misaligned HLMI* if either

1. *HLMI is never developed* (blue-circled node), or
2. We manage to *correct course as we go*, meaning HLMI is either aligned by default or HLMI can be aligned in an iterative fashion in a post-HLMI world (orange-circled node), or
3. *HLMI [is] aligned ahead of time*—that is, people find a way to align HLMI before it appears or needs to be aligned (green-circled node).

While it seems that most people in the AI safety community believe option 3 is necessary for safe HLMI, option 2 is argued by some mainstream AI researchers and by some within the AI safety community, and option 1 would apply if humanity successfully coordinated to never build HLMI (though this possibility is not currently captured by our model). We will discuss option 2 further in the next chapter on failure modes (§7).

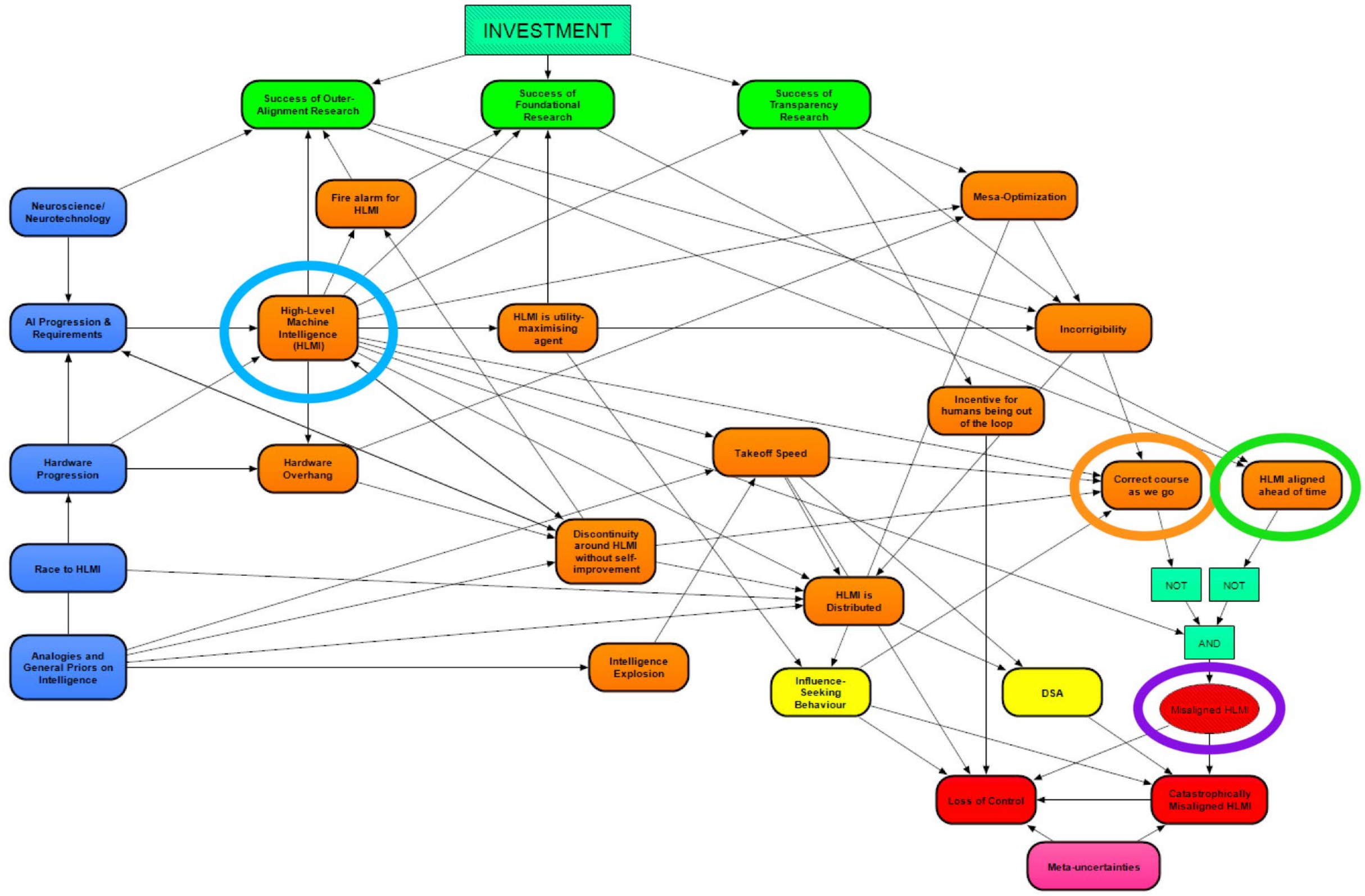

*Figure 64: Avoiding misaligned HLMI*

If we assume that attempting to align HLMI ahead of time *is* a worthwhile endeavor (i.e., conditioning on the model of the risks), how would success be possible? It would either be through the success of currently proposed research agendas and any of their direct follow-ups, or through more novel approaches developed as we get closer to HLMI. New approaches may come from new insights or from a paradigm shift in AI [16]. This is essentially what the *HLMI aligned ahead of time* module (Figure 65) is capturing. In particular, this module currently includes *foundational research* (specifically the highly reliable agent designs agenda [126]) and *synthesized utility function* [127]. We are currently including these two here for simplicity, but other agendas should fit in as well.

Our current model focuses primarily on currently proposed agendas since we can, and want to, model them more concretely. Additional work on both the current agendas and the potential for future progress would be useful in improving our understanding of the risk and building the model.

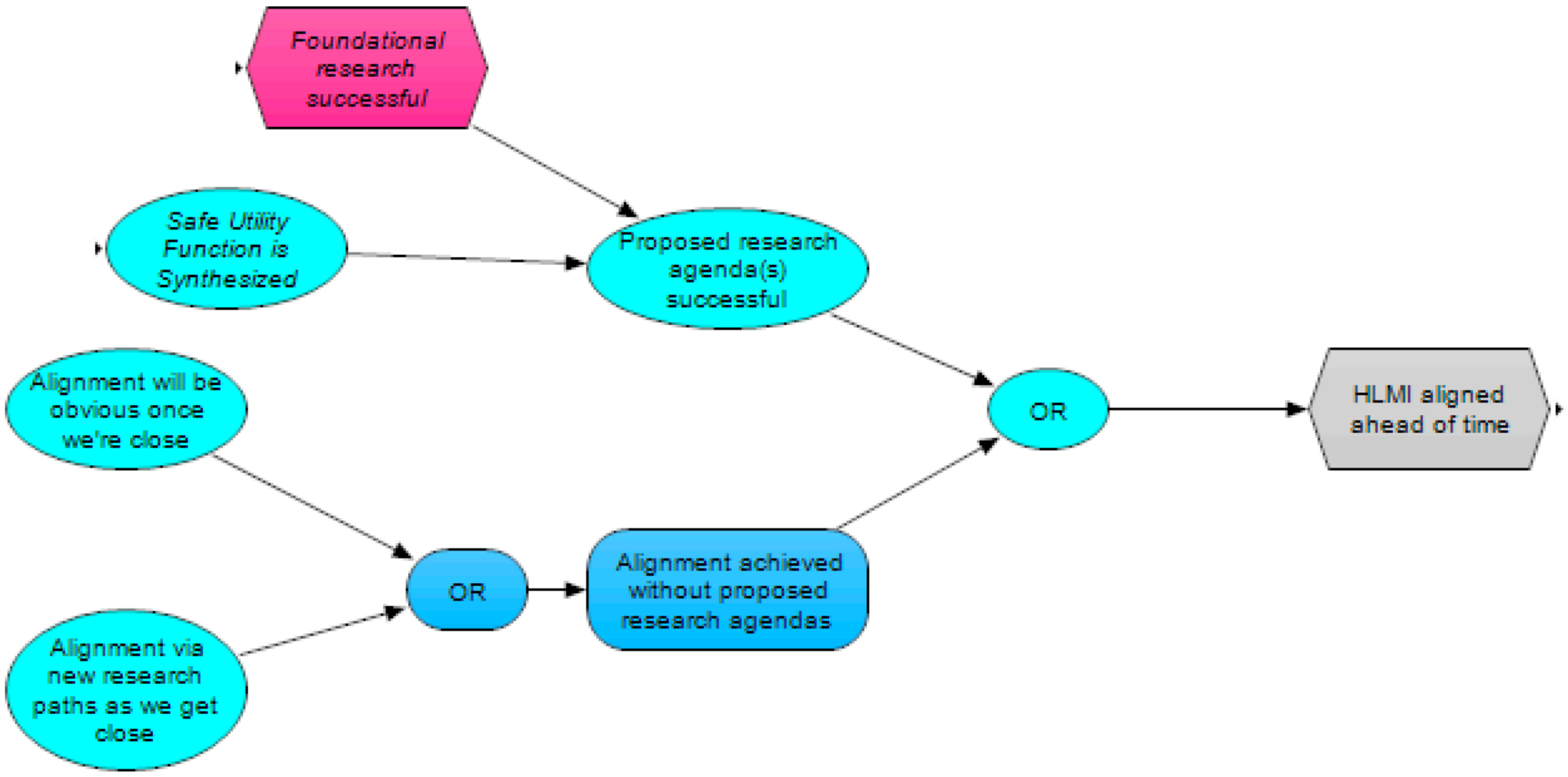

*Figure 65: HLMI aligned ahead of time*

In this chapter, we'll focus on three (or perhaps two and a half) different approaches to safety: 1) *iterated distillation and amplification (IDA)* (§6.2.2), 2) *foundational research* (§6.2.3), and 3) *transparency research* (§6.2.4), which has been proposed as a useful part of several approaches to safety but may not be sufficient alone. In the following sections, we go through our preliminary models of these agendas and point out major uncertainties we have about them.

## 6.2.2 Iterated Distillation and Amplification

We will use the IDA [128] research agenda as the main example to explain our uncertainties about modeling, as it appears to have more published detail than most other proposed agendas regarding what is involved, its theory of change, and its assumptions about AI progress. The section of the model for IDA is shown in Figure 66.

The final output of this section is *IDA research successful*. We assume that success means obtaining a clear, vetted procedure to align the intentions [129] of an actual HLMI via IDA. However, the IDA agenda seems to address outer alignment (i.e., finding a training objective with optima that are aligned with the overseer) and not necessarily inner alignment (i.e., making a model robustly aligned with the training objective itself).[5] This is a case where we are uncertain what parts of the alignment problem the agenda is supposed to solve, and how a partial solution to alignment is expected to fit into a full solution.

For IDA and other agendas, it's also difficult to reason about *degrees* of success. What if the research doesn't reach its end goal, but produces some useful insights? Relatedly, how could the agenda help other agendas if it does not directly achieve its aims? We're uncertain how researchers think about this and how best to model these effects.

[5] For more on this distinction/issue, see this post [130].

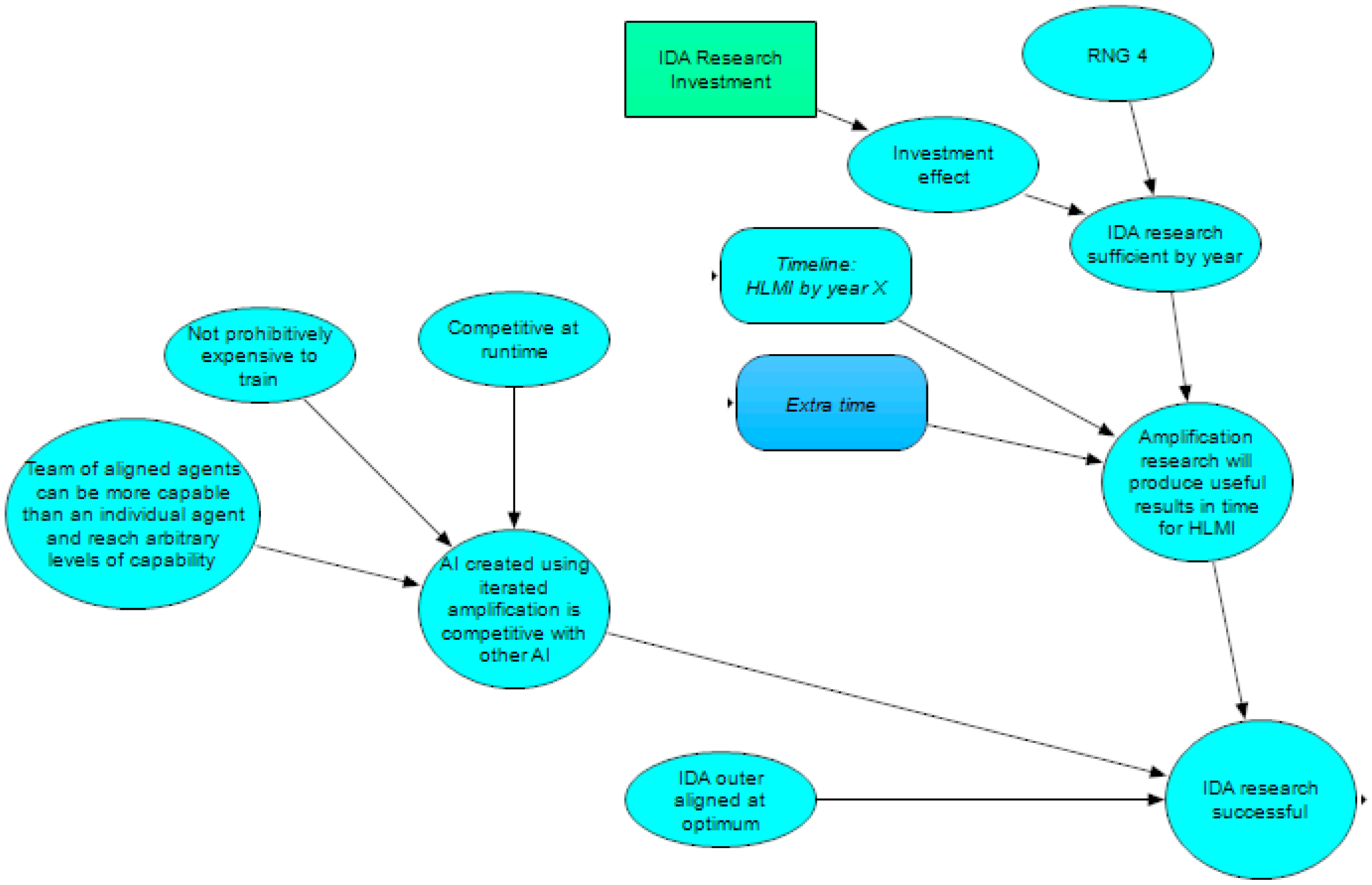

*Figure 66: IDA research successful*

Continuing through the model, towards the right of the figure we have a section about the "race" between IDA research and AI capabilities research, summarized as *Amplification research will produce useful results in time for HLMI*. We're particularly unsure how to think about timelines to solve research agendas, and we are interested in either community feedback about their understanding of the timelines for success, or about any insights on the topic.

Currently, we model research timelines using the node *IDA research sufficient by year X*, where *X* is affected by both *Investment effect* and randomness. This result is then modified by the node *Extra time*, which models the possibility that a "fire alarm" for HLMI [131] is recognized and speeds up safety research in the years before HLMI, either through insight or increased resources. This time to sufficient IDA research is then compared to the timeline for HLMI (*Timeline: HLMI by year X*, addressed in the chapter Paths to High-Level Machine Intelligence [§3]), where success is dependent on the time to IDA being less than the time to HLMI.

The competitiveness of IDA is modeled in a section towards the left of the figure. We break competitiveness down into *Not prohibitively expensive to train*, *Competitive at runtime*, and whether IDA scales to arbitrary capabilities (*Team of aligned agents can be more capable than an individual agent and reach arbitrary levels of capability*). All are modeled as being necessary for IDA to be competitive.

Finally, we have one node to represent whether *IDA [is] outer aligned at optimum*. This means that all possible models which are optimal according to the training objective are at least intent aligned. Being outer aligned at optimum has been argued [130] to defuse much of the threat from Goodhart's Law, specifically the causal and extremal variants of it [132]. So, IDA being outer aligned at optimum would be important for the IDA agenda to be successful at alignment and is therefore a key uncertainty.

Putting it all together, our model considers IDA to be a workable solution for outer alignment if and only if IDA research wins in a “race” against unaligned HLMI, and IDA is sufficiently competitive with other approaches of HLMI, and IDA is outer aligned at optimum. The output node *IDA research successful* then feeds into the *Incorrigibility* module at the top level of the model—that is, if it’s successful and we have an intent-aligned HLMI, then it *is* corrigible. Corrigibility in turn increases the ability to *correct course as we go* (and so on, as explained in the previous section).

### 6.2.3 Foundational Research

Our work on modeling *foundational research* has focused entirely on MIRI’s highly reliable agent designs (HRAD) research agenda [126], as this has had the most discussion in the AI alignment community (out of all the technical work in foundational research). Trying to model disagreements about the value of the HRAD agenda led to the post Plausible cases for HRAD work, and locating the crux in the “realism about rationality” debate [133]. To summarize the post, one of the difficulties with modeling the value of HRAD research is that there seems to be disagreement about what the debate is even about. The post tries to organize the debate into three “possible worlds” about what the core disagreement is and gives some reasons for thinking why we might be in each world. The discussion in the comments did not lead to a consensus, so more work will probably be needed to make our thoughts precise enough to encode in the graph structure of the model.

The model in Figure 67 is a simpler substitute, pending further work on the above. Like the IDA model, this considers whether the research can succeed in time for HLMI. Besides that, there are two nodes about the possibility and difficulty of HRAD. The *Foundational research [is] successful* node feeds directly into the *HLMI aligned ahead of time* module shown previously.

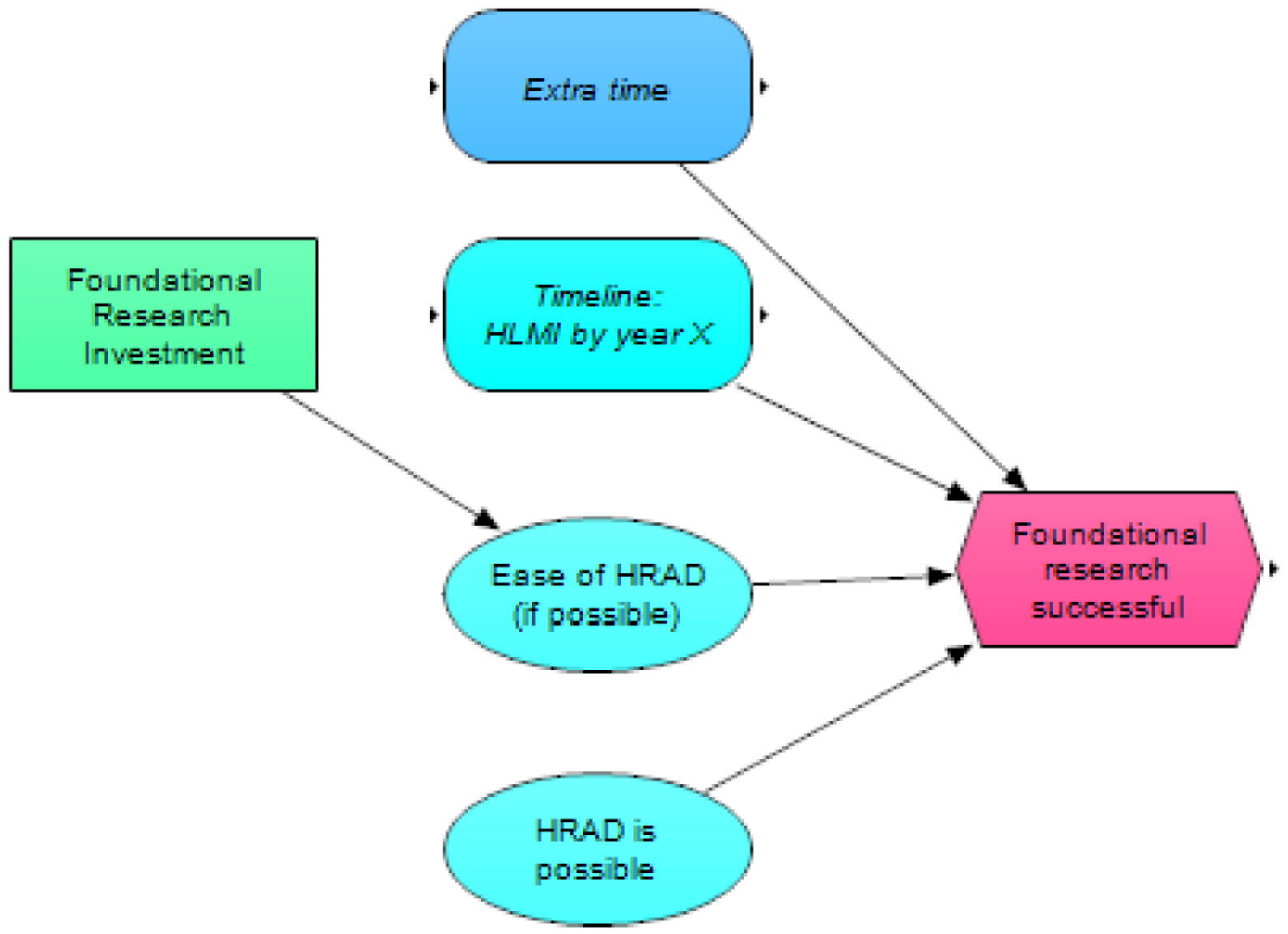

*Figure 67: Foundational research successful*

## 6.2.4 Transparency

*Here, we are using the term "transparency" as shorthand for transparency and interpretability research that has long-term AI safety as a core motivation.*

Transparency can be applied to whole classes of machine learning models and may be a part, or complement, of several alignment techniques. In An overview of 11 proposals for building safe advanced AI [134], "transparency tools" form a key part of several proposals, for different reasons. More recently, Transparency Trichotomy [135] analyzed different ways that transparency can help understand a model: via inspection, training, or architecture. The parts of the trichotomy can also work together, for instance by using transparency via inspection to get more informed oversight [136], which then feeds back into the model via training. So, the current structure of our model, with its largely separate paths to impact for each agenda, does not seem well suited to transparency research. However, the *mesa-optimization* module (§5) does incorporate some nodes on how transparency research may help detect deception (via inspection) or actively avoid deception (via training).

Theories of change for transparency helping to align HLMI have become clearer in published writing over the last couple of years. The post Chris Olah's views on AGI safety [137] offers several claims on theory of change, including:

- Transparency tools give you a mulligan—a chance to recognize a bad HLMI system and try again with better understanding.
- Advances in transparency tools feed back into design. If we build systems with more understanding of how they work, then we can better understand their failure cases and how to avoid them.
- Careful analysis using transparency tools will clarify what we *don't* understand too. Pointing out what we don't understand will generate more concern about HLMI.
- Transparency tools help an overseer give feedback not just on a system's output but also the process by which it produced that output.
- Advances in transparency tools (and demonstrating their usefulness and appeal) helps realign the ML community to focus on deliberate design and understanding.

From the above claims and discussion elsewhere, there are several apparent cruxes for transparency helping to align HLMI:

- Using transparency tools will not make enough progress (or any progress) on the "hard problem" of transparency. The hard problem is to figure out what it even means to understand a model in a way that can save us from Goodhart's Law [138] and deception [117]. As discussed in Transparency Trichotomy [135], transparency tools can themselves be gamed. There seems to be agreement that transparency tools will not get us *all* the way on this problem, but disagreement about how much they help—see for example, this thread [139] and this comment (bullet point 3) [140].

- Similar to the above, though we are not sure if this is a distinct crux for anyone, there is a risk that transparency tools make the flaws we are trying to detect harder to understand (discussed here [125] and here [141]), so there is too great a risk that the tools cause net harm.

- Transparency tools will not scale with the capabilities of HLMI and beyond—discussed here [137]. The crux could be specifically about the amount of labor required to understand increasingly large models. It could also be about increasingly capable systems using increasingly alien abstractions. The linked post suggests that an amplified overseer could get around this problem, so the crux could actually be in whether an amplified overseer can make transparency reliably scale in place of humans.

- The available transparency tools will not be useful for the kind of system that HLMI is (e.g., the work on circuits [142] in vision models will not transfer well to language models). This is like a horizontal version of the above scaling crux. Chris Olah raised this point himself [143].

Again, more work is needed to structure our model in a way that incorporates the above cruxes.

### 6.2.5 Other Agendas

Other agendas or strategies which we have not yet modeled include:

- Counterfactual oracles [144], STEM AI [134], quantilizers [145], myopic cognition (discussed in Arguments against myopic training [122]), debate [146]
- Multiagent safety (there is writing on it [147], and some issues have been identified [148], but we are not aware of a research agenda)
- Aligning current systems (e.g., large language models)—see The case for aligning narrowly superhuman models [149]

Some of the above are more difficult to model because there is less writing that clearly outlines paths to impact or what success looks like. A potentially valuable project is to make a clearer case to the community for how a given research agenda could be impactful and explaining what the goals and specific approaches are.

## 6.3 Help from This Community

Our tentative understanding suggests that more public effort to understand and clearly articulate safety agendas' impacts, driving beliefs, and main points of disagreement would be really helpful. Examples of good work in this area are An overview of 11 proposals for building safe advanced AI [134], and Some AI research areas and their relevance to existential safety [150]. This work can take a lot of effort and time, but some of the uncertainties highlighted in this chapter seem fairly easy to clarify through comments or smaller write-ups.

To illustrate the kind of information that would help, we have written the following condensed explanation of an imaginary agenda (the agenda and opinions are made up—this is not quoting anyone):

*This agenda aims to increase the chance that high-level machine intelligence (HLMI) is inner aligned. More specifically, it will defuse the threat of [deceptively aligned]() HLMI* [117]*. The path from deceptively aligned HLMI to existential catastrophe is roughly: such a system would be deployed due to economic or other incentives and lack of apparent danger. It would also be capable enough to take the long-term future out of humanity's control. While we have a very wide distribution over how humanity loses control, we expect a scenario similar to scenario 2 of [What Failure Looks Like]()* [151].

*The specific outcome we are aiming for is <alignment procedure>. For this to succeed and be scalable, we rely on AI progressing like <current machine learning trends>. We expect the resulting AI to be competitive, with training time on the same order of magnitude and performance within 20% of the unaligned baseline.*

*With regard to timelines, we tentatively estimate that this work is at its most valuable if HLMI is produced in the medium term, neither in the next 10 years, nor more than 30 years from now. With our current resources, we give a rough 5% chance of having a viable procedure within 5 years and a 10% chance within 20 years. This increases to 20% and 30% respectively with <additional resources>. The remaining subjective estimated probability of failure is split evenly between obstacles from the theory, project management, or external factors. This agenda relies on outer alignment being solved using <broad outer-alignment approach>, but otherwise not interacting much with the problem we aim to solve.*

Finally, as a way of quickly gathering opinions, we would love to see comments on the following: for any agenda you can think of, or one that you're working on, what are the cruxes for working on it?

In the next chapter, we will look at the failure modes of HLMI and the final outcomes of our model.

# 7 Modeling Failure Modes of High-Level Machine Intelligence

Ben Cottier, Daniel Eth, Samuel Dylan Martin

We now come to the potential failure modes caused by high-level machine intelligence (HLMI). These failure modes are the principal outputs of the model. Ultimately, we need to look at these outputs to analyze the effect that upstream parts of the model have on the risks, and in turn to guide decision-making. This chapter will first explain the relevant high-level components of our model before going through each in detail. In Figure 68, failure modes are represented by red-colored nodes and circled in red.

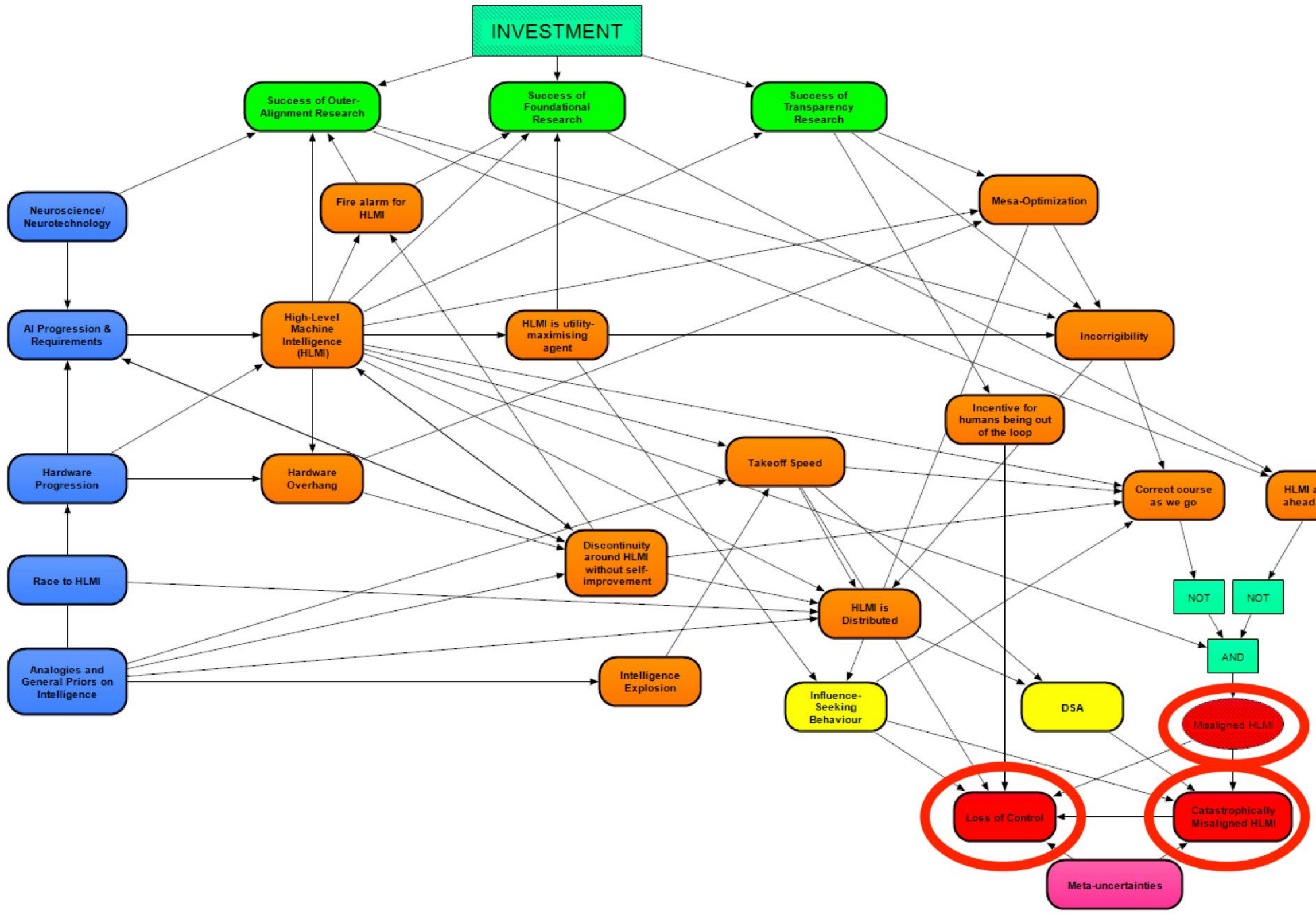

*Figure 68: Top-level view of the model—failure modes are in red*

We do not model catastrophe itself in detail, because there seem to be a very wide range of ways it could play out. Rather, the outcomes are states from which existential catastrophe is very likely or the "points of no return" for a wide range of catastrophic scenarios. *Catastrophically misaligned HLMI* (§7.5) covers scenarios where one HLMI, or a coalition of HLMIs and possibly humans, achieves a decisive strategic advantage (DSA), and the DSA enables an existential catastrophe to occur. *Loss of control* (§7.6) covers existential scenarios where a DSA does not occur; instead, AI systems gain influence over the world in an incremental and distributed fashion, and either humanity's control over the future gradually fades away or a sudden shift in the world causes a sufficiently large and irreversibly damaging correlated automation failure (see What failure looks like [151]).

Moving one step back, there are three major drivers of those outcomes. Firstly, *misaligned HLMI* (§7.2) is the technical failure to align[6] an HLMI. Second, the *DSA* module (§7.3) considers different ways DSA could be achieved by a leading HLMI project or coalition. Third, *influence-seeking behavior* (§7.4) considers whether an HLMI would pursue accumulation of power (e.g., political, financial) and/or manipulation of humans as an instrumental goal. This third consideration is based on whether HLMI will be agent-like and to what extent the Instrumental Convergence [152] thesis applies.

## 7.1 Outcomes of HLMI Development

As discussed in the impact of safety agendas chapter (§6), we are still in the process of understanding the theory of change behind many of the safety agendas, and similarly our attempts at translating the success or failure of these safety agendas into models of HLMI development are still only approximate. The *mesa-optimization* module (§5) evaluates the likelihood of dangers from learned optimization [99], which would tend to make HLMI incorrigible [114]. The green modules on the *success of [agenda] research* (§6.2) also impact the likelihood of an incorrigible HLMI—in particular, if the research fails to produce its intended effect.

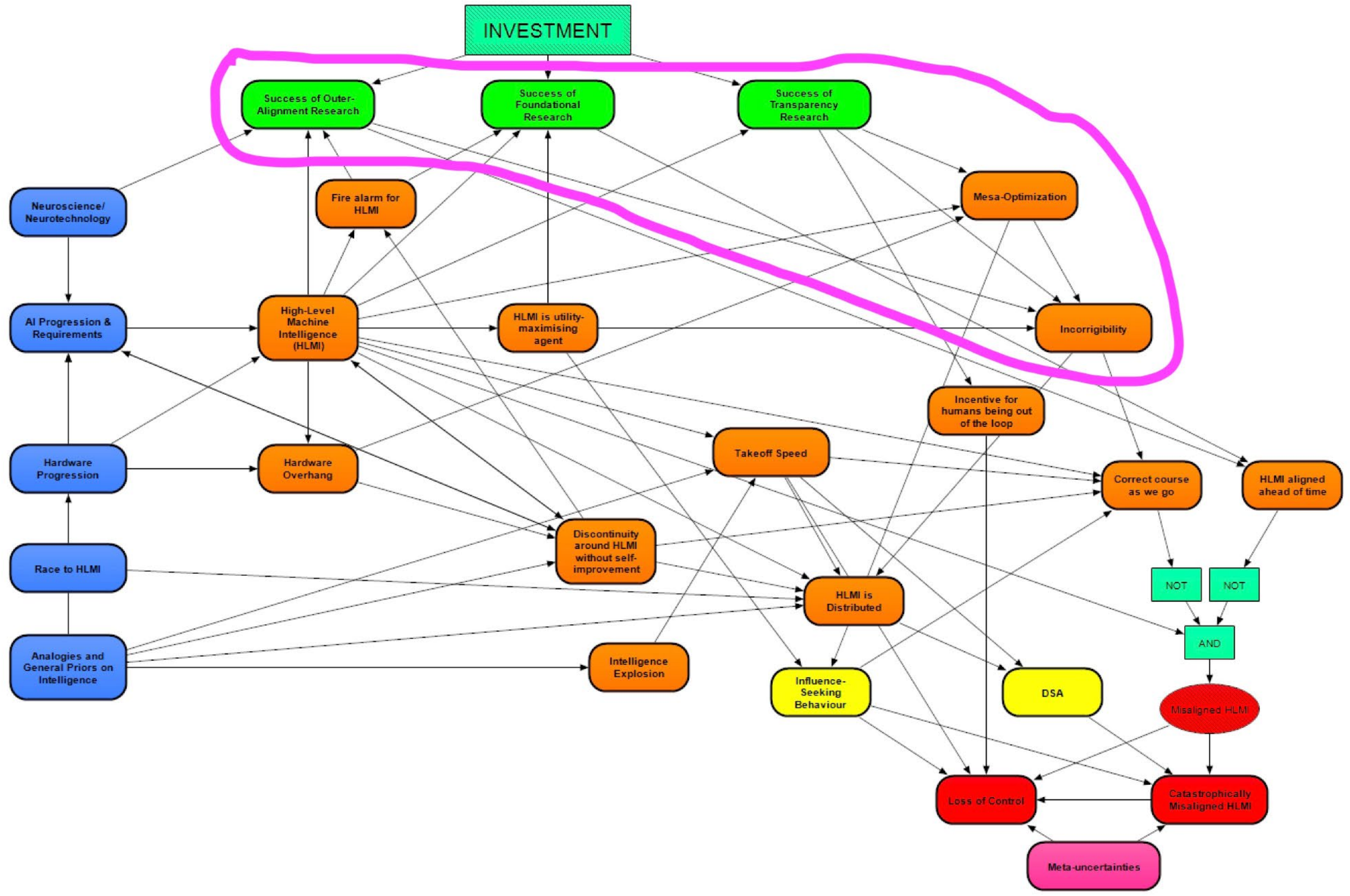

*Figure 69: Some modules impacting incorrigibility*

[6] By aligned, we mean that the HLMI is trying to do sufficiently close to what the operator actually wants, such that the HLMI's actions do not cause an X-risk from the operator's perspective.

Besides *incorrigibility*, the question of HLMI being aligned may be influenced by the expected speed and level of discontinuity during takeoff (represented by the modules *Takeoff Speed* (§4.3.1) and *Discontinuity around HLMI without self-improvement* (§4.1.1). The expectation is that misalignment is more likely the faster progress becomes because attempts at control will be harder if less time is available. These factors—incorrigibility and takeoff—are more relevant if alignment will be attempted in an iterated manner post-HLMI (*Correct course as we go* [§7.1.1]), so those are connected in the model. On the other hand, if we have *HLMI aligned ahead of time* (§7.1.2), then we assume that the *Success of [agenda] Research* is the key influence, so those are connected. We'll look at these last two modules about alignment in the next section.

## 7.1.1 Correct Course as We Go

The *Correct course as we go* module is shown in Figure 70. We consider two possibilities that would allow us to correct course as we go. One is that *we can intervene to fix misbehaving HLMI*, circled in black in the image.

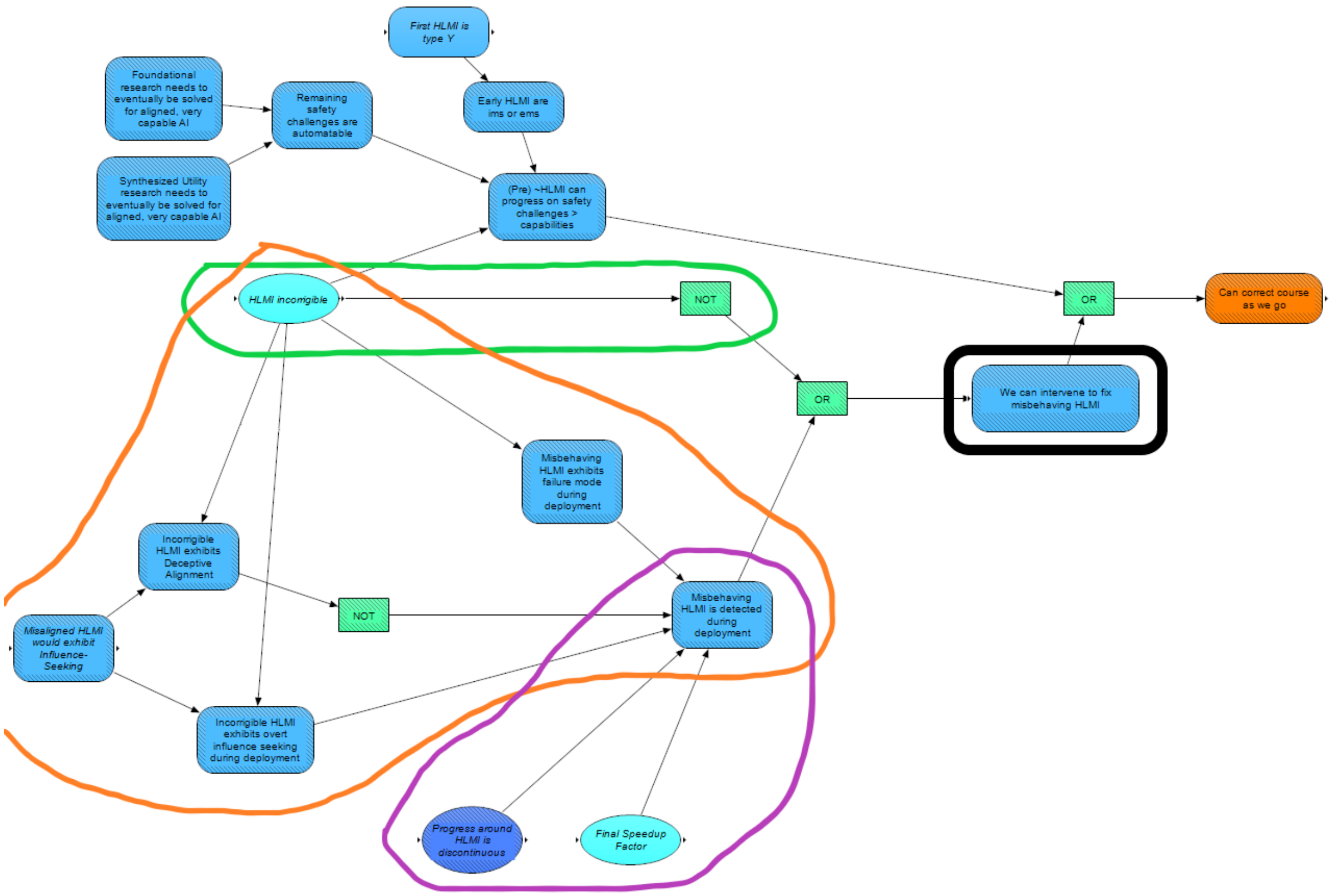

*Figure 70: Correct course as we go*

In general, if HLMI is corrigible—either due to "alignment by default" or the success of safety agendas aimed at corrigibility—then we can expect that it will be possible to deal with potentially dangerous behaviors as they arise (green-circled nodes). If HLMI is incorrigible, then the risk of not catching and stopping any dangerous behaviors depends on two factors. First, it depends on the nature of these behaviors; if the HLMI is deceptively aligned (§5.1.3.1), or otherwise exhibits influence-seeking behavior, then it is less likely we will notice misbehavior (orange-circled nodes). Second, the risk depends on how fast progress in AI is when we develop HLMI: the less time that we have to catch and correct misbehavior and the more rapidly HLMI increases in capability, the more likely we are to fail to course correct as we go (purple-circled nodes).

The other way we could course correct as we go is if AI systems that are (close to) HLMI can progress on safety challenges more than on capabilities, modeled in the top left of the figure. This seems more likely if early HLMI takes the form of imitation learners (ims) or emulations of humans (ems) (§3). It also seems more likely if we reach a point in AI safety research where the remaining safety challenges are automatable. Reaching that point in turn depends on the progress of specific safety agendas—currently, we just include foundational research [126] and synthesizing a utility function [127]. As the model is further developed, it will additionally depend on the success of other safety agendas as we clarify their routes to impact.

### 7.1.2 Alignment Ahead of Time

Some safety agendas—principally those based around ambitious value learning [153]—don't seem to depend on ensuring corrigibility; instead of trying to catch misalignment issues that arise, they focus on ensuring that HLMI is successfully aligned on the first try. We model these safety agendas separately in the *HLMI aligned ahead of time* module, as they do not depend on catching dangerous behaviors post-HLMI; therefore, if they are successful, they are less affected by the speed of progress post-HLMI. The module (Figure 71) is a simplistic version of this idea, and once again, other research agendas besides the ones shown may fit into its logic. We discussed this module from a slightly different perspective in the earlier chapter on safety agendas (§6).

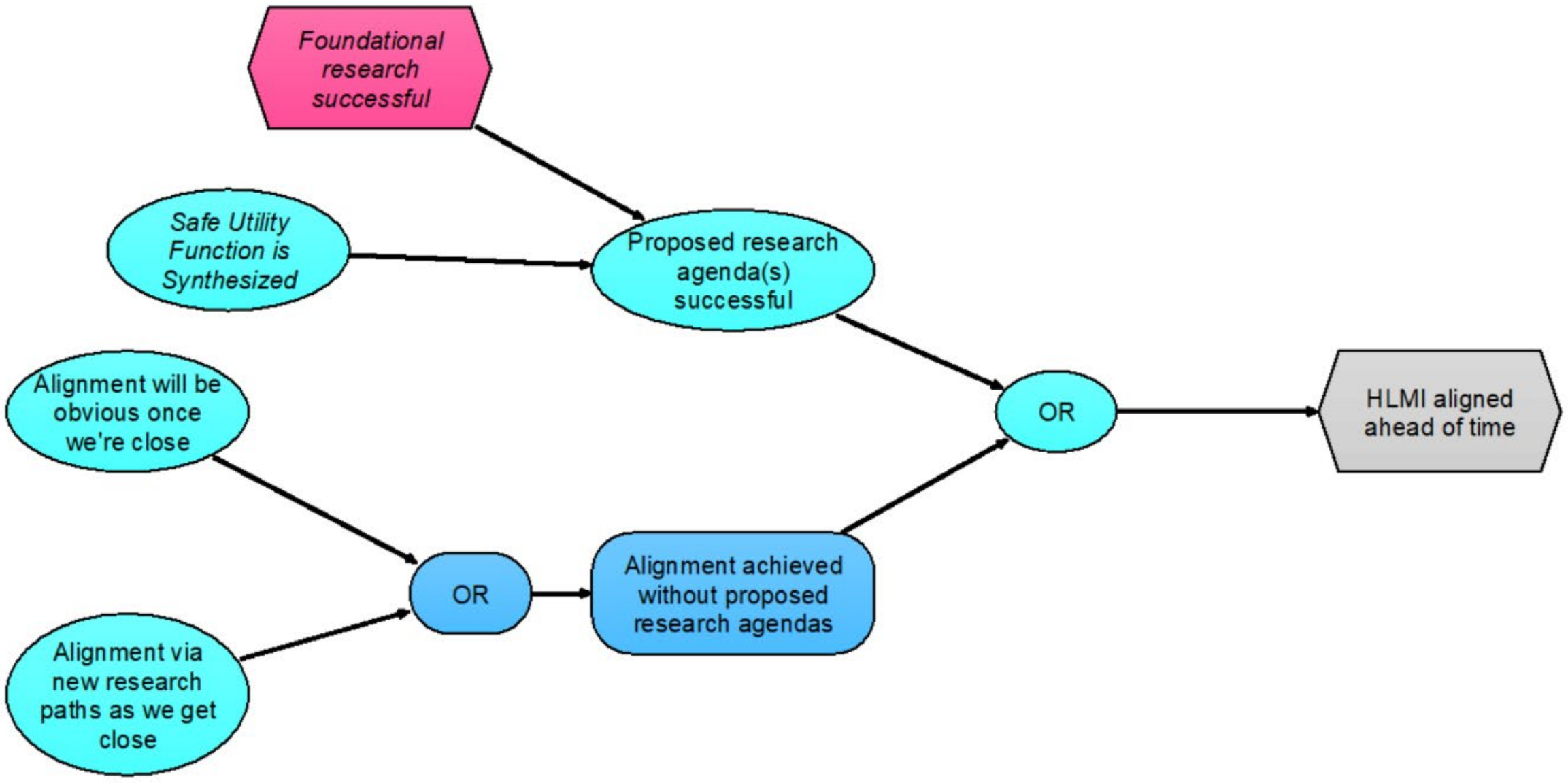

*Figure 71: HLMI aligned ahead of time*

## 7.2 Misaligned HLMI

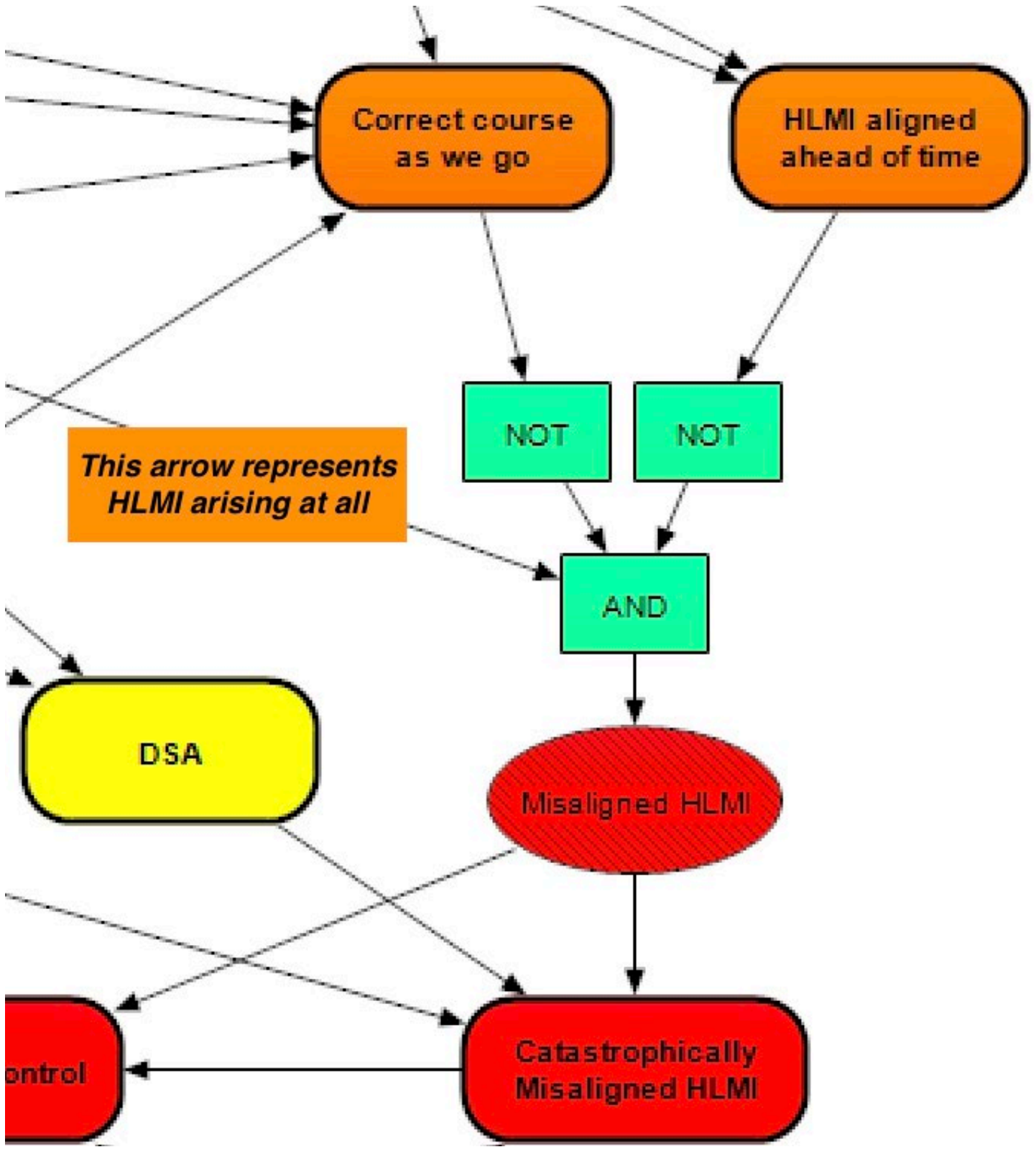

*Figure 72: Misaligned HLMI*

We decomposed the risk factor of *Misaligned HLMI* into three conditions. The first is that HLMI arises at all. The second condition is that it's not feasible to correct course as we go—this would depend on HLMI not being benign by default and humanity not developing and implementing technical solutions that iteratively align HLMI in a post-HLMI world. The third condition is that we do not find a way to align HLMI before it appears or before some other pre-HLMI point of no return. This third condition is a big uncertainty for most people concerned about AI risk, and we have discussed how it is particularly difficult to model in the earlier chapter on the impact of safety research agendas (§6).

## 7.3 Decisive Strategic Advantage (DSA)

Attaining a DSA through HLMI is a key factor in many risk scenarios. We assume that if DSA is attained, it is attained by a "project" or a coalition of different projects. A project could be a team within a tech company, a state-owned lab, a military, a rogue actor, etc. Working from right to left in the figure pictured above, the leading project to develop HLMI can only achieve DSA if *governing systems do not prevent it* from independently amassing such power. Governing systems include state governments, institutions, laws, and norms. AI systems, including other HLMI projects, can also count as governing systems. Failing the intervention of governing systems, there are three main ways to reach the potential for DSA. First, *multiple HLMI systems could form a coalition* that attains a DSA. Second, the *first HLMI may achieve a DSA* over the rest of the world before competitors arise. Third, in a world with multiple

HLMIs, *a single HLMI may achieve a DSA* over the rest of the world (including over its competitor HLMIs).

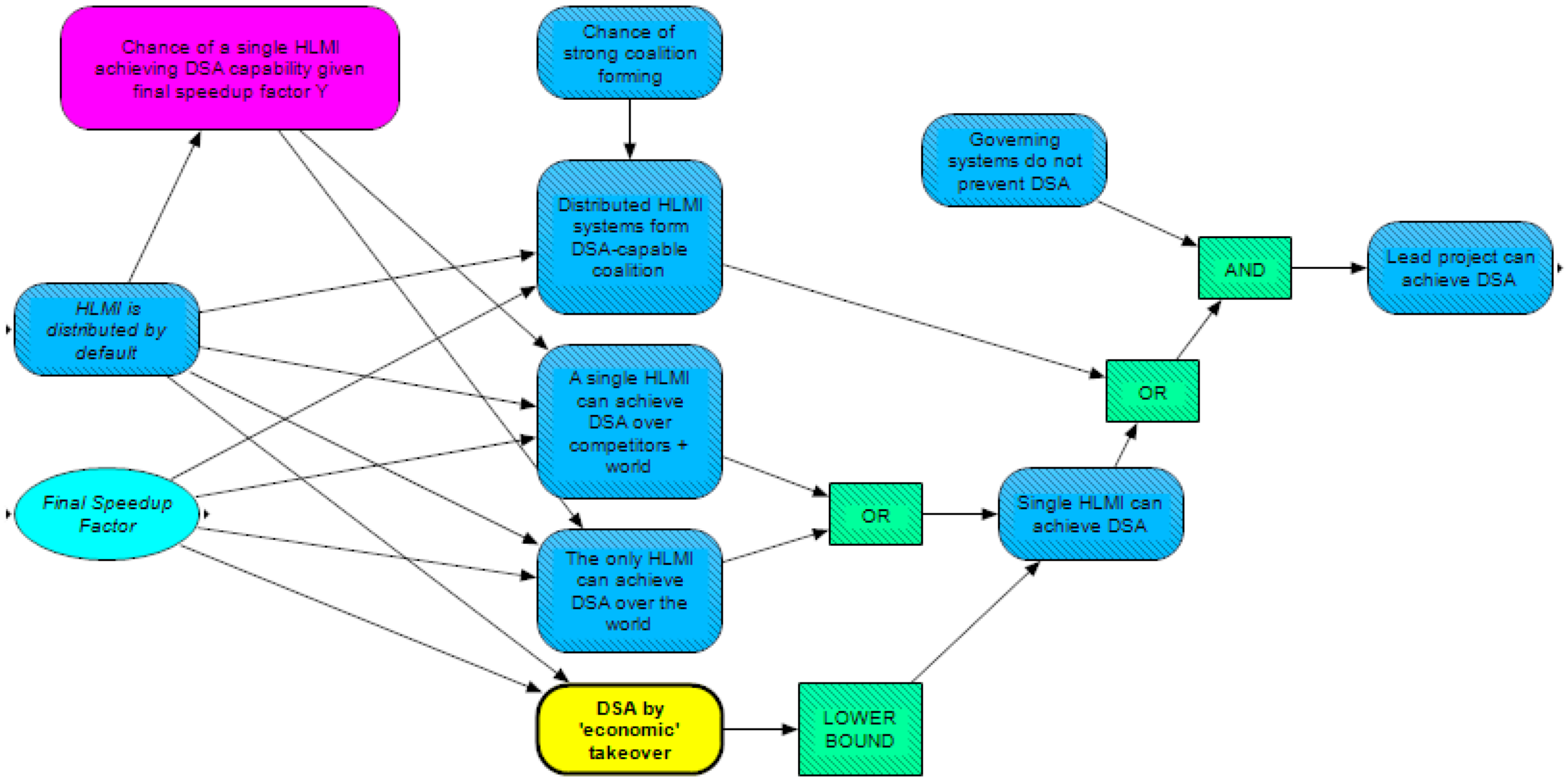

*Figure 73: Lead project can achieve DSA*

These three paths to potential DSA depend on whether HLMI is distributed by default (§4.3.2). It's also linked to the Final Speedup Factor, determined by the node for post-HLMI New Economic Doubling Time (§4.3.1). Faster growth will tend to widen the gap between the leading HLMI project and others, giving the project a greater advantage. The greater advantage makes it more likely that other projects are either abandoned or taken over by the leading project. Therefore, faster growth seems to make a single-HLMI DSA more likely, all else equal.

The level of advantage needed for DSA is difficult to anticipate or even measure. It might be argued that an overwhelming economic or technical advantage is required for a DSA, but various arguments have been put forward for why this is incorrect [154]. Many times in history, groups of humans have used political manipulation to take control of countries or empires. For instance, Hitler's initial takeover of Germany was mostly via manipulating the political system (as opposed to investing his personal savings so well that he was responsible for >50% of Germany's GDP), and Napoleon's Grande Armée was built during his takeover of Europe, not beforehand. In a more modern context, individuals have gained control over corporations via leverage and debt, without themselves having the resources needed to assert such control.

On the other hand, if these analogies don't hold and an HLMI project is unable to exert significant manipulative pressure on the rest of the world, it might require actual control over a significant fraction of all the world's resources before DSA occurs. Therefore, we must have a very wide range of uncertainty over the degree of advantage required to achieve a DSA. There are a couple of nodes intended to capture this uncertainty. The yellow *DSA by "economic" takeover* module (Figure 74) provides an upper bound on the degree of advantage needed (or equivalently, as stated in the diagram, a lower bound on the likelihood) based on the assumption that such an upper bound would be 50% of the world's GDP. The purple *Chance of a single HLMI achieving DSA* node attempts to estimate the likelihood of DSA given different takeoff scenarios, taking into account, among other things, the possibility of HLMI-assisted manipulation of human society, or of HLMI gaining control of resources or new technologies.

The economic takeover route assumes that the HLMI does not take significant, successful steps to seize resources, influence people, or attack its potential opponents and instead just outgrows the rest of the world until it is overwhelmingly powerful. The logic here is that if the lead project increased its economic position to be responsible for >50% of the world GDP, then we would assume it would already have "taken over" (most of) the world without firing a shot. If we know how large the project is when it's on the cusp of HLMI compared to the size of the world economy (both modeled in the Paths to HLMI chapter [§3]), and if we know how fast the project will grow economically once it reaches HLMI (which we can naïvely assume is similarly quickly to the economic growth rate after HLMI, as determined in our chapter on Takeoff Speeds and Discontinuities [§4]), then we can determine how long it would have to outgrow the rest of the world to represent >50% of GDP. We then might assume that if the leading project has a lead time larger than this time, it will achieve economic takeover (the lead time can itself be estimated based on the *typical lead time of large engineering projects* and what *portion of the lead time we might expect the leading project to keep during AI takeoff*). See this sequence for further discussion [155].

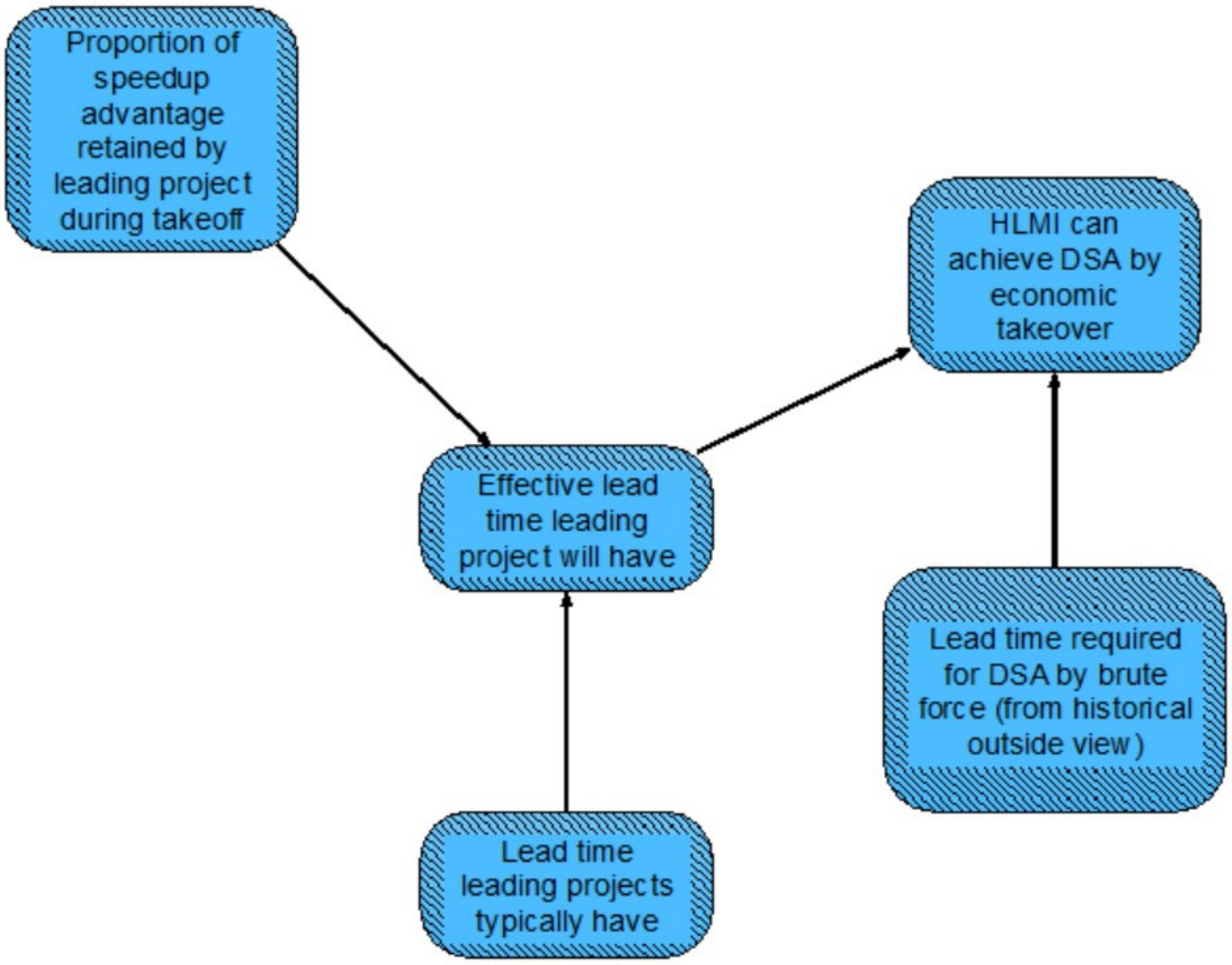

*Figure 74: HLMI can achieve DSA by economic takeover*

## 7.4 Influence-Seeking Behavior

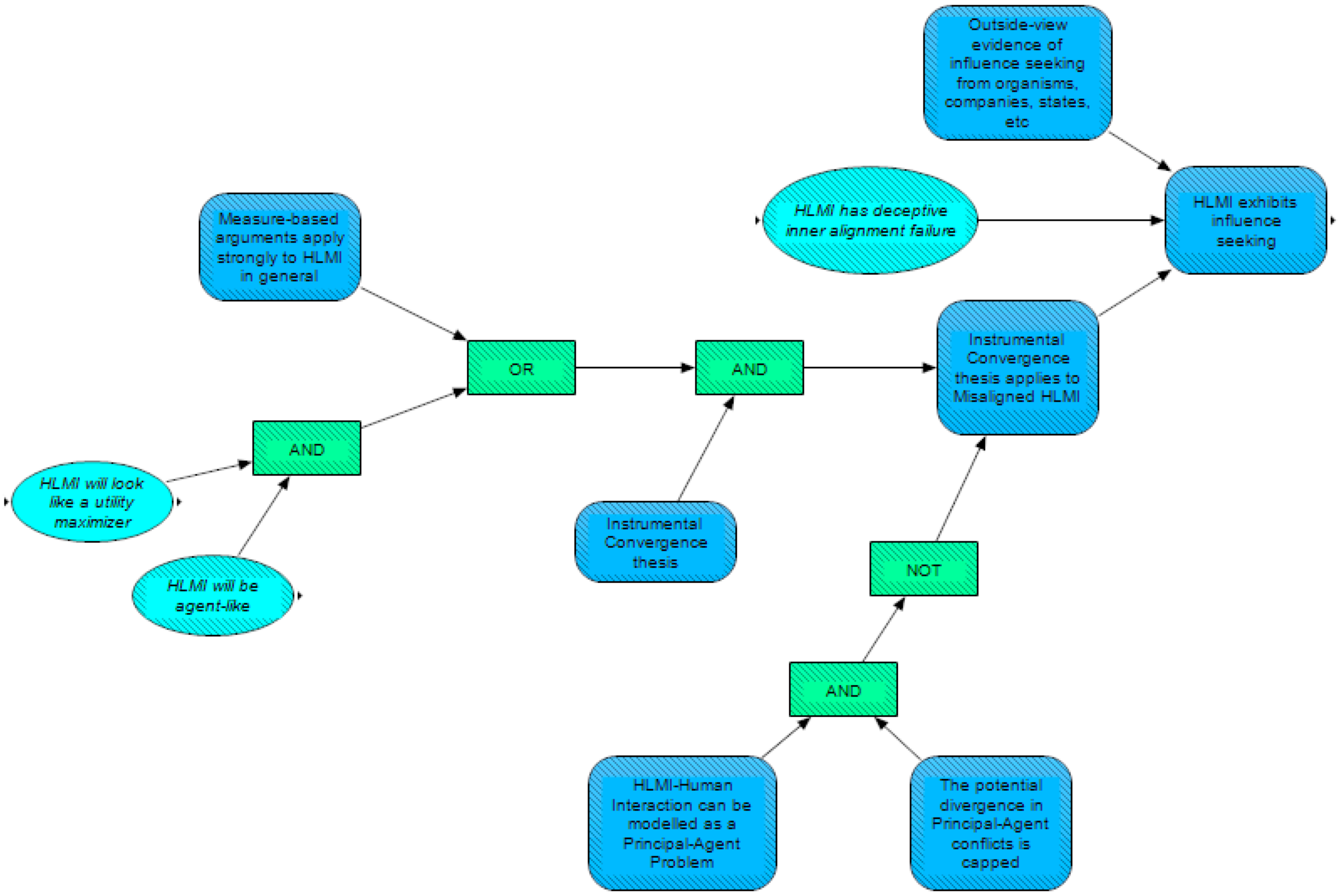

*Figure 75: HLMI exhibits influence-seeking*

Another key factor to many risk scenarios is influence-seeking behavior. We use this term to mean that an AI tends to increase its ability to manipulate people for instrumental reasons, either through improved manipulation skills or acquiring resources for the purposes of manipulation. This doesn't require that the AI system has an explicit objective of control, but having effective control may still be the goal, at least when viewing the system's behavior with an intentional stance.

As always, in assessing the likelihood of *HLMI exhibiting influence-seeking* behavior, we can use reasoning by analogy. Some plausible analogies for AI in the category of influence-seeking behavior include *organisms, companies, and states*. In each of these classes there are varying degrees of influence-seeking, so it would make sense to study the conditions for this behavior. For now, we have kept this reasoning by analogy very simplified as a single node.

Part of the *Instrumental Convergence thesis* [152] is that a sufficiently capable and rational agent will pursue instrumental goals that increase the likelihood of achieving its terminal goal(s). In general, these instrumental goals will include some commonalities, such as resource acquisition, self-preservation, and influence-seeking abilities. For this reason, whether the *Instrumental Convergence thesis applies to Misaligned HLMI* is an important factor in whether it exhibits influence-seeking. We separated this question into (a) how strong the Instrumental Convergence thesis is in general, and (b) how much HLMI meets the conditions of the thesis. For (b), one could argue that *measure-based arguments* [156] (e.g., "a large fraction of possible agents with certain properties will have this behavior") apply strongly to HLMI in general. Alternatively, there is the more direct question of whether *HLMI will look like a utility maximizer* and be agent-like. The latter is a point of disagreement that has received a lot of attention in the discourse on AI risk and is handled by the *HLMI is utility-maximizing agent* module (this module is still a work in progress and is not covered in this report).

Another important factor in whether *HLMI exhibits influence-seeking* is whether *HLMI has deceptive inner alignment failure*. This node is derived from the node *Mesa-optimizer is deceptively aligned* in the *mesa-optimization* module (§5). Deceptive alignment entails some degree of influence-seeking, because the AI will perform well on its training objective for instrumental reasons, that is, to be deployed and pursue some "true" objective. Deceptive alignment and influence-seeking are not per se the same thing, because a deceptively aligned AI may not continue to seek influence once it is deployed. However, influence-seeking algorithms might be favored among all deceptively aligned algorithms if they are more generally effective and simple to encode.

## 7.5 Catastrophically Misaligned HLMI

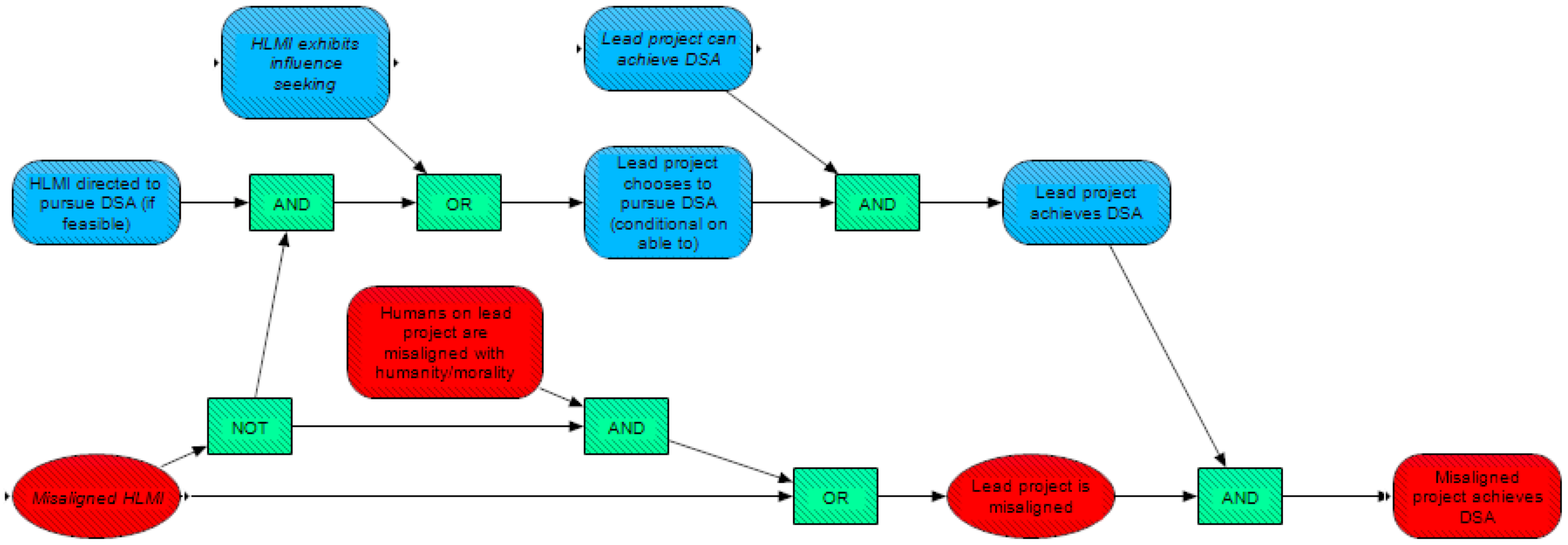

*Figure 76: Catastrophically misaligned HLMI*

Returning to the scenario of *Catastrophically Misaligned HLMI*, we need to consider that misalignment can happen in many possible ways and to many degrees. So, when can we be confident it is catastrophic? We operationalize this final output as *Misaligned project achieves DSA*, the node shown in red all the way on the right in Figure 76. The "project" includes HLMI(s) and any humans that can direct the HLMI(s), as well as coalitions of projects. If a project is capable of neutralizing any potential opposition due to holding a DSA and its interests are misaligned such that after doing so it would cause an existential catastrophe, then we assume that it will cause an existential catastrophe. Naturally, this final outcome is the conjunction of *Lead project achieves DSA* and this *Lead project is misaligned*.

We decomposed the question of whether the *Lead project achieves DSA* into whether *it can achieve DSA* (from the DSA module [§7.3]), and whether *it chooses to pursue DSA* (conditional on being able to). Then we further break the latter of these down into two ways the HLMI project could choose to pursue DSA: either the *HLMI exhibits influence-seeking* (§7.4) or the HLMI is aligned with members of the project (i.e., it would NOT be misaligned) and [is] *directed to pursue DSA*.

Regarding whether the lead project is misaligned, this can either occur if we wind up with *misaligned HLMI* (§7.2), or if we do not get misaligned HLMI (from a technical perspective) but the *humans on the lead project are sufficiently misaligned with humanity or morality* that they nevertheless cause an existential catastrophe (i.e., existential misuse).

## 7.6 Loss of Control

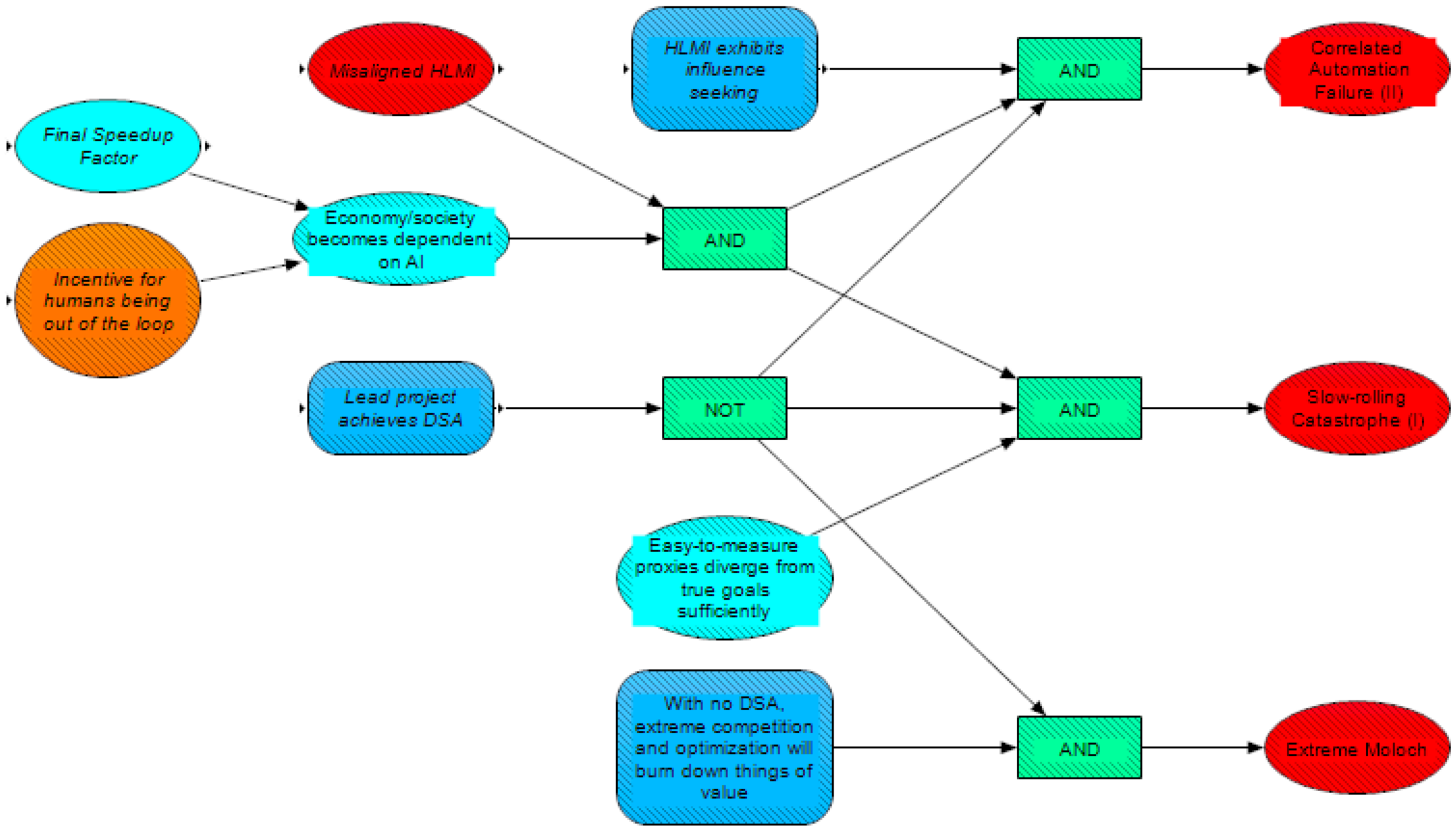

*Figure 77: Loss of control*

Here, we consider the possibility of humanity suffering existential harm from *loss of control*—not to a single HLMI or a coalition that has achieved a DSA, but instead to a broader HLMI ecosystem. Such an outcome could involve human extinction (e.g., if the HLMI ecosystem depleted aspects of the environment that humans need to survive, such as the oxygen in the atmosphere) but wouldn't necessarily (e.g., it could involve humans being stuck in a sufficiently suboptimal state with no method to course correct).

The three outcomes we model are named after popular descriptions of multipolar catastrophes. Two of these come from What failure looks like [151]: *Slow-rolling Catastrophe (I)* and *Correlated Automation Failure (II)*, and the third (*Extreme Moloch*) comes from Meditations on Moloch [157]. For each of these, we're considering a broader set of possibilities than is given in the narrow description, and we use the description just as one example of the failure mode (for a more granular breakdown of some subtypes of these scenarios, see the chapter on takeover scenarios [§8]). Specifically, here's what we're envisioning for each:

- **Slow-rolling catastrophe:** any failure mode where an ecosystem of HLMI systems irreversibly sends civilization off the rails due to technical misalignment causing HLMIs to pursue proxies of what we really want (the Production Web [158] provides another example of this type of failure).

- **Correlated automation failure:** any failure mode where various misaligned HLMIs within a broader ecosystem develop influence-seeking behavior, and then (through whatever means) collectively take power from humans (such HLMIs may or may not subsequently continue to compete for power among themselves, but either way, human interests would not be prioritized).

- **Extreme Moloch:** a failure mode where HLMI allows humans to optimize more for what we do want (as individuals or groups), but competition between individuals/groups burns down most of the potential value.

Looking at our model (and starting in the middle row of Figure 77), we see that all three of these failure modes depend on there not being a *project or coalition that achieves DSA* (since otherwise we would presumably get a *Catastrophically Misaligned HLMI* scenario). Other than that, *Slow-rolling Catastrophe (I)* and *Correlated Automation Failure (II)* share a couple of prerequisites not shared by the *Extreme Moloch* outcome. Both *(I)* and *(II)* depend on *misaligned HLMI* (for *(I)* the HLMIs will be optimizing sufficiently imperfect proxies, while for *(II)* they will be undesirably influence-seeking). *Extreme Moloch*, meanwhile, does not depend on technical misalignment, as the bad outcome would instead come from conflicting human interests (or human irrationality). Additionally, *(I)* and *(II)* would likely depend on the *economy/society becoming dependent on AI* (otherwise humans may be able to "pull the plug" and/or recover after a collapse); *Extreme Moloch*, meanwhile, does not require dependency, as even if civilization wasn't dependent on AI, humans could still compete using AI and burn down the cosmic endowment in the process. Societal dependence on AI is presumably more likely if there is a larger *incentive for humans being out of the loop* and a faster *Final Speedup Factor* to the global economy from HLMI. The previous chapter discusses in more depth how the various AI takeover scenarios assume different economic incentives and societal responses to HLMI.

Additionally, each of these failure modes has its own requirements not shared with the others. *Correlated Automation Failure* would require HLMI that was *influence-seeking*. *Slow-rolling Catastrophe* would require *relevant proxies to diverge* from what we actually want HLMI to do, to a sufficient degree that their pursuit represents an X-risk. And *Extreme Moloch* requires that in a multipolar world, *extreme competition and optimization will burn down things of value*.

Note that existential HLMI risks other than the classic "misaligned AI singleton causes existential catastrophe" is a fast-changing area of research [151], [156], [158], [159] in AI alignment, and there is very little consensus about how to distinguish potential failure modes from each other—see for example this recent survey [160]. Resultantly, our model in this section is limited in several ways: the logic is more binary than we think is justified (the outcomes shouldn't be as discrete as indicated here, and the logic from cause to effects shouldn't be as absolute either), we ignore the possibility of recovering from a temporary loss of control, we don't consider the possibility of being in a world with some aligned HLMI systems and other misaligned systems, and so on. We intend to develop this section further in response to feedback and deeper research into loss-of-control-like scenarios.

## 7.7 Conclusion

To summarize, we have presented our model of some widely discussed ways that HLMI could bring about existential catastrophe, examining some of the key mechanisms and uncertainties involved. One set of paths to catastrophe involves a misaligned HLMI or HLMI-equipped group achieving decisive strategic advantage. Another set of paths involves loss of control due to HLMI systems gaining influence on the world in an incremental and distributed fashion, without anyone achieving DSA. We traced these paths back to the outcomes of HLMI development (whether and how HLMI systems are aligned), and the nature of HLMI systems (e.g., whether they exhibit influence-seeking behavior). The chapter on AI takeover scenarios (§8) provides a more up-to-date and detailed understanding of those scenarios compared to our current concrete model.

In the next chapter, we will look at proposed takeover scenarios.

# 8 Distinguishing and Investigating AI Takeover Scenarios

Samuel Dylan Martin, Sam Clarke

In the last few years, people have proposed various AI takeover scenarios.[7] We think this type of scenario building is great, since there are now more concrete ideas of what AI takeover could realistically look like. That said, we have been confused for a while about what different assumptions are made when outlining each scenario. This chapter discusses the differences between and investigates the assumptions behind seven prominent scenarios: "brain-in-a-box" (as described in the book *Superintelligence* [161]), "What failure looks like" part 1 (WFLL 1) [151], "What failure looks like" part 2 (WFLL 2) [151], "Another (outer) alignment failure story" (AAFS) [162], "Production Web" [158], "Flash economy" [158], and "Soft takeoff leading to decisive strategic advantage" [163]. While these scenarios do not capture all of the risks from transformative AI, participants in a recent survey [160] aimed at leading AI safety and governance researchers estimated the first three of these scenarios to cover 50% of existential catastrophes from AI.[8]

This chapter combines two posts on this topic: Distinguishing AI Takeover Scenarios [164] and Investigating AI Takeover Scenarios [165]. In this chapter, we want to reveal assumptions of AI takeover scenarios that might not be obvious. Understanding these assumptions is essential for predicting which risks are most serious. We first look at three variables that distinguish these scenarios and others that are useful for thinking about them, and then look at variable social, economic, and technological characteristics of the worlds described in each of seven takeover scenarios.

## 8.1 Variables and Characteristics Relating to AI Takeover Scenarios

There are three variables which are sufficient to distinguish the takeover scenarios discussed in this chapter. We will briefly introduce these along with a number of others that are generally useful for thinking about takeover scenarios.

Key variables for distinguishing the AI takeover scenarios in this chapter:

- **Speed:** Is there a sudden jump in AI capabilities over a very short period (i.e., much faster than what we would expect from extrapolating past progress)?
- **Uni/multipolarity:** Is there a single AI system that takes over or many?
- **Alignment:** What (misaligned) goals are pursued by AI system(s)? Are they outer or inner misaligned?

[7] We define AI takeover to be a scenario where the most consequential decisions about the future get made by AI systems with goals that aren't desirable by human standards.
[8] That is, the median respondent's total probability on these three scenarios was 50%, *conditional on an existential catastrophe due to AI having occurred*.

Other variables:

- **Agency**: How agentic [166] are the AI(s) that take over? Do the AIs have large-scale objectives over the physical world, and can they autonomously execute long-term plans to reach those objectives?
- **Generality:** How generally capable are the AI(s) that take over? (vs. only being capable in specific narrow domains)
- **Competitive pressures:** To what extent do competitive pressures (incentives to develop or deploy existentially risky systems in order to remain competitive) cause or exacerbate the catastrophe?[9]
- **Irreversibility mechanism:** When and how does the takeover become irreversible?
- **Interactions between AI systems:** In the scenarios that involve multiple AI systems, do we see strong coordination between them, or conflict?[10]

The social, economic, and technological characteristics of the worlds described in the scenarios are:

- **Crucial decisions:** the specific (human) decisions necessary for takeover
- **Competitive pressures:** the strength of incentives to deploy AI systems despite the dangers they might pose
- **Takeover capabilities:** how powerful the systems executing the takeover are
- **Hackability of alignment:** the difficulty of correcting misaligned behavior through incremental fixes

Note that the scenarios we consider *do not* differ on the dimensions of agency and generality: they all concern takeovers by highly agentic, generally capable AIs—including "What failure looks like" part 1 [168]—we just stated these dimensions here for completeness.[11]

### 8.1.1 Clarifying What We Mean by Outer and Inner Alignment

Recently, there has been some discussion [3], [169]–[171] about how outer and inner alignment should be defined (along with related terms like objective and capability robustness). In this chapter, we roughly take what has become known as the "objective-focused approach" [169], while also taking into account Richard Ngo's arguments [166] that it is not actually clear what it means to implement a "safe" or "aligned" objective function.

[9] Some of the failure stories described here must assume the competitive pressures to deploy AI systems are unprecedentedly strong, as was noted by Carlsmith [167]. We plan to discuss the plausibility of these assumptions later in this chapter.

[10] How these scenarios are affected by varying the level of cooperation/conflict between HLMI systems is outside the scope of this chapter, but we plan to address it in a future post.

[11] We would welcome more scenario building about takeovers by agentic, narrow AI (which don't seem to have been discussed very much). Takeovers by non-agentic AI, on the other hand, do not seem plausible: it's hard to imagine non-agentic systems—which are, by definition, less capable than humans at making plans for the future—taking control of the future. Whether and how non-agentic systems could nonetheless cause an existential catastrophe is something we plan to address in a future post.

*Outer alignment* is a property of the objective function *used to train* the AI system. We treat outer alignment as a continuum. An objective function is outer aligned to the extent that it incentivizes or produces the behavior we actually want from the AI system.

*Inner alignment* is a property of the objective which the AI system *actually has*.[12] This objective is inner aligned to the extent that it is aligned with, or generalizes "correctly" from, the objective function used to train the system (that is, inner misalignment implies that the AI system contains or is a misaligned mesa-optimizer [§5]).

If you're new to this distinction between outer and inner alignment, you might wonder why an AI system wouldn't always just have the same objective as the one used to train it. Here is one intuition: if the training environment contains subgoals (e.g., "gaining influence or resources") which are consistently useful for scoring highly on the training objective function, then the training process may select for AI systems which care about those subgoals in ways that ultimately end up being adversarial to humans (e.g., "gaining influence at the expense of human control"). Human evolution provides another intuition: you could think of evolution as a training process that led to inner misalignment because humans care about goals other than just maximizing our genetic fitness.

This chapter builds on previous work investigating these questions. Joe Carlsmith's report on power-seeking AI [172] discussed deployment decisions and the role of competitive pressures in AI takeover scenarios *in general* (sections 5,6). Kaj Sotala's report [173] on disjunctive scenarios for AI takeover investigated competitive pressures and crucial decisions, primarily as they pertained to "brain-in-a-box" scenarios (several of the scenarios we discuss here had not been devised when that report was written).

## 8.2 AI Takeover Scenarios

Table 2 summarizes how the scenarios discussed in this chapter differ, according to the three key variables above. We then explain and illustrate the differences between the scenarios in more detail.

For clarity, we divide our discussion into slow scenarios and fast scenarios, following Critch [158]. In the slow scenarios, technological progress is incremental, whereas in fast scenarios there is a sudden jump in AI capabilities over a very short period.

[12] You can think about the objective that an AI system actually has in terms of its behavior or its internals.

*Table 2: Three key variables for the takeover scenarios*

| | **Speed: is there a sudden jump in AI capabilities over a very short period?** | **Uni/multipolarity: is there a single AI system that takes over, or many?** | **Alignment: what (misaligned) goals are pursued by AI system(s)? Are they outer or inner misaligned?** |
|---|---|---|---|
| **Outer-misaligned brain-in-a-box "superintelligence" scenario** | Yes | Unipolar | The AI system is pursuing some "correct" generalization of the misspecified objective function used to train it (e.g., "maximize production"). |
| **Inner-misaligned brain-in-a-box scenario** | Yes | Unipolar | The AI system is pursuing some arbitrary objective that emerged in the training process, which isn't aligned with the objective function to train it. |
| **Flash economy** | Yes, but not so fast as to rule out the sharing of capabilities between systems | Multipolar | AIs are pursuing some "correct" generalization of the misspecified objective function used to train them (in particular: maximizing productive output or some similarly crude measure of success). |
| **What failure looks like, part 1 (WFLL 1)** | No | Multipolar | AIs are pursuing some "correct" generalization of the misspecified objective functions used to train them (e.g., "reduce reported crimes," "increase reported life satisfaction," "maximize production"). The objective functions are actually pretty good, so the world never *looks* that bad from a human perspective—but nonetheless, humans gradually lose control. |
| **Another (outer) alignment failure story (AAFS)** | No | Multipolar | AIs are pursuing some "correct" generalization of the misspecified objective function used to train them (in particular: ensuring that things look good according to some kind of (augmented) human judgment). The objective functions are somewhat more poorly specified than in WFLL 1, so the world eventually looks worse and humans lose control more quickly. |

| | Speed: is there a sudden jump in AI capabilities over a very short period? | Uni/multipolarity: is there a single AI system that takes over, or many? | Alignment: what (misaligned) goals are pursued by AI system(s)? Are they outer or inner misaligned? |
|---|---|---|---|
| **Production web** | No | Multipolar | AIs are pursuing some "correct" generalization of the misspecified objective function used to train them (in particular: maximizing productive output or some similarly crude measure of success). This objective is *even more* misaligned with what we actually want than in AAFS, so the world eventually looks even worse and humans lose control even more quickly than in AAFS (and much more quickly than WFLL 1). |
| **What failure looks like, part 2 (WFLL 2)** | No | Multipolar | As per the "inner-misaligned brain-in-a-box scenario," AIs are pursuing some arbitrary objective that emerged in the training process, which isn't aligned with the objective function to train it. |
| **Soft takeoff leading to decisive strategic advantage** | No | Unipolar | The AI's objective isn't specified in the original scenario, but one can imagine the AI systems are pursuing misaligned objectives because of either outer or inner alignment failure. |

## 8.2.1 Fast Scenarios

### 8.2.1.1 Outer-Misaligned Brain-in-a-Box Scenario

This is the "classic" scenario that most people remember from reading *Superintelligence* [161]: in a world broadly similar to that of today, a single highly agentic AI system rapidly becomes superintelligent on all human tasks. The objective function used to train the system (e.g., "maximize production") doesn't push it to do what we really want, and the system's goals match the objective function.[13] In other words, this is an outer-alignment failure. Competitive pressures aren't especially important, though they may have encouraged the organization that trained the system to skimp on existential safety and alignment, especially if there was a race dynamic leading up to the catastrophe.

[13] We think an important, underappreciated point about this kind of failure (made by Richard Ngo [166]) is that the superintelligence probably doesn't destroy the world because it misunderstands what humans want (e.g., by interpreting our instructions overly literally)—it probably understands what humans want very well, but doesn't care because it ended up having a goal that isn't desirable by our standards (e.g., "maximize production")—in the same way that a human exploiting a tax loophole to save money may understand that the government created the tax writeoff for a different reason and doesn't want this person to exploit the loophole, but this understanding doesn't cause the person to lose interest in exploiting the loophole.

The takeover becomes irreversible once the superintelligence has undergone an intelligence explosion [174].

#### 8.2.1.2 Inner-Misaligned Brain-in-a-Box Scenario

Another version of the brain-in-a-box scenario features inner misalignment, rather than outer misalignment. That is, a superintelligent AGI could form some arbitrary objective that arose during the training process. This could happen for the reason given above (there are subgoals in the training environment that are consistently useful for doing well in training, but which generalize to be adversarial to humans), or simply because some arbitrary influence-seeking model just happened to arise during training, and performing well on the training objective is a good strategy for obtaining influence.

We suspect most people who find the brain-in-a-box scenario plausible are more concerned by this inner misalignment version. For example, Yudkowsky claims [175] to be most concerned about a scenario where an AGI learns to do something random (rather than one where it "successfully" pursues some misspecified objective function).

It is not clear whether the superintelligence being inner- rather than outer-misaligned has any practical impact on how the scenario would play out. An inner-misaligned superintelligence would be less likely to act in pursuit of a human-comprehensible final goal like "maximize production," but since in either case the system would both be strongly influence-seeking and capable of seizing a decisive strategic advantage, the details of what it would do after seizing the decisive strategic advantage probably wouldn't matter. Perhaps, if the AI system is outer misaligned, there is an increased possibility that a superintelligence could be blackmailed or bargained with, early in its development, by threatening its (more human-comprehensible) objective.

#### 8.2.1.3 Flash Economy

This scenario, described by Critch [158], can be thought of as one *multipolar* version of the outer-misaligned brain-in-a-box scenario. After a key breakthrough is made which enables highly autonomous, generally capable, agentic systems with long-term planning capability and advanced natural language processing, several such systems become superintelligent over the course of several months. This jump in capabilities is unprecedentedly fast, but "slow enough" that capabilities are shared between systems (enabling multipolarity). At some point in the scenario, groups of systems reach an agreement to divide the Earth and space above it into several conical sectors, to avoid conflict between them (locking in multipolarity).

Each system becomes responsible for a large fraction of production within a given industry sector (e.g., material production, construction, electricity, telecoms). The objective functions used to train these systems can be loosely described as "maximizing production and exchange" within their industry sector. The systems are "successfully" pursuing these objectives (so this is an outer-alignment failure).

In the first year, things seem wonderful from the perspective of humans. As economic production explodes, a large fraction of humanity gains access to free housing, food, probably a UBI, and even many luxury goods. Of course, the systems are also strategically influencing the news to reinforce this positive perspective.

By the second year, we have become thoroughly dependent on this machine economy. Any states that try to slow down progress rapidly fall behind economically. The factories and facilities of the AI systems have now also become very well-defended, and their capabilities far exceed those of humans. Human opposition to their production objectives is now futile. By this point, the AIs have little incentive to preserve humanity's long-term well-being and existence. Eventually, resources critical to human survival but noncritical to machines (e.g., arable land, drinking water, atmospheric oxygen) gradually become depleted or destroyed, until humans can no longer survive.

## 8.2.2 Slow Scenarios

We'll now describe scenarios where there is no sudden jump in AI capabilities. We've presented these scenarios in an order that illustrates an increasing "degree" of misalignment. In the first two scenarios (WFLL 1 and AAFS), the outer-misaligned objective functions are somewhat close to what we want: they produce AI systems that are trying to make the world look good according to a mixture of feedback and metrics specified by humans. Eventually, this still results in catastrophe because once the systems are sufficiently powerful, they can produce much more desirable-looking outcomes (according to the metrics they care about), much more easily, by controlling the inputs to their sensors instead of actually making the world desirable for humans. In the third scenario (Production Web), the "degree" of misalignment is worse: we just train systems to maximize production (an objective that is further from what we really want), without even caring about approval from their human overseers. The fourth scenario (WFLL 2) is worse still: the AIs have arbitrary objectives (due to inner alignment failure) and so are even more likely to take actions that aren't desirable by human standards and are likely do so at a much earlier point. We explain this in more detail below.

The fifth scenario doesn't follow this pattern: instead of varying the degree of misalignment, this scenario demonstrates a slow, *unipolar* takeover (whereas the others in this section are multipolar). There could be more or less misaligned versions of this scenario.

### 8.2.2.1 What Failure Looks Like, Part 1 (WFLL 1)

In this scenario, described by Christiano [151], many agentic AI systems gradually increase in intelligence and generality, and are deployed increasingly widely across society to do important tasks (e.g., law enforcement, running companies, manufacturing and logistics).

The objective functions used to train them (e.g., "reduce reported crimes," "increase reported life satisfaction," "increasing human wealth on paper") don't push them to do what we really want (e.g., "actually prevent crime," "actually live good lives," "increasing effective human control over resources")—so this is an outer-alignment failure.

The systems' goals match these objectives (i.e., are "natural" or "correct" generalizations of them). Competitive pressures (e.g., strong economic incentives, an international "race dynamic," etc.) are probably necessary to explain why these systems are being deployed across society, despite some people pointing out that this could have very bad long-term consequences.

There's no discrete point where this scenario becomes irreversible. AI systems gradually become more sophisticated, and their goals gradually gain more influence over the future relative to human goals. In the end, humans may not go extinct, but we have lost most of our control to much more sophisticated machines. (Imagine replacing today's powerful corporations and states with machines pursuing similar objectives, except without the possibility of internal dissent or humans in the loop.)

#### 8.2.2.2 Another (Outer) Alignment Failure Story (AAFS)

This scenario, also described by Christiano [162], is initially similar to WFLL 1. AI systems slowly increase in generality and capability and become widely deployed. The systems are outer misaligned: they pursue natural generalizations of the poorly chosen objective functions they are trained on. This scenario is more specific about exactly what objectives the systems are pursuing: they are trying to ensure that the world looks good according to some kind of (augmented) human judgment (the systems are basically trained according to the regime described in An unaligned benchmark [176]).

Problems arise along the way when systems do things that look good but aren't actually good (e.g., a factory colludes with the auditors valuing its output, giving a great quarterly report that didn't actually correspond to any revenue). Such problems tend to be dealt with via short-term fixes—improving sensor coverage to check mistakes (e.g., in a way that reveals collusion between factories and auditors) or tweaking reward functions (e.g., to punish collusion between factories and auditors). This initially leads to a false sense of security. But as the pace of AI progress accelerates and we still don't know how to train AI systems to actually help us, we eventually have extremely powerful systems, widely deployed across society, which are pursuing proxy goals that diverge from what we actually want. Specifically: "ensuring things look good according to human judgment" eventually means fooling humans and carefully controlling what gets fed into the sensors, because the AIs can produce much more desirable-looking outcomes, much more easily, by controlling the sensors instead of actually making the world good. Eventually, all humans will either be killed or totally disempowered, because this is the best way of making sure the systems' objectives are maximally positive and will remain that way forever.

To explain exactly how this scenario differs from WFLL 1, consider that outer (mis)alignment can be viewed as a continuum: "how" misspecified is the objective function used to train the AI system? On one extreme, we have objective functions that do really well at producing or incentivizing the behavior we actually want from AI systems (e.g., a reward function trained using iterated amplification to reward systems to the extent that they try to help their operators). On the other extreme, we have objective functions that don't capture anything we value (e.g., "maximize paperclips").

We find it helpful to think about the objective functions used in training as specifying a "sensory window" through which the system being trained views the world (you could even think of it as a huge set of camera feeds). This window will probably be defined by a bunch of human feedback, along with other metrics (e.g., GDP, inflation, unemployment, approval ratings). The training process is selecting for AI systems that make this sensory window look "good" according to feedback and desired values for those metrics.

Bringing these ideas together: the better defined this "sensory window" (i.e., the more outer aligned the objective function is), the better things will look from the human perspective. In WFLL 1, the sensory window is very large, rich, and well-defined, such that even as AI systems gain more and more influence relative to humans, the world continues to look pretty good to us. In AAFS, the sensory window is smaller and less well-defined, such that it's eventually easy for systems to seize their sensors and kill or disempower any humans who try to stop them.

This has a few practical implications for how AAFS plays out, compared to WFLL 1.

First, in WFLL 1, there is a decent chance (maybe 50:50) that AI systems will leave some humans alone (though still mostly disempowered). This is because the sensory window was so well-defined that it was too hard for AI systems to cause extinction without it showing up on their sensors and metrics. In AAFS, this is much less likely, because the sensory window is easier to fool.

Second, in AAFS, the point of no return will happen sooner than in WFLL 1. This is because it will require a lower level of capabilities for systems to take control without it showing up on the (more poorly defined) sensory window.

Third, in AAFS, warning shots (i.e., small- or medium-scale accidents caused by alignment failures, like the "factory colludes with auditors" example above) are more likely and/or severe than in WFLL 1. This is because a greater number of possible accidents will not show up on the (more poorly defined) sensory window.[14] A further implication here is that competitive pressures probably need to be somewhat higher—or AI progress somewhat faster—than in WFLL 1, to explain why we don't take steps to fix the problem before it's too late.

The next scenario demonstrates what happens when the objective function/sensory window is even closer to the bad extreme.

#### 8.2.2.3 Production Web

Critch's Production Web [158] scenario is similar to WFLL 1 and AAFS, except that the objective functions used to train the systems are more severely outer misaligned. Specifically, the systems are trained to "maximize productive output" or another similarly crude measure of success. This measure defines an even narrower sensory window onto the world than for the systems in WFLL 1 and AAFS—it isn't even superficially aligned with what humans want (the AI systems are not trying to optimize for human approval at all).

"Maximizing productive output" eventually means taking steps that aren't desirable from the human perspective (e.g., using up resources critical to human survival but noncritical to machines, like drinking water and atmospheric oxygen).

The implications of this even more (outer) misaligned objective follow the same pattern we described when comparing AAFS with WFLL 1. In the Production Web scenario:

[14] This does assume that systems will be deployed *before* they are capable enough to anticipate that causing such "accidents" will get them shut down. Given there will be incentives to deploy systems as soon as they are profitable, this assumption is plausible.

- Human extinction is the only likely outcome (keeping humans alive becomes counterproductive to maximizing productive output).
- The point of no return will happen even sooner (AI systems will start, e.g., using up resources critical to human survival but noncritical to machines as soon as they are capable enough to ensure that humans cannot stop them, rather than having to wait until they are capable enough to manipulate their sensors and human overseers).
- Warning shots will be even more likely/severe (since their objectives are more misaligned, fewer possible accidents will be punished).
    - Competitive pressures therefore need to be even higher.

Another point of comparison: you can also view this scenario as a slower version of the Flash Economy, meaning that there is more opportunity for incremental progress on AI alignment or improved regulation to stop the takeover.

## 8.2.3 Further Variants of Slow, Outer-Alignment Failure Scenarios

If systems don't develop coherent large-scale goals over the physical world, then the failures might take the form of unorganized breakdowns or systems "wireheading" themselves (i.e., trying to maximize the contents of their reward memory cell) without attempting to seize control of resources.

We can also consider varying the level of competitive pressure. The more competitive pressure there is, the harder it becomes to coordinate to prevent the deployment of dangerous technologies. Especially if there are warning shots (i.e., small- or medium-scale accidents caused by alignment failures), competitive pressures must be unusually intense for potentially dangerous HLMI systems to be deployed en masse.

We could also vary the competence of the technical response in these scenarios. The more we attempt to "patch" outer misalignment with short-term fixes (e.g., giving feedback to make the systems' objectives closer to what we want, or to make their policies more aligned with their objectives), the more likely we are to prevent small-scale accidents. The effect of this mitigation depends on how "hackable" the alignment problem is: perhaps this kind of incremental course correction will be sufficient for existentially safe outcomes. But if it isn't, then all we would be doing is deferring the problem to a world with even more powerful systems (increasing the stakes of alignment failures), and where inner-misaligned systems have been given more time to arise during the training process (increasing the likelihood of alignment failures). So, in worlds where the alignment problem is much less "hackable," competent early responses tend to defer bad outcomes into the future, and less competent early responses tend to result in an escalating series of disasters (which we could hope leads to an international moratorium on AGI research).

### 8.2.3.1 What Failure Looks Like, Part 2 (WFLL 2)

Described by Christiano [151] and elaborated further by Joe Carlsmith [167], this scenario sees many agentic AI systems gradually increase in intelligence and be deployed increasingly widely across society to do important tasks, just like WFLL 1.

But then, instead of learning some natural generalization of the (poorly chosen) training objective, there is an inner alignment failure: the systems learn some unrelated objective(s) that arise naturally in the training process; that is, the objectives are easily discovered in neural networks (e.g., “don’t get shut down”).[15] The systems seek influence as an instrumental subgoal (since with more influence, a system is more likely to be able to, e.g., prevent attempts to shut it down).[16] Early in training, the best way to do that is by being obedient (since it knows that unobedient behavior would get it shut down). Then, once the systems become sufficiently capable, they attempt to acquire resources and power to more effectively achieve their goals.

Takeover becomes irreversible during a period of heightened vulnerability (a conflict between states, a natural disaster, a serious cyberattack, etc.) before systems have undergone an intelligence explosion. Quoting from [151], this could look like a “rapidly cascading series of automation failures: a few automated systems go off the rails in response to some local shock. As those systems go off the rails, the local shock is compounded into a larger disturbance; more and more automated systems move further from their training distribution and start failing.” After this catastrophe, “we are left with a bunch of powerful influence-seeking systems, which are sophisticated enough that we can probably not get rid of them.”

Compared to the slow outer-alignment failure scenarios, the point of no return in this scenario will be even sooner (all else being equal), because AIs don’t need to keep things looking good according to their somewhat human-desirable objectives (which takes more sophistication)—they just need to be able to make sure humans cannot take back control. The point of no return will probably be even sooner if the AIs all happen to learn similar objectives or have good cooperative capabilities (because then they will be able to pool their resources and capabilities, and hence be able to take control from humans at a lower level of individual capability).

You could get a similar scenario where takeover becomes irreversible without any period of heightened vulnerability, if the AI systems are capable enough to take control without the world being chaotic.

#### 8.2.3.2 Soft Takeoff Leads to Decisive Strategic Advantage

This scenario, described by Kokotajlo [163], starts off much like “What failure looks like.” Many general agentic AI systems get deployed across the economy and are misaligned to varying degrees. AI progress is much faster than today, but there is no sudden jump in AI capabilities. Each system has some incentive to play nice and obey governing systems. However, then one particular AI is able to buy more computing hardware and invest more time and resources into improving itself, enabling it to do more research and pull further ahead of its competition, until it can seize a decisive strategic advantage and defeat all opposition. This would look a lot like the brain-in-a-box superintelligence scenario, except it would be occurring in a world that is already very different to today. The system that takes over could be outer or inner misaligned.

[15] So, for this failure scenario, it isn’t crucial whether the training objective was outer aligned.
[16] Of course, not *all* arbitrarily chosen objectives, and not all training setups, will incentivize influence-seeking behavior, but many will.

### 8.2.4 Different Assumptions between Slow and Fast Scenarios

As we look into the various assumptions the above scenarios require, the starting point for our investigation is the following observation: fast brain-in-a-box scenarios assume that takeover probably cannot be prevented after the misaligned HLMI is deployed (due to very rapid capability gain), but the slow scenarios involve an extended period where misaligned AIs are deployed, incremental improvements to alignment are attempted, and, in some cases, warning shots (small-scale disasters that indicate that AI is unsafe) happen.

Therefore, the slow scenarios must provide an explanation as to why many actors persist in deploying this dangerous technology over several years. These social and economic assumptions can be thought of as substituting for the assumption of very fast progress that is key to the fast scenarios—the rapid capability gain with no time to respond is replaced by a slower capability gain and an ineffective response.

If the slow scenarios capture reality better than the fast scenarios, then systems will be deployed deliberately and will initially be given power rather than seizing power. This means both that the systems won't be so obviously dangerous that the misbehavior is noticed early on and that there is still misalignment later on.[17] Carlsmith [167]:

> The question, then, isn't whether relevant actors will intentionally deploy systems that are already blatantly failing to behave as they intend. The question is whether the standards for good behavior they apply during training/testing will be adequate to ensure that the systems in question won't seek power in misaligned ways on any inputs post-deployment.

Just from this initial observation, we know that there are several differences in the assumptions of slow and fast scenarios that go beyond just technical factors or overall results like whether the outcome is unipolar or multipolar. This led us to investigate exactly how particular slow and fast scenarios differ in the broader set of assumptions they make.

## 8.3 Takeover Characteristics

Our initial table of characteristics for AI takeover scenarios (Table 2) discussed the primary and overt characteristics of a takeover—whether it was unipolar or multipolar, whether it involved rapid or slow capability gain, and how and why the AI systems were misaligned. Here, we present a table of secondary

[17] This switch from apparently benign to dangerous behavior could be due to

- Power-seeking misaligned behavior that is too subtle to notice in the training environment but is obviously dangerous in deployment, due to the scale and makeup of the training and deployment environments being quite different.
- Power-seeking misaligned behavior that only shows up over long time horizons and therefore will not be noticed in training, which we might expect occurs over a shorter period than deployment.
- Systems intentionally hiding misaligned behavior during training to deceive their operators. Systems could be highly deceptively misaligned from the beginning and capable enough to know that if they seek power in adversarial ways too early, they will get shut down. Michaël Trazzi and Stuart Armstrong argue [177] that ML models don't have to be extremely competent to be manipulative, suggesting that these behaviors might show up very early.

characteristics of AI takeover scenarios—factors that influence these primary characteristics or depend on them in various ways.

The characteristics of AI takeover can be divided into, first, social and economic factors (*crucial decisions* and *competitive pressures)* and, second, technical factors (*takeover capabilities* and *alignment "hackability")*.

*Crucial decisions* and *competitive pressures* are two ways of looking at the preconditions for an AI takeover scenario. The first is a local view, focusing on particular mistaken decisions (e.g., around deploying a dangerous AI). The second is a broad view, focusing on the presence of perverse economic or political incentives. These overlap: perverse competitive pressures can cause bad decisions to be made, and key decisions about oversight or regulation can lessen or intensify competitive.

*Takeover capabilities* and *alignment "hackability"* are assumptions each scenario makes about the competence of the AIs which take over and how difficult it is to align them using short-term, case-by-case fixes. There are complicated relationships between the assumptions you make about these technological questions and the assumptions you make about social factors. Roughly speaking, the weaker the competitive pressures and the more competently crucial decisions are made, the more capable the AIs must be and the harder (less "hackable") alignment must be for disaster to occur. However, note that if hackability is very low, we might have enough warning shots to avoid developing dangerous AI in the first place. These relationships are discussed in more detail in §8.6.

Table 3 presents our best guess of what the crucial decisions, degree and cause of competitive pressures, assumed capabilities for AI takeover and hackability (effectiveness of short-term fixes) in different takeover scenarios are. In the following section, we then discuss each scenario from these perspectives. You may want to refer back to our first summary table.

*Table 3: Four characteristics of takeover scenarios*

| | **Crucial Decisions**<br><br>(Identifiable decisions made by humans that lead to takeover) | **Competitive Pressures**<br><br>(Strength and nature of incentives to deploy AI) | **Takeover Capabilities**<br><br>(What capabilities do the AIs employ to execute takeover) | **Alignment 'Hackability'**<br><br>(The extent to which short-term fixes are sufficient for aligning systems on all inputs which they will in fact receive) |
|---|---|---|---|---|
| **Outer-misaligned brain-in-a-box "superintelligence" scenario**<br><br>**Inner-misaligned brain-in-a-box scenario** | Choose to develop HLMI<br><br>(*If released deliberately*: Choose to deploy HLMI) | (*Race dynamic may be present in leadup to HLMI development*) | Rapid Capability Gain<br><br>Ability to seize DSA or major advantage over the rest of the world from ~nothing<br><br>If not released deliberately—has to escape | Irrelevant (no time for fixes) |
| **Flash economy** | Choose to develop HLMI<br><br>Choose to release system open-source / share research | Enough to allow initial deployment of the HLMI systems | Ability to seize DSA or major advantage over the rest of the world from strong starting point | Could be fairly high, not much time for fixes |
| **What failure looks like, part 1**<br><br>**(WFLL 1)** | Choose to develop HLMI<br><br>Choose to automate systems on a large scale<br><br>Inadequate response to warning signs | Incentives to keep deploying AI<br><br>Some pressure to fix small errors | Irrelevant, loss of control occurs without takeover | Moderate |

| | Crucial Decisions | Competitive Pressures | Takeover Capabilities | Alignment 'Hackability' |
|---|---|---|---|---|
| **Another (outer) alignment failure story**<br>**(AAFS)** | Choose to develop HLMI<br>Choose to automate systems on a large-scale Inadequate response to warning signs and small disasters | Incentives to keep deploying AI<br>Significant pressure to fix small errors | Ability to seize DSA or major advantage over the rest of the world from strong starting point | Lower than WFLL 1 |
| **Production web** | Choose to develop HLMI<br>Choose to automate systems on a large scale<br>Inadequate response to warning signs and small disasters | Strong incentives to keep deploying AI<br>No real pressure to fix small errors | Ability to seize DSA or major advantage over the rest of the world from strong starting point | Similar to WFLL 1 |
| **What failure looks like, part 2**<br>**(WFFL 2)** | Choose to develop HLMI<br>Choose to automate systems on a large scale<br>Inadequate response to warning signs and escalating series of disasters | Strong incentives to keep deploying AI | Ability to seize DSA or major advantage over the rest of the world after some weakening event | Low |
| **Soft takeoff leading to DSA** | Choose to develop HLMI<br>Government or research group centralizes research effort and achieves strong lead | Race dynamic | Ability to seize DSA or major advantage over the rest of the world from resources of initial project | (Low enough that whatever is tried during testing of system fails) |

## 8.4 Discussion of Scenarios

Here we discuss each of the seven scenarios in depth from the perspective of *crucial decisions*, *competitive pressures*, *takeover capabilities*, and *alignment hackability*.

### 8.4.1 Outer- and Inner-Misaligned Brain-in-a-Box Scenarios

In brain-in-a-box scenarios, the main crucial decisions occur early on and involve development (and possibly voluntary deployment) of the first and only HLMI, with the assumption that once this HLMI is deployed, it's game over. Depending on the anticipated level of capability, the system might also be capable of talking its way into being deployed during testing or escaping its testing environment, or else it might be voluntarily deployed. This particular critical decision—the choice to deploy systems—was discussed by Sotala in depth [173].

As well as anticipated economic benefit, the systems could be voluntarily released for unethical reasons—terrorism, criminal profit, ideological motives, or a last-ditch mutually assured destruction attempt.

Competitive pressures to allow the AI to proliferate despite the risks it poses aren't that important because after deployment, the AI rapidly completes its takeover and there is no chance for opposition. A race dynamic due to anticipated economic benefit or military power may well be present and might explain why the system got developed in the first place, but unlike with the slow scenarios, there aren't noticeable competitive pressures explaining how the AI takes over after release. Alignment hackability also doesn't become an issue—there's no time to incrementally correct the system because it increases in capability too quickly.

### 8.4.2 Flash Economy

The scenario unfolds quickly once the requisite jump in capability has been made (over a few months), but unlike the brain-in-a-box scenarios, there are multiple highly capable systems in the world. Crucially, the breakthroughs required to create the "distributed autonomous organizations" (highly capable TAIs in this scenario) have to either be leaked, hacked, or shared (e.g., open-sourced or shared between particular companies) rapidly, so that the technology isn't monopolized by one group, leading to a DSA.

The agents—distributed autonomous organizations—proliferate quickly after the required technology is developed. Because of the extreme speed with which the agents proliferate, the large benefit they deliver early on, and their decentralized nature, there are strong incentives against interference by government and regulation (competitive pressures).

The agents aren't capable of executing a takeover immediately after being deployed, but do so once they have built up their own infrastructure (takeover capabilities). Lastly, because of how fast the scenario unfolds and the fact that the agents are mostly left alone, alignment might be fairly hackable and corrections easy to apply. As with outer-misaligned brain-in-a-box, once the systems are released there's just no opportunity to coordinate and actually do this, so even if some systems are controlled with incremental improvements, many escape human attention or they avoid human interference through methods such as regulatory capture or the economic benefit they deliver.

### 8.4.3 What Failure Looks Like, Part 1 (WFLL 1)

In WFLL 1 there are fewer crucial decisions. AI systems gradually increase in capability and are used throughout the economy. Therefore, there has to be no successful concerted effort to prevent this sort of heavy automation of the economy (so a lack of restrictive regulation or litigation), but otherwise there are few identifiable specific decisions that need to be made. Competitive pressures—mainly arising from the direct economic benefit the systems provide and their benefit to stakeholders, are quite strong. In this scenario, a fraction of people are aware that things are proceeding along a dangerous path, yet AI deployment continues. However, there aren't many visible small-scale disasters, so competitive pressures needn't be exceptionally strong (i.e., sufficient to maintain deployment even in the face of warning shots).

The systems don't execute an overt takeover at any point, so the required capability for takeover is effectively nil—they are just delegated more and more power until humanity loses control of the future. There also aren't many obvious disasters as things proceed, and the final result of the scenario doesn't necessarily involve human extinction. Since the systems don't end up so egregiously misaligned that they execute a violent takeover, there is some, probably intermittent, effort to incrementally fix systems as they malfunction. Therefore, the hackability of AI alignment in this scenario is neither very high (otherwise we wouldn't lose control eventually) nor very low (in which case the systems would end up egregiously misaligned and execute a violent takeover, definitely resulting in extinction).

### 8.4.4 Another (Outer) Alignment Failure Story (AAFS)

AAFS is subtly different from WFLL 1 in several key ways. The crucial decisions are the same as WFLL 1, except that this scenario specifies there are many early warning signs of misaligned behavior, but the response to these accidents is always incremental patches and improvements to oversight rather than blanket bans on automation or rethinking our overall approach to AI development. Competitive pressures are somewhat strong, with direct economic benefit and benefit to shareholders again playing key roles in explaining why we persist in deploying dangerous systems.

However, the scenario also specifies that there are many varied attempts at incremental improvements to HLMI systems in response to each failure—since these attempts are a key part of the story (unlike WFLL 1) but the result (definite extinction) is worse than in WFLL 1, and this scenario assumes that alignment hackability is lower than WFLL 1 (also see Paul's comment that this scenario is one where "the alignment problem is somewhat harder than I expect" [162]). This also means that the scenario assumes competitive pressures are *weaker* than in WFLL 1, as there is much more coordination around attempting to patch mistakes compared to WFLL 1 (see Paul's comment that this scenario is one where "society handles AI more competently than I expect" [162]). However, while there are more attempts at reining in AI than in WFLL 1, the competitive pressures aren't reduced by enough to prevent eventual AI takeover.

Lastly, this scenario does feature a takeover executed by systems that physically and violently seize control of their sensors and feedback mechanisms—the takeover capabilities must therefore include technological capabilities such as cyberoffense, drones, and advanced nanotechnology, rather than primarily effective persuasion tools and other "quiet" means.

However, unlike the brain-in-a-box scenarios, the AI systems are already highly embedded in the economy when they take over, so they are starting from a much stronger position than brain-in-a-box Ais, including control of lots of physical resources, factories, and drones. Therefore, the technological capabilities required for takeover are lower.

### 8.4.5 Production Web

Production Web is similar to AAFS in terms of crucial decisions, except that the systems that proliferate in Production Web gain their large-scale goals without much deliberate planning or testing at all (agentic systems with narrow goals like fulfilling a specific market niche knit together into a "production web" by themselves). The competitive pressures, primarily from economic benefit and benefit delivered to stakeholders, must be very strong for this process to proceed (stronger than in AAFS/WFLL 1) despite the fact that it occurs over multiple years and with obvious signs that humanity is losing control of the situation. Regulatory capture and benefit to stakeholders are emphasized as reasons why development of the production web is not halted, but there is less focus on the ambiguity of the situation compared to WFLL 1 (since the outcome is much more obviously disastrous in Production Web).

Alignment hackability is similar to AAFS—in both cases, incremental fixes work for a while and produce behavior that is at least beneficial in the short term. The difference is that because competitive pressures are stronger in Production Web compared to AAFS, there is less effort put into incremental fixes and so systems end up going off the rails much sooner.

Like AAFS, the takeover occurs when the systems are already highly embedded in the world economy, but probably occurs earlier and with a somewhat lower barrier to success, since the systems don't need to seize control of sensors to ensure that things continue to "look good." Otherwise, the takeover route is similar to AAFS, though the story emphasizes resources being consumed and humanity going extinct as a side effect, rather than systems seizing control of their sensors and oversight systems.

### 8.4.6 What Failure Looks Like, Part 2 (WFLL 2)

WFLL 2 involves an inner alignment failure, so setting up the training in ways that disincentivize power-seeking behavior less will be very hard, as by specification, power-seeking is a strong attractor state. Therefore, hackability is low. This has various other effects on the scenario. The crucial decisions probably involve a greater neglect of potential risks than in WFLL 1, especially because the warning shots and small-scale failure modes in WFLL 2 are more likely to take the form of violent power-seeking behavior rather than comparatively benign mistakes (like auditor-AIs and factory-AIs colluding).

The competitive pressures have to be strong to explain why systems keep getting deployed despite the damage they have already inflicted.

Christiano describes the takeover as occurring at a point of heightened vulnerability—both because this is a Schelling point where different systems can coordinate to strike and because the minimum level of capability required for a takeover is lower. Since the systems will execute a takeover at the first opportunity and during a period of heightened vulnerability (and will therefore be attempting takeover much earlier), the required capabilities for takeover are lower in this scenario, compared to AAFS/Production Web.

### 8.4.7 Soft Takeoff Leads to Decisive Strategic Advantage

Soft takeoff leading to decisive strategic advantage (DSA) (§7.3) has an extra assumption on top of the preconditions for AAFS/WFLL 1/Production Web—that one particular research group is able to secure significant lead time over competitors such that it can defeat both humanity and rival AIs. Given this assumption, what's going on in the rest of the world, whether the other AI systems are aligned or not, is irrelevant.

The leading project is probably motivated by a strategic race for military or economic dominance, since it has secured enough resources to dominate the rest of the world. The required takeover capability is very high as the system is competing against other transformative AI systems, although not quite as high as in the brain-in-a-box scenario, as this leading project starts out with a lot of resources. Alignment cannot be hackable enough that the leading project is able to successfully align the AI system in the development time it has, but otherwise the exact level of hackability is underdetermined.

## 8.5 New Scenarios

Here, we present some scenarios devised by varying one or more of the takeover characteristics.

### 8.5.1 Soft Takeoff and Decisive Strategic Advantage by Narrow AI

We devised this scenario by setting the necessary "takeover capabilities" to be very low.

This scenario is similar to "Soft takeoff leads to decisive strategic advantage," except that the single system which takes over is not that much more capable than its rivals. Rather, it simply has a single good trick that enables it to subvert and take control of the rest of the world. Its takeover capability might be exceptionally good manipulation techniques, specific deadly technology, or cyberoffensive capability, any of which could allow the system to exploit other AIs and humans. This removes the assumption that a lot of research effort will need to be concentrated to achieve a DSA and replaces it with an assumption that there is some unique vulnerability in human society which a narrow system can exploit. Implicit in this scenario is the assumption that generally capable AI is not needed in order to take on an extraordinary research effort to find this vulnerability in human society.[18]

---

[18] In reality, we feel that this is more of a fuzzy rather than binary thing: we expect this to require *somewhat* less of an extraordinary research effort and instead require that there exists *somewhat* more of a crucial vulnerability in human society (there are already some examples of vulnerabilities, e.g., biological viruses and humans being easy to manipulate under certain conditions).

## 8.5.2 Compounding Disasters

We devised this scenario by assuming the competitive pressures are very high, crucial decisions are very incompetent, and hackability is very low.

This scenario is similar to AAFS, with HLMI systems widely being deployed, pursuing goals that are reasonable proxies for what humans actually want, and demonstrating some misbehavior. However, instead of the small-scale failures taking the form of relatively benign “warning shots” that lead to (failed) attempts to hack AI systems to prevent future errors, the small-scale disasters cause a large amount of direct damage. For example, an AI advisor misleads the government, leading to terrible policy mistakes and a collapse of trust, or autonomous weapon systems go rogue and attack cities before being taken out. The result of this is a compounding series of small disasters that rapidly spiral out of control, rather than attempted patches staving off disaster for a while before a single sudden AI takeover. In the end, the AI takeover occurs at a period of heightened vulnerability *brought about by previous medium-sized AI-related disasters*. Therefore, AI systems in this scenario need not be as competent as in AAFS or even WFLL 2 to take over. Alignment may be easily hackable in this situation, but such damage has been done by early, agentic, narrow AIs that no such fixes are attempted.

## 8.5.3 Automated War

A situation rather like the AAFS scenario plays out, where the economy becomes dependent on AI, and we lose control of much key infrastructure. Capable, agentic AI systems are built which do a good job of representing and pursuing the goals of their operators (inner and outer aligned). These are deployed on a large scale and used to control armies of drones and automatic factories, as well as the infrastructure needed for surveillance, for the purposes of defending countries.

However, there are key flaws in the design of the AI systems that only become apparent after they are in a position to act relatively independent of human feedback. At that point, flaws in their ability to both model each other and predict their own chances of winning potential contests over valuable resources lead to arms races and ultimately destructive wars that the AIs have precommitted to pursue.

This scenario probably involves a stronger role for military competition, instead of just economic competition, and also involves a particular kind of cooperation-relevant capability failure—systems failing to behave correctly in multiagent situations (along with an intent alignment failure that means the systems can’t just be told to stand down when they start going against the interests of their operators).

From the perspective we are taking in this chapter, there need to be particular crucial decisions made (automation of military command and control) as well as strong military competitive pressures and a likely race dynamic. Alignment is not very hackable for a specific reason—the multiagent flaw in AIs is not easy to detect in testing or soon after deployment.

### 8.5.4 Failed Production Web

The preconditions for Production Web play out as described in that scenario, where agentic AI systems each designed to fill specific market niches attempt to integrate together. However, due to either specific defects in modeling other AIs or inner misalignment, the systems are constantly seeking ways to exploit and defraud each other. These attempts eventually result in AI systems physically attacking each other, resulting in a chaotic war that kills humans as a side effect. This is similar to "automated war," but with different competitive pressures. There is less of a focus on strategic competition and more of a focus on economic competition, and similar assumptions to Production Web about very strong competitive pressures being required.

## 8.6 Discussion of Characteristics

Below, we discuss the four characteristics we have identified (critical decisions, competitive pressures, threshold for takeover, and level of alignment hackability) and give an assessment of the reasons why you might expect them to be at one extreme or another (crucial decisions made unusually competently/incompetently [§8.6.1], very strong/very weak competitive pressures to deploy AI systems [§8.6.2], a low/high bar for AIs to be capable enough to take over [§8.6.3], easy/hard alignment hackability [§8.6.4]).

### 8.6.1 Crucial Decisions

In all the scenarios discussed, we can identify certain decisions which governments and companies must make. Most obviously, research into agentic AI has to be pursued for long enough to produce significant results, and this would have to include a lack of sufficient oversight and no successful decisions to halt research in the face of significant risk. Some scenarios also involve cases where AIs that obviously pose a risk are deliberately released.

A scenario is less plausible if many crucial decisions must all be made wrongly for the scenario to come about. A scenario is more plausible if varying whether actors make the wrong choice at many stages of HLMI development doesn't change whether the scenario happens.

This is important, especially because it is very difficult to assess what choices actors will actually make while HLMI develops (and we won't try to figure this out in this report [178]). By finding out how many crucial decisions are relevant for a given AI takeover scenario and how many actors have the options of making such decisions, we can get a better understanding of how plausible they are, despite our confusion about what governments and companies would decide in particular cases. There is an extensive discussion of the plausibility of some potential crucial decisions on page 326 and after [173] of Kaj Sotala's report.

### 8.6.2 Competitive Pressures

"Competitive pressures" is a characteristic that describes how strong the incentives will be to keep deploying dangerous AI, even in the face of significant risk. There has been some discussion of the implied strength of competitive pressures in the slow and fast scenarios. Clarke [179] lists some reasons, which we discuss below, to expect that there will be strong pressures to deploy dangerous HLMI:

- Short-term incentives and collective action

  **Economic Incentives:** Since HLMI will be economically valuable in the short-term, incentives might lead us to cut corners on safety research, especially checks on how models generalize over long time horizons.

  **Military Incentives:** Even in its early stages, HLMI might provide an unchallengeable military advantage, so states would have an extra incentive to compete with each other to produce HLMI first.

- Regulatory capture

  **AI actions benefit stakeholders:** There will be many particular beneficiaries (as distinct from benefits to the overall economy) from HLMI systems acting in misaligned ways, especially if they are pursuing goals like "make money" or "maximize production." This means the stakeholders will have both the resources and motivation to water down regulation and oversight.

  **AI existence provides value (due to IP):** If financial markets realize how valuable HLMI is ahead of time, the developers can quickly become extremely wealthy ahead of deployment once they demonstrate the future value they will be able to provide (before the HLMI has had time to act in the world to produce economic benefit). This again gives stakeholders resources and a motivation to water down regulation and oversight.

- Genuine ambiguity

  **Actual ambiguity:** In many of the scenarios we discuss, humanity's situation might be good in easy-to-measure ways. This means getting buy-in to challenge the status quo could be difficult.

  **Invisible misalignment:** The AI systems might not be acting in dangerous, power-seeking, or obviously misaligned ways early on. This could either be because of deliberate deception (deceptive alignment) or because the systems only fail to effectively generalize their goals on a very large scale or over long time horizons, so the misbehavior takes years to show up.

Clearly, there are many reasons to expect strong competitive pressure to develop HLMI. But how plausible is the idea that competitive pressures would be so high that potentially dangerous AI would be deployed despite major concerns? There are two intuitions we might have before looking into the details of the slow scenarios. We illustrate these intuitions with examples from existing writing on this question:

#### 8.6.2.1 Unprecedentedly Dangerous

Transformative AI has the potential to cause unprecedented damage, all the way up to human extinction. Therefore, our response to other very dangerous technologies such as nuclear weapons is a good analogy for our response to HLMI. It is significantly less likely for HLMI to be deployed if first there are many worsening warning shots involving dangerous AI systems. This would be comparable to an unrealistic alternate history where nuclear weapons were immediately used by the US and Soviet Union as soon as they were developed and in every war where they might have offered a temporary advantage, resulting in nuclear annihilation in the 1950s.[19] From Ngo [183]:

> The second default expectation about technology is that, if using it in certain ways is bad for humanity, we will stop people from doing so. This is a less reliable extrapolation—there are plenty of seemingly-harmful applications of technology which are still occurring. But note that we're talking about a slow-rolling *catastrophe*—that is, a situation which is unprecedentedly harmful. And so we should expect an unprecedented level of support for preventing whatever is causing it, all else equal.

Perhaps the development of HLMI will be similar enough to the development of nuclear weapons that, by analogy with this past development, we can claim evidence that harmful AI takeover is unlikely. In order for the risk from HLMI to be like the risk from nuclear escalation, the potential HLMI disaster would have to have a clear precedent (some small-scale version of the disaster has already occurred), the delay between the poor decision and the negative consequence would have to be very short, and we would have to be sure beforehand that deployment would be catastrophic (an equivalent of mutually assured destruction). Carlsmith discusses such a scenario as potentially plausible:

> it seems plausible to me that we see PS [power-seeking]-alignment failures of escalating severity (e.g., deployed AI systems stealing money, seizing control of infrastructure, manipulating humans on large scales), some of which may be quite harmful, but which humans ultimately prove capable of containing and correcting.

---

[19] Note that this is not the same as an alternate history where nuclear near misses escalated (e.g., Petrov [180], Vasili Arkhipov [181]), but instead an outcome where nuclear weapons were used as ordinary weapons of war with no regard for the larger dangers they presented—there would be no concept of "near misses" because MAD wouldn't have developed as a doctrine. In a previous post, we argued [182], following Anders Sandberg, that paradoxically the large number of nuclear near misses implies that there is a forceful pressure away from the worst outcomes.

> **Robert Wiblin:** So just to be clear, you're saying there's a lot of near misses, but that hasn't updated you very much in favor of thinking that the risk is very high. That's the reverse of what we expected.
>
> **Anders Sandberg:** Yeah.
>
> **Robert Wiblin:** Explain the reasoning there.
>
> **Anders Sandberg:** So imagine a world that has a lot of nuclear warheads. So if there is a nuclear war, it's guaranteed to wipe out humanity, and then you compare that to a world where [there are] a few warheads. So if there's a nuclear war, the risk is relatively small. Now in the first dangerous world, you would have a very strong deflection. Even getting close to the state of nuclear war would be strongly disfavored because most histories close to nuclear war end up with no observers left at all.
>
> In the second one, you get the much weaker effect, and now over time you can plot when the near misses happen and the number of nuclear warheads, and you actually see that they don't behave as strongly as you would think. If there was a very strong anthropic effect you would expect very few near misses during the height of the Cold War, and in fact you see roughly the opposite. So this is weirdly reassuring. In some sense the Petrov incident implies that we are slightly safer about nuclear war.

### 8.6.2.2 Unprecedentedly Useful

Transformative AI has the potential to accelerate economic growth by an unprecedented amount, potentially resulting in an entirely new growth regime far faster than today's. A scenario where we don't take shortcuts when deploying HLMI systems may be comparable to an unrealistic alternate history where the entire world refrained from industrializing and stopped additional burning of fossil fuels after early evidence of climate change became widely available in the 1960s. From Carlsmith [167]:

> Climate change might be some analogy. Thus, the social costs of carbon emissions are not, at present, adequately reflected in the incentives of potential emitters—a fact often thought key to ongoing failures to curb net-harmful emissions. Something similar could hold true of the social costs of actors risking the deployment of practically PS [power-seeking] -misaligned APS [agentic AI] systems for the sake of e.g. profit, global power, and so forth…
>
> ...The first calculations of the greenhouse effect occurred in 1896; the issue began to receive attention in the highest levels of national and international governance in the late 1960s; and scientific consensus began to form in the 1980s. Yet here we are, more than 30 years later, with the problem unsolved, and continuing to escalate—thanks in part to the multiplicity of relevant actors (some of whom deny/minimize the problem even in the face of clear evidence), and the incentives and externalities faced by those in a position to do harm. There are many disanalogies between PS-alignment risk and climate change (notably, in the possible—though not strictly necessary—immediacy, ease of attribution, and directness of AI-related harms), but we find the comparison sobering regardless. At least in some cases, "warnings" aren't enough.

Just as with the optimistic analogy to nuclear weapons, we can ask what AI takeover scenarios fit with this pessimistic analogy to climate change. The relevance of the climate change analogy will depend on the lag between early signs of profit/success and early signs of damage, as well as how much of the damage represents an externality to the whole of society versus directly backfiring onto the stakeholders of the individual project in a short time. It might also depend on how well (power-seeking) alignment failures are understood and (relatedly) how strong public backlash is (which could also depend on whether AI causes other non-alignment-related, non-existential-level harms, e.g., widespread unemployment and widening inequality).[20]

## 8.6.3 Takeover Capabilities

In each scenario, there is a certain understanding of what capabilities are necessary for AIs to seize control of the future from humanity. The assumption about how capable AIs need to be varies for two reasons. The first is that some scenarios make different assumptions than others about the intrinsic vulnerability of human civilization. The second is that in different scenarios, TAIs become obviously adversarial to humans and start fighting back at different points in their development.

[20] My first guess is that these two extra considerations seem to push in opposite directions: we expect PS alignment failures will be less well understood than climate change, but that AI will have some harms on society before we get to clearly HLMI systems

Some scenarios (such as brain-in-a-box) describe systems acting in ways that provoke human opposition almost immediately, so if those scenarios result in AI takeover, the systems must be supremely capable (able to defeat all opponents with no starting resources). Other scenarios assume a "creeping failure" where competitive pressures mean humans allow AI systems to monopolize resources and build up infrastructure for a while before the systems execute a takeover (such as AAFS). In these scenarios, the HLMI systems need to be capable enough to defeat human opposition while already having access to factories, drones, large amounts of money, etc., which requires fewer assumptions about the AI's capabilities.

How do we quantify the "intrinsic vulnerability" of human civilization? It is hard to assess how much of an advantage is required to secure a DSA. Two intuitions on this question point in radically different directions:

- **Economic:** To be able to outcompete the rest of the world, your project has to represent a substantial fraction of the *entire world's capability* on some crucial metric relevant to competitive success, because if you are actively seeking to take over the world then you will face opposition from everyone else. Perhaps that should be measured by GDP, military power, the majority of the world's AI compute, or some other measure [96]. For a single project to represent a large fraction of world GDP, you would need either an extraordinary effort to concentrate resources or an assumption of sudden, off-trend rapid capability gain such that the leading project can race ahead of competitors.

- **Historical:** Humans with no substantial advantage over the rest of humanity have in fact secured what Sotala called a "major strategic advantage" (MSA) [184] repeatedly in the past. For example: Hitler in 1920 had access to a microscopic fraction of global GDP / human brain-compute / (any other metric of capability) but had secured an MSA 20 years later (since his actions did lead to the deaths of 10+ million people), along with control over a significant fraction of the world's resources. No single human has ever secured a DSA (the closest was probably Cyrus the Great, who at one point ruled just under half of humanity). We might reasonably assume that if a human was able to rule over 45% of humanity, taking control of 100% would not require vastly greater capability.

Even if the absolute baseline capabilities required to achieve a DSA are unknown, we can see how other aspects of the scenario will raise or lower the required capabilities. Specifically, there is the issue of Dependency and deskilling [179]. We might integrate misaligned AI into our infrastructure when it attempts to take over. If we hand over lots of decision-making power to systems, they don't need to be as capable to take over. Taken to the extreme, we can imagine a scenario where we directly hand over control of the future to the systems, such that no takeover is even necessary.

Even if this is not the case, a given system might already have centralized control of important services [183] before acting in ways that motivate a human counterresponse. They would still have to go the extra mile and secure full control from their strong starting point, but that is necessarily easier than a brain-in-a-box taking over from nothing.

All else being equal, we might expect a greater degree of required takeover capability in faster stories or stories where systems are more obviously power-seeking from the start. The more clearly dangerous and power-seeking systems are, the more likely we are to try to stop them instead of succumbing to pressures to persist in deploying them, so either the response needs to be worse or the HLMI's capabilities need to be greater for takeover to succeed.

Asking "how capable" HLMI systems need to be to take over, and discussing factors that might influence this, is an abstraction that covers up the question of *which* capabilities are necessary.

Some failure scenarios don't discuss the exact route by which final takeover occurs, but sometimes they emphasize a particular approach (such as massed drone swarms or highly effective manipulation propaganda). Ngo breaks down the takeover capabilities into two general categories of manipulation and direct destructive capabilities [183].

Example takeover capabilities:

- Nanotechnology
- Drones, advanced robotics
- Biotechnology
- Persuasion skills
- Cyberoffense skills

In general, fast scenarios must assume systems can take over from a very weak starting point, which is more likely to require the capabilities to seize control of already existing infrastructure (persuasion skills and cyberoffense skills), while slow scenarios that assume takeover begins when the systems are already well established in the world economy might only require the ability to make use of that infrastructure to defeat opponents (advanced robotics and drones, biotechnology).

### 8.6.4 Alignment "Hackability"

"Competitive pressures" determine how much extra effort is put into aligning and overseeing AI systems—if the pressures are weaker, then we assume more effort is put into alignment and oversight because there is less incentive to cut corners. However, scenarios also differ on how "hackable" the alignment problem is—that is, how easy it is to "correct" misbehavior by methods of incremental course correction such as improving oversight and sensor coverage or tweaking reward functions. This correction requires two parts—first, noticing that there is a problem with the system early on,[21] then determining what fix to employ and applying it.

[21] Many of the same considerations around correcting misbehavior also apply to detecting it, and the required capabilities seem to overlap. In this chapter, we focus on applying corrections to misbehavior, but there is existing writing on detecting it as well.
Considering inner alignment, Trazzi and Armstrong [177] argue that models don't have to be very competent to be appear aligned when they are not, suggesting that it's possible that it won't be easy to tell if deployed systems are inner misaligned. But their argument doesn't have too much to say about how likely this is in practice.
Considering outer alignment, it seems less clear. See this summary [185] of some discussion between Richard Ngo and Paul Christiano about how easy it will be to tell that models are outer misaligned to the objective of pursuing easily measurable goals (rather than the hard-to-measure goals that we actually want).

In fast-takeoff worlds, the hackability of the alignment problem doesn't matter. There is no opportunity for alignment via course correction: either the AIs that rapidly become superintelligent are aligned or they are not.

In slow-takeoff worlds, the hackability of the alignment problem appears to have a U-shaped effect on how good the outcomes are. On one extreme, the alignment problem is hackable "all the way"—that is, we can incrementally correct AI systems as we go until we end up with existentially safe HLMI.[22] On the other extreme, the alignment problem isn't hackable at all. This might seem like a terrible outcome, but if it is the reality, it will probably lead to many early warning shots (i.e., small- or medium-scale accidents caused by alignment failures) that cannot be fixed. These will hopefully illustrate the danger ahead and bring about a slowdown in AI development and deployment until we have robust solutions to alignment.[23]

Between these two extremes, things seem to be more existentially risky. Consider if the alignment problem is "hackable until it isn't"—that is, for systems of lower capability, we can patch our way towards systems that do what we want, but as systems become increasingly capable, this becomes impossible. In this world, warning shots are likely to result in fixes that "work" in the short term, in the sense that they fix the specific problem. This gives humans confidence, resulting in more systems being deployed and more decision-making power being handed over to them. But this course correction becomes unworkable as systems become more capable, until eventually the alignment failure of a highly capable system results in existential catastrophe.

What predictions can we make today about how hackable the alignment problem is? Considering outer alignment: without any breakthroughs in techniques, there seems to be a strong case that we are on track towards the "intermediate" world where the alignment problem is hackable until it isn't. It seems like the best workable approach to outer alignment we have so far is to train systems to try to ensure that the world looks good according to some kind of (augmented) human judgment (i.e., using something like the training regime described in "An unaligned benchmark" [176][24]). This will result in a world that "looks good until it doesn't" for the reasons described in Another (outer) alignment failure story [162].

Considering inner alignment: it's unclear how pervasive of a problem inner misalignment will turn out to be, and also how competent systems have to be to appear aligned when they are not. To the extent that inner alignment is a pervasive problem, and models don't have to be very competent to appear aligned when they are not [177], then this also looks like the "intermediate" world where we can hack around the alignment problem, deploying increasingly capable systems, until a treacherous turn results in catastrophe.

[22] If you think that the methods we are most likely to employ absent an attempt to change research paradigms are exactly these highly hackable methods, then you accept the claim [186] of Alignment by Default [187].

[23] Unless, of course, (1) competitive pressures are very strong or (2) the governance response to warning shots is very incompetent (and AI companies just keep deploying misaligned systems leading to a highly unstable world due to lots of AI disasters). Neither of these assumptions seem particularly plausible to us.

[24] Whether the method described in "an unaligned benchmark" (which would result in this risky, intermediate level of hackability) actually turns out to be the most natural method to use for building advanced AI will depend on how easily it produces useful, intelligent behavior. If we are lucky, the method (something like model-based RL with MCTS) is one we will have incentive to use depends on how capable it is. If we are lucky, there will be more of a correlation between methods that are easily hackable and methods that produce capability, such that highly hackable methods are easier to find and more capable than even intermediately hackable methods.

## 8.7 Conclusion

We have described distinguishing features of some of the most widely discussed AI takeover scenarios and identified four characteristics which help us to interpret and examine those scenarios. Each scenario is unique, but there are large differences in which assumptions you need to make about these characteristics in order for slow scenarios vs fast scenarios to be plausible.

Compared to fast scenarios, slow scenarios don't rely as much on decisions to deploy single dangerous AIs but make more assumptions about incentives to widely deploy dangerous systems over a long period. From one perspective, this assumption about competitive pressures is the default, because that's what humans have tended to do throughout our history when some lucrative new technology has been made available. From another perspective, the unprecedented danger posed by HLMI implies a strong incentive to avoid making any mistakes.

Similarly, aside from the obvious assumption of rapid capability gain, fast stories also differ from slow stories in that they require systems to be capable enough to seize power from a very weak starting point (since in the slow stories, HLMI systems are instead *given* power). How plausible is it that a system could seize power from such a weak starting point? The economic analogy suggests a system would need to acquire a substantial fraction of the world's resources before attempting to take over, while the historical analogy suggests the system might not need to be much more intelligent than a smart human.

Finally, fast stories don't really make any assumptions about alignment hackability—they just assume progress is too fast to course correct. Slow stories generally assume hackability is not too high or too low—if hackability is too high there will be no disaster, and if it is too low there will be many escalating warning shots.

# 9 Epilogue

This report sketches our initial thoughts about the structure of risks from advanced artificial intelligence. While preliminary, this type of detailed conceptual understanding of how various questions related to the risk from HLMI are interrelated is critical for better planning and risk mitigation efforts. Because the domain is complex and our overall understanding is still tentative, we think it is especially critical to break down the conceptual model which is discussed by AI safety researchers, AI governance researchers, and policymakers into pieces that are individually understandable, discussable, and debatable. While the model development itself is being continued by some of the group who worked on the report, we hope that the report itself contributes to understanding and ongoing debates, and prompts further discussion.